%% file: PaperForReview.tex
\documentclass[10pt,twocolumn,letterpaper]{article}

\usepackage{wacv}              

\usepackage{graphicx}
\usepackage{amsmath}
\usepackage{amssymb}
\usepackage{booktabs}

\usepackage{caption}
\usepackage{subcaption}
\usepackage{tikz}
\usepackage{multicol}

\usepackage[pagebackref,breaklinks,colorlinks]{hyperref}

\usepackage[capitalize]{cleveref}
\crefname{section}{Sec.}{Secs.}
\Crefname{section}{Section}{Sections}
\Crefname{table}{Table}{Tables}
\crefname{table}{Tab.}{Tabs.}

\def\wacvPaperID{1589} 
\def\confName{WACV}
\def\confYear{2024}

\newcommand*{\affaddr}[1]{#1} 
\newcommand*{\affmark}[1][*]{\textsuperscript{#1}}
\newcommand*{\email}[1]{\texttt{#1}}

\begin{document}

\title{Self-Supervised Video Transformers for Isolated Sign Language Recognition}

\author{%
Marcelo Sandoval-Casta\~neda\affmark[1]\affmark[*], Yanhong Li\affmark[2], Diane Brentari\affmark[2], Karen Livescu\affmark[1], and Gregory Shakhnarovich\affmark[1]\\
\affaddr{\affmark[1]Toyota Technological Institute at Chicago, IL, USA}\\
\affaddr{\affmark[2]University of Chicago, IL, USA}\\
\email{\affmark[*]marcelo@ttic.edu}\\
}

\maketitle

\begin{abstract}
   This paper presents an in-depth analysis of various self-supervision methods for isolated sign language recognition (ISLR). We consider four recently introduced transformer-based approaches to self-supervised learning from videos, and four pre-training data regimes, and study all the combinations on the WLASL2000 dataset.
   Our findings reveal that MaskFeat achieves performance superior to pose-based and supervised video models, with a top-1 accuracy of 79.02\% on gloss-based WLASL2000. Furthermore, we analyze these models' ability to produce representations of ASL signs using linear probing on diverse phonological features. This study underscores the value of architecture and pre-training task choices in ISLR. Specifically, our results on WLASL2000 highlight the power of masked reconstruction pre-training, and our linear probing results demonstrate the importance of hierarchical vision transformers for sign language representation.
\end{abstract}

\input{sections/introduction}

\input{sections/related-work}

\input{sections/method}

\input{sections/experimental-setup}

\input{sections/experimental-results}

\input{sections/conclusion}


{\small
\bibliographystyle{ieee_fullname}
\bibliography{egbib}
}

\clearpage

\appendix
\input{sections/appendix}

\end{document}

%% file: sections/introduction.tex
\section{Introduction}
\label{sec:intro}

American Sign Language (ASL) is the predominant language of Deaf communities in the United States, with an estimated 500,000 native users~\cite{ethnologue}. However, natural language processing has largely focused on spoken and written language only~\cite{yin2021including}. Recently, there is an increasing body of work in NLP tasks related to sign language, with particular focus on isolated sign language recognition (ISLR) and sign language translation (SLT).

\begin{figure}
    \centering
    \includegraphics[width=\linewidth]{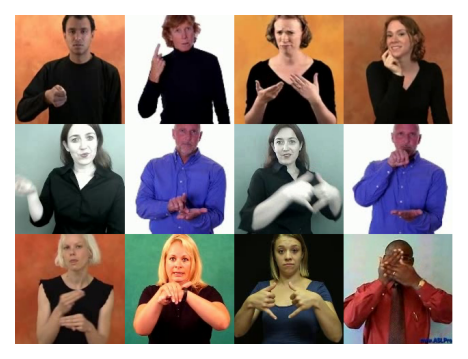}
    \caption{Sample frames from isolated sign language video extracted from the WLASL2000 dataset~\cite{Li2020WLASL}.}
    \label{fig:wlasl-samples}
\end{figure}

In ISLR, the input consists of dictionary-style videos which have only one signer articulating one individual sign, and the task is to classify this video according to the sign label. Figure~\ref{fig:wlasl-samples} shows frames extracted from such videos, which are typically two or three seconds long, have a solid color background, and show people signing while standing facing the camera. ISLR labels are usually words in English or sign language glosses, which are typically a combination of morpheme translations into English along with differentiating phonological features like handshape and location.

\begin{figure}
    \centering
    \includegraphics[width=\linewidth]{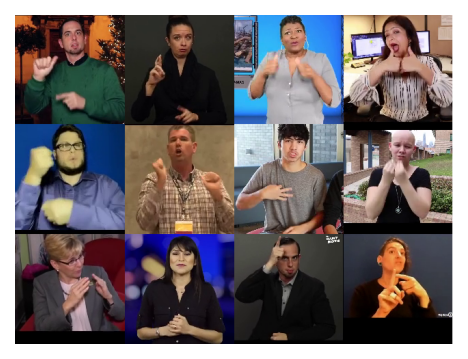}
    \caption{Sample frames extracted from ASL to English translation dataset OpenASL~\cite{Shi2022OpenASL}.}
    \label{fig:oasl-samples}
\end{figure}

On the other hand, in SLT the input consists of videos that contain continuous signing, and the task is to produce (usually text) translations in a given target language, typically the lingua franca of the region where the source sign language is predominant or English. The sources of these videos are highly variable, ranging from news sources to personal video blogs. Figure~\ref{fig:oasl-samples} shows example frames from sign language translation videos. Unlike ISLR, these videos are significantly longer, and can include all sorts of backgrounds, positions, and number of people, depending on the source it was extracted from.

In this paper, we focus on ISLR as a task to evaluate our models, as translation remains extremely difficult, with very poor state-of-the-art performance. Moreover, SLT models often rely on ISLR models as feature extractors~\cite{chen2022simple,muller2022findings,shi2022ttic}. Despite their usage in a language task, these models have not yet been evaluated in their ability to encode linguistically relevant features. In the case of sign languages, Brentari's prosodic model of sign language phonology~\cite{brentari1998prosodic} provides a set of phonological features that can uniquely characterize up to 70\% of signs in ASL~\cite{sehyr2021asllex}. These encode multiple characteristics of all signs, such as handshapes, movement, and symmetry.

Naturally, the most common modality of sign language data is video. Current state-of-the-art video classification tasks are dominated by vision transformers. Recent advances in this area have come from various self-supervised pre-training tasks, especially when dealing with smaller classification datasets. These are either based on some form of contrastive pre-training~\cite{pan2021videomoco,Kuang_2021_ICCV,Ranasinghe2021SVT} or masked reconstruction~\cite{Tong2022VideoMAE,Wang2021BEVT,Wei2022MaskFeat}, following the nature of advances in other fields like text, single images, and speech. These self-supervision methods have not yet been explored for sign language tasks. Self-supervision requires large datasets to function effectively, and large sign language datasets did not exist until very recently. In this paper, we study multiple video transformers and self-supervision settings to understand the effect of self-supervision for sign language processing tasks.

The contributions of this paper are the following:
\begin{itemize}
    \item \textbf{Establish a new state of the art on gloss-based WLASL2000.} MViTv2 with MaskFeat pre-training achieves 79.02\% accuracy, surpassing pose-based state of the art by 1.59\%.
    \item \textbf{Evaluate the effect of multiple self-supervision tasks on ISLR performance.} We compare pixel reconstruction (VideoMAE~\cite{Tong2022VideoMAE}), feature reconstruction (MaskFeat~\cite{Wei2022MaskFeat}), BERT pre-training for videos (BEVT~\cite{Wang2021BEVT}), and DINO pre-training for videos (SVT~\cite{Ranasinghe2021SVT}).
    \item \textbf{Introduce phonological features as a tool to analyze sign language representations produced by self-supervised models.} We use this to better characterize the strengths and limitations of architectures and pre-training tasks.
\end{itemize}

%% file: sections/related-work.tex
\section{Related Work}
\label{sec:related-work}

\subsection{Isolated Sign Language Recognition}
\label{sec:related-work-islr}

The most common approach to ISLR is based on convolutional neural networks (CNNs), such as I3D~\cite{carreira2017quo}, R3D~\cite{hara2017learning}, and S3D~\cite{xie2018rethinking}. These models receive RGB videos as input, extract features from them, and use these features to assign a label for the corresponding sign~\cite{Albanie2020BSL1K, Varol2021BSL1KImp, zuo2023natural}. Other approaches employ pose-based models to mitigate variations naturally present in video that are not related to the task of identifying signs~\cite{jiang2021skeleton, jiang2021sign, Dafnis2022Bidirectional}.

Choosing the right pre-training tasks and set-ups has largely been the determining factor in ISLR performance, all of which involve supervision on other datasets. The most common pre-training datasets include Kinetics400~\cite{carreira2017quo} and larger isolated sign language datasets in a language different from the target~\cite{Albanie2020BSL1K, Varol2021BSL1KImp, Novopoltsev2023Finetuning}. Recent approaches also include auxiliary tasks during ISLR training. These include weakly-supervised phonological features~\cite{kezar2023improving} and word embeddings from written language models~\cite{zuo2023natural}.

Our approach is most similar to~\cite{Novopoltsev2023Finetuning}, which uses a VideoSwin transformer with supervised pre-training on human action recognition with Kinetics400, then Russian Sign Language ISLR with a proprietary dataset, and finally fine-tuning on WLASL2000. In contrast, we pre-train on either human action videos or exclusively on ASL videos (or a combination of both) in a self-supervised manner, and then fine-tune for ISLR on WLASL2000.

\subsection{Self-Supervised Video Transformers}
\label{sec:related-work:-self-supervised-video-transformers}

In video self-supervision with transformers, there are four dominant approaches that can be broadly summarized by their choice of objective during pre-training:
masked reconstruction of pixels~\cite{Tong2022VideoMAE, feichtenhofer2022maest, Wang2023VideoMAE2}, DINO-based~\cite{caron2021emerging} representation learning using a teacher-student setup~\cite{Ranasinghe2021SVT, wang2022mvd}, masked reconstruction of hand-crafted features, most often HOG,~\cite{Wei2022MaskFeat, sun2023mme}, and BERT-like~\cite{devlin2018bert} masked token prediction using codewords generated by a discrete VAE~\cite{Wang2021BEVT, tan2021VIMPAC}. We choose one model from each category, specifically VideoMAE~\cite{Tong2022VideoMAE} for masked reconstruction over pixels, SVT~\cite{Ranasinghe2021SVT} for DINO-like representation learning, MaskFeat~\cite{Wei2022MaskFeat} for masked reconstruction over HOG, and BEVT~\cite{Wang2021BEVT} for BERT-like masked token prediction.

Additionally, there has also been a significant amount of work on custom architectures for vision transformers. ViT~\cite{dosovitskiy2020image} remains the most common architecture, which uses standard multi-head attention~\cite{vaswani2017attention}, with a fixed patch size of $16 \times 16$ pixels. Other custom architectures for vision have been proposed recently, most importantly VideoSwin~\cite{liu2022video} and MViT~\cite{fan2021multiscale, li2022mvitv2}. These use modified multi-head attention modules: window-based attention for VideoSwin and pooling attention for MViT, with much smaller input patches of $4 \times 4$ pixels. Our model choices also reflect this landscape, since VideoMAE and SVT are based on ViT, BEVT is based on VideoSwin, and MaskFeat is based on MViTv2.

%% file: sections/method.tex
\section{Method}
\label{sec:method}

For our experiments, we select one model from each method identified in Section~\ref{sec:related-work:-self-supervised-video-transformers}. We take the best performing architecture and set-up described in their respective papers.

{\bf VideoMAE~\cite{Tong2022VideoMAE}.} We pre-train a standard ViT with masked reconstruction over pixels, where the masked regions are tubes across time (i.e. the same 2D mask is applied to every frame), and with an extremely high masking ratio of 90\%. As a standard ViT, this model divides frames into patches of $16 \times 16$ pixels.

{\bf SVT~\cite{Ranasinghe2021SVT}.} We extend a standard ViT previously DINO pre-trained on ImageNet with further DINO pre-training on video. That is, we take a mean teacher~\cite{tarvainen2017mean} with a global view of the data and a student with narrower spatio-temporal slices, and the objective of the student is to produce representations from these narrower views that are as close as possible to the mean teacher. As a standard ViT, its patch size is $16 \times 16$ pixels.

{\bf MaskFeat~\cite{Wei2022MaskFeat}.} We pre-train a MViTv2 model for masked reconstruction over HOG, where the masked regions are segments over time of up to half the video length, composed of multiple adjacent patches, and with a much lower masking ratio compared to VideoMAE, 40\%. As a MViT-based model, the patch size is $4 \times 4$ pixels.

{\bf BEVT~\cite{Wang2021BEVT}.} We train a VideoSwin to predict labels or ``words'' for masked regions in a video. These labels are generated by the codebook of an external discrete VAE (dVAE), separately trained on Conceptual Captions~\cite{sharma2018conceptual} as part of DALL-E~\cite{ramesh2021zero}. The masking strategy is the same as in MaskFeat, but with a slightly higher masking ratio of 50\%. VideoSwin also takes as input patches of $4 \times 4$ pixels.

%% file: sections/experimental-setup.tex
\section{Experimental Setup}
\label{sec:experimental-setup}

\subsection{Datasets}
\label{sec:experimental-setup-datasets}

We rely on four different datasets, two for pre-training, one for fine-tuning, and one for linear probing of phonological features.

The first of these is Kinetics400~\cite{carreira2017quo}, the de facto standard dataset of human action videos. It contains 400 categories of human actions, with around 600 hours of video in total. We use it exclusively for self-supervised pre-training, ignoring the labels of the videos.

Additionally, we use OpenASL~\cite{Shi2022OpenASL}, one of the largest and most diverse ASL translation datasets, with 288 hours of ASL videos paired with English text translation and more than 200 signers. Since we only use it for self-supervised pre-training, we use OpenASL videos without their subtitle counterparts.

We fine-tune all of our models using WLASL2000~\cite{Li2020WLASL}. The original version of this dataset consisted of 14 hours of isolated sign language videos in ASL, with 2000 labels in total. However, recent analysis of this dataset~\cite{Dafnis2022Bidirectional} revealed major weaknesses and inconsistencies that are a product of relying on English translations as sign names. For this reason, we use the manually-corrected version of WLASL2000~\cite{neidle2022alternative}, with only 1535 labels based on ASL glosses instead of English translations and 12 hours of video.

For our linear probing experiments, we draw from sign language phonology, that is, the study of abstract grammatical components that are combined into meaningful utterances~\cite{brentari2019sign}. We extend the gloss-based WLASL2000 labels with phonological features extracted from ASL-LEX 2.0~\cite{sehyr2021asllex}. Due to inconsistent labeling across datasets and dialect variations, we only map a subset of 916 WLASL labels with their corresponding features where an unambiguous correspondence can be found. That is, we strip the morpheme translation part of each gloss label from both WLASL2000 and ASL-LEX 2.0, and only map one to another when there is a unique exact match. This is partially inspired by previous work~\cite{tavella2022wlasl,kezar2023improving}, with two key differences. First, we rely on manually labeled glosses instead of treating the original WLASL2000 labels as gloss approximations, and second, we do not use these additional features as a weak supervision signal, but as a tool help us characterize our self-supervised models.

For a detailed description of each phonological feature used in this study, see Table~\ref{tab:phono-features-description}. Note that the number of classes in Table~\ref{tab:phono-features-description} might be smaller than the original ASL-LEX 2.0 labels, as a product of unambiguous data available in WLASL2000. Additionally, Table~\ref{tab:phono-categorization} groups the phonological features we extract from ASL-LEX 2.0 into broad categories according to the aspect they describe.

\begin{table*}[]
\centering
\begin{tabular}{llc}
\hline
\textbf{Feature} & \textbf{Description} & \textbf{Classes} \\
\hline
Sign Type & Describes whether a sign is one-handed or two-handed and its symmetry properties. & 6 \\
Major Location & General location of the dominant hand in relation to the body. & 5 \\
First Minor Location & Subdivisions within each minor location except ``neutral'' location. & 35 \\
Second Minor Location & Used when the dominant hand moves while signing. & 30 \\
Contact & Describes whether the dominant hand touches the major location during signing. & 2 \\
Handshape & Shape of the dominant signing hand. & 49 \\
Non-Dom. Handshape & Shape of the non-dominant hand. & 45 \\
Selected Fingers & Describes combinations of finger movement and position in the dominant hand. & 10 \\
Flexion & Describes selected finger positions at morpheme onset. & 8 \\
Thumb Position & Describes whether the thumb is in contact with other fingers in the dominant hand. & 2 \\
Path Movement & Path followed by the dominant hand. & 8 \\
Wrist Twist & Describes whether ulnar rotation is present in a sign's movement. & 2 \\
Repeated Movement & Describes whether a sign involves movement repetition of any kind. & 2 \\
\hline
\end{tabular}
\caption{Description of all phonological features used in this study.}
\label{tab:phono-features-description}
\end{table*}

\begin{table}[]
\centering
\begin{tabular}{ll}
\hline
\textbf{Main Category} & \textbf{Features} \\
\hline
Sign Type & Sign Type \\
\hline
Location & Major Location, Minor Location, \\
  &Second Minor Location, Contact \\
\hline
Hand Configuration & Thumb Position, Flexion, \\
  &Handshape, Selected Fingers, \\
    & Non-dominant Handshape \\
\hline
Movement & Path Movement, Wrist Twist, \\
  & Repeated Movement \\
\hline
\end{tabular}
\caption{Broad cateogrization of phonological features included in this study.}
\label{tab:phono-categorization}
\end{table}

\subsection{Pre-Training}
\label{sec:experimental-setup-pretraining}

For each of our four models, we explore four pre-training configurations. The first is to pre-train each model in Kinetics400 using the pre-training setup and hyperparameters from the model's original paper. The second is to keep the setup and hyperparameters, but replace Kinetics400 with OpenASL. Our third configuration is to first pre-train the model on Kinetics400 as in the corresponding paper, and add a second pre-training stage on OpenASL, with all the hyperparameters staying the same except learning rate, which we instead reduce to a tenth of the original. Last, our fourth configuration is to pre-train a model on the union of Kinetics400 and OpenASL, keeping each model's original pre-training setup and hyperparameters.

\subsection{Evaluation}
\label{sec:experimental-setup-evaluation}

To evaluate our models, we conduct fine-tuning and layer-wise linear probing. We fine-tune on the gloss-based version of WLASL2000, and compare models by top-1 accuracy. For linear probing, we freeze the model and train softmax classifiers with the output of each layer as input, as an attempt to measure whether phonological features are encoded well in our trained models. We do so before and after fine-tuning on WLASL2000. This is inspired by previous work that analyzes linguistic properties of speech models~\cite{pasad2021layer,ji2022predicting,pasad2023comparative}. We select the models to analyze based on ISLR fine-tuning performance.

%% file: sections/experimental-results.tex
\section{Experimental Results}
\label{sec:experimental-results}

\subsection{Fine-Tuning}
\label{sec:experimental-results-fine-tuning}

Our fine-tuning results can be found in Table~\ref{tab:wlasl-performance}. We include the results of pose-based Bi-GCN on gloss-based WLASL2000, as reported in~\cite{Dafnis2022Bidirectional}, which is state of the art in the gloss-based setup. Additionally, we report results from video-based I3D models with supervised pre-training~\cite{Albanie2020BSL1K, Varol2021BSL1KImp}, which is state of the art on the original WLASL2000 with publicly available datasets and without additional signals, and that we train on gloss-based WLASL2000 for this paper.

We can see in Table~\ref{tab:wlasl-performance} that there is no clear superior pre-training dataset combination across all models and pre-training tasks. Both models that deal with reconstruction over masked regions, VideoMAE and MaskFeat, benefit from the combination of Kinetics and OpenASL. However, they favour different datasets in our single-dataset setting, albeit with similar behaviour for their respective performances. In the case of VideoMAE, OpenASL pre-training achieves much lower accuracy than that of Kinetics400 pre-training (16.19 vs. 67.25), and the benefit of our two-stage pretraining results in an increase of 2.07 after pre-training on both Kinetics400 and OpenASL. For MaskFeat, this is inverted: Kinetics400-only pre-training yields an accuracy of 12.50, much lower than OpenASL-only at 74.68. Two-stage pre-training pushes this number further, up to 79.02, 4.34 points above the best single-dataset MaskFeat. We hypothesize that these differences can be at least partially attributable to the pre-training objective. HOG may allow the model to mitigate the less-relevant variation over pixels, while focusing on aspects that are more important to sign language and are captured by HOG, especially shapes.

\begin{table}[]
\centering
\begin{tabular}{lcc}
\hline
\textbf{Model} & \textbf{Pre-Training} & \textbf{Top 1 Acc.} \\
\hline
\textit{Bi-GCN} & \textit{Pose} & \textit{77.43} \\
\textit{I3D} & \textit{K400 (Sup.)} & \textit{55.02} \\
\textit{I3D} & \textit{BSL-1K (Sup.)} & \textit{67.60} \\
\hline
VideoMAE & No Pre-Training & 0.47 \\
VideoMAE & K400 & 67.25 \\
VideoMAE & OpenASL & 16.19 \\
VideoMAE & Two-Stage & 69.32 \\
VideoMAE & Mixed & 59.28 \\
\hline
SVT & No Pre-Training & 8.52 \\
SVT & K400 & 19.34 \\
SVT & OpenASL & 13.36 \\
SVT & Two-Stage & 16.06 \\
SVT & Mixed & 11.22 \\
\hline
MaskFeat & No Pre-Training & 1.17 \\
MaskFeat & K400 & 12.50 \\
MaskFeat & OpenASL & 74.68 \\
\textbf{MaskFeat} & \textbf{Two-Stage} & \textbf{79.02} \\
MaskFeat & Mixed & 75.74 \\
\hline
BEVT & No Pre-Training & 0.47 \\
BEVT & K400 & 53.07 \\
BEVT & OpenASL & 60.73 \\
BEVT & Two-Stage & 58.69 \\
BEVT & Mixed & 59.63 \\
\hline
\end{tabular}
\caption{Top 1 accuracy on gloss-based WLASL2000 after fine-tuning for various models and pre-training datasets. Bi-GCN and I3D in \textit{italics} are supervised classification pre-training and pose-based baselines for reference. We take the best-performing word-based WLASL I3D setups and train them for gloss-based WLASL instead. Best performing model in \textbf{bold}. Two-Stage refers to our configuration where a model is first pre-trained with Kinetics400 and then pre-trained with OpenASL, and Mixed refers to our configuration where a model is trained on both Kinetics400 and OpenASL at the same time.}
\label{tab:wlasl-performance}
\end{table}

On the other hand, BEVT performs worse when combining both Kinetics400 and OpenASL, and the best performance is obtained through OpenASL pre-training only. Since the dVAE used in BEVT is frozen and obtained from training over a more diverse set of images, the addition of Kinetics400 leads to optimizing over code word labels that are entirely irrelevant to the task of ISLR, especially since Kinetics400 is larger than OpenASL. This, in turn, diminishes the final accuracy of BEVT, though not by much (60.73 for OpenASL only, and 58.69 for two-stage pre-training or 59.63 for mixed pre-training).

Lastly, SVT does not achieve comparable performance to our baselines under any setting. This is likely a product of its pre-training procedure: it is not necessarily meaningful to map a global view of a sentence in ASL to a smaller slice, especially across time, of the same sentence (which could be a single sign, or a few sign within the sentence). Therefore, it only manages to reach 19.34 accuracy with Kinetics400, with any OpenASL deteriorating its performance further.

We also find that no model benefits from mixed pre-training from scratch. The reason is model-dependent, but we see this happen across the board. For VideoMAE, we find that mixed pre-training performs significantly worse than both Kinetics400-only and two-stage pre-training. This suggests that VideoMAE benefits from first having a diverse enough set of videos and then narrowing down the domain to sign language in the second stage, as opposed to having a single dataset where sign language is highly represented. For SVT, any inclusion of sign language data deteriorates performance significantly. The case of MaskFeat is similar to the case of VideoMAE, but the improvement found from including Kinetics400 is not as significant as in VideoMAE. For BEVT, mixed pre-training deteriorates performance in comparison to OpenASL as a product of adding unrelated data to the training process, which is also found in the deterioration found in two-stage pre-training.

Some of our models surpass current state-of-the-art I3Ds in ISLR, in particular VideoMAE, pre-trained on Kinetics400 and OpenASL, and MaskFeat, pre-trained on both OpenASL alone and on Kinetics400 and OpenASL, using only a quarter of the frames used in I3D and without requiring additional supervision. Even more so, MaskFeat, on Kinetics400 and OpenASL, manages to perform better than Bi-GCN, which not only relies on external pose estimation, but also uses 150 timesteps for classification.

\subsection{Linear Probing}
\label{sec:experimental-results-linear-probing}

\begin{figure}
    \begin{subfigure}[b]{\linewidth}
        \centering
        \renewcommand\sffamily{}
        \input{figures/majorlocation-nox.pgf}
        \caption{Major Location (5 classes)}
        \label{fig:probing-maj-loc}
    \end{subfigure}

    \begin{subfigure}[b]{\linewidth}
        \centering
        \renewcommand\sffamily{}
        \input{figures/minorlocation-nox.pgf}
        \caption{Minor Location (35 classes)}
        \label{fig:probing-min-loc}
    \end{subfigure}

    \begin{subfigure}[b]{\linewidth}
        \centering
        \renewcommand\sffamily{}
        \input{figures/handshape-nox.pgf}
        \caption{Handshape (49 classes)}
        \label{fig:probing-handshape}
    \end{subfigure}

    \begin{subfigure}[b]{\linewidth}
        \centering
        \renewcommand\sffamily{}
        \input{figures/selectedfinger-nox.pgf}
        \caption{Selected Finger (10 classes)}
        \label{fig:probing-sel-finger}
    \end{subfigure}

    \begin{subfigure}[b]{\linewidth}
        \centering
        \renewcommand\sffamily{}
        \input{figures/pathmovement.pgf}
        \caption{Path Movement (8 classes)}
        \label{fig:probing-path-mov}
    \end{subfigure}
    \centering
    \renewcommand\sffamily{}
    \input{figures/probing-legend.pgf}
    \caption{Linear probing accuracy of pre-trained models across layers on select phonological features. Note the different y-axis scales for different features.}
    \label{fig:probing}
\end{figure}
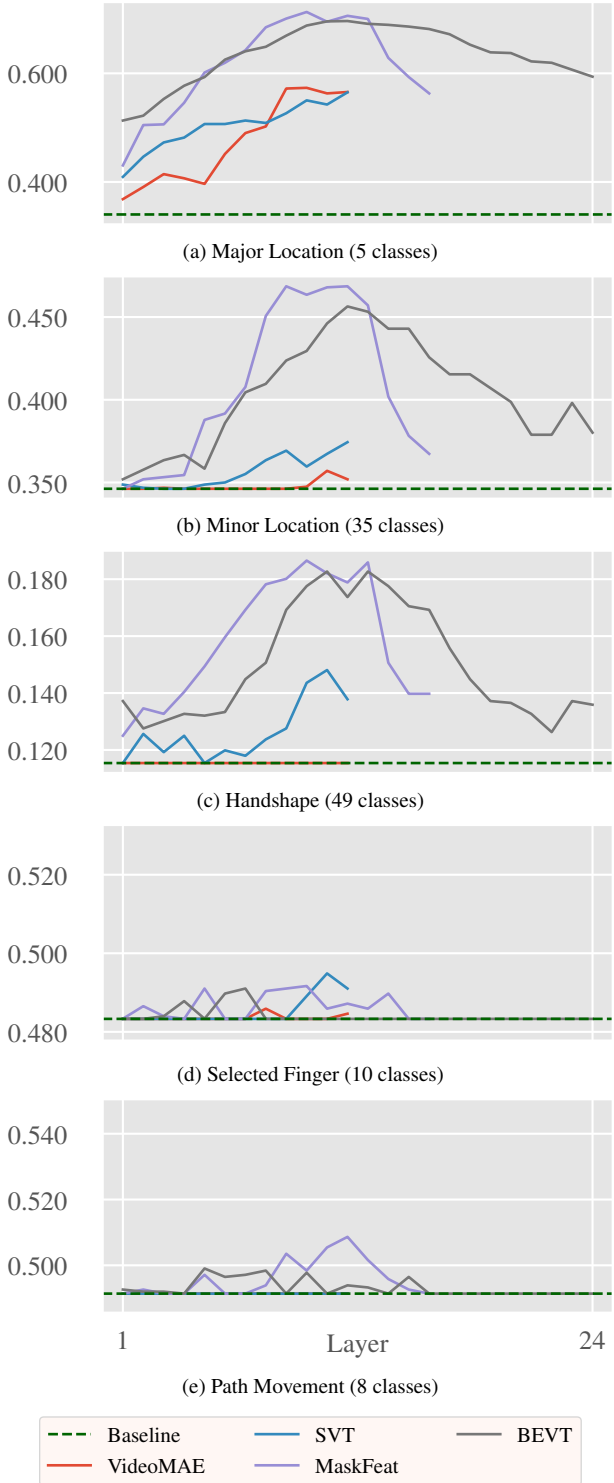

\begin{figure*}
    \caption*{\hspace{0.5in} Before Fine-Tuning \hspace{2in} After Fine-Tuning}
    \begin{subfigure}[b]{\linewidth}
        \centering
        \renewcommand\sffamily{}
        \input{figures/comparison-minorlocation-nox.pgf}
        \caption{Minor Location (35 classes)}
        \label{fig:comparison-min-loc}
    \end{subfigure}

    \begin{subfigure}[b]{\linewidth}
        \centering
        \renewcommand\sffamily{}
        \input{figures/comparison-handshape-nox.pgf}
        \caption{Handshape (49 classes)}
        \label{fig:comparison-handshape}
    \end{subfigure}

    \begin{subfigure}[b]{\linewidth}
        \centering
        \renewcommand\sffamily{}
        \input{figures/comparison-pathmovement.pgf}
        \caption{Path Movement (8 classes)}
        \label{fig:comparison-path-mov}
    \end{subfigure}
    
    \centering
    \renewcommand\sffamily{}
    \input{figures/comparison-legend.pgf}
    \caption{Comparison between pre-trained and fine-tuned models on select phonological features. Figures on the left are before fine-tuning and figures on the right are after fine-tuning. For our naive baseline, we take a model that predicts the most common label in the training set at all times.}
    \label{fig:comparison}
\end{figure*}

We evaluate each model based on the setting that achieves best fine-tuning accuracy on gloss-based WLASL2000, described in Section~\ref{sec:experimental-results-fine-tuning}. For our naive baseline, we take a model that predicts the most common label in the training set at all times. Thus, for VideoMAE we take the model with two-stage pre-training, for SVT we take the one with Kinetics400-only pre-training, for MaskFeat we take the one with two-stage pre-training, and for BEVT we take the model with OpenASL-only pre-training.

We first analyze the results of our linear probing experiment on the encoders of our chosen models before fine-tuning. Figure~\ref{fig:probing} shows our most meaningful findings, and graphs for all phonological features can be found in Appendix~\ref{sec:appendix-comparison-graphs}. Notably, hierarchical vision transformers (MViTv2 in MaskFeat and VideoSwin in BEVT) perform significantly better than the plain ViTs found in VideoMAE and SVT across all features where models achieve non-trivial performance. Even though VideoMAE achieves a much higher WLASL2000 fine-tuning accuracy, it produces less linguistically relevant representations than those from BEVT, for instance.

Additionally, BEVT and MaskFeat display a trend similar to that observed in self-supervised speech models~\cite{pasad2021layer}. Specifically, the middle layers of these models provide representations that better capture linguistic features compared to their initial or final layers. Much like in self-supervised speech models, it is likely that initial layers tend to focus on lower-level features in the input, and that final layers produce features that are most relevant to the corresponding pre-training tasks, at the expense of some linguistically relevant characteristics. We do not see the same phenomenon occur with VideoMAE and SVT. Layers that produce representations most related to our phonological features are the final layers. In the case of SVT, this is to be expected, as the pre-training task is explicitly to produce meaningfully different representations for each video in the dataset. However, VideoMAE's task being closely related to BEVT's and MaskFeat's makes this behavior surprising. For VideoMAE, final encoder layers do not diverge from producing representations that capture linguistic features in favor of task-related features, with this behavior likely occurring entirely in the decoder.

Comparing across phonological features also reveals interesting characteristics of our models. All of our models can provide meaningful representations for Major Location, as shown in Figure~\ref{fig:probing-maj-loc}. Major Location refers to where the dominant signing hand is broadly located, and takes one of five values: head, arm, body, non-dominant hand, or neutral space. In contrast, when we shift to the more specific Minor Location, which divides every Major Location except neutral space into eight smaller locations, it is evident that plain ViT-based models are largely unable to capture this feature. This effect can be seen in Figure~\ref{fig:probing-min-loc}. This is likely a product of the input patch size for each model, as VideoSwin and MViTv2 take in much smaller patches than plain ViT, and are thus unable to differentiate locations at this level of granularity.

When looking at hand configuration features, this deterioration is more dramatic and occurs across all models. For Handshape in Figure~\ref{fig:probing-handshape}, for example, most models are able to capture this feature, with the exception of VideoMAE. However, the graph for Selected Finger, in Figure~\ref{fig:probing-sel-finger} shows that none of our models provide meaningful representations in terms of this feature. This is likely a product of Selected Fingers being a feature that requires finer-grained differences that are not relevant to any of our pre-training tasks, such as differences in positions between fingers of the same hand. Additionally, hand configuration features largely rely on making sense of 3D information, particularly as the hand configuration is invariant to the orientation of the hand in relation to the camera.

Along the same lines, most movement-related phonological features are not properly captured by any of our models either. Path Movement, in Figure~\ref{fig:probing-path-mov}, is an example of this. Path Movement refers to the shape followed by the motion of the dominant hand. None of our models are able to produce meaningful representations for it, as shown by how linear probing accuracies remain close to the baseline accuracy. Movement features, and especially Path Movement, also require 3D information, and can use different planes as reference for signs that have the same type of movement.

We then also compare models' linear probing performance before and after fine-tuning on gloss-based WLASL2000. Our most significant findings are shown in Figure~\ref{fig:comparison}, and results for all phonological features can be found in Appendix~\ref{sec:appendix-comparison-graphs}. We have also included linear probing accuracy of an I3D model trained on WLASL2000 for an additional comparison. Most strikingly, we find that fine-tuning plain ViTs on supervised data does not necessarily produce better representations. This is partially consistent with previous work on ViTs~\cite{caron2021emerging}, which claims that the most important step in producing meaningful representations with vision transformers is self-supervised pre-training. However, we find that unlike claims in previous work on self-supervised video models~\cite{Tong2022VideoMAE}, the right transformer architecture is just as important as the right pre-training task, at least for producing sign language representations. Plain ViTs seem to be insufficient for this, despite some of them showing good performance in ISLR. In contrast, hierarchical video transformers with a reconstruction task almost always find significant improvement in linear probing accuracy after fine-tuning on ISLR, even if ISLR performance is inferior to that of plain ViTs (as is the case for our BEVT model).

Alongside improving the quality of models in terms of their ability to represent linguistic features, fine-tuning also changes the shape of our hierarchical models' linear probing curves. Though the fine-tuning task is not explicitly to predict any of these features, the task of classifying isolated signs benefits from having representations that capture them. Thus, as we move towards the final layers of these models, linear probing accuracy almost always monotonically increases. This is, again, consistent with what has been reported for self-supervised speech models~\cite{pasad2021layer}, where fine-tuning on speech recognition modifies the shape of the linear probing curves in similar ways.

The superiority of hierarchical vision transformers over ViT remains consistent after fine-tuning, too. As Figure~\ref{fig:comparison} shows, VideoMAE and SVT consistently perform worse than BEVT and MaskFeat across all features, and perhaps more importantly, do not achieve performance comparable to the I3D baseline for any feature. In comparison, both BEVT and MaskFeat generally perform comparably to or better than I3D for all phonological features.

Fine-tuning also allows some models which were previously unable to capture movement and time-dependent features to do so. An example of this is Path Movement in Figure~\ref{fig:comparison-path-mov}, where linear probing performance greatly improves for MaskFeat and BEVT. This is at least partially related to model architecture, as shown by the fact that VideoMAE and SVT still struggle in this setting. However, we suspect this is most significantly due to video sampling approaches during pre-training. As self-supervised video pre-training pipelines are typically designed with broad human action videos in mind, frames from these videos are sampled sparsely across time. However, this is insufficient to capture time-dependent properties of signs from continuous signing videos. Sixteen frames sampled uniformly from multiple full sentences is very little, and it leads to our models only getting one or two frames per sign from any given continuous signing video. This likely misses key linguistic information. Fine-tuning on ISLR then mitigates this, as we have much shorter videos all containing only one sign. During fine-tuning, the models get as input sixteen frames all corresponding to the same sign, and thus are able to capture motion-based features much better than from continuous signing.

%% file: figures/majorlocation-nox.pgf
\begingroup%
\makeatletter%
\begin{pgfpicture}%
\pgfpathrectangle{\pgfpointorigin}{\pgfqpoint{3.250000in}{1.250000in}}%
\pgfusepath{use as bounding box, clip}%
\begin{pgfscope}%
\pgfsetbuttcap%
\pgfsetmiterjoin%
\definecolor{currentfill}{rgb}{1.000000,1.000000,1.000000}%
\pgfsetfillcolor{currentfill}%
\pgfsetlinewidth{0.000000pt}%
\definecolor{currentstroke}{rgb}{0.500000,0.500000,0.500000}%
\pgfsetstrokecolor{currentstroke}%
\pgfsetdash{}{0pt}%
\pgfpathmoveto{\pgfqpoint{0.000000in}{0.000000in}}%
\pgfpathlineto{\pgfqpoint{3.250000in}{0.000000in}}%
\pgfpathlineto{\pgfqpoint{3.250000in}{1.250000in}}%
\pgfpathlineto{\pgfqpoint{0.000000in}{1.250000in}}%
\pgfpathlineto{\pgfqpoint{0.000000in}{0.000000in}}%
\pgfpathclose%
\pgfusepath{fill}%
\end{pgfscope}%
\begin{pgfscope}%
\pgfsetbuttcap%
\pgfsetmiterjoin%
\definecolor{currentfill}{rgb}{0.898039,0.898039,0.898039}%
\pgfsetfillcolor{currentfill}%
\pgfsetlinewidth{0.000000pt}%
\definecolor{currentstroke}{rgb}{0.000000,0.000000,0.000000}%
\pgfsetstrokecolor{currentstroke}%
\pgfsetstrokeopacity{0.000000}%
\pgfsetdash{}{0pt}%
\pgfpathmoveto{\pgfqpoint{0.536502in}{0.041670in}}%
\pgfpathlineto{\pgfqpoint{3.208330in}{0.041670in}}%
\pgfpathlineto{\pgfqpoint{3.208330in}{1.208330in}}%
\pgfpathlineto{\pgfqpoint{0.536502in}{1.208330in}}%
\pgfpathlineto{\pgfqpoint{0.536502in}{0.041670in}}%
\pgfpathclose%
\pgfusepath{fill}%
\end{pgfscope}%
\begin{pgfscope}%
\pgfpathrectangle{\pgfqpoint{0.536502in}{0.041670in}}{\pgfqpoint{2.671828in}{1.166660in}}%
\pgfusepath{clip}%
\pgfsetrectcap%
\pgfsetroundjoin%
\pgfsetlinewidth{0.803000pt}%
\definecolor{currentstroke}{rgb}{1.000000,1.000000,1.000000}%
\pgfsetstrokecolor{currentstroke}%
\pgfsetdash{}{0pt}%
\pgfpathmoveto{\pgfqpoint{0.643375in}{0.041670in}}%
\pgfpathlineto{\pgfqpoint{0.643375in}{1.208330in}}%
\pgfusepath{stroke}%
\end{pgfscope}%
\begin{pgfscope}%
\pgfpathrectangle{\pgfqpoint{0.536502in}{0.041670in}}{\pgfqpoint{2.671828in}{1.166660in}}%
\pgfusepath{clip}%
\pgfsetrectcap%
\pgfsetroundjoin%
\pgfsetlinewidth{0.803000pt}%
\definecolor{currentstroke}{rgb}{1.000000,1.000000,1.000000}%
\pgfsetstrokecolor{currentstroke}%
\pgfsetdash{}{0pt}%
\pgfpathmoveto{\pgfqpoint{3.101457in}{0.041670in}}%
\pgfpathlineto{\pgfqpoint{3.101457in}{1.208330in}}%
\pgfusepath{stroke}%
\end{pgfscope}%
\begin{pgfscope}%
\pgfpathrectangle{\pgfqpoint{0.536502in}{0.041670in}}{\pgfqpoint{2.671828in}{1.166660in}}%
\pgfusepath{clip}%
\pgfsetrectcap%
\pgfsetroundjoin%
\pgfsetlinewidth{0.803000pt}%
\definecolor{currentstroke}{rgb}{1.000000,1.000000,1.000000}%
\pgfsetstrokecolor{currentstroke}%
\pgfsetdash{}{0pt}%
\pgfpathmoveto{\pgfqpoint{0.536502in}{0.264826in}}%
\pgfpathlineto{\pgfqpoint{3.208330in}{0.264826in}}%
\pgfusepath{stroke}%
\end{pgfscope}%
\begin{pgfscope}%
\definecolor{textcolor}{rgb}{0.333333,0.333333,0.333333}%
\pgfsetstrokecolor{textcolor}%
\pgfsetfillcolor{textcolor}%
\pgftext[x=0.041670in, y=0.212065in, left, base]{\color{textcolor}\sffamily\fontsize{10.000000}{12.000000}\selectfont 0.400}%
\end{pgfscope}%
\begin{pgfscope}%
\pgfpathrectangle{\pgfqpoint{0.536502in}{0.041670in}}{\pgfqpoint{2.671828in}{1.166660in}}%
\pgfusepath{clip}%
\pgfsetrectcap%
\pgfsetroundjoin%
\pgfsetlinewidth{0.803000pt}%
\definecolor{currentstroke}{rgb}{1.000000,1.000000,1.000000}%
\pgfsetstrokecolor{currentstroke}%
\pgfsetdash{}{0pt}%
\pgfpathmoveto{\pgfqpoint{0.536502in}{0.833811in}}%
\pgfpathlineto{\pgfqpoint{3.208330in}{0.833811in}}%
\pgfusepath{stroke}%
\end{pgfscope}%
\begin{pgfscope}%
\definecolor{textcolor}{rgb}{0.333333,0.333333,0.333333}%
\pgfsetstrokecolor{textcolor}%
\pgfsetfillcolor{textcolor}%
\pgftext[x=0.041670in, y=0.781049in, left, base]{\color{textcolor}\sffamily\fontsize{10.000000}{12.000000}\selectfont 0.600}%
\end{pgfscope}%
\begin{pgfscope}%
\pgfpathrectangle{\pgfqpoint{0.536502in}{0.041670in}}{\pgfqpoint{2.671828in}{1.166660in}}%
\pgfusepath{clip}%
\pgfsetrectcap%
\pgfsetroundjoin%
\pgfsetlinewidth{1.003750pt}%
\definecolor{currentstroke}{rgb}{0.886275,0.290196,0.200000}%
\pgfsetstrokecolor{currentstroke}%
\pgfsetdash{}{0pt}%
\pgfpathmoveto{\pgfqpoint{0.643375in}{0.174795in}}%
\pgfpathlineto{\pgfqpoint{0.750249in}{0.238582in}}%
\pgfpathlineto{\pgfqpoint{0.857122in}{0.306015in}}%
\pgfpathlineto{\pgfqpoint{0.963995in}{0.284145in}}%
\pgfpathlineto{\pgfqpoint{1.070868in}{0.254985in}}%
\pgfpathlineto{\pgfqpoint{1.177741in}{0.411720in}}%
\pgfpathlineto{\pgfqpoint{1.284614in}{0.521070in}}%
\pgfpathlineto{\pgfqpoint{1.391487in}{0.555697in}}%
\pgfpathlineto{\pgfqpoint{1.498360in}{0.754350in}}%
\pgfpathlineto{\pgfqpoint{1.605233in}{0.757995in}}%
\pgfpathlineto{\pgfqpoint{1.712107in}{0.728835in}}%
\pgfpathlineto{\pgfqpoint{1.818980in}{0.736125in}}%
\pgfusepath{stroke}%
\end{pgfscope}%
\begin{pgfscope}%
\pgfpathrectangle{\pgfqpoint{0.536502in}{0.041670in}}{\pgfqpoint{2.671828in}{1.166660in}}%
\pgfusepath{clip}%
\pgfsetrectcap%
\pgfsetroundjoin%
\pgfsetlinewidth{1.003750pt}%
\definecolor{currentstroke}{rgb}{0.203922,0.541176,0.741176}%
\pgfsetstrokecolor{currentstroke}%
\pgfsetdash{}{0pt}%
\pgfpathmoveto{\pgfqpoint{0.643375in}{0.291435in}}%
\pgfpathlineto{\pgfqpoint{0.750249in}{0.397140in}}%
\pgfpathlineto{\pgfqpoint{0.857122in}{0.471862in}}%
\pgfpathlineto{\pgfqpoint{0.963995in}{0.497377in}}%
\pgfpathlineto{\pgfqpoint{1.070868in}{0.568455in}}%
\pgfpathlineto{\pgfqpoint{1.177741in}{0.568455in}}%
\pgfpathlineto{\pgfqpoint{1.284614in}{0.586680in}}%
\pgfpathlineto{\pgfqpoint{1.391487in}{0.573922in}}%
\pgfpathlineto{\pgfqpoint{1.498360in}{0.624952in}}%
\pgfpathlineto{\pgfqpoint{1.605233in}{0.692385in}}%
\pgfpathlineto{\pgfqpoint{1.712107in}{0.670515in}}%
\pgfpathlineto{\pgfqpoint{1.818980in}{0.734302in}}%
\pgfusepath{stroke}%
\end{pgfscope}%
\begin{pgfscope}%
\pgfpathrectangle{\pgfqpoint{0.536502in}{0.041670in}}{\pgfqpoint{2.671828in}{1.166660in}}%
\pgfusepath{clip}%
\pgfsetrectcap%
\pgfsetroundjoin%
\pgfsetlinewidth{1.003750pt}%
\definecolor{currentstroke}{rgb}{0.596078,0.556863,0.835294}%
\pgfsetstrokecolor{currentstroke}%
\pgfsetdash{}{0pt}%
\pgfpathmoveto{\pgfqpoint{0.643375in}{0.351577in}}%
\pgfpathlineto{\pgfqpoint{0.750249in}{0.562987in}}%
\pgfpathlineto{\pgfqpoint{0.857122in}{0.566632in}}%
\pgfpathlineto{\pgfqpoint{0.963995in}{0.679627in}}%
\pgfpathlineto{\pgfqpoint{1.070868in}{0.838185in}}%
\pgfpathlineto{\pgfqpoint{1.177741in}{0.889215in}}%
\pgfpathlineto{\pgfqpoint{1.284614in}{0.954825in}}%
\pgfpathlineto{\pgfqpoint{1.391487in}{1.075110in}}%
\pgfpathlineto{\pgfqpoint{1.498360in}{1.120672in}}%
\pgfpathlineto{\pgfqpoint{1.605233in}{1.155300in}}%
\pgfpathlineto{\pgfqpoint{1.712107in}{1.104270in}}%
\pgfpathlineto{\pgfqpoint{1.818980in}{1.135252in}}%
\pgfpathlineto{\pgfqpoint{1.925853in}{1.118850in}}%
\pgfpathlineto{\pgfqpoint{2.032726in}{0.914730in}}%
\pgfpathlineto{\pgfqpoint{2.139599in}{0.814492in}}%
\pgfpathlineto{\pgfqpoint{2.246472in}{0.728835in}}%
\pgfusepath{stroke}%
\end{pgfscope}%
\begin{pgfscope}%
\pgfpathrectangle{\pgfqpoint{0.536502in}{0.041670in}}{\pgfqpoint{2.671828in}{1.166660in}}%
\pgfusepath{clip}%
\pgfsetrectcap%
\pgfsetroundjoin%
\pgfsetlinewidth{1.003750pt}%
\definecolor{currentstroke}{rgb}{0.466667,0.466667,0.466667}%
\pgfsetstrokecolor{currentstroke}%
\pgfsetdash{}{0pt}%
\pgfpathmoveto{\pgfqpoint{0.643375in}{0.586680in}}%
\pgfpathlineto{\pgfqpoint{0.750249in}{0.612195in}}%
\pgfpathlineto{\pgfqpoint{0.857122in}{0.699675in}}%
\pgfpathlineto{\pgfqpoint{0.963995in}{0.768930in}}%
\pgfpathlineto{\pgfqpoint{1.070868in}{0.814492in}}%
\pgfpathlineto{\pgfqpoint{1.177741in}{0.905617in}}%
\pgfpathlineto{\pgfqpoint{1.284614in}{0.949357in}}%
\pgfpathlineto{\pgfqpoint{1.391487in}{0.973050in}}%
\pgfpathlineto{\pgfqpoint{1.498360in}{1.031370in}}%
\pgfpathlineto{\pgfqpoint{1.605233in}{1.084222in}}%
\pgfpathlineto{\pgfqpoint{1.712107in}{1.106092in}}%
\pgfpathlineto{\pgfqpoint{1.818980in}{1.107915in}}%
\pgfpathlineto{\pgfqpoint{1.925853in}{1.093335in}}%
\pgfpathlineto{\pgfqpoint{2.032726in}{1.087867in}}%
\pgfpathlineto{\pgfqpoint{2.139599in}{1.078755in}}%
\pgfpathlineto{\pgfqpoint{2.246472in}{1.065997in}}%
\pgfpathlineto{\pgfqpoint{2.353345in}{1.038660in}}%
\pgfpathlineto{\pgfqpoint{2.460218in}{0.983985in}}%
\pgfpathlineto{\pgfqpoint{2.567091in}{0.943890in}}%
\pgfpathlineto{\pgfqpoint{2.673964in}{0.940245in}}%
\pgfpathlineto{\pgfqpoint{2.780838in}{0.896505in}}%
\pgfpathlineto{\pgfqpoint{2.887711in}{0.889215in}}%
\pgfpathlineto{\pgfqpoint{2.994584in}{0.852765in}}%
\pgfpathlineto{\pgfqpoint{3.101457in}{0.816315in}}%
\pgfusepath{stroke}%
\end{pgfscope}%
\begin{pgfscope}%
\pgfpathrectangle{\pgfqpoint{0.536502in}{0.041670in}}{\pgfqpoint{2.671828in}{1.166660in}}%
\pgfusepath{clip}%
\pgfsetbuttcap%
\pgfsetroundjoin%
\pgfsetlinewidth{1.003750pt}%
\definecolor{currentstroke}{rgb}{0.000000,0.392157,0.000000}%
\pgfsetstrokecolor{currentstroke}%
\pgfsetdash{{3.700000pt}{1.600000pt}}{0.000000pt}%
\pgfpathmoveto{\pgfqpoint{0.536502in}{0.094700in}}%
\pgfpathlineto{\pgfqpoint{3.208330in}{0.094700in}}%
\pgfusepath{stroke}%
\end{pgfscope}%
\begin{pgfscope}%
\pgfsetrectcap%
\pgfsetmiterjoin%
\pgfsetlinewidth{1.003750pt}%
\definecolor{currentstroke}{rgb}{1.000000,1.000000,1.000000}%
\pgfsetstrokecolor{currentstroke}%
\pgfsetdash{}{0pt}%
\pgfpathmoveto{\pgfqpoint{0.536502in}{0.041670in}}%
\pgfpathlineto{\pgfqpoint{0.536502in}{1.208330in}}%
\pgfusepath{stroke}%
\end{pgfscope}%
\begin{pgfscope}%
\pgfsetrectcap%
\pgfsetmiterjoin%
\pgfsetlinewidth{1.003750pt}%
\definecolor{currentstroke}{rgb}{1.000000,1.000000,1.000000}%
\pgfsetstrokecolor{currentstroke}%
\pgfsetdash{}{0pt}%
\pgfpathmoveto{\pgfqpoint{3.208330in}{0.041670in}}%
\pgfpathlineto{\pgfqpoint{3.208330in}{1.208330in}}%
\pgfusepath{stroke}%
\end{pgfscope}%
\begin{pgfscope}%
\pgfsetrectcap%
\pgfsetmiterjoin%
\pgfsetlinewidth{1.003750pt}%
\definecolor{currentstroke}{rgb}{1.000000,1.000000,1.000000}%
\pgfsetstrokecolor{currentstroke}%
\pgfsetdash{}{0pt}%
\pgfpathmoveto{\pgfqpoint{0.536502in}{0.041670in}}%
\pgfpathlineto{\pgfqpoint{3.208330in}{0.041670in}}%
\pgfusepath{stroke}%
\end{pgfscope}%
\begin{pgfscope}%
\pgfsetrectcap%
\pgfsetmiterjoin%
\pgfsetlinewidth{1.003750pt}%
\definecolor{currentstroke}{rgb}{1.000000,1.000000,1.000000}%
\pgfsetstrokecolor{currentstroke}%
\pgfsetdash{}{0pt}%
\pgfpathmoveto{\pgfqpoint{0.536502in}{1.208330in}}%
\pgfpathlineto{\pgfqpoint{3.208330in}{1.208330in}}%
\pgfusepath{stroke}%
\end{pgfscope}%
\end{pgfpicture}%
\makeatother%
\endgroup%

%% file: figures/minorlocation-nox.pgf
\begingroup%
\makeatletter%
\begin{pgfpicture}%
\pgfpathrectangle{\pgfpointorigin}{\pgfqpoint{3.250000in}{1.250000in}}%
\pgfusepath{use as bounding box, clip}%
\begin{pgfscope}%
\pgfsetbuttcap%
\pgfsetmiterjoin%
\definecolor{currentfill}{rgb}{1.000000,1.000000,1.000000}%
\pgfsetfillcolor{currentfill}%
\pgfsetlinewidth{0.000000pt}%
\definecolor{currentstroke}{rgb}{0.500000,0.500000,0.500000}%
\pgfsetstrokecolor{currentstroke}%
\pgfsetdash{}{0pt}%
\pgfpathmoveto{\pgfqpoint{0.000000in}{0.000000in}}%
\pgfpathlineto{\pgfqpoint{3.250000in}{0.000000in}}%
\pgfpathlineto{\pgfqpoint{3.250000in}{1.250000in}}%
\pgfpathlineto{\pgfqpoint{0.000000in}{1.250000in}}%
\pgfpathlineto{\pgfqpoint{0.000000in}{0.000000in}}%
\pgfpathclose%
\pgfusepath{fill}%
\end{pgfscope}%
\begin{pgfscope}%
\pgfsetbuttcap%
\pgfsetmiterjoin%
\definecolor{currentfill}{rgb}{0.898039,0.898039,0.898039}%
\pgfsetfillcolor{currentfill}%
\pgfsetlinewidth{0.000000pt}%
\definecolor{currentstroke}{rgb}{0.000000,0.000000,0.000000}%
\pgfsetstrokecolor{currentstroke}%
\pgfsetstrokeopacity{0.000000}%
\pgfsetdash{}{0pt}%
\pgfpathmoveto{\pgfqpoint{0.536502in}{0.041670in}}%
\pgfpathlineto{\pgfqpoint{3.208330in}{0.041670in}}%
\pgfpathlineto{\pgfqpoint{3.208330in}{1.208330in}}%
\pgfpathlineto{\pgfqpoint{0.536502in}{1.208330in}}%
\pgfpathlineto{\pgfqpoint{0.536502in}{0.041670in}}%
\pgfpathclose%
\pgfusepath{fill}%
\end{pgfscope}%
\begin{pgfscope}%
\pgfpathrectangle{\pgfqpoint{0.536502in}{0.041670in}}{\pgfqpoint{2.671828in}{1.166660in}}%
\pgfusepath{clip}%
\pgfsetrectcap%
\pgfsetroundjoin%
\pgfsetlinewidth{0.803000pt}%
\definecolor{currentstroke}{rgb}{1.000000,1.000000,1.000000}%
\pgfsetstrokecolor{currentstroke}%
\pgfsetdash{}{0pt}%
\pgfpathmoveto{\pgfqpoint{0.643375in}{0.041670in}}%
\pgfpathlineto{\pgfqpoint{0.643375in}{1.208330in}}%
\pgfusepath{stroke}%
\end{pgfscope}%
\begin{pgfscope}%
\pgfpathrectangle{\pgfqpoint{0.536502in}{0.041670in}}{\pgfqpoint{2.671828in}{1.166660in}}%
\pgfusepath{clip}%
\pgfsetrectcap%
\pgfsetroundjoin%
\pgfsetlinewidth{0.803000pt}%
\definecolor{currentstroke}{rgb}{1.000000,1.000000,1.000000}%
\pgfsetstrokecolor{currentstroke}%
\pgfsetdash{}{0pt}%
\pgfpathmoveto{\pgfqpoint{3.101457in}{0.041670in}}%
\pgfpathlineto{\pgfqpoint{3.101457in}{1.208330in}}%
\pgfusepath{stroke}%
\end{pgfscope}%
\begin{pgfscope}%
\pgfpathrectangle{\pgfqpoint{0.536502in}{0.041670in}}{\pgfqpoint{2.671828in}{1.166660in}}%
\pgfusepath{clip}%
\pgfsetrectcap%
\pgfsetroundjoin%
\pgfsetlinewidth{0.803000pt}%
\definecolor{currentstroke}{rgb}{1.000000,1.000000,1.000000}%
\pgfsetstrokecolor{currentstroke}%
\pgfsetdash{}{0pt}%
\pgfpathmoveto{\pgfqpoint{0.536502in}{0.127630in}}%
\pgfpathlineto{\pgfqpoint{3.208330in}{0.127630in}}%
\pgfusepath{stroke}%
\end{pgfscope}%
\begin{pgfscope}%
\definecolor{textcolor}{rgb}{0.333333,0.333333,0.333333}%
\pgfsetstrokecolor{textcolor}%
\pgfsetfillcolor{textcolor}%
\pgftext[x=0.041670in, y=0.074868in, left, base]{\color{textcolor}\sffamily\fontsize{10.000000}{12.000000}\selectfont 0.350}%
\end{pgfscope}%
\begin{pgfscope}%
\pgfpathrectangle{\pgfqpoint{0.536502in}{0.041670in}}{\pgfqpoint{2.671828in}{1.166660in}}%
\pgfusepath{clip}%
\pgfsetrectcap%
\pgfsetroundjoin%
\pgfsetlinewidth{0.803000pt}%
\definecolor{currentstroke}{rgb}{1.000000,1.000000,1.000000}%
\pgfsetstrokecolor{currentstroke}%
\pgfsetdash{}{0pt}%
\pgfpathmoveto{\pgfqpoint{0.536502in}{0.560918in}}%
\pgfpathlineto{\pgfqpoint{3.208330in}{0.560918in}}%
\pgfusepath{stroke}%
\end{pgfscope}%
\begin{pgfscope}%
\definecolor{textcolor}{rgb}{0.333333,0.333333,0.333333}%
\pgfsetstrokecolor{textcolor}%
\pgfsetfillcolor{textcolor}%
\pgftext[x=0.041670in, y=0.508156in, left, base]{\color{textcolor}\sffamily\fontsize{10.000000}{12.000000}\selectfont 0.400}%
\end{pgfscope}%
\begin{pgfscope}%
\pgfpathrectangle{\pgfqpoint{0.536502in}{0.041670in}}{\pgfqpoint{2.671828in}{1.166660in}}%
\pgfusepath{clip}%
\pgfsetrectcap%
\pgfsetroundjoin%
\pgfsetlinewidth{0.803000pt}%
\definecolor{currentstroke}{rgb}{1.000000,1.000000,1.000000}%
\pgfsetstrokecolor{currentstroke}%
\pgfsetdash{}{0pt}%
\pgfpathmoveto{\pgfqpoint{0.536502in}{0.994206in}}%
\pgfpathlineto{\pgfqpoint{3.208330in}{0.994206in}}%
\pgfusepath{stroke}%
\end{pgfscope}%
\begin{pgfscope}%
\definecolor{textcolor}{rgb}{0.333333,0.333333,0.333333}%
\pgfsetstrokecolor{textcolor}%
\pgfsetfillcolor{textcolor}%
\pgftext[x=0.041670in, y=0.941444in, left, base]{\color{textcolor}\sffamily\fontsize{10.000000}{12.000000}\selectfont 0.450}%
\end{pgfscope}%
\begin{pgfscope}%
\pgfpathrectangle{\pgfqpoint{0.536502in}{0.041670in}}{\pgfqpoint{2.671828in}{1.166660in}}%
\pgfusepath{clip}%
\pgfsetrectcap%
\pgfsetroundjoin%
\pgfsetlinewidth{1.003750pt}%
\definecolor{currentstroke}{rgb}{0.886275,0.290196,0.200000}%
\pgfsetstrokecolor{currentstroke}%
\pgfsetdash{}{0pt}%
\pgfpathmoveto{\pgfqpoint{0.643375in}{0.094700in}}%
\pgfpathlineto{\pgfqpoint{0.750249in}{0.094700in}}%
\pgfpathlineto{\pgfqpoint{0.857122in}{0.099855in}}%
\pgfpathlineto{\pgfqpoint{0.963995in}{0.094700in}}%
\pgfpathlineto{\pgfqpoint{1.070868in}{0.094700in}}%
\pgfpathlineto{\pgfqpoint{1.177741in}{0.094700in}}%
\pgfpathlineto{\pgfqpoint{1.284614in}{0.094700in}}%
\pgfpathlineto{\pgfqpoint{1.391487in}{0.094700in}}%
\pgfpathlineto{\pgfqpoint{1.498360in}{0.094700in}}%
\pgfpathlineto{\pgfqpoint{1.605233in}{0.105410in}}%
\pgfpathlineto{\pgfqpoint{1.712107in}{0.188735in}}%
\pgfpathlineto{\pgfqpoint{1.818980in}{0.144295in}}%
\pgfusepath{stroke}%
\end{pgfscope}%
\begin{pgfscope}%
\pgfpathrectangle{\pgfqpoint{0.536502in}{0.041670in}}{\pgfqpoint{2.671828in}{1.166660in}}%
\pgfusepath{clip}%
\pgfsetrectcap%
\pgfsetroundjoin%
\pgfsetlinewidth{1.003750pt}%
\definecolor{currentstroke}{rgb}{0.203922,0.541176,0.741176}%
\pgfsetstrokecolor{currentstroke}%
\pgfsetdash{}{0pt}%
\pgfpathmoveto{\pgfqpoint{0.643375in}{0.116520in}}%
\pgfpathlineto{\pgfqpoint{0.750249in}{0.099855in}}%
\pgfpathlineto{\pgfqpoint{0.857122in}{0.094700in}}%
\pgfpathlineto{\pgfqpoint{0.963995in}{0.094700in}}%
\pgfpathlineto{\pgfqpoint{1.070868in}{0.116520in}}%
\pgfpathlineto{\pgfqpoint{1.177741in}{0.127630in}}%
\pgfpathlineto{\pgfqpoint{1.284614in}{0.172070in}}%
\pgfpathlineto{\pgfqpoint{1.391487in}{0.244284in}}%
\pgfpathlineto{\pgfqpoint{1.498360in}{0.294279in}}%
\pgfpathlineto{\pgfqpoint{1.605233in}{0.210954in}}%
\pgfpathlineto{\pgfqpoint{1.712107in}{0.277614in}}%
\pgfpathlineto{\pgfqpoint{1.818980in}{0.338719in}}%
\pgfusepath{stroke}%
\end{pgfscope}%
\begin{pgfscope}%
\pgfpathrectangle{\pgfqpoint{0.536502in}{0.041670in}}{\pgfqpoint{2.671828in}{1.166660in}}%
\pgfusepath{clip}%
\pgfsetrectcap%
\pgfsetroundjoin%
\pgfsetlinewidth{1.003750pt}%
\definecolor{currentstroke}{rgb}{0.596078,0.556863,0.835294}%
\pgfsetstrokecolor{currentstroke}%
\pgfsetdash{}{0pt}%
\pgfpathmoveto{\pgfqpoint{0.643375in}{0.094700in}}%
\pgfpathlineto{\pgfqpoint{0.750249in}{0.144295in}}%
\pgfpathlineto{\pgfqpoint{0.857122in}{0.155405in}}%
\pgfpathlineto{\pgfqpoint{0.963995in}{0.166515in}}%
\pgfpathlineto{\pgfqpoint{1.070868in}{0.455373in}}%
\pgfpathlineto{\pgfqpoint{1.177741in}{0.488703in}}%
\pgfpathlineto{\pgfqpoint{1.284614in}{0.627578in}}%
\pgfpathlineto{\pgfqpoint{1.391487in}{0.999761in}}%
\pgfpathlineto{\pgfqpoint{1.498360in}{1.155300in}}%
\pgfpathlineto{\pgfqpoint{1.605233in}{1.110860in}}%
\pgfpathlineto{\pgfqpoint{1.712107in}{1.149745in}}%
\pgfpathlineto{\pgfqpoint{1.818980in}{1.155300in}}%
\pgfpathlineto{\pgfqpoint{1.925853in}{1.055310in}}%
\pgfpathlineto{\pgfqpoint{2.032726in}{0.577583in}}%
\pgfpathlineto{\pgfqpoint{2.139599in}{0.372049in}}%
\pgfpathlineto{\pgfqpoint{2.246472in}{0.277614in}}%
\pgfusepath{stroke}%
\end{pgfscope}%
\begin{pgfscope}%
\pgfpathrectangle{\pgfqpoint{0.536502in}{0.041670in}}{\pgfqpoint{2.671828in}{1.166660in}}%
\pgfusepath{clip}%
\pgfsetrectcap%
\pgfsetroundjoin%
\pgfsetlinewidth{1.003750pt}%
\definecolor{currentstroke}{rgb}{0.466667,0.466667,0.466667}%
\pgfsetstrokecolor{currentstroke}%
\pgfsetdash{}{0pt}%
\pgfpathmoveto{\pgfqpoint{0.643375in}{0.144295in}}%
\pgfpathlineto{\pgfqpoint{0.750249in}{0.194290in}}%
\pgfpathlineto{\pgfqpoint{0.857122in}{0.244284in}}%
\pgfpathlineto{\pgfqpoint{0.963995in}{0.272059in}}%
\pgfpathlineto{\pgfqpoint{1.070868in}{0.199845in}}%
\pgfpathlineto{\pgfqpoint{1.177741in}{0.438708in}}%
\pgfpathlineto{\pgfqpoint{1.284614in}{0.599803in}}%
\pgfpathlineto{\pgfqpoint{1.391487in}{0.644242in}}%
\pgfpathlineto{\pgfqpoint{1.498360in}{0.766452in}}%
\pgfpathlineto{\pgfqpoint{1.605233in}{0.816447in}}%
\pgfpathlineto{\pgfqpoint{1.712107in}{0.960876in}}%
\pgfpathlineto{\pgfqpoint{1.818980in}{1.049756in}}%
\pgfpathlineto{\pgfqpoint{1.925853in}{1.021981in}}%
\pgfpathlineto{\pgfqpoint{2.032726in}{0.933101in}}%
\pgfpathlineto{\pgfqpoint{2.139599in}{0.933101in}}%
\pgfpathlineto{\pgfqpoint{2.246472in}{0.783117in}}%
\pgfpathlineto{\pgfqpoint{2.353345in}{0.694237in}}%
\pgfpathlineto{\pgfqpoint{2.460218in}{0.694237in}}%
\pgfpathlineto{\pgfqpoint{2.567091in}{0.622023in}}%
\pgfpathlineto{\pgfqpoint{2.673964in}{0.549808in}}%
\pgfpathlineto{\pgfqpoint{2.780838in}{0.377604in}}%
\pgfpathlineto{\pgfqpoint{2.887711in}{0.377604in}}%
\pgfpathlineto{\pgfqpoint{2.994584in}{0.544253in}}%
\pgfpathlineto{\pgfqpoint{3.101457in}{0.388714in}}%
\pgfusepath{stroke}%
\end{pgfscope}%
\begin{pgfscope}%
\pgfpathrectangle{\pgfqpoint{0.536502in}{0.041670in}}{\pgfqpoint{2.671828in}{1.166660in}}%
\pgfusepath{clip}%
\pgfsetbuttcap%
\pgfsetroundjoin%
\pgfsetlinewidth{1.003750pt}%
\definecolor{currentstroke}{rgb}{0.000000,0.392157,0.000000}%
\pgfsetstrokecolor{currentstroke}%
\pgfsetdash{{3.700000pt}{1.600000pt}}{0.000000pt}%
\pgfpathmoveto{\pgfqpoint{0.536502in}{0.094700in}}%
\pgfpathlineto{\pgfqpoint{3.208330in}{0.094700in}}%
\pgfusepath{stroke}%
\end{pgfscope}%
\begin{pgfscope}%
\pgfsetrectcap%
\pgfsetmiterjoin%
\pgfsetlinewidth{1.003750pt}%
\definecolor{currentstroke}{rgb}{1.000000,1.000000,1.000000}%
\pgfsetstrokecolor{currentstroke}%
\pgfsetdash{}{0pt}%
\pgfpathmoveto{\pgfqpoint{0.536502in}{0.041670in}}%
\pgfpathlineto{\pgfqpoint{0.536502in}{1.208330in}}%
\pgfusepath{stroke}%
\end{pgfscope}%
\begin{pgfscope}%
\pgfsetrectcap%
\pgfsetmiterjoin%
\pgfsetlinewidth{1.003750pt}%
\definecolor{currentstroke}{rgb}{1.000000,1.000000,1.000000}%
\pgfsetstrokecolor{currentstroke}%
\pgfsetdash{}{0pt}%
\pgfpathmoveto{\pgfqpoint{3.208330in}{0.041670in}}%
\pgfpathlineto{\pgfqpoint{3.208330in}{1.208330in}}%
\pgfusepath{stroke}%
\end{pgfscope}%
\begin{pgfscope}%
\pgfsetrectcap%
\pgfsetmiterjoin%
\pgfsetlinewidth{1.003750pt}%
\definecolor{currentstroke}{rgb}{1.000000,1.000000,1.000000}%
\pgfsetstrokecolor{currentstroke}%
\pgfsetdash{}{0pt}%
\pgfpathmoveto{\pgfqpoint{0.536502in}{0.041670in}}%
\pgfpathlineto{\pgfqpoint{3.208330in}{0.041670in}}%
\pgfusepath{stroke}%
\end{pgfscope}%
\begin{pgfscope}%
\pgfsetrectcap%
\pgfsetmiterjoin%
\pgfsetlinewidth{1.003750pt}%
\definecolor{currentstroke}{rgb}{1.000000,1.000000,1.000000}%
\pgfsetstrokecolor{currentstroke}%
\pgfsetdash{}{0pt}%
\pgfpathmoveto{\pgfqpoint{0.536502in}{1.208330in}}%
\pgfpathlineto{\pgfqpoint{3.208330in}{1.208330in}}%
\pgfusepath{stroke}%
\end{pgfscope}%
\end{pgfpicture}%
\makeatother%
\endgroup%

%% file: figures/handshape-nox.pgf
\begingroup%
\makeatletter%
\begin{pgfpicture}%
\pgfpathrectangle{\pgfpointorigin}{\pgfqpoint{3.250000in}{1.250000in}}%
\pgfusepath{use as bounding box, clip}%
\begin{pgfscope}%
\pgfsetbuttcap%
\pgfsetmiterjoin%
\definecolor{currentfill}{rgb}{1.000000,1.000000,1.000000}%
\pgfsetfillcolor{currentfill}%
\pgfsetlinewidth{0.000000pt}%
\definecolor{currentstroke}{rgb}{0.500000,0.500000,0.500000}%
\pgfsetstrokecolor{currentstroke}%
\pgfsetdash{}{0pt}%
\pgfpathmoveto{\pgfqpoint{0.000000in}{0.000000in}}%
\pgfpathlineto{\pgfqpoint{3.250000in}{0.000000in}}%
\pgfpathlineto{\pgfqpoint{3.250000in}{1.250000in}}%
\pgfpathlineto{\pgfqpoint{0.000000in}{1.250000in}}%
\pgfpathlineto{\pgfqpoint{0.000000in}{0.000000in}}%
\pgfpathclose%
\pgfusepath{fill}%
\end{pgfscope}%
\begin{pgfscope}%
\pgfsetbuttcap%
\pgfsetmiterjoin%
\definecolor{currentfill}{rgb}{0.898039,0.898039,0.898039}%
\pgfsetfillcolor{currentfill}%
\pgfsetlinewidth{0.000000pt}%
\definecolor{currentstroke}{rgb}{0.000000,0.000000,0.000000}%
\pgfsetstrokecolor{currentstroke}%
\pgfsetstrokeopacity{0.000000}%
\pgfsetdash{}{0pt}%
\pgfpathmoveto{\pgfqpoint{0.536502in}{0.041670in}}%
\pgfpathlineto{\pgfqpoint{3.208330in}{0.041670in}}%
\pgfpathlineto{\pgfqpoint{3.208330in}{1.208330in}}%
\pgfpathlineto{\pgfqpoint{0.536502in}{1.208330in}}%
\pgfpathlineto{\pgfqpoint{0.536502in}{0.041670in}}%
\pgfpathclose%
\pgfusepath{fill}%
\end{pgfscope}%
\begin{pgfscope}%
\pgfpathrectangle{\pgfqpoint{0.536502in}{0.041670in}}{\pgfqpoint{2.671828in}{1.166660in}}%
\pgfusepath{clip}%
\pgfsetrectcap%
\pgfsetroundjoin%
\pgfsetlinewidth{0.803000pt}%
\definecolor{currentstroke}{rgb}{1.000000,1.000000,1.000000}%
\pgfsetstrokecolor{currentstroke}%
\pgfsetdash{}{0pt}%
\pgfpathmoveto{\pgfqpoint{0.643375in}{0.041670in}}%
\pgfpathlineto{\pgfqpoint{0.643375in}{1.208330in}}%
\pgfusepath{stroke}%
\end{pgfscope}%
\begin{pgfscope}%
\pgfpathrectangle{\pgfqpoint{0.536502in}{0.041670in}}{\pgfqpoint{2.671828in}{1.166660in}}%
\pgfusepath{clip}%
\pgfsetrectcap%
\pgfsetroundjoin%
\pgfsetlinewidth{0.803000pt}%
\definecolor{currentstroke}{rgb}{1.000000,1.000000,1.000000}%
\pgfsetstrokecolor{currentstroke}%
\pgfsetdash{}{0pt}%
\pgfpathmoveto{\pgfqpoint{3.101457in}{0.041670in}}%
\pgfpathlineto{\pgfqpoint{3.101457in}{1.208330in}}%
\pgfusepath{stroke}%
\end{pgfscope}%
\begin{pgfscope}%
\pgfpathrectangle{\pgfqpoint{0.536502in}{0.041670in}}{\pgfqpoint{2.671828in}{1.166660in}}%
\pgfusepath{clip}%
\pgfsetrectcap%
\pgfsetroundjoin%
\pgfsetlinewidth{0.803000pt}%
\definecolor{currentstroke}{rgb}{1.000000,1.000000,1.000000}%
\pgfsetstrokecolor{currentstroke}%
\pgfsetdash{}{0pt}%
\pgfpathmoveto{\pgfqpoint{0.536502in}{0.163281in}}%
\pgfpathlineto{\pgfqpoint{3.208330in}{0.163281in}}%
\pgfusepath{stroke}%
\end{pgfscope}%
\begin{pgfscope}%
\definecolor{textcolor}{rgb}{0.333333,0.333333,0.333333}%
\pgfsetstrokecolor{textcolor}%
\pgfsetfillcolor{textcolor}%
\pgftext[x=0.041670in, y=0.110520in, left, base]{\color{textcolor}\sffamily\fontsize{10.000000}{12.000000}\selectfont 0.120}%
\end{pgfscope}%
\begin{pgfscope}%
\pgfpathrectangle{\pgfqpoint{0.536502in}{0.041670in}}{\pgfqpoint{2.671828in}{1.166660in}}%
\pgfusepath{clip}%
\pgfsetrectcap%
\pgfsetroundjoin%
\pgfsetlinewidth{0.803000pt}%
\definecolor{currentstroke}{rgb}{1.000000,1.000000,1.000000}%
\pgfsetstrokecolor{currentstroke}%
\pgfsetdash{}{0pt}%
\pgfpathmoveto{\pgfqpoint{0.536502in}{0.461460in}}%
\pgfpathlineto{\pgfqpoint{3.208330in}{0.461460in}}%
\pgfusepath{stroke}%
\end{pgfscope}%
\begin{pgfscope}%
\definecolor{textcolor}{rgb}{0.333333,0.333333,0.333333}%
\pgfsetstrokecolor{textcolor}%
\pgfsetfillcolor{textcolor}%
\pgftext[x=0.041670in, y=0.408699in, left, base]{\color{textcolor}\sffamily\fontsize{10.000000}{12.000000}\selectfont 0.140}%
\end{pgfscope}%
\begin{pgfscope}%
\pgfpathrectangle{\pgfqpoint{0.536502in}{0.041670in}}{\pgfqpoint{2.671828in}{1.166660in}}%
\pgfusepath{clip}%
\pgfsetrectcap%
\pgfsetroundjoin%
\pgfsetlinewidth{0.803000pt}%
\definecolor{currentstroke}{rgb}{1.000000,1.000000,1.000000}%
\pgfsetstrokecolor{currentstroke}%
\pgfsetdash{}{0pt}%
\pgfpathmoveto{\pgfqpoint{0.536502in}{0.759639in}}%
\pgfpathlineto{\pgfqpoint{3.208330in}{0.759639in}}%
\pgfusepath{stroke}%
\end{pgfscope}%
\begin{pgfscope}%
\definecolor{textcolor}{rgb}{0.333333,0.333333,0.333333}%
\pgfsetstrokecolor{textcolor}%
\pgfsetfillcolor{textcolor}%
\pgftext[x=0.041670in, y=0.706878in, left, base]{\color{textcolor}\sffamily\fontsize{10.000000}{12.000000}\selectfont 0.160}%
\end{pgfscope}%
\begin{pgfscope}%
\pgfpathrectangle{\pgfqpoint{0.536502in}{0.041670in}}{\pgfqpoint{2.671828in}{1.166660in}}%
\pgfusepath{clip}%
\pgfsetrectcap%
\pgfsetroundjoin%
\pgfsetlinewidth{0.803000pt}%
\definecolor{currentstroke}{rgb}{1.000000,1.000000,1.000000}%
\pgfsetstrokecolor{currentstroke}%
\pgfsetdash{}{0pt}%
\pgfpathmoveto{\pgfqpoint{0.536502in}{1.057818in}}%
\pgfpathlineto{\pgfqpoint{3.208330in}{1.057818in}}%
\pgfusepath{stroke}%
\end{pgfscope}%
\begin{pgfscope}%
\definecolor{textcolor}{rgb}{0.333333,0.333333,0.333333}%
\pgfsetstrokecolor{textcolor}%
\pgfsetfillcolor{textcolor}%
\pgftext[x=0.041670in, y=1.005057in, left, base]{\color{textcolor}\sffamily\fontsize{10.000000}{12.000000}\selectfont 0.180}%
\end{pgfscope}%
\begin{pgfscope}%
\pgfpathrectangle{\pgfqpoint{0.536502in}{0.041670in}}{\pgfqpoint{2.671828in}{1.166660in}}%
\pgfusepath{clip}%
\pgfsetrectcap%
\pgfsetroundjoin%
\pgfsetlinewidth{1.003750pt}%
\definecolor{currentstroke}{rgb}{0.886275,0.290196,0.200000}%
\pgfsetstrokecolor{currentstroke}%
\pgfsetdash{}{0pt}%
\pgfpathmoveto{\pgfqpoint{0.643375in}{0.094700in}}%
\pgfpathlineto{\pgfqpoint{0.750249in}{0.094700in}}%
\pgfpathlineto{\pgfqpoint{0.857122in}{0.094700in}}%
\pgfpathlineto{\pgfqpoint{0.963995in}{0.094700in}}%
\pgfpathlineto{\pgfqpoint{1.070868in}{0.094700in}}%
\pgfpathlineto{\pgfqpoint{1.177741in}{0.094700in}}%
\pgfpathlineto{\pgfqpoint{1.284614in}{0.094700in}}%
\pgfpathlineto{\pgfqpoint{1.391487in}{0.094700in}}%
\pgfpathlineto{\pgfqpoint{1.498360in}{0.094700in}}%
\pgfpathlineto{\pgfqpoint{1.605233in}{0.094700in}}%
\pgfpathlineto{\pgfqpoint{1.712107in}{0.094700in}}%
\pgfpathlineto{\pgfqpoint{1.818980in}{0.094700in}}%
\pgfusepath{stroke}%
\end{pgfscope}%
\begin{pgfscope}%
\pgfpathrectangle{\pgfqpoint{0.536502in}{0.041670in}}{\pgfqpoint{2.671828in}{1.166660in}}%
\pgfusepath{clip}%
\pgfsetrectcap%
\pgfsetroundjoin%
\pgfsetlinewidth{1.003750pt}%
\definecolor{currentstroke}{rgb}{0.203922,0.541176,0.741176}%
\pgfsetstrokecolor{currentstroke}%
\pgfsetdash{}{0pt}%
\pgfpathmoveto{\pgfqpoint{0.643375in}{0.094700in}}%
\pgfpathlineto{\pgfqpoint{0.750249in}{0.247383in}}%
\pgfpathlineto{\pgfqpoint{0.857122in}{0.151813in}}%
\pgfpathlineto{\pgfqpoint{0.963995in}{0.237826in}}%
\pgfpathlineto{\pgfqpoint{1.070868in}{0.094700in}}%
\pgfpathlineto{\pgfqpoint{1.177741in}{0.161370in}}%
\pgfpathlineto{\pgfqpoint{1.284614in}{0.132699in}}%
\pgfpathlineto{\pgfqpoint{1.391487in}{0.218712in}}%
\pgfpathlineto{\pgfqpoint{1.498360in}{0.276054in}}%
\pgfpathlineto{\pgfqpoint{1.605233in}{0.514980in}}%
\pgfpathlineto{\pgfqpoint{1.712107in}{0.581879in}}%
\pgfpathlineto{\pgfqpoint{1.818980in}{0.428966in}}%
\pgfusepath{stroke}%
\end{pgfscope}%
\begin{pgfscope}%
\pgfpathrectangle{\pgfqpoint{0.536502in}{0.041670in}}{\pgfqpoint{2.671828in}{1.166660in}}%
\pgfusepath{clip}%
\pgfsetrectcap%
\pgfsetroundjoin%
\pgfsetlinewidth{1.003750pt}%
\definecolor{currentstroke}{rgb}{0.596078,0.556863,0.835294}%
\pgfsetstrokecolor{currentstroke}%
\pgfsetdash{}{0pt}%
\pgfpathmoveto{\pgfqpoint{0.643375in}{0.237826in}}%
\pgfpathlineto{\pgfqpoint{0.750249in}{0.381181in}}%
\pgfpathlineto{\pgfqpoint{0.857122in}{0.352510in}}%
\pgfpathlineto{\pgfqpoint{0.963995in}{0.467194in}}%
\pgfpathlineto{\pgfqpoint{1.070868in}{0.600993in}}%
\pgfpathlineto{\pgfqpoint{1.177741in}{0.753905in}}%
\pgfpathlineto{\pgfqpoint{1.284614in}{0.897260in}}%
\pgfpathlineto{\pgfqpoint{1.391487in}{1.031059in}}%
\pgfpathlineto{\pgfqpoint{1.498360in}{1.059730in}}%
\pgfpathlineto{\pgfqpoint{1.605233in}{1.155300in}}%
\pgfpathlineto{\pgfqpoint{1.712107in}{1.088401in}}%
\pgfpathlineto{\pgfqpoint{1.818980in}{1.040616in}}%
\pgfpathlineto{\pgfqpoint{1.925853in}{1.145743in}}%
\pgfpathlineto{\pgfqpoint{2.032726in}{0.620107in}}%
\pgfpathlineto{\pgfqpoint{2.139599in}{0.457637in}}%
\pgfpathlineto{\pgfqpoint{2.246472in}{0.457637in}}%
\pgfusepath{stroke}%
\end{pgfscope}%
\begin{pgfscope}%
\pgfpathrectangle{\pgfqpoint{0.536502in}{0.041670in}}{\pgfqpoint{2.671828in}{1.166660in}}%
\pgfusepath{clip}%
\pgfsetrectcap%
\pgfsetroundjoin%
\pgfsetlinewidth{1.003750pt}%
\definecolor{currentstroke}{rgb}{0.466667,0.466667,0.466667}%
\pgfsetstrokecolor{currentstroke}%
\pgfsetdash{}{0pt}%
\pgfpathmoveto{\pgfqpoint{0.643375in}{0.419409in}}%
\pgfpathlineto{\pgfqpoint{0.750249in}{0.276054in}}%
\pgfpathlineto{\pgfqpoint{0.857122in}{0.314282in}}%
\pgfpathlineto{\pgfqpoint{0.963995in}{0.352510in}}%
\pgfpathlineto{\pgfqpoint{1.070868in}{0.342953in}}%
\pgfpathlineto{\pgfqpoint{1.177741in}{0.362067in}}%
\pgfpathlineto{\pgfqpoint{1.284614in}{0.534094in}}%
\pgfpathlineto{\pgfqpoint{1.391487in}{0.620107in}}%
\pgfpathlineto{\pgfqpoint{1.498360in}{0.897260in}}%
\pgfpathlineto{\pgfqpoint{1.605233in}{1.021502in}}%
\pgfpathlineto{\pgfqpoint{1.712107in}{1.097958in}}%
\pgfpathlineto{\pgfqpoint{1.818980in}{0.964160in}}%
\pgfpathlineto{\pgfqpoint{1.925853in}{1.097958in}}%
\pgfpathlineto{\pgfqpoint{2.032726in}{1.021502in}}%
\pgfpathlineto{\pgfqpoint{2.139599in}{0.916374in}}%
\pgfpathlineto{\pgfqpoint{2.246472in}{0.897260in}}%
\pgfpathlineto{\pgfqpoint{2.353345in}{0.696563in}}%
\pgfpathlineto{\pgfqpoint{2.460218in}{0.534094in}}%
\pgfpathlineto{\pgfqpoint{2.567091in}{0.419409in}}%
\pgfpathlineto{\pgfqpoint{2.673964in}{0.409852in}}%
\pgfpathlineto{\pgfqpoint{2.780838in}{0.352510in}}%
\pgfpathlineto{\pgfqpoint{2.887711in}{0.256940in}}%
\pgfpathlineto{\pgfqpoint{2.994584in}{0.419409in}}%
\pgfpathlineto{\pgfqpoint{3.101457in}{0.400295in}}%
\pgfusepath{stroke}%
\end{pgfscope}%
\begin{pgfscope}%
\pgfpathrectangle{\pgfqpoint{0.536502in}{0.041670in}}{\pgfqpoint{2.671828in}{1.166660in}}%
\pgfusepath{clip}%
\pgfsetbuttcap%
\pgfsetroundjoin%
\pgfsetlinewidth{1.003750pt}%
\definecolor{currentstroke}{rgb}{0.000000,0.392157,0.000000}%
\pgfsetstrokecolor{currentstroke}%
\pgfsetdash{{3.700000pt}{1.600000pt}}{0.000000pt}%
\pgfpathmoveto{\pgfqpoint{0.536502in}{0.094700in}}%
\pgfpathlineto{\pgfqpoint{3.208330in}{0.094700in}}%
\pgfusepath{stroke}%
\end{pgfscope}%
\begin{pgfscope}%
\pgfsetrectcap%
\pgfsetmiterjoin%
\pgfsetlinewidth{1.003750pt}%
\definecolor{currentstroke}{rgb}{1.000000,1.000000,1.000000}%
\pgfsetstrokecolor{currentstroke}%
\pgfsetdash{}{0pt}%
\pgfpathmoveto{\pgfqpoint{0.536502in}{0.041670in}}%
\pgfpathlineto{\pgfqpoint{0.536502in}{1.208330in}}%
\pgfusepath{stroke}%
\end{pgfscope}%
\begin{pgfscope}%
\pgfsetrectcap%
\pgfsetmiterjoin%
\pgfsetlinewidth{1.003750pt}%
\definecolor{currentstroke}{rgb}{1.000000,1.000000,1.000000}%
\pgfsetstrokecolor{currentstroke}%
\pgfsetdash{}{0pt}%
\pgfpathmoveto{\pgfqpoint{3.208330in}{0.041670in}}%
\pgfpathlineto{\pgfqpoint{3.208330in}{1.208330in}}%
\pgfusepath{stroke}%
\end{pgfscope}%
\begin{pgfscope}%
\pgfsetrectcap%
\pgfsetmiterjoin%
\pgfsetlinewidth{1.003750pt}%
\definecolor{currentstroke}{rgb}{1.000000,1.000000,1.000000}%
\pgfsetstrokecolor{currentstroke}%
\pgfsetdash{}{0pt}%
\pgfpathmoveto{\pgfqpoint{0.536502in}{0.041670in}}%
\pgfpathlineto{\pgfqpoint{3.208330in}{0.041670in}}%
\pgfusepath{stroke}%
\end{pgfscope}%
\begin{pgfscope}%
\pgfsetrectcap%
\pgfsetmiterjoin%
\pgfsetlinewidth{1.003750pt}%
\definecolor{currentstroke}{rgb}{1.000000,1.000000,1.000000}%
\pgfsetstrokecolor{currentstroke}%
\pgfsetdash{}{0pt}%
\pgfpathmoveto{\pgfqpoint{0.536502in}{1.208330in}}%
\pgfpathlineto{\pgfqpoint{3.208330in}{1.208330in}}%
\pgfusepath{stroke}%
\end{pgfscope}%
\end{pgfpicture}%
\makeatother%
\endgroup%

%% file: figures/selectedfinger-nox.pgf
\begingroup%
\makeatletter%
\begin{pgfpicture}%
\pgfpathrectangle{\pgfpointorigin}{\pgfqpoint{3.250000in}{1.250000in}}%
\pgfusepath{use as bounding box, clip}%
\begin{pgfscope}%
\pgfsetbuttcap%
\pgfsetmiterjoin%
\definecolor{currentfill}{rgb}{1.000000,1.000000,1.000000}%
\pgfsetfillcolor{currentfill}%
\pgfsetlinewidth{0.000000pt}%
\definecolor{currentstroke}{rgb}{0.500000,0.500000,0.500000}%
\pgfsetstrokecolor{currentstroke}%
\pgfsetdash{}{0pt}%
\pgfpathmoveto{\pgfqpoint{0.000000in}{0.000000in}}%
\pgfpathlineto{\pgfqpoint{3.250000in}{0.000000in}}%
\pgfpathlineto{\pgfqpoint{3.250000in}{1.250000in}}%
\pgfpathlineto{\pgfqpoint{0.000000in}{1.250000in}}%
\pgfpathlineto{\pgfqpoint{0.000000in}{0.000000in}}%
\pgfpathclose%
\pgfusepath{fill}%
\end{pgfscope}%
\begin{pgfscope}%
\pgfsetbuttcap%
\pgfsetmiterjoin%
\definecolor{currentfill}{rgb}{0.898039,0.898039,0.898039}%
\pgfsetfillcolor{currentfill}%
\pgfsetlinewidth{0.000000pt}%
\definecolor{currentstroke}{rgb}{0.000000,0.000000,0.000000}%
\pgfsetstrokecolor{currentstroke}%
\pgfsetstrokeopacity{0.000000}%
\pgfsetdash{}{0pt}%
\pgfpathmoveto{\pgfqpoint{0.536502in}{0.076719in}}%
\pgfpathlineto{\pgfqpoint{3.208330in}{0.076719in}}%
\pgfpathlineto{\pgfqpoint{3.208330in}{1.208330in}}%
\pgfpathlineto{\pgfqpoint{0.536502in}{1.208330in}}%
\pgfpathlineto{\pgfqpoint{0.536502in}{0.076719in}}%
\pgfpathclose%
\pgfusepath{fill}%
\end{pgfscope}%
\begin{pgfscope}%
\pgfpathrectangle{\pgfqpoint{0.536502in}{0.076719in}}{\pgfqpoint{2.671828in}{1.131611in}}%
\pgfusepath{clip}%
\pgfsetrectcap%
\pgfsetroundjoin%
\pgfsetlinewidth{0.803000pt}%
\definecolor{currentstroke}{rgb}{1.000000,1.000000,1.000000}%
\pgfsetstrokecolor{currentstroke}%
\pgfsetdash{}{0pt}%
\pgfpathmoveto{\pgfqpoint{0.643375in}{0.076719in}}%
\pgfpathlineto{\pgfqpoint{0.643375in}{1.208330in}}%
\pgfusepath{stroke}%
\end{pgfscope}%
\begin{pgfscope}%
\pgfpathrectangle{\pgfqpoint{0.536502in}{0.076719in}}{\pgfqpoint{2.671828in}{1.131611in}}%
\pgfusepath{clip}%
\pgfsetrectcap%
\pgfsetroundjoin%
\pgfsetlinewidth{0.803000pt}%
\definecolor{currentstroke}{rgb}{1.000000,1.000000,1.000000}%
\pgfsetstrokecolor{currentstroke}%
\pgfsetdash{}{0pt}%
\pgfpathmoveto{\pgfqpoint{3.101457in}{0.076719in}}%
\pgfpathlineto{\pgfqpoint{3.101457in}{1.208330in}}%
\pgfusepath{stroke}%
\end{pgfscope}%
\begin{pgfscope}%
\pgfpathrectangle{\pgfqpoint{0.536502in}{0.076719in}}{\pgfqpoint{2.671828in}{1.131611in}}%
\pgfusepath{clip}%
\pgfsetrectcap%
\pgfsetroundjoin%
\pgfsetlinewidth{0.803000pt}%
\definecolor{currentstroke}{rgb}{1.000000,1.000000,1.000000}%
\pgfsetstrokecolor{currentstroke}%
\pgfsetdash{}{0pt}%
\pgfpathmoveto{\pgfqpoint{0.536502in}{0.123600in}}%
\pgfpathlineto{\pgfqpoint{3.208330in}{0.123600in}}%
\pgfusepath{stroke}%
\end{pgfscope}%
\begin{pgfscope}%
\definecolor{textcolor}{rgb}{0.333333,0.333333,0.333333}%
\pgfsetstrokecolor{textcolor}%
\pgfsetfillcolor{textcolor}%
\pgftext[x=0.041670in, y=0.070838in, left, base]{\color{textcolor}\sffamily\fontsize{10.000000}{12.000000}\selectfont 0.480}%
\end{pgfscope}%
\begin{pgfscope}%
\pgfpathrectangle{\pgfqpoint{0.536502in}{0.076719in}}{\pgfqpoint{2.671828in}{1.131611in}}%
\pgfusepath{clip}%
\pgfsetrectcap%
\pgfsetroundjoin%
\pgfsetlinewidth{0.803000pt}%
\definecolor{currentstroke}{rgb}{1.000000,1.000000,1.000000}%
\pgfsetstrokecolor{currentstroke}%
\pgfsetdash{}{0pt}%
\pgfpathmoveto{\pgfqpoint{0.536502in}{0.535095in}}%
\pgfpathlineto{\pgfqpoint{3.208330in}{0.535095in}}%
\pgfusepath{stroke}%
\end{pgfscope}%
\begin{pgfscope}%
\definecolor{textcolor}{rgb}{0.333333,0.333333,0.333333}%
\pgfsetstrokecolor{textcolor}%
\pgfsetfillcolor{textcolor}%
\pgftext[x=0.041670in, y=0.482334in, left, base]{\color{textcolor}\sffamily\fontsize{10.000000}{12.000000}\selectfont 0.500}%
\end{pgfscope}%
\begin{pgfscope}%
\pgfpathrectangle{\pgfqpoint{0.536502in}{0.076719in}}{\pgfqpoint{2.671828in}{1.131611in}}%
\pgfusepath{clip}%
\pgfsetrectcap%
\pgfsetroundjoin%
\pgfsetlinewidth{0.803000pt}%
\definecolor{currentstroke}{rgb}{1.000000,1.000000,1.000000}%
\pgfsetstrokecolor{currentstroke}%
\pgfsetdash{}{0pt}%
\pgfpathmoveto{\pgfqpoint{0.536502in}{0.946590in}}%
\pgfpathlineto{\pgfqpoint{3.208330in}{0.946590in}}%
\pgfusepath{stroke}%
\end{pgfscope}%
\begin{pgfscope}%
\definecolor{textcolor}{rgb}{0.333333,0.333333,0.333333}%
\pgfsetstrokecolor{textcolor}%
\pgfsetfillcolor{textcolor}%
\pgftext[x=0.041670in, y=0.893829in, left, base]{\color{textcolor}\sffamily\fontsize{10.000000}{12.000000}\selectfont 0.520}%
\end{pgfscope}%
\begin{pgfscope}%
\pgfpathrectangle{\pgfqpoint{0.536502in}{0.076719in}}{\pgfqpoint{2.671828in}{1.131611in}}%
\pgfusepath{clip}%
\pgfsetrectcap%
\pgfsetroundjoin%
\pgfsetlinewidth{1.003750pt}%
\definecolor{currentstroke}{rgb}{0.886275,0.290196,0.200000}%
\pgfsetstrokecolor{currentstroke}%
\pgfsetdash{}{0pt}%
\pgfpathmoveto{\pgfqpoint{0.643375in}{0.191497in}}%
\pgfpathlineto{\pgfqpoint{0.750249in}{0.191497in}}%
\pgfpathlineto{\pgfqpoint{0.857122in}{0.191497in}}%
\pgfpathlineto{\pgfqpoint{0.963995in}{0.191497in}}%
\pgfpathlineto{\pgfqpoint{1.070868in}{0.191497in}}%
\pgfpathlineto{\pgfqpoint{1.177741in}{0.191497in}}%
\pgfpathlineto{\pgfqpoint{1.284614in}{0.191497in}}%
\pgfpathlineto{\pgfqpoint{1.391487in}{0.244938in}}%
\pgfpathlineto{\pgfqpoint{1.498360in}{0.191497in}}%
\pgfpathlineto{\pgfqpoint{1.605233in}{0.191497in}}%
\pgfpathlineto{\pgfqpoint{1.712107in}{0.191497in}}%
\pgfpathlineto{\pgfqpoint{1.818980in}{0.218560in}}%
\pgfusepath{stroke}%
\end{pgfscope}%
\begin{pgfscope}%
\pgfpathrectangle{\pgfqpoint{0.536502in}{0.076719in}}{\pgfqpoint{2.671828in}{1.131611in}}%
\pgfusepath{clip}%
\pgfsetrectcap%
\pgfsetroundjoin%
\pgfsetlinewidth{1.003750pt}%
\definecolor{currentstroke}{rgb}{0.203922,0.541176,0.741176}%
\pgfsetstrokecolor{currentstroke}%
\pgfsetdash{}{0pt}%
\pgfpathmoveto{\pgfqpoint{0.643375in}{0.191497in}}%
\pgfpathlineto{\pgfqpoint{0.750249in}{0.191497in}}%
\pgfpathlineto{\pgfqpoint{0.857122in}{0.191497in}}%
\pgfpathlineto{\pgfqpoint{0.963995in}{0.191497in}}%
\pgfpathlineto{\pgfqpoint{1.070868in}{0.191497in}}%
\pgfpathlineto{\pgfqpoint{1.177741in}{0.191497in}}%
\pgfpathlineto{\pgfqpoint{1.284614in}{0.191497in}}%
\pgfpathlineto{\pgfqpoint{1.391487in}{0.191497in}}%
\pgfpathlineto{\pgfqpoint{1.498360in}{0.191497in}}%
\pgfpathlineto{\pgfqpoint{1.605233in}{0.310883in}}%
\pgfpathlineto{\pgfqpoint{1.712107in}{0.429583in}}%
\pgfpathlineto{\pgfqpoint{1.818980in}{0.350450in}}%
\pgfusepath{stroke}%
\end{pgfscope}%
\begin{pgfscope}%
\pgfpathrectangle{\pgfqpoint{0.536502in}{0.076719in}}{\pgfqpoint{2.671828in}{1.131611in}}%
\pgfusepath{clip}%
\pgfsetrectcap%
\pgfsetroundjoin%
\pgfsetlinewidth{1.003750pt}%
\definecolor{currentstroke}{rgb}{0.596078,0.556863,0.835294}%
\pgfsetstrokecolor{currentstroke}%
\pgfsetdash{}{0pt}%
\pgfpathmoveto{\pgfqpoint{0.643375in}{0.192182in}}%
\pgfpathlineto{\pgfqpoint{0.750249in}{0.258127in}}%
\pgfpathlineto{\pgfqpoint{0.857122in}{0.205371in}}%
\pgfpathlineto{\pgfqpoint{0.963995in}{0.191497in}}%
\pgfpathlineto{\pgfqpoint{1.070868in}{0.350450in}}%
\pgfpathlineto{\pgfqpoint{1.177741in}{0.191497in}}%
\pgfpathlineto{\pgfqpoint{1.284614in}{0.191497in}}%
\pgfpathlineto{\pgfqpoint{1.391487in}{0.337261in}}%
\pgfpathlineto{\pgfqpoint{1.498360in}{0.350450in}}%
\pgfpathlineto{\pgfqpoint{1.605233in}{0.363639in}}%
\pgfpathlineto{\pgfqpoint{1.712107in}{0.244938in}}%
\pgfpathlineto{\pgfqpoint{1.818980in}{0.271316in}}%
\pgfpathlineto{\pgfqpoint{1.925853in}{0.244938in}}%
\pgfpathlineto{\pgfqpoint{2.032726in}{0.324072in}}%
\pgfpathlineto{\pgfqpoint{2.139599in}{0.191497in}}%
\pgfpathlineto{\pgfqpoint{2.246472in}{0.191497in}}%
\pgfusepath{stroke}%
\end{pgfscope}%
\begin{pgfscope}%
\pgfpathrectangle{\pgfqpoint{0.536502in}{0.076719in}}{\pgfqpoint{2.671828in}{1.131611in}}%
\pgfusepath{clip}%
\pgfsetrectcap%
\pgfsetroundjoin%
\pgfsetlinewidth{1.003750pt}%
\definecolor{currentstroke}{rgb}{0.466667,0.466667,0.466667}%
\pgfsetstrokecolor{currentstroke}%
\pgfsetdash{}{0pt}%
\pgfpathmoveto{\pgfqpoint{0.643375in}{0.192182in}}%
\pgfpathlineto{\pgfqpoint{0.750249in}{0.191497in}}%
\pgfpathlineto{\pgfqpoint{0.857122in}{0.205371in}}%
\pgfpathlineto{\pgfqpoint{0.963995in}{0.284505in}}%
\pgfpathlineto{\pgfqpoint{1.070868in}{0.192182in}}%
\pgfpathlineto{\pgfqpoint{1.177741in}{0.324072in}}%
\pgfpathlineto{\pgfqpoint{1.284614in}{0.350450in}}%
\pgfpathlineto{\pgfqpoint{1.391487in}{0.191497in}}%
\pgfpathlineto{\pgfqpoint{1.498360in}{0.191497in}}%
\pgfpathlineto{\pgfqpoint{1.605233in}{0.191497in}}%
\pgfpathlineto{\pgfqpoint{1.712107in}{0.191497in}}%
\pgfpathlineto{\pgfqpoint{1.818980in}{0.191497in}}%
\pgfpathlineto{\pgfqpoint{1.925853in}{0.191497in}}%
\pgfpathlineto{\pgfqpoint{2.032726in}{0.191497in}}%
\pgfpathlineto{\pgfqpoint{2.139599in}{0.191497in}}%
\pgfpathlineto{\pgfqpoint{2.246472in}{0.191497in}}%
\pgfpathlineto{\pgfqpoint{2.353345in}{0.191497in}}%
\pgfpathlineto{\pgfqpoint{2.460218in}{0.191497in}}%
\pgfpathlineto{\pgfqpoint{2.567091in}{0.191497in}}%
\pgfpathlineto{\pgfqpoint{2.673964in}{0.191497in}}%
\pgfpathlineto{\pgfqpoint{2.780838in}{0.191497in}}%
\pgfpathlineto{\pgfqpoint{2.887711in}{0.191497in}}%
\pgfpathlineto{\pgfqpoint{2.994584in}{0.191497in}}%
\pgfpathlineto{\pgfqpoint{3.101457in}{0.191497in}}%
\pgfusepath{stroke}%
\end{pgfscope}%
\begin{pgfscope}%
\pgfpathrectangle{\pgfqpoint{0.536502in}{0.076719in}}{\pgfqpoint{2.671828in}{1.131611in}}%
\pgfusepath{clip}%
\pgfsetbuttcap%
\pgfsetroundjoin%
\pgfsetlinewidth{1.003750pt}%
\definecolor{currentstroke}{rgb}{0.000000,0.392157,0.000000}%
\pgfsetstrokecolor{currentstroke}%
\pgfsetdash{{3.700000pt}{1.600000pt}}{0.000000pt}%
\pgfpathmoveto{\pgfqpoint{0.536502in}{0.191497in}}%
\pgfpathlineto{\pgfqpoint{3.208330in}{0.191497in}}%
\pgfusepath{stroke}%
\end{pgfscope}%
\begin{pgfscope}%
\pgfsetrectcap%
\pgfsetmiterjoin%
\pgfsetlinewidth{1.003750pt}%
\definecolor{currentstroke}{rgb}{1.000000,1.000000,1.000000}%
\pgfsetstrokecolor{currentstroke}%
\pgfsetdash{}{0pt}%
\pgfpathmoveto{\pgfqpoint{0.536502in}{0.076719in}}%
\pgfpathlineto{\pgfqpoint{0.536502in}{1.208330in}}%
\pgfusepath{stroke}%
\end{pgfscope}%
\begin{pgfscope}%
\pgfsetrectcap%
\pgfsetmiterjoin%
\pgfsetlinewidth{1.003750pt}%
\definecolor{currentstroke}{rgb}{1.000000,1.000000,1.000000}%
\pgfsetstrokecolor{currentstroke}%
\pgfsetdash{}{0pt}%
\pgfpathmoveto{\pgfqpoint{3.208330in}{0.076719in}}%
\pgfpathlineto{\pgfqpoint{3.208330in}{1.208330in}}%
\pgfusepath{stroke}%
\end{pgfscope}%
\begin{pgfscope}%
\pgfsetrectcap%
\pgfsetmiterjoin%
\pgfsetlinewidth{1.003750pt}%
\definecolor{currentstroke}{rgb}{1.000000,1.000000,1.000000}%
\pgfsetstrokecolor{currentstroke}%
\pgfsetdash{}{0pt}%
\pgfpathmoveto{\pgfqpoint{0.536502in}{0.076719in}}%
\pgfpathlineto{\pgfqpoint{3.208330in}{0.076719in}}%
\pgfusepath{stroke}%
\end{pgfscope}%
\begin{pgfscope}%
\pgfsetrectcap%
\pgfsetmiterjoin%
\pgfsetlinewidth{1.003750pt}%
\definecolor{currentstroke}{rgb}{1.000000,1.000000,1.000000}%
\pgfsetstrokecolor{currentstroke}%
\pgfsetdash{}{0pt}%
\pgfpathmoveto{\pgfqpoint{0.536502in}{1.208330in}}%
\pgfpathlineto{\pgfqpoint{3.208330in}{1.208330in}}%
\pgfusepath{stroke}%
\end{pgfscope}%
\end{pgfpicture}%
\makeatother%
\endgroup%

%% file: figures/pathmovement.pgf
\begingroup%
\makeatletter%
\begin{pgfpicture}%
\pgfpathrectangle{\pgfpointorigin}{\pgfqpoint{3.250000in}{1.450000in}}%
\pgfusepath{use as bounding box, clip}%
\begin{pgfscope}%
\pgfsetbuttcap%
\pgfsetmiterjoin%
\definecolor{currentfill}{rgb}{1.000000,1.000000,1.000000}%
\pgfsetfillcolor{currentfill}%
\pgfsetlinewidth{0.000000pt}%
\definecolor{currentstroke}{rgb}{0.500000,0.500000,0.500000}%
\pgfsetstrokecolor{currentstroke}%
\pgfsetdash{}{0pt}%
\pgfpathmoveto{\pgfqpoint{0.000000in}{0.000000in}}%
\pgfpathlineto{\pgfqpoint{3.250000in}{0.000000in}}%
\pgfpathlineto{\pgfqpoint{3.250000in}{1.450000in}}%
\pgfpathlineto{\pgfqpoint{0.000000in}{1.450000in}}%
\pgfpathlineto{\pgfqpoint{0.000000in}{0.000000in}}%
\pgfpathclose%
\pgfusepath{fill}%
\end{pgfscope}%
\begin{pgfscope}%
\pgfsetbuttcap%
\pgfsetmiterjoin%
\definecolor{currentfill}{rgb}{0.898039,0.898039,0.898039}%
\pgfsetfillcolor{currentfill}%
\pgfsetlinewidth{0.000000pt}%
\definecolor{currentstroke}{rgb}{0.000000,0.000000,0.000000}%
\pgfsetstrokecolor{currentstroke}%
\pgfsetstrokeopacity{0.000000}%
\pgfsetdash{}{0pt}%
\pgfpathmoveto{\pgfqpoint{0.536502in}{0.288143in}}%
\pgfpathlineto{\pgfqpoint{3.208330in}{0.288143in}}%
\pgfpathlineto{\pgfqpoint{3.208330in}{1.408330in}}%
\pgfpathlineto{\pgfqpoint{0.536502in}{1.408330in}}%
\pgfpathlineto{\pgfqpoint{0.536502in}{0.288143in}}%
\pgfpathclose%
\pgfusepath{fill}%
\end{pgfscope}%
\begin{pgfscope}%
\pgfpathrectangle{\pgfqpoint{0.536502in}{0.288143in}}{\pgfqpoint{2.671828in}{1.120188in}}%
\pgfusepath{clip}%
\pgfsetrectcap%
\pgfsetroundjoin%
\pgfsetlinewidth{0.803000pt}%
\definecolor{currentstroke}{rgb}{1.000000,1.000000,1.000000}%
\pgfsetstrokecolor{currentstroke}%
\pgfsetdash{}{0pt}%
\pgfpathmoveto{\pgfqpoint{0.643375in}{0.288143in}}%
\pgfpathlineto{\pgfqpoint{0.643375in}{1.408330in}}%
\pgfusepath{stroke}%
\end{pgfscope}%
\begin{pgfscope}%
\definecolor{textcolor}{rgb}{0.333333,0.333333,0.333333}%
\pgfsetstrokecolor{textcolor}%
\pgfsetfillcolor{textcolor}%
\pgftext[x=0.643375in,y=0.190920in,,top]{\color{textcolor}\sffamily\fontsize{10.000000}{12.000000}\selectfont 1}%
\end{pgfscope}%
\begin{pgfscope}%
\pgfpathrectangle{\pgfqpoint{0.536502in}{0.288143in}}{\pgfqpoint{2.671828in}{1.120188in}}%
\pgfusepath{clip}%
\pgfsetrectcap%
\pgfsetroundjoin%
\pgfsetlinewidth{0.803000pt}%
\definecolor{currentstroke}{rgb}{1.000000,1.000000,1.000000}%
\pgfsetstrokecolor{currentstroke}%
\pgfsetdash{}{0pt}%
\pgfpathmoveto{\pgfqpoint{3.101457in}{0.288143in}}%
\pgfpathlineto{\pgfqpoint{3.101457in}{1.408330in}}%
\pgfusepath{stroke}%
\end{pgfscope}%
\begin{pgfscope}%
\definecolor{textcolor}{rgb}{0.333333,0.333333,0.333333}%
\pgfsetstrokecolor{textcolor}%
\pgfsetfillcolor{textcolor}%
\pgftext[x=3.101457in,y=0.190920in,,top]{\color{textcolor}\sffamily\fontsize{10.000000}{12.000000}\selectfont 24}%
\end{pgfscope}%
\begin{pgfscope}%
\definecolor{textcolor}{rgb}{0.333333,0.333333,0.333333}%
\pgfsetstrokecolor{textcolor}%
\pgfsetfillcolor{textcolor}%
\pgftext[x=1.872416in,y=0.176124in,,top]{\color{textcolor}\sffamily\fontsize{10.000000}{12.000000}\selectfont Layer}%
\end{pgfscope}%
\begin{pgfscope}%
\pgfpathrectangle{\pgfqpoint{0.536502in}{0.288143in}}{\pgfqpoint{2.671828in}{1.120188in}}%
\pgfusepath{clip}%
\pgfsetrectcap%
\pgfsetroundjoin%
\pgfsetlinewidth{0.803000pt}%
\definecolor{currentstroke}{rgb}{1.000000,1.000000,1.000000}%
\pgfsetstrokecolor{currentstroke}%
\pgfsetdash{}{0pt}%
\pgfpathmoveto{\pgfqpoint{0.536502in}{0.537383in}}%
\pgfpathlineto{\pgfqpoint{3.208330in}{0.537383in}}%
\pgfusepath{stroke}%
\end{pgfscope}%
\begin{pgfscope}%
\definecolor{textcolor}{rgb}{0.333333,0.333333,0.333333}%
\pgfsetstrokecolor{textcolor}%
\pgfsetfillcolor{textcolor}%
\pgftext[x=0.041670in, y=0.484621in, left, base]{\color{textcolor}\sffamily\fontsize{10.000000}{12.000000}\selectfont 0.500}%
\end{pgfscope}%
\begin{pgfscope}%
\pgfpathrectangle{\pgfqpoint{0.536502in}{0.288143in}}{\pgfqpoint{2.671828in}{1.120188in}}%
\pgfusepath{clip}%
\pgfsetrectcap%
\pgfsetroundjoin%
\pgfsetlinewidth{0.803000pt}%
\definecolor{currentstroke}{rgb}{1.000000,1.000000,1.000000}%
\pgfsetstrokecolor{currentstroke}%
\pgfsetdash{}{0pt}%
\pgfpathmoveto{\pgfqpoint{0.536502in}{0.882056in}}%
\pgfpathlineto{\pgfqpoint{3.208330in}{0.882056in}}%
\pgfusepath{stroke}%
\end{pgfscope}%
\begin{pgfscope}%
\definecolor{textcolor}{rgb}{0.333333,0.333333,0.333333}%
\pgfsetstrokecolor{textcolor}%
\pgfsetfillcolor{textcolor}%
\pgftext[x=0.041670in, y=0.829294in, left, base]{\color{textcolor}\sffamily\fontsize{10.000000}{12.000000}\selectfont 0.520}%
\end{pgfscope}%
\begin{pgfscope}%
\pgfpathrectangle{\pgfqpoint{0.536502in}{0.288143in}}{\pgfqpoint{2.671828in}{1.120188in}}%
\pgfusepath{clip}%
\pgfsetrectcap%
\pgfsetroundjoin%
\pgfsetlinewidth{0.803000pt}%
\definecolor{currentstroke}{rgb}{1.000000,1.000000,1.000000}%
\pgfsetstrokecolor{currentstroke}%
\pgfsetdash{}{0pt}%
\pgfpathmoveto{\pgfqpoint{0.536502in}{1.226729in}}%
\pgfpathlineto{\pgfqpoint{3.208330in}{1.226729in}}%
\pgfusepath{stroke}%
\end{pgfscope}%
\begin{pgfscope}%
\definecolor{textcolor}{rgb}{0.333333,0.333333,0.333333}%
\pgfsetstrokecolor{textcolor}%
\pgfsetfillcolor{textcolor}%
\pgftext[x=0.041670in, y=1.173967in, left, base]{\color{textcolor}\sffamily\fontsize{10.000000}{12.000000}\selectfont 0.540}%
\end{pgfscope}%
\begin{pgfscope}%
\pgfpathrectangle{\pgfqpoint{0.536502in}{0.288143in}}{\pgfqpoint{2.671828in}{1.120188in}}%
\pgfusepath{clip}%
\pgfsetrectcap%
\pgfsetroundjoin%
\pgfsetlinewidth{1.003750pt}%
\definecolor{currentstroke}{rgb}{0.886275,0.290196,0.200000}%
\pgfsetstrokecolor{currentstroke}%
\pgfsetdash{}{0pt}%
\pgfpathmoveto{\pgfqpoint{0.643375in}{0.389173in}}%
\pgfpathlineto{\pgfqpoint{0.750249in}{0.389173in}}%
\pgfpathlineto{\pgfqpoint{0.857122in}{0.389173in}}%
\pgfpathlineto{\pgfqpoint{0.963995in}{0.389173in}}%
\pgfpathlineto{\pgfqpoint{1.070868in}{0.389173in}}%
\pgfpathlineto{\pgfqpoint{1.177741in}{0.389173in}}%
\pgfpathlineto{\pgfqpoint{1.284614in}{0.389173in}}%
\pgfpathlineto{\pgfqpoint{1.391487in}{0.389173in}}%
\pgfpathlineto{\pgfqpoint{1.498360in}{0.389173in}}%
\pgfpathlineto{\pgfqpoint{1.605233in}{0.389173in}}%
\pgfpathlineto{\pgfqpoint{1.712107in}{0.389173in}}%
\pgfpathlineto{\pgfqpoint{1.818980in}{0.389173in}}%
\pgfusepath{stroke}%
\end{pgfscope}%
\begin{pgfscope}%
\pgfpathrectangle{\pgfqpoint{0.536502in}{0.288143in}}{\pgfqpoint{2.671828in}{1.120188in}}%
\pgfusepath{clip}%
\pgfsetrectcap%
\pgfsetroundjoin%
\pgfsetlinewidth{1.003750pt}%
\definecolor{currentstroke}{rgb}{0.203922,0.541176,0.741176}%
\pgfsetstrokecolor{currentstroke}%
\pgfsetdash{}{0pt}%
\pgfpathmoveto{\pgfqpoint{0.643375in}{0.389173in}}%
\pgfpathlineto{\pgfqpoint{0.750249in}{0.389173in}}%
\pgfpathlineto{\pgfqpoint{0.857122in}{0.389173in}}%
\pgfpathlineto{\pgfqpoint{0.963995in}{0.389173in}}%
\pgfpathlineto{\pgfqpoint{1.070868in}{0.389173in}}%
\pgfpathlineto{\pgfqpoint{1.177741in}{0.389173in}}%
\pgfpathlineto{\pgfqpoint{1.284614in}{0.389173in}}%
\pgfpathlineto{\pgfqpoint{1.391487in}{0.389173in}}%
\pgfpathlineto{\pgfqpoint{1.498360in}{0.389173in}}%
\pgfpathlineto{\pgfqpoint{1.605233in}{0.389173in}}%
\pgfpathlineto{\pgfqpoint{1.712107in}{0.389173in}}%
\pgfpathlineto{\pgfqpoint{1.818980in}{0.389173in}}%
\pgfusepath{stroke}%
\end{pgfscope}%
\begin{pgfscope}%
\pgfpathrectangle{\pgfqpoint{0.536502in}{0.288143in}}{\pgfqpoint{2.671828in}{1.120188in}}%
\pgfusepath{clip}%
\pgfsetrectcap%
\pgfsetroundjoin%
\pgfsetlinewidth{1.003750pt}%
\definecolor{currentstroke}{rgb}{0.596078,0.556863,0.835294}%
\pgfsetstrokecolor{currentstroke}%
\pgfsetdash{}{0pt}%
\pgfpathmoveto{\pgfqpoint{0.643375in}{0.389173in}}%
\pgfpathlineto{\pgfqpoint{0.750249in}{0.410421in}}%
\pgfpathlineto{\pgfqpoint{0.857122in}{0.389173in}}%
\pgfpathlineto{\pgfqpoint{0.963995in}{0.389173in}}%
\pgfpathlineto{\pgfqpoint{1.070868in}{0.487702in}}%
\pgfpathlineto{\pgfqpoint{1.177741in}{0.389173in}}%
\pgfpathlineto{\pgfqpoint{1.284614in}{0.389173in}}%
\pgfpathlineto{\pgfqpoint{1.391487in}{0.432501in}}%
\pgfpathlineto{\pgfqpoint{1.498360in}{0.598104in}}%
\pgfpathlineto{\pgfqpoint{1.605233in}{0.509782in}}%
\pgfpathlineto{\pgfqpoint{1.712107in}{0.631224in}}%
\pgfpathlineto{\pgfqpoint{1.818980in}{0.686425in}}%
\pgfpathlineto{\pgfqpoint{1.925853in}{0.564983in}}%
\pgfpathlineto{\pgfqpoint{2.032726in}{0.465622in}}%
\pgfpathlineto{\pgfqpoint{2.139599in}{0.410421in}}%
\pgfpathlineto{\pgfqpoint{2.246472in}{0.389173in}}%
\pgfusepath{stroke}%
\end{pgfscope}%
\begin{pgfscope}%
\pgfpathrectangle{\pgfqpoint{0.536502in}{0.288143in}}{\pgfqpoint{2.671828in}{1.120188in}}%
\pgfusepath{clip}%
\pgfsetrectcap%
\pgfsetroundjoin%
\pgfsetlinewidth{1.003750pt}%
\definecolor{currentstroke}{rgb}{0.466667,0.466667,0.466667}%
\pgfsetstrokecolor{currentstroke}%
\pgfsetdash{}{0pt}%
\pgfpathmoveto{\pgfqpoint{0.643375in}{0.410421in}}%
\pgfpathlineto{\pgfqpoint{0.750249in}{0.399381in}}%
\pgfpathlineto{\pgfqpoint{0.857122in}{0.399381in}}%
\pgfpathlineto{\pgfqpoint{0.963995in}{0.389173in}}%
\pgfpathlineto{\pgfqpoint{1.070868in}{0.520823in}}%
\pgfpathlineto{\pgfqpoint{1.177741in}{0.476662in}}%
\pgfpathlineto{\pgfqpoint{1.284614in}{0.487702in}}%
\pgfpathlineto{\pgfqpoint{1.391487in}{0.509782in}}%
\pgfpathlineto{\pgfqpoint{1.498360in}{0.389173in}}%
\pgfpathlineto{\pgfqpoint{1.605233in}{0.498742in}}%
\pgfpathlineto{\pgfqpoint{1.712107in}{0.389173in}}%
\pgfpathlineto{\pgfqpoint{1.818980in}{0.432501in}}%
\pgfpathlineto{\pgfqpoint{1.925853in}{0.421461in}}%
\pgfpathlineto{\pgfqpoint{2.032726in}{0.389173in}}%
\pgfpathlineto{\pgfqpoint{2.139599in}{0.476662in}}%
\pgfpathlineto{\pgfqpoint{2.246472in}{0.389173in}}%
\pgfpathlineto{\pgfqpoint{2.353345in}{0.389173in}}%
\pgfpathlineto{\pgfqpoint{2.460218in}{0.389173in}}%
\pgfpathlineto{\pgfqpoint{2.567091in}{0.389173in}}%
\pgfpathlineto{\pgfqpoint{2.673964in}{0.389173in}}%
\pgfpathlineto{\pgfqpoint{2.780838in}{0.389173in}}%
\pgfpathlineto{\pgfqpoint{2.887711in}{0.389173in}}%
\pgfpathlineto{\pgfqpoint{2.994584in}{0.389173in}}%
\pgfpathlineto{\pgfqpoint{3.101457in}{0.389173in}}%
\pgfusepath{stroke}%
\end{pgfscope}%
\begin{pgfscope}%
\pgfpathrectangle{\pgfqpoint{0.536502in}{0.288143in}}{\pgfqpoint{2.671828in}{1.120188in}}%
\pgfusepath{clip}%
\pgfsetbuttcap%
\pgfsetroundjoin%
\pgfsetlinewidth{1.003750pt}%
\definecolor{currentstroke}{rgb}{0.000000,0.392157,0.000000}%
\pgfsetstrokecolor{currentstroke}%
\pgfsetdash{{3.700000pt}{1.600000pt}}{0.000000pt}%
\pgfpathmoveto{\pgfqpoint{0.536502in}{0.389173in}}%
\pgfpathlineto{\pgfqpoint{3.208330in}{0.389173in}}%
\pgfusepath{stroke}%
\end{pgfscope}%
\begin{pgfscope}%
\pgfsetrectcap%
\pgfsetmiterjoin%
\pgfsetlinewidth{1.003750pt}%
\definecolor{currentstroke}{rgb}{1.000000,1.000000,1.000000}%
\pgfsetstrokecolor{currentstroke}%
\pgfsetdash{}{0pt}%
\pgfpathmoveto{\pgfqpoint{0.536502in}{0.288143in}}%
\pgfpathlineto{\pgfqpoint{0.536502in}{1.408330in}}%
\pgfusepath{stroke}%
\end{pgfscope}%
\begin{pgfscope}%
\pgfsetrectcap%
\pgfsetmiterjoin%
\pgfsetlinewidth{1.003750pt}%
\definecolor{currentstroke}{rgb}{1.000000,1.000000,1.000000}%
\pgfsetstrokecolor{currentstroke}%
\pgfsetdash{}{0pt}%
\pgfpathmoveto{\pgfqpoint{3.208330in}{0.288143in}}%
\pgfpathlineto{\pgfqpoint{3.208330in}{1.408330in}}%
\pgfusepath{stroke}%
\end{pgfscope}%
\begin{pgfscope}%
\pgfsetrectcap%
\pgfsetmiterjoin%
\pgfsetlinewidth{1.003750pt}%
\definecolor{currentstroke}{rgb}{1.000000,1.000000,1.000000}%
\pgfsetstrokecolor{currentstroke}%
\pgfsetdash{}{0pt}%
\pgfpathmoveto{\pgfqpoint{0.536502in}{0.288143in}}%
\pgfpathlineto{\pgfqpoint{3.208330in}{0.288143in}}%
\pgfusepath{stroke}%
\end{pgfscope}%
\begin{pgfscope}%
\pgfsetrectcap%
\pgfsetmiterjoin%
\pgfsetlinewidth{1.003750pt}%
\definecolor{currentstroke}{rgb}{1.000000,1.000000,1.000000}%
\pgfsetstrokecolor{currentstroke}%
\pgfsetdash{}{0pt}%
\pgfpathmoveto{\pgfqpoint{0.536502in}{1.408330in}}%
\pgfpathlineto{\pgfqpoint{3.208330in}{1.408330in}}%
\pgfusepath{stroke}%
\end{pgfscope}%
\end{pgfpicture}%
\makeatother%
\endgroup%

%% file: figures/probing-legend.pgf
\begingroup%
\makeatletter%
\begin{pgfpicture}%
\pgfpathrectangle{\pgfpointorigin}{\pgfqpoint{3.250000in}{0.500000in}}%
\pgfusepath{use as bounding box, clip}%
\begin{pgfscope}%
\pgfsetbuttcap%
\pgfsetmiterjoin%
\definecolor{currentfill}{rgb}{1.000000,1.000000,1.000000}%
\pgfsetfillcolor{currentfill}%
\pgfsetlinewidth{0.000000pt}%
\definecolor{currentstroke}{rgb}{0.500000,0.500000,0.500000}%
\pgfsetstrokecolor{currentstroke}%
\pgfsetdash{}{0pt}%
\pgfpathmoveto{\pgfqpoint{0.000000in}{0.000000in}}%
\pgfpathlineto{\pgfqpoint{3.250000in}{0.000000in}}%
\pgfpathlineto{\pgfqpoint{3.250000in}{0.500000in}}%
\pgfpathlineto{\pgfqpoint{0.000000in}{0.500000in}}%
\pgfpathlineto{\pgfqpoint{0.000000in}{0.000000in}}%
\pgfpathclose%
\pgfusepath{fill}%
\end{pgfscope}%
\begin{pgfscope}%
\pgfsetbuttcap%
\pgfsetmiterjoin%
\definecolor{currentfill}{rgb}{1.000000,0.960784,0.949020}%
\pgfsetfillcolor{currentfill}%
\pgfsetfillopacity{0.800000}%
\pgfsetlinewidth{0.501875pt}%
\definecolor{currentstroke}{rgb}{0.800000,0.800000,0.800000}%
\pgfsetstrokecolor{currentstroke}%
\pgfsetstrokeopacity{0.800000}%
\pgfsetdash{}{0pt}%
\pgfpathmoveto{\pgfqpoint{0.230024in}{0.070248in}}%
\pgfpathlineto{\pgfqpoint{3.019976in}{0.070248in}}%
\pgfpathquadraticcurveto{\pgfqpoint{3.042198in}{0.070248in}}{\pgfqpoint{3.042198in}{0.092470in}}%
\pgfpathlineto{\pgfqpoint{3.042198in}{0.407530in}}%
\pgfpathquadraticcurveto{\pgfqpoint{3.042198in}{0.429752in}}{\pgfqpoint{3.019976in}{0.429752in}}%
\pgfpathlineto{\pgfqpoint{0.230024in}{0.429752in}}%
\pgfpathquadraticcurveto{\pgfqpoint{0.207802in}{0.429752in}}{\pgfqpoint{0.207802in}{0.407530in}}%
\pgfpathlineto{\pgfqpoint{0.207802in}{0.092470in}}%
\pgfpathquadraticcurveto{\pgfqpoint{0.207802in}{0.070248in}}{\pgfqpoint{0.230024in}{0.070248in}}%
\pgfpathlineto{\pgfqpoint{0.230024in}{0.070248in}}%
\pgfpathclose%
\pgfusepath{stroke,fill}%
\end{pgfscope}%
\begin{pgfscope}%
\pgfsetbuttcap%
\pgfsetroundjoin%
\pgfsetlinewidth{1.003750pt}%
\definecolor{currentstroke}{rgb}{0.000000,0.392157,0.000000}%
\pgfsetstrokecolor{currentstroke}%
\pgfsetdash{{3.700000pt}{1.600000pt}}{0.000000pt}%
\pgfpathmoveto{\pgfqpoint{0.252246in}{0.339778in}}%
\pgfpathlineto{\pgfqpoint{0.363357in}{0.339778in}}%
\pgfpathlineto{\pgfqpoint{0.474468in}{0.339778in}}%
\pgfusepath{stroke}%
\end{pgfscope}%
\begin{pgfscope}%
\definecolor{textcolor}{rgb}{0.000000,0.000000,0.000000}%
\pgfsetstrokecolor{textcolor}%
\pgfsetfillcolor{textcolor}%
\pgftext[x=0.563357in,y=0.300890in,left,base]{\color{textcolor}\sffamily\fontsize{8.000000}{9.600000}\selectfont Baseline}%
\end{pgfscope}%
\begin{pgfscope}%
\pgfsetrectcap%
\pgfsetroundjoin%
\pgfsetlinewidth{1.003750pt}%
\definecolor{currentstroke}{rgb}{0.886275,0.290196,0.200000}%
\pgfsetstrokecolor{currentstroke}%
\pgfsetdash{}{0pt}%
\pgfpathmoveto{\pgfqpoint{0.252246in}{0.176693in}}%
\pgfpathlineto{\pgfqpoint{0.363357in}{0.176693in}}%
\pgfpathlineto{\pgfqpoint{0.474468in}{0.176693in}}%
\pgfusepath{stroke}%
\end{pgfscope}%
\begin{pgfscope}%
\definecolor{textcolor}{rgb}{0.000000,0.000000,0.000000}%
\pgfsetstrokecolor{textcolor}%
\pgfsetfillcolor{textcolor}%
\pgftext[x=0.563357in,y=0.137804in,left,base]{\color{textcolor}\sffamily\fontsize{8.000000}{9.600000}\selectfont VideoMAE}%
\end{pgfscope}%
\begin{pgfscope}%
\pgfsetrectcap%
\pgfsetroundjoin%
\pgfsetlinewidth{1.003750pt}%
\definecolor{currentstroke}{rgb}{0.203922,0.541176,0.741176}%
\pgfsetstrokecolor{currentstroke}%
\pgfsetdash{}{0pt}%
\pgfpathmoveto{\pgfqpoint{1.338965in}{0.339778in}}%
\pgfpathlineto{\pgfqpoint{1.450076in}{0.339778in}}%
\pgfpathlineto{\pgfqpoint{1.561187in}{0.339778in}}%
\pgfusepath{stroke}%
\end{pgfscope}%
\begin{pgfscope}%
\definecolor{textcolor}{rgb}{0.000000,0.000000,0.000000}%
\pgfsetstrokecolor{textcolor}%
\pgfsetfillcolor{textcolor}%
\pgftext[x=1.650076in,y=0.300890in,left,base]{\color{textcolor}\sffamily\fontsize{8.000000}{9.600000}\selectfont SVT}%
\end{pgfscope}%
\begin{pgfscope}%
\pgfsetrectcap%
\pgfsetroundjoin%
\pgfsetlinewidth{1.003750pt}%
\definecolor{currentstroke}{rgb}{0.596078,0.556863,0.835294}%
\pgfsetstrokecolor{currentstroke}%
\pgfsetdash{}{0pt}%
\pgfpathmoveto{\pgfqpoint{1.338965in}{0.176693in}}%
\pgfpathlineto{\pgfqpoint{1.450076in}{0.176693in}}%
\pgfpathlineto{\pgfqpoint{1.561187in}{0.176693in}}%
\pgfusepath{stroke}%
\end{pgfscope}%
\begin{pgfscope}%
\definecolor{textcolor}{rgb}{0.000000,0.000000,0.000000}%
\pgfsetstrokecolor{textcolor}%
\pgfsetfillcolor{textcolor}%
\pgftext[x=1.650076in,y=0.137804in,left,base]{\color{textcolor}\sffamily\fontsize{8.000000}{9.600000}\selectfont MaskFeat}%
\end{pgfscope}%
\begin{pgfscope}%
\pgfsetrectcap%
\pgfsetroundjoin%
\pgfsetlinewidth{1.003750pt}%
\definecolor{currentstroke}{rgb}{0.466667,0.466667,0.466667}%
\pgfsetstrokecolor{currentstroke}%
\pgfsetdash{}{0pt}%
\pgfpathmoveto{\pgfqpoint{2.396333in}{0.339778in}}%
\pgfpathlineto{\pgfqpoint{2.507444in}{0.339778in}}%
\pgfpathlineto{\pgfqpoint{2.618555in}{0.339778in}}%
\pgfusepath{stroke}%
\end{pgfscope}%
\begin{pgfscope}%
\definecolor{textcolor}{rgb}{0.000000,0.000000,0.000000}%
\pgfsetstrokecolor{textcolor}%
\pgfsetfillcolor{textcolor}%
\pgftext[x=2.707444in,y=0.300890in,left,base]{\color{textcolor}\sffamily\fontsize{8.000000}{9.600000}\selectfont BEVT}%
\end{pgfscope}%
\end{pgfpicture}%
\makeatother%
\endgroup%

%% file: figures/comparison-minorlocation-nox.pgf
\begingroup%
\makeatletter%
\begin{pgfpicture}%
\pgfpathrectangle{\pgfpointorigin}{\pgfqpoint{7.000000in}{1.250000in}}%
\pgfusepath{use as bounding box, clip}%
\begin{pgfscope}%
\pgfsetbuttcap%
\pgfsetmiterjoin%
\definecolor{currentfill}{rgb}{1.000000,1.000000,1.000000}%
\pgfsetfillcolor{currentfill}%
\pgfsetlinewidth{0.000000pt}%
\definecolor{currentstroke}{rgb}{0.500000,0.500000,0.500000}%
\pgfsetstrokecolor{currentstroke}%
\pgfsetdash{}{0pt}%
\pgfpathmoveto{\pgfqpoint{0.000000in}{0.000000in}}%
\pgfpathlineto{\pgfqpoint{7.000000in}{0.000000in}}%
\pgfpathlineto{\pgfqpoint{7.000000in}{1.250000in}}%
\pgfpathlineto{\pgfqpoint{0.000000in}{1.250000in}}%
\pgfpathlineto{\pgfqpoint{0.000000in}{0.000000in}}%
\pgfpathclose%
\pgfusepath{fill}%
\end{pgfscope}%
\begin{pgfscope}%
\pgfsetbuttcap%
\pgfsetmiterjoin%
\definecolor{currentfill}{rgb}{0.898039,0.898039,0.898039}%
\pgfsetfillcolor{currentfill}%
\pgfsetlinewidth{0.000000pt}%
\definecolor{currentstroke}{rgb}{0.000000,0.000000,0.000000}%
\pgfsetstrokecolor{currentstroke}%
\pgfsetstrokeopacity{0.000000}%
\pgfsetdash{}{0pt}%
\pgfpathmoveto{\pgfqpoint{0.536502in}{0.041670in}}%
\pgfpathlineto{\pgfqpoint{6.958330in}{0.041670in}}%
\pgfpathlineto{\pgfqpoint{6.958330in}{1.208189in}}%
\pgfpathlineto{\pgfqpoint{0.536502in}{1.208189in}}%
\pgfpathlineto{\pgfqpoint{0.536502in}{0.041670in}}%
\pgfpathclose%
\pgfusepath{fill}%
\end{pgfscope}%
\begin{pgfscope}%
\pgfpathrectangle{\pgfqpoint{0.536502in}{0.041670in}}{\pgfqpoint{6.421828in}{1.166519in}}%
\pgfusepath{clip}%
\pgfsetrectcap%
\pgfsetroundjoin%
\pgfsetlinewidth{0.803000pt}%
\definecolor{currentstroke}{rgb}{1.000000,1.000000,1.000000}%
\pgfsetstrokecolor{currentstroke}%
\pgfsetdash{}{0pt}%
\pgfpathmoveto{\pgfqpoint{0.662421in}{0.041670in}}%
\pgfpathlineto{\pgfqpoint{0.662421in}{1.208189in}}%
\pgfusepath{stroke}%
\end{pgfscope}%
\begin{pgfscope}%
\pgfpathrectangle{\pgfqpoint{0.536502in}{0.041670in}}{\pgfqpoint{6.421828in}{1.166519in}}%
\pgfusepath{clip}%
\pgfsetrectcap%
\pgfsetroundjoin%
\pgfsetlinewidth{0.803000pt}%
\definecolor{currentstroke}{rgb}{1.000000,1.000000,1.000000}%
\pgfsetstrokecolor{currentstroke}%
\pgfsetdash{}{0pt}%
\pgfpathmoveto{\pgfqpoint{3.558539in}{0.041670in}}%
\pgfpathlineto{\pgfqpoint{3.558539in}{1.208189in}}%
\pgfusepath{stroke}%
\end{pgfscope}%
\begin{pgfscope}%
\pgfpathrectangle{\pgfqpoint{0.536502in}{0.041670in}}{\pgfqpoint{6.421828in}{1.166519in}}%
\pgfusepath{clip}%
\pgfsetrectcap%
\pgfsetroundjoin%
\pgfsetlinewidth{0.803000pt}%
\definecolor{currentstroke}{rgb}{1.000000,1.000000,1.000000}%
\pgfsetstrokecolor{currentstroke}%
\pgfsetdash{}{0pt}%
\pgfpathmoveto{\pgfqpoint{3.810375in}{0.041670in}}%
\pgfpathlineto{\pgfqpoint{3.810375in}{1.208189in}}%
\pgfusepath{stroke}%
\end{pgfscope}%
\begin{pgfscope}%
\pgfpathrectangle{\pgfqpoint{0.536502in}{0.041670in}}{\pgfqpoint{6.421828in}{1.166519in}}%
\pgfusepath{clip}%
\pgfsetrectcap%
\pgfsetroundjoin%
\pgfsetlinewidth{0.803000pt}%
\definecolor{currentstroke}{rgb}{1.000000,1.000000,1.000000}%
\pgfsetstrokecolor{currentstroke}%
\pgfsetdash{}{0pt}%
\pgfpathmoveto{\pgfqpoint{6.832412in}{0.041670in}}%
\pgfpathlineto{\pgfqpoint{6.832412in}{1.208189in}}%
\pgfusepath{stroke}%
\end{pgfscope}%
\begin{pgfscope}%
\pgfpathrectangle{\pgfqpoint{0.536502in}{0.041670in}}{\pgfqpoint{6.421828in}{1.166519in}}%
\pgfusepath{clip}%
\pgfsetrectcap%
\pgfsetroundjoin%
\pgfsetlinewidth{0.803000pt}%
\definecolor{currentstroke}{rgb}{1.000000,1.000000,1.000000}%
\pgfsetstrokecolor{currentstroke}%
\pgfsetdash{}{0pt}%
\pgfpathmoveto{\pgfqpoint{0.536502in}{0.220417in}}%
\pgfpathlineto{\pgfqpoint{6.958330in}{0.220417in}}%
\pgfusepath{stroke}%
\end{pgfscope}%
\begin{pgfscope}%
\definecolor{textcolor}{rgb}{0.333333,0.333333,0.333333}%
\pgfsetstrokecolor{textcolor}%
\pgfsetfillcolor{textcolor}%
\pgftext[x=0.041670in, y=0.167656in, left, base]{\color{textcolor}\sffamily\fontsize{10.000000}{12.000000}\selectfont 0.400}%
\end{pgfscope}%
\begin{pgfscope}%
\pgfpathrectangle{\pgfqpoint{0.536502in}{0.041670in}}{\pgfqpoint{6.421828in}{1.166519in}}%
\pgfusepath{clip}%
\pgfsetrectcap%
\pgfsetroundjoin%
\pgfsetlinewidth{0.803000pt}%
\definecolor{currentstroke}{rgb}{1.000000,1.000000,1.000000}%
\pgfsetstrokecolor{currentstroke}%
\pgfsetdash{}{0pt}%
\pgfpathmoveto{\pgfqpoint{0.536502in}{0.687791in}}%
\pgfpathlineto{\pgfqpoint{6.958330in}{0.687791in}}%
\pgfusepath{stroke}%
\end{pgfscope}%
\begin{pgfscope}%
\definecolor{textcolor}{rgb}{0.333333,0.333333,0.333333}%
\pgfsetstrokecolor{textcolor}%
\pgfsetfillcolor{textcolor}%
\pgftext[x=0.041670in, y=0.635030in, left, base]{\color{textcolor}\sffamily\fontsize{10.000000}{12.000000}\selectfont 0.600}%
\end{pgfscope}%
\begin{pgfscope}%
\pgfpathrectangle{\pgfqpoint{0.536502in}{0.041670in}}{\pgfqpoint{6.421828in}{1.166519in}}%
\pgfusepath{clip}%
\pgfsetrectcap%
\pgfsetroundjoin%
\pgfsetlinewidth{0.803000pt}%
\definecolor{currentstroke}{rgb}{1.000000,1.000000,1.000000}%
\pgfsetstrokecolor{currentstroke}%
\pgfsetdash{}{0pt}%
\pgfpathmoveto{\pgfqpoint{0.536502in}{1.155165in}}%
\pgfpathlineto{\pgfqpoint{6.958330in}{1.155165in}}%
\pgfusepath{stroke}%
\end{pgfscope}%
\begin{pgfscope}%
\definecolor{textcolor}{rgb}{0.333333,0.333333,0.333333}%
\pgfsetstrokecolor{textcolor}%
\pgfsetfillcolor{textcolor}%
\pgftext[x=0.041670in, y=1.102404in, left, base]{\color{textcolor}\sffamily\fontsize{10.000000}{12.000000}\selectfont 0.800}%
\end{pgfscope}%
\begin{pgfscope}%
\pgfpathrectangle{\pgfqpoint{0.536502in}{0.041670in}}{\pgfqpoint{6.421828in}{1.166519in}}%
\pgfusepath{clip}%
\pgfsetrectcap%
\pgfsetroundjoin%
\pgfsetlinewidth{1.003750pt}%
\definecolor{currentstroke}{rgb}{0.886275,0.290196,0.200000}%
\pgfsetstrokecolor{currentstroke}%
\pgfsetdash{}{0pt}%
\pgfpathmoveto{\pgfqpoint{0.662421in}{0.094694in}}%
\pgfpathlineto{\pgfqpoint{0.788339in}{0.094694in}}%
\pgfpathlineto{\pgfqpoint{0.914257in}{0.096084in}}%
\pgfpathlineto{\pgfqpoint{1.040175in}{0.094694in}}%
\pgfpathlineto{\pgfqpoint{1.166093in}{0.094694in}}%
\pgfpathlineto{\pgfqpoint{1.292011in}{0.094694in}}%
\pgfpathlineto{\pgfqpoint{1.417930in}{0.094694in}}%
\pgfpathlineto{\pgfqpoint{1.543848in}{0.094694in}}%
\pgfpathlineto{\pgfqpoint{1.669766in}{0.094694in}}%
\pgfpathlineto{\pgfqpoint{1.795684in}{0.097582in}}%
\pgfpathlineto{\pgfqpoint{1.921602in}{0.120052in}}%
\pgfpathlineto{\pgfqpoint{2.047521in}{0.108068in}}%
\pgfusepath{stroke}%
\end{pgfscope}%
\begin{pgfscope}%
\pgfpathrectangle{\pgfqpoint{0.536502in}{0.041670in}}{\pgfqpoint{6.421828in}{1.166519in}}%
\pgfusepath{clip}%
\pgfsetrectcap%
\pgfsetroundjoin%
\pgfsetlinewidth{1.003750pt}%
\definecolor{currentstroke}{rgb}{0.886275,0.290196,0.200000}%
\pgfsetstrokecolor{currentstroke}%
\pgfsetdash{}{0pt}%
\pgfpathmoveto{\pgfqpoint{3.810375in}{0.094694in}}%
\pgfpathlineto{\pgfqpoint{3.936293in}{0.094694in}}%
\pgfpathlineto{\pgfqpoint{4.062212in}{0.094694in}}%
\pgfpathlineto{\pgfqpoint{4.188130in}{0.094694in}}%
\pgfpathlineto{\pgfqpoint{4.314048in}{0.094694in}}%
\pgfpathlineto{\pgfqpoint{4.439966in}{0.094694in}}%
\pgfpathlineto{\pgfqpoint{4.565884in}{0.096084in}}%
\pgfpathlineto{\pgfqpoint{4.691803in}{0.094694in}}%
\pgfpathlineto{\pgfqpoint{4.817721in}{0.118554in}}%
\pgfpathlineto{\pgfqpoint{4.943639in}{0.106570in}}%
\pgfpathlineto{\pgfqpoint{5.069557in}{0.097582in}}%
\pgfpathlineto{\pgfqpoint{5.195475in}{0.117056in}}%
\pgfusepath{stroke}%
\end{pgfscope}%
\begin{pgfscope}%
\pgfpathrectangle{\pgfqpoint{0.536502in}{0.041670in}}{\pgfqpoint{6.421828in}{1.166519in}}%
\pgfusepath{clip}%
\pgfsetrectcap%
\pgfsetroundjoin%
\pgfsetlinewidth{1.003750pt}%
\definecolor{currentstroke}{rgb}{0.203922,0.541176,0.741176}%
\pgfsetstrokecolor{currentstroke}%
\pgfsetdash{}{0pt}%
\pgfpathmoveto{\pgfqpoint{0.662421in}{0.100578in}}%
\pgfpathlineto{\pgfqpoint{0.788339in}{0.096084in}}%
\pgfpathlineto{\pgfqpoint{0.914257in}{0.094694in}}%
\pgfpathlineto{\pgfqpoint{1.040175in}{0.094694in}}%
\pgfpathlineto{\pgfqpoint{1.166093in}{0.100578in}}%
\pgfpathlineto{\pgfqpoint{1.292011in}{0.103574in}}%
\pgfpathlineto{\pgfqpoint{1.417930in}{0.115558in}}%
\pgfpathlineto{\pgfqpoint{1.543848in}{0.135032in}}%
\pgfpathlineto{\pgfqpoint{1.669766in}{0.148514in}}%
\pgfpathlineto{\pgfqpoint{1.795684in}{0.126044in}}%
\pgfpathlineto{\pgfqpoint{1.921602in}{0.144020in}}%
\pgfpathlineto{\pgfqpoint{2.047521in}{0.160497in}}%
\pgfusepath{stroke}%
\end{pgfscope}%
\begin{pgfscope}%
\pgfpathrectangle{\pgfqpoint{0.536502in}{0.041670in}}{\pgfqpoint{6.421828in}{1.166519in}}%
\pgfusepath{clip}%
\pgfsetrectcap%
\pgfsetroundjoin%
\pgfsetlinewidth{1.003750pt}%
\definecolor{currentstroke}{rgb}{0.203922,0.541176,0.741176}%
\pgfsetstrokecolor{currentstroke}%
\pgfsetdash{}{0pt}%
\pgfpathmoveto{\pgfqpoint{3.810375in}{0.094694in}}%
\pgfpathlineto{\pgfqpoint{3.936293in}{0.094694in}}%
\pgfpathlineto{\pgfqpoint{4.062212in}{0.094694in}}%
\pgfpathlineto{\pgfqpoint{4.188130in}{0.094694in}}%
\pgfpathlineto{\pgfqpoint{4.314048in}{0.094694in}}%
\pgfpathlineto{\pgfqpoint{4.439966in}{0.094694in}}%
\pgfpathlineto{\pgfqpoint{4.565884in}{0.094694in}}%
\pgfpathlineto{\pgfqpoint{4.691803in}{0.094694in}}%
\pgfpathlineto{\pgfqpoint{4.817721in}{0.094694in}}%
\pgfpathlineto{\pgfqpoint{4.943639in}{0.094694in}}%
\pgfpathlineto{\pgfqpoint{5.069557in}{0.094694in}}%
\pgfpathlineto{\pgfqpoint{5.195475in}{0.094694in}}%
\pgfusepath{stroke}%
\end{pgfscope}%
\begin{pgfscope}%
\pgfpathrectangle{\pgfqpoint{0.536502in}{0.041670in}}{\pgfqpoint{6.421828in}{1.166519in}}%
\pgfusepath{clip}%
\pgfsetrectcap%
\pgfsetroundjoin%
\pgfsetlinewidth{1.003750pt}%
\definecolor{currentstroke}{rgb}{0.596078,0.556863,0.835294}%
\pgfsetstrokecolor{currentstroke}%
\pgfsetdash{}{0pt}%
\pgfpathmoveto{\pgfqpoint{0.662421in}{0.094694in}}%
\pgfpathlineto{\pgfqpoint{0.788339in}{0.108068in}}%
\pgfpathlineto{\pgfqpoint{0.914257in}{0.111064in}}%
\pgfpathlineto{\pgfqpoint{1.040175in}{0.114060in}}%
\pgfpathlineto{\pgfqpoint{1.166093in}{0.191955in}}%
\pgfpathlineto{\pgfqpoint{1.292011in}{0.200943in}}%
\pgfpathlineto{\pgfqpoint{1.417930in}{0.238393in}}%
\pgfpathlineto{\pgfqpoint{1.543848in}{0.338759in}}%
\pgfpathlineto{\pgfqpoint{1.669766in}{0.380703in}}%
\pgfpathlineto{\pgfqpoint{1.795684in}{0.368719in}}%
\pgfpathlineto{\pgfqpoint{1.921602in}{0.379205in}}%
\pgfpathlineto{\pgfqpoint{2.047521in}{0.380703in}}%
\pgfpathlineto{\pgfqpoint{2.173439in}{0.353739in}}%
\pgfpathlineto{\pgfqpoint{2.299357in}{0.224911in}}%
\pgfpathlineto{\pgfqpoint{2.425275in}{0.169485in}}%
\pgfpathlineto{\pgfqpoint{2.551193in}{0.144020in}}%
\pgfusepath{stroke}%
\end{pgfscope}%
\begin{pgfscope}%
\pgfpathrectangle{\pgfqpoint{0.536502in}{0.041670in}}{\pgfqpoint{6.421828in}{1.166519in}}%
\pgfusepath{clip}%
\pgfsetrectcap%
\pgfsetroundjoin%
\pgfsetlinewidth{1.003750pt}%
\definecolor{currentstroke}{rgb}{0.596078,0.556863,0.835294}%
\pgfsetstrokecolor{currentstroke}%
\pgfsetdash{}{0pt}%
\pgfpathmoveto{\pgfqpoint{3.810375in}{0.096084in}}%
\pgfpathlineto{\pgfqpoint{3.936293in}{0.109566in}}%
\pgfpathlineto{\pgfqpoint{4.062212in}{0.117056in}}%
\pgfpathlineto{\pgfqpoint{4.188130in}{0.154505in}}%
\pgfpathlineto{\pgfqpoint{4.314048in}{0.217421in}}%
\pgfpathlineto{\pgfqpoint{4.439966in}{0.299811in}}%
\pgfpathlineto{\pgfqpoint{4.565884in}{0.352241in}}%
\pgfpathlineto{\pgfqpoint{4.691803in}{0.451108in}}%
\pgfpathlineto{\pgfqpoint{4.817721in}{0.491554in}}%
\pgfpathlineto{\pgfqpoint{4.943639in}{0.552972in}}%
\pgfpathlineto{\pgfqpoint{5.069557in}{0.564956in}}%
\pgfpathlineto{\pgfqpoint{5.195475in}{0.620382in}}%
\pgfpathlineto{\pgfqpoint{5.321394in}{0.755201in}}%
\pgfpathlineto{\pgfqpoint{5.447312in}{0.931964in}}%
\pgfpathlineto{\pgfqpoint{5.573230in}{1.155165in}}%
\pgfpathlineto{\pgfqpoint{5.699148in}{1.039820in}}%
\pgfusepath{stroke}%
\end{pgfscope}%
\begin{pgfscope}%
\pgfpathrectangle{\pgfqpoint{0.536502in}{0.041670in}}{\pgfqpoint{6.421828in}{1.166519in}}%
\pgfusepath{clip}%
\pgfsetrectcap%
\pgfsetroundjoin%
\pgfsetlinewidth{1.003750pt}%
\definecolor{currentstroke}{rgb}{0.466667,0.466667,0.466667}%
\pgfsetstrokecolor{currentstroke}%
\pgfsetdash{}{0pt}%
\pgfpathmoveto{\pgfqpoint{0.662421in}{0.108068in}}%
\pgfpathlineto{\pgfqpoint{0.788339in}{0.121550in}}%
\pgfpathlineto{\pgfqpoint{0.914257in}{0.135032in}}%
\pgfpathlineto{\pgfqpoint{1.040175in}{0.142522in}}%
\pgfpathlineto{\pgfqpoint{1.166093in}{0.123048in}}%
\pgfpathlineto{\pgfqpoint{1.292011in}{0.187461in}}%
\pgfpathlineto{\pgfqpoint{1.417930in}{0.230903in}}%
\pgfpathlineto{\pgfqpoint{1.543848in}{0.242887in}}%
\pgfpathlineto{\pgfqpoint{1.669766in}{0.275843in}}%
\pgfpathlineto{\pgfqpoint{1.795684in}{0.289325in}}%
\pgfpathlineto{\pgfqpoint{1.921602in}{0.328273in}}%
\pgfpathlineto{\pgfqpoint{2.047521in}{0.352241in}}%
\pgfpathlineto{\pgfqpoint{2.173439in}{0.344751in}}%
\pgfpathlineto{\pgfqpoint{2.299357in}{0.320783in}}%
\pgfpathlineto{\pgfqpoint{2.425275in}{0.320783in}}%
\pgfpathlineto{\pgfqpoint{2.551193in}{0.280337in}}%
\pgfpathlineto{\pgfqpoint{2.677112in}{0.256369in}}%
\pgfpathlineto{\pgfqpoint{2.803030in}{0.256369in}}%
\pgfpathlineto{\pgfqpoint{2.928948in}{0.236895in}}%
\pgfpathlineto{\pgfqpoint{3.054866in}{0.217421in}}%
\pgfpathlineto{\pgfqpoint{3.180784in}{0.170983in}}%
\pgfpathlineto{\pgfqpoint{3.306703in}{0.170983in}}%
\pgfpathlineto{\pgfqpoint{3.432621in}{0.215923in}}%
\pgfpathlineto{\pgfqpoint{3.558539in}{0.173979in}}%
\pgfusepath{stroke}%
\end{pgfscope}%
\begin{pgfscope}%
\pgfpathrectangle{\pgfqpoint{0.536502in}{0.041670in}}{\pgfqpoint{6.421828in}{1.166519in}}%
\pgfusepath{clip}%
\pgfsetrectcap%
\pgfsetroundjoin%
\pgfsetlinewidth{1.003750pt}%
\definecolor{currentstroke}{rgb}{0.466667,0.466667,0.466667}%
\pgfsetstrokecolor{currentstroke}%
\pgfsetdash{}{0pt}%
\pgfpathmoveto{\pgfqpoint{3.810375in}{0.106570in}}%
\pgfpathlineto{\pgfqpoint{3.936293in}{0.118554in}}%
\pgfpathlineto{\pgfqpoint{4.062212in}{0.157501in}}%
\pgfpathlineto{\pgfqpoint{4.188130in}{0.170983in}}%
\pgfpathlineto{\pgfqpoint{4.314048in}{0.179971in}}%
\pgfpathlineto{\pgfqpoint{4.439966in}{0.227907in}}%
\pgfpathlineto{\pgfqpoint{4.565884in}{0.305803in}}%
\pgfpathlineto{\pgfqpoint{4.691803in}{0.340257in}}%
\pgfpathlineto{\pgfqpoint{4.817721in}{0.388192in}}%
\pgfpathlineto{\pgfqpoint{4.943639in}{0.439124in}}%
\pgfpathlineto{\pgfqpoint{5.069557in}{0.503538in}}%
\pgfpathlineto{\pgfqpoint{5.195475in}{0.572446in}}%
\pgfpathlineto{\pgfqpoint{5.321394in}{0.656333in}}%
\pgfpathlineto{\pgfqpoint{5.447312in}{0.731233in}}%
\pgfpathlineto{\pgfqpoint{5.573230in}{0.764189in}}%
\pgfpathlineto{\pgfqpoint{5.699148in}{0.738723in}}%
\pgfpathlineto{\pgfqpoint{5.825066in}{0.804635in}}%
\pgfpathlineto{\pgfqpoint{5.950984in}{0.824109in}}%
\pgfpathlineto{\pgfqpoint{6.076903in}{0.830101in}}%
\pgfpathlineto{\pgfqpoint{6.202821in}{0.849575in}}%
\pgfpathlineto{\pgfqpoint{6.328739in}{0.873542in}}%
\pgfpathlineto{\pgfqpoint{6.454657in}{0.881032in}}%
\pgfpathlineto{\pgfqpoint{6.580575in}{0.969414in}}%
\pgfpathlineto{\pgfqpoint{6.706494in}{0.976904in}}%
\pgfusepath{stroke}%
\end{pgfscope}%
\begin{pgfscope}%
\pgfpathrectangle{\pgfqpoint{0.536502in}{0.041670in}}{\pgfqpoint{6.421828in}{1.166519in}}%
\pgfusepath{clip}%
\pgfsetbuttcap%
\pgfsetroundjoin%
\pgfsetlinewidth{1.003750pt}%
\definecolor{currentstroke}{rgb}{0.000000,0.392157,0.000000}%
\pgfsetstrokecolor{currentstroke}%
\pgfsetdash{{3.700000pt}{1.600000pt}}{0.000000pt}%
\pgfpathmoveto{\pgfqpoint{0.536502in}{0.094694in}}%
\pgfpathlineto{\pgfqpoint{6.958330in}{0.094694in}}%
\pgfusepath{stroke}%
\end{pgfscope}%
\begin{pgfscope}%
\pgfpathrectangle{\pgfqpoint{0.536502in}{0.041670in}}{\pgfqpoint{6.421828in}{1.166519in}}%
\pgfusepath{clip}%
\pgfsetbuttcap%
\pgfsetroundjoin%
\pgfsetlinewidth{1.003750pt}%
\definecolor{currentstroke}{rgb}{0.803922,0.521569,0.247059}%
\pgfsetstrokecolor{currentstroke}%
\pgfsetdash{{3.700000pt}{1.600000pt}}{0.000000pt}%
\pgfpathmoveto{\pgfqpoint{0.536502in}{1.002334in}}%
\pgfpathlineto{\pgfqpoint{6.958330in}{1.002334in}}%
\pgfusepath{stroke}%
\end{pgfscope}%
\begin{pgfscope}%
\pgfpathrectangle{\pgfqpoint{0.536502in}{0.041670in}}{\pgfqpoint{6.421828in}{1.166519in}}%
\pgfusepath{clip}%
\pgfsetrectcap%
\pgfsetroundjoin%
\pgfsetlinewidth{6.022500pt}%
\definecolor{currentstroke}{rgb}{1.000000,1.000000,1.000000}%
\pgfsetstrokecolor{currentstroke}%
\pgfsetdash{}{0pt}%
\pgfpathmoveto{\pgfqpoint{3.684457in}{0.041670in}}%
\pgfpathlineto{\pgfqpoint{3.684457in}{1.208189in}}%
\pgfusepath{stroke}%
\end{pgfscope}%
\begin{pgfscope}%
\pgfsetrectcap%
\pgfsetmiterjoin%
\pgfsetlinewidth{1.003750pt}%
\definecolor{currentstroke}{rgb}{1.000000,1.000000,1.000000}%
\pgfsetstrokecolor{currentstroke}%
\pgfsetdash{}{0pt}%
\pgfpathmoveto{\pgfqpoint{0.536502in}{0.041670in}}%
\pgfpathlineto{\pgfqpoint{0.536502in}{1.208189in}}%
\pgfusepath{stroke}%
\end{pgfscope}%
\begin{pgfscope}%
\pgfsetrectcap%
\pgfsetmiterjoin%
\pgfsetlinewidth{1.003750pt}%
\definecolor{currentstroke}{rgb}{1.000000,1.000000,1.000000}%
\pgfsetstrokecolor{currentstroke}%
\pgfsetdash{}{0pt}%
\pgfpathmoveto{\pgfqpoint{6.958330in}{0.041670in}}%
\pgfpathlineto{\pgfqpoint{6.958330in}{1.208189in}}%
\pgfusepath{stroke}%
\end{pgfscope}%
\begin{pgfscope}%
\pgfsetrectcap%
\pgfsetmiterjoin%
\pgfsetlinewidth{1.003750pt}%
\definecolor{currentstroke}{rgb}{1.000000,1.000000,1.000000}%
\pgfsetstrokecolor{currentstroke}%
\pgfsetdash{}{0pt}%
\pgfpathmoveto{\pgfqpoint{0.536502in}{0.041670in}}%
\pgfpathlineto{\pgfqpoint{6.958330in}{0.041670in}}%
\pgfusepath{stroke}%
\end{pgfscope}%
\begin{pgfscope}%
\pgfsetrectcap%
\pgfsetmiterjoin%
\pgfsetlinewidth{1.003750pt}%
\definecolor{currentstroke}{rgb}{1.000000,1.000000,1.000000}%
\pgfsetstrokecolor{currentstroke}%
\pgfsetdash{}{0pt}%
\pgfpathmoveto{\pgfqpoint{0.536502in}{1.208189in}}%
\pgfpathlineto{\pgfqpoint{6.958330in}{1.208189in}}%
\pgfusepath{stroke}%
\end{pgfscope}%
\end{pgfpicture}%
\makeatother%
\endgroup%

%% file: figures/comparison-handshape-nox.pgf
\begingroup%
\makeatletter%
\begin{pgfpicture}%
\pgfpathrectangle{\pgfpointorigin}{\pgfqpoint{7.000000in}{1.250000in}}%
\pgfusepath{use as bounding box, clip}%
\begin{pgfscope}%
\pgfsetbuttcap%
\pgfsetmiterjoin%
\definecolor{currentfill}{rgb}{1.000000,1.000000,1.000000}%
\pgfsetfillcolor{currentfill}%
\pgfsetlinewidth{0.000000pt}%
\definecolor{currentstroke}{rgb}{0.500000,0.500000,0.500000}%
\pgfsetstrokecolor{currentstroke}%
\pgfsetdash{}{0pt}%
\pgfpathmoveto{\pgfqpoint{0.000000in}{0.000000in}}%
\pgfpathlineto{\pgfqpoint{7.000000in}{0.000000in}}%
\pgfpathlineto{\pgfqpoint{7.000000in}{1.250000in}}%
\pgfpathlineto{\pgfqpoint{0.000000in}{1.250000in}}%
\pgfpathlineto{\pgfqpoint{0.000000in}{0.000000in}}%
\pgfpathclose%
\pgfusepath{fill}%
\end{pgfscope}%
\begin{pgfscope}%
\pgfsetbuttcap%
\pgfsetmiterjoin%
\definecolor{currentfill}{rgb}{0.898039,0.898039,0.898039}%
\pgfsetfillcolor{currentfill}%
\pgfsetlinewidth{0.000000pt}%
\definecolor{currentstroke}{rgb}{0.000000,0.000000,0.000000}%
\pgfsetstrokecolor{currentstroke}%
\pgfsetstrokeopacity{0.000000}%
\pgfsetdash{}{0pt}%
\pgfpathmoveto{\pgfqpoint{0.536502in}{0.041670in}}%
\pgfpathlineto{\pgfqpoint{6.958330in}{0.041670in}}%
\pgfpathlineto{\pgfqpoint{6.958330in}{1.208330in}}%
\pgfpathlineto{\pgfqpoint{0.536502in}{1.208330in}}%
\pgfpathlineto{\pgfqpoint{0.536502in}{0.041670in}}%
\pgfpathclose%
\pgfusepath{fill}%
\end{pgfscope}%
\begin{pgfscope}%
\pgfpathrectangle{\pgfqpoint{0.536502in}{0.041670in}}{\pgfqpoint{6.421828in}{1.166660in}}%
\pgfusepath{clip}%
\pgfsetrectcap%
\pgfsetroundjoin%
\pgfsetlinewidth{0.803000pt}%
\definecolor{currentstroke}{rgb}{1.000000,1.000000,1.000000}%
\pgfsetstrokecolor{currentstroke}%
\pgfsetdash{}{0pt}%
\pgfpathmoveto{\pgfqpoint{0.662421in}{0.041670in}}%
\pgfpathlineto{\pgfqpoint{0.662421in}{1.208330in}}%
\pgfusepath{stroke}%
\end{pgfscope}%
\begin{pgfscope}%
\pgfpathrectangle{\pgfqpoint{0.536502in}{0.041670in}}{\pgfqpoint{6.421828in}{1.166660in}}%
\pgfusepath{clip}%
\pgfsetrectcap%
\pgfsetroundjoin%
\pgfsetlinewidth{0.803000pt}%
\definecolor{currentstroke}{rgb}{1.000000,1.000000,1.000000}%
\pgfsetstrokecolor{currentstroke}%
\pgfsetdash{}{0pt}%
\pgfpathmoveto{\pgfqpoint{3.558539in}{0.041670in}}%
\pgfpathlineto{\pgfqpoint{3.558539in}{1.208330in}}%
\pgfusepath{stroke}%
\end{pgfscope}%
\begin{pgfscope}%
\pgfpathrectangle{\pgfqpoint{0.536502in}{0.041670in}}{\pgfqpoint{6.421828in}{1.166660in}}%
\pgfusepath{clip}%
\pgfsetrectcap%
\pgfsetroundjoin%
\pgfsetlinewidth{0.803000pt}%
\definecolor{currentstroke}{rgb}{1.000000,1.000000,1.000000}%
\pgfsetstrokecolor{currentstroke}%
\pgfsetdash{}{0pt}%
\pgfpathmoveto{\pgfqpoint{3.810375in}{0.041670in}}%
\pgfpathlineto{\pgfqpoint{3.810375in}{1.208330in}}%
\pgfusepath{stroke}%
\end{pgfscope}%
\begin{pgfscope}%
\pgfpathrectangle{\pgfqpoint{0.536502in}{0.041670in}}{\pgfqpoint{6.421828in}{1.166660in}}%
\pgfusepath{clip}%
\pgfsetrectcap%
\pgfsetroundjoin%
\pgfsetlinewidth{0.803000pt}%
\definecolor{currentstroke}{rgb}{1.000000,1.000000,1.000000}%
\pgfsetstrokecolor{currentstroke}%
\pgfsetdash{}{0pt}%
\pgfpathmoveto{\pgfqpoint{6.832412in}{0.041670in}}%
\pgfpathlineto{\pgfqpoint{6.832412in}{1.208330in}}%
\pgfusepath{stroke}%
\end{pgfscope}%
\begin{pgfscope}%
\pgfpathrectangle{\pgfqpoint{0.536502in}{0.041670in}}{\pgfqpoint{6.421828in}{1.166660in}}%
\pgfusepath{clip}%
\pgfsetrectcap%
\pgfsetroundjoin%
\pgfsetlinewidth{0.803000pt}%
\definecolor{currentstroke}{rgb}{1.000000,1.000000,1.000000}%
\pgfsetstrokecolor{currentstroke}%
\pgfsetdash{}{0pt}%
\pgfpathmoveto{\pgfqpoint{0.536502in}{0.240814in}}%
\pgfpathlineto{\pgfqpoint{6.958330in}{0.240814in}}%
\pgfusepath{stroke}%
\end{pgfscope}%
\begin{pgfscope}%
\definecolor{textcolor}{rgb}{0.333333,0.333333,0.333333}%
\pgfsetstrokecolor{textcolor}%
\pgfsetfillcolor{textcolor}%
\pgftext[x=0.041670in, y=0.188053in, left, base]{\color{textcolor}\sffamily\fontsize{10.000000}{12.000000}\selectfont 0.200}%
\end{pgfscope}%
\begin{pgfscope}%
\pgfpathrectangle{\pgfqpoint{0.536502in}{0.041670in}}{\pgfqpoint{6.421828in}{1.166660in}}%
\pgfusepath{clip}%
\pgfsetrectcap%
\pgfsetroundjoin%
\pgfsetlinewidth{0.803000pt}%
\definecolor{currentstroke}{rgb}{1.000000,1.000000,1.000000}%
\pgfsetstrokecolor{currentstroke}%
\pgfsetdash{}{0pt}%
\pgfpathmoveto{\pgfqpoint{0.536502in}{0.586237in}}%
\pgfpathlineto{\pgfqpoint{6.958330in}{0.586237in}}%
\pgfusepath{stroke}%
\end{pgfscope}%
\begin{pgfscope}%
\definecolor{textcolor}{rgb}{0.333333,0.333333,0.333333}%
\pgfsetstrokecolor{textcolor}%
\pgfsetfillcolor{textcolor}%
\pgftext[x=0.041670in, y=0.533476in, left, base]{\color{textcolor}\sffamily\fontsize{10.000000}{12.000000}\selectfont 0.400}%
\end{pgfscope}%
\begin{pgfscope}%
\pgfpathrectangle{\pgfqpoint{0.536502in}{0.041670in}}{\pgfqpoint{6.421828in}{1.166660in}}%
\pgfusepath{clip}%
\pgfsetrectcap%
\pgfsetroundjoin%
\pgfsetlinewidth{0.803000pt}%
\definecolor{currentstroke}{rgb}{1.000000,1.000000,1.000000}%
\pgfsetstrokecolor{currentstroke}%
\pgfsetdash{}{0pt}%
\pgfpathmoveto{\pgfqpoint{0.536502in}{0.931661in}}%
\pgfpathlineto{\pgfqpoint{6.958330in}{0.931661in}}%
\pgfusepath{stroke}%
\end{pgfscope}%
\begin{pgfscope}%
\definecolor{textcolor}{rgb}{0.333333,0.333333,0.333333}%
\pgfsetstrokecolor{textcolor}%
\pgfsetfillcolor{textcolor}%
\pgftext[x=0.041670in, y=0.878899in, left, base]{\color{textcolor}\sffamily\fontsize{10.000000}{12.000000}\selectfont 0.600}%
\end{pgfscope}%
\begin{pgfscope}%
\pgfpathrectangle{\pgfqpoint{0.536502in}{0.041670in}}{\pgfqpoint{6.421828in}{1.166660in}}%
\pgfusepath{clip}%
\pgfsetrectcap%
\pgfsetroundjoin%
\pgfsetlinewidth{1.003750pt}%
\definecolor{currentstroke}{rgb}{0.886275,0.290196,0.200000}%
\pgfsetstrokecolor{currentstroke}%
\pgfsetdash{}{0pt}%
\pgfpathmoveto{\pgfqpoint{0.662421in}{0.094700in}}%
\pgfpathlineto{\pgfqpoint{0.788339in}{0.094700in}}%
\pgfpathlineto{\pgfqpoint{0.914257in}{0.094700in}}%
\pgfpathlineto{\pgfqpoint{1.040175in}{0.094700in}}%
\pgfpathlineto{\pgfqpoint{1.166093in}{0.094700in}}%
\pgfpathlineto{\pgfqpoint{1.292011in}{0.094700in}}%
\pgfpathlineto{\pgfqpoint{1.417930in}{0.094700in}}%
\pgfpathlineto{\pgfqpoint{1.543848in}{0.094700in}}%
\pgfpathlineto{\pgfqpoint{1.669766in}{0.094700in}}%
\pgfpathlineto{\pgfqpoint{1.795684in}{0.094700in}}%
\pgfpathlineto{\pgfqpoint{1.921602in}{0.094700in}}%
\pgfpathlineto{\pgfqpoint{2.047521in}{0.094700in}}%
\pgfusepath{stroke}%
\end{pgfscope}%
\begin{pgfscope}%
\pgfpathrectangle{\pgfqpoint{0.536502in}{0.041670in}}{\pgfqpoint{6.421828in}{1.166660in}}%
\pgfusepath{clip}%
\pgfsetrectcap%
\pgfsetroundjoin%
\pgfsetlinewidth{1.003750pt}%
\definecolor{currentstroke}{rgb}{0.886275,0.290196,0.200000}%
\pgfsetstrokecolor{currentstroke}%
\pgfsetdash{}{0pt}%
\pgfpathmoveto{\pgfqpoint{3.810375in}{0.094700in}}%
\pgfpathlineto{\pgfqpoint{3.936293in}{0.097995in}}%
\pgfpathlineto{\pgfqpoint{4.062212in}{0.094700in}}%
\pgfpathlineto{\pgfqpoint{4.188130in}{0.094700in}}%
\pgfpathlineto{\pgfqpoint{4.314048in}{0.094700in}}%
\pgfpathlineto{\pgfqpoint{4.439966in}{0.094700in}}%
\pgfpathlineto{\pgfqpoint{4.565884in}{0.094700in}}%
\pgfpathlineto{\pgfqpoint{4.691803in}{0.094700in}}%
\pgfpathlineto{\pgfqpoint{4.817721in}{0.094700in}}%
\pgfpathlineto{\pgfqpoint{4.943639in}{0.094700in}}%
\pgfpathlineto{\pgfqpoint{5.069557in}{0.094700in}}%
\pgfpathlineto{\pgfqpoint{5.195475in}{0.094700in}}%
\pgfusepath{stroke}%
\end{pgfscope}%
\begin{pgfscope}%
\pgfpathrectangle{\pgfqpoint{0.536502in}{0.041670in}}{\pgfqpoint{6.421828in}{1.166660in}}%
\pgfusepath{clip}%
\pgfsetrectcap%
\pgfsetroundjoin%
\pgfsetlinewidth{1.003750pt}%
\definecolor{currentstroke}{rgb}{0.203922,0.541176,0.741176}%
\pgfsetstrokecolor{currentstroke}%
\pgfsetdash{}{0pt}%
\pgfpathmoveto{\pgfqpoint{0.662421in}{0.094700in}}%
\pgfpathlineto{\pgfqpoint{0.788339in}{0.112387in}}%
\pgfpathlineto{\pgfqpoint{0.914257in}{0.101316in}}%
\pgfpathlineto{\pgfqpoint{1.040175in}{0.111280in}}%
\pgfpathlineto{\pgfqpoint{1.166093in}{0.094700in}}%
\pgfpathlineto{\pgfqpoint{1.292011in}{0.102423in}}%
\pgfpathlineto{\pgfqpoint{1.417930in}{0.099102in}}%
\pgfpathlineto{\pgfqpoint{1.543848in}{0.109066in}}%
\pgfpathlineto{\pgfqpoint{1.669766in}{0.115709in}}%
\pgfpathlineto{\pgfqpoint{1.795684in}{0.143387in}}%
\pgfpathlineto{\pgfqpoint{1.921602in}{0.151137in}}%
\pgfpathlineto{\pgfqpoint{2.047521in}{0.133423in}}%
\pgfusepath{stroke}%
\end{pgfscope}%
\begin{pgfscope}%
\pgfpathrectangle{\pgfqpoint{0.536502in}{0.041670in}}{\pgfqpoint{6.421828in}{1.166660in}}%
\pgfusepath{clip}%
\pgfsetrectcap%
\pgfsetroundjoin%
\pgfsetlinewidth{1.003750pt}%
\definecolor{currentstroke}{rgb}{0.203922,0.541176,0.741176}%
\pgfsetstrokecolor{currentstroke}%
\pgfsetdash{}{0pt}%
\pgfpathmoveto{\pgfqpoint{3.810375in}{0.094700in}}%
\pgfpathlineto{\pgfqpoint{3.936293in}{0.094700in}}%
\pgfpathlineto{\pgfqpoint{4.062212in}{0.104638in}}%
\pgfpathlineto{\pgfqpoint{4.188130in}{0.094700in}}%
\pgfpathlineto{\pgfqpoint{4.314048in}{0.094700in}}%
\pgfpathlineto{\pgfqpoint{4.439966in}{0.094700in}}%
\pgfpathlineto{\pgfqpoint{4.565884in}{0.094700in}}%
\pgfpathlineto{\pgfqpoint{4.691803in}{0.094700in}}%
\pgfpathlineto{\pgfqpoint{4.817721in}{0.094700in}}%
\pgfpathlineto{\pgfqpoint{4.943639in}{0.094700in}}%
\pgfpathlineto{\pgfqpoint{5.069557in}{0.094700in}}%
\pgfpathlineto{\pgfqpoint{5.195475in}{0.094700in}}%
\pgfusepath{stroke}%
\end{pgfscope}%
\begin{pgfscope}%
\pgfpathrectangle{\pgfqpoint{0.536502in}{0.041670in}}{\pgfqpoint{6.421828in}{1.166660in}}%
\pgfusepath{clip}%
\pgfsetrectcap%
\pgfsetroundjoin%
\pgfsetlinewidth{1.003750pt}%
\definecolor{currentstroke}{rgb}{0.596078,0.556863,0.835294}%
\pgfsetstrokecolor{currentstroke}%
\pgfsetdash{}{0pt}%
\pgfpathmoveto{\pgfqpoint{0.662421in}{0.111280in}}%
\pgfpathlineto{\pgfqpoint{0.788339in}{0.127887in}}%
\pgfpathlineto{\pgfqpoint{0.914257in}{0.124566in}}%
\pgfpathlineto{\pgfqpoint{1.040175in}{0.137851in}}%
\pgfpathlineto{\pgfqpoint{1.166093in}{0.153351in}}%
\pgfpathlineto{\pgfqpoint{1.292011in}{0.171065in}}%
\pgfpathlineto{\pgfqpoint{1.417930in}{0.187672in}}%
\pgfpathlineto{\pgfqpoint{1.543848in}{0.203172in}}%
\pgfpathlineto{\pgfqpoint{1.669766in}{0.206493in}}%
\pgfpathlineto{\pgfqpoint{1.795684in}{0.217564in}}%
\pgfpathlineto{\pgfqpoint{1.921602in}{0.209815in}}%
\pgfpathlineto{\pgfqpoint{2.047521in}{0.204279in}}%
\pgfpathlineto{\pgfqpoint{2.173439in}{0.216457in}}%
\pgfpathlineto{\pgfqpoint{2.299357in}{0.155565in}}%
\pgfpathlineto{\pgfqpoint{2.425275in}{0.136744in}}%
\pgfpathlineto{\pgfqpoint{2.551193in}{0.136744in}}%
\pgfusepath{stroke}%
\end{pgfscope}%
\begin{pgfscope}%
\pgfpathrectangle{\pgfqpoint{0.536502in}{0.041670in}}{\pgfqpoint{6.421828in}{1.166660in}}%
\pgfusepath{clip}%
\pgfsetrectcap%
\pgfsetroundjoin%
\pgfsetlinewidth{1.003750pt}%
\definecolor{currentstroke}{rgb}{0.596078,0.556863,0.835294}%
\pgfsetstrokecolor{currentstroke}%
\pgfsetdash{}{0pt}%
\pgfpathmoveto{\pgfqpoint{3.810375in}{0.116816in}}%
\pgfpathlineto{\pgfqpoint{3.936293in}{0.122352in}}%
\pgfpathlineto{\pgfqpoint{4.062212in}{0.130101in}}%
\pgfpathlineto{\pgfqpoint{4.188130in}{0.147815in}}%
\pgfpathlineto{\pgfqpoint{4.314048in}{0.168851in}}%
\pgfpathlineto{\pgfqpoint{4.439966in}{0.212029in}}%
\pgfpathlineto{\pgfqpoint{4.565884in}{0.204279in}}%
\pgfpathlineto{\pgfqpoint{4.691803in}{0.239707in}}%
\pgfpathlineto{\pgfqpoint{4.817721in}{0.258528in}}%
\pgfpathlineto{\pgfqpoint{4.943639in}{0.307242in}}%
\pgfpathlineto{\pgfqpoint{5.069557in}{0.307242in}}%
\pgfpathlineto{\pgfqpoint{5.195475in}{0.405776in}}%
\pgfpathlineto{\pgfqpoint{5.321394in}{0.640486in}}%
\pgfpathlineto{\pgfqpoint{5.447312in}{0.835341in}}%
\pgfpathlineto{\pgfqpoint{5.573230in}{1.132050in}}%
\pgfpathlineto{\pgfqpoint{5.699148in}{1.155300in}}%
\pgfusepath{stroke}%
\end{pgfscope}%
\begin{pgfscope}%
\pgfpathrectangle{\pgfqpoint{0.536502in}{0.041670in}}{\pgfqpoint{6.421828in}{1.166660in}}%
\pgfusepath{clip}%
\pgfsetrectcap%
\pgfsetroundjoin%
\pgfsetlinewidth{1.003750pt}%
\definecolor{currentstroke}{rgb}{0.466667,0.466667,0.466667}%
\pgfsetstrokecolor{currentstroke}%
\pgfsetdash{}{0pt}%
\pgfpathmoveto{\pgfqpoint{0.662421in}{0.132316in}}%
\pgfpathlineto{\pgfqpoint{0.788339in}{0.115709in}}%
\pgfpathlineto{\pgfqpoint{0.914257in}{0.120137in}}%
\pgfpathlineto{\pgfqpoint{1.040175in}{0.124566in}}%
\pgfpathlineto{\pgfqpoint{1.166093in}{0.123459in}}%
\pgfpathlineto{\pgfqpoint{1.292011in}{0.125673in}}%
\pgfpathlineto{\pgfqpoint{1.417930in}{0.145601in}}%
\pgfpathlineto{\pgfqpoint{1.543848in}{0.155565in}}%
\pgfpathlineto{\pgfqpoint{1.669766in}{0.187672in}}%
\pgfpathlineto{\pgfqpoint{1.795684in}{0.202065in}}%
\pgfpathlineto{\pgfqpoint{1.921602in}{0.210922in}}%
\pgfpathlineto{\pgfqpoint{2.047521in}{0.195422in}}%
\pgfpathlineto{\pgfqpoint{2.173439in}{0.210922in}}%
\pgfpathlineto{\pgfqpoint{2.299357in}{0.202065in}}%
\pgfpathlineto{\pgfqpoint{2.425275in}{0.189886in}}%
\pgfpathlineto{\pgfqpoint{2.551193in}{0.187672in}}%
\pgfpathlineto{\pgfqpoint{2.677112in}{0.164422in}}%
\pgfpathlineto{\pgfqpoint{2.803030in}{0.145601in}}%
\pgfpathlineto{\pgfqpoint{2.928948in}{0.132316in}}%
\pgfpathlineto{\pgfqpoint{3.054866in}{0.131209in}}%
\pgfpathlineto{\pgfqpoint{3.180784in}{0.124566in}}%
\pgfpathlineto{\pgfqpoint{3.306703in}{0.113495in}}%
\pgfpathlineto{\pgfqpoint{3.432621in}{0.132316in}}%
\pgfpathlineto{\pgfqpoint{3.558539in}{0.130101in}}%
\pgfusepath{stroke}%
\end{pgfscope}%
\begin{pgfscope}%
\pgfpathrectangle{\pgfqpoint{0.536502in}{0.041670in}}{\pgfqpoint{6.421828in}{1.166660in}}%
\pgfusepath{clip}%
\pgfsetrectcap%
\pgfsetroundjoin%
\pgfsetlinewidth{1.003750pt}%
\definecolor{currentstroke}{rgb}{0.466667,0.466667,0.466667}%
\pgfsetstrokecolor{currentstroke}%
\pgfsetdash{}{0pt}%
\pgfpathmoveto{\pgfqpoint{3.810375in}{0.133423in}}%
\pgfpathlineto{\pgfqpoint{3.936293in}{0.121244in}}%
\pgfpathlineto{\pgfqpoint{4.062212in}{0.125673in}}%
\pgfpathlineto{\pgfqpoint{4.188130in}{0.136744in}}%
\pgfpathlineto{\pgfqpoint{4.314048in}{0.150030in}}%
\pgfpathlineto{\pgfqpoint{4.439966in}{0.166637in}}%
\pgfpathlineto{\pgfqpoint{4.565884in}{0.221993in}}%
\pgfpathlineto{\pgfqpoint{4.691803in}{0.237493in}}%
\pgfpathlineto{\pgfqpoint{4.817721in}{0.275135in}}%
\pgfpathlineto{\pgfqpoint{4.943639in}{0.298385in}}%
\pgfpathlineto{\pgfqpoint{5.069557in}{0.336027in}}%
\pgfpathlineto{\pgfqpoint{5.195475in}{0.368134in}}%
\pgfpathlineto{\pgfqpoint{5.321394in}{0.373669in}}%
\pgfpathlineto{\pgfqpoint{5.447312in}{0.452275in}}%
\pgfpathlineto{\pgfqpoint{5.573230in}{0.555238in}}%
\pgfpathlineto{\pgfqpoint{5.699148in}{0.544167in}}%
\pgfpathlineto{\pgfqpoint{5.825066in}{0.595094in}}%
\pgfpathlineto{\pgfqpoint{5.950984in}{0.644915in}}%
\pgfpathlineto{\pgfqpoint{6.076903in}{0.652665in}}%
\pgfpathlineto{\pgfqpoint{6.202821in}{0.698057in}}%
\pgfpathlineto{\pgfqpoint{6.328739in}{0.712450in}}%
\pgfpathlineto{\pgfqpoint{6.454657in}{0.733485in}}%
\pgfpathlineto{\pgfqpoint{6.580575in}{0.794377in}}%
\pgfpathlineto{\pgfqpoint{6.706494in}{0.880733in}}%
\pgfusepath{stroke}%
\end{pgfscope}%
\begin{pgfscope}%
\pgfpathrectangle{\pgfqpoint{0.536502in}{0.041670in}}{\pgfqpoint{6.421828in}{1.166660in}}%
\pgfusepath{clip}%
\pgfsetbuttcap%
\pgfsetroundjoin%
\pgfsetlinewidth{1.003750pt}%
\definecolor{currentstroke}{rgb}{0.000000,0.392157,0.000000}%
\pgfsetstrokecolor{currentstroke}%
\pgfsetdash{{3.700000pt}{1.600000pt}}{0.000000pt}%
\pgfpathmoveto{\pgfqpoint{0.536502in}{0.094700in}}%
\pgfpathlineto{\pgfqpoint{6.958330in}{0.094700in}}%
\pgfusepath{stroke}%
\end{pgfscope}%
\begin{pgfscope}%
\pgfpathrectangle{\pgfqpoint{0.536502in}{0.041670in}}{\pgfqpoint{6.421828in}{1.166660in}}%
\pgfusepath{clip}%
\pgfsetbuttcap%
\pgfsetroundjoin%
\pgfsetlinewidth{1.003750pt}%
\definecolor{currentstroke}{rgb}{0.803922,0.521569,0.247059}%
\pgfsetstrokecolor{currentstroke}%
\pgfsetdash{{3.700000pt}{1.600000pt}}{0.000000pt}%
\pgfpathmoveto{\pgfqpoint{0.536502in}{0.944959in}}%
\pgfpathlineto{\pgfqpoint{6.958330in}{0.944959in}}%
\pgfusepath{stroke}%
\end{pgfscope}%
\begin{pgfscope}%
\pgfpathrectangle{\pgfqpoint{0.536502in}{0.041670in}}{\pgfqpoint{6.421828in}{1.166660in}}%
\pgfusepath{clip}%
\pgfsetrectcap%
\pgfsetroundjoin%
\pgfsetlinewidth{6.022500pt}%
\definecolor{currentstroke}{rgb}{1.000000,1.000000,1.000000}%
\pgfsetstrokecolor{currentstroke}%
\pgfsetdash{}{0pt}%
\pgfpathmoveto{\pgfqpoint{3.684457in}{0.041670in}}%
\pgfpathlineto{\pgfqpoint{3.684457in}{1.208330in}}%
\pgfusepath{stroke}%
\end{pgfscope}%
\begin{pgfscope}%
\pgfsetrectcap%
\pgfsetmiterjoin%
\pgfsetlinewidth{1.003750pt}%
\definecolor{currentstroke}{rgb}{1.000000,1.000000,1.000000}%
\pgfsetstrokecolor{currentstroke}%
\pgfsetdash{}{0pt}%
\pgfpathmoveto{\pgfqpoint{0.536502in}{0.041670in}}%
\pgfpathlineto{\pgfqpoint{0.536502in}{1.208330in}}%
\pgfusepath{stroke}%
\end{pgfscope}%
\begin{pgfscope}%
\pgfsetrectcap%
\pgfsetmiterjoin%
\pgfsetlinewidth{1.003750pt}%
\definecolor{currentstroke}{rgb}{1.000000,1.000000,1.000000}%
\pgfsetstrokecolor{currentstroke}%
\pgfsetdash{}{0pt}%
\pgfpathmoveto{\pgfqpoint{6.958330in}{0.041670in}}%
\pgfpathlineto{\pgfqpoint{6.958330in}{1.208330in}}%
\pgfusepath{stroke}%
\end{pgfscope}%
\begin{pgfscope}%
\pgfsetrectcap%
\pgfsetmiterjoin%
\pgfsetlinewidth{1.003750pt}%
\definecolor{currentstroke}{rgb}{1.000000,1.000000,1.000000}%
\pgfsetstrokecolor{currentstroke}%
\pgfsetdash{}{0pt}%
\pgfpathmoveto{\pgfqpoint{0.536502in}{0.041670in}}%
\pgfpathlineto{\pgfqpoint{6.958330in}{0.041670in}}%
\pgfusepath{stroke}%
\end{pgfscope}%
\begin{pgfscope}%
\pgfsetrectcap%
\pgfsetmiterjoin%
\pgfsetlinewidth{1.003750pt}%
\definecolor{currentstroke}{rgb}{1.000000,1.000000,1.000000}%
\pgfsetstrokecolor{currentstroke}%
\pgfsetdash{}{0pt}%
\pgfpathmoveto{\pgfqpoint{0.536502in}{1.208330in}}%
\pgfpathlineto{\pgfqpoint{6.958330in}{1.208330in}}%
\pgfusepath{stroke}%
\end{pgfscope}%
\end{pgfpicture}%
\makeatother%
\endgroup%

%% file: figures/comparison-pathmovement.pgf
\begingroup%
\makeatletter%
\begin{pgfpicture}%
\pgfpathrectangle{\pgfpointorigin}{\pgfqpoint{7.000000in}{1.550000in}}%
\pgfusepath{use as bounding box, clip}%
\begin{pgfscope}%
\pgfsetbuttcap%
\pgfsetmiterjoin%
\definecolor{currentfill}{rgb}{1.000000,1.000000,1.000000}%
\pgfsetfillcolor{currentfill}%
\pgfsetlinewidth{0.000000pt}%
\definecolor{currentstroke}{rgb}{0.500000,0.500000,0.500000}%
\pgfsetstrokecolor{currentstroke}%
\pgfsetdash{}{0pt}%
\pgfpathmoveto{\pgfqpoint{0.000000in}{0.000000in}}%
\pgfpathlineto{\pgfqpoint{7.000000in}{0.000000in}}%
\pgfpathlineto{\pgfqpoint{7.000000in}{1.550000in}}%
\pgfpathlineto{\pgfqpoint{0.000000in}{1.550000in}}%
\pgfpathlineto{\pgfqpoint{0.000000in}{0.000000in}}%
\pgfpathclose%
\pgfusepath{fill}%
\end{pgfscope}%
\begin{pgfscope}%
\pgfsetbuttcap%
\pgfsetmiterjoin%
\definecolor{currentfill}{rgb}{0.898039,0.898039,0.898039}%
\pgfsetfillcolor{currentfill}%
\pgfsetlinewidth{0.000000pt}%
\definecolor{currentstroke}{rgb}{0.000000,0.000000,0.000000}%
\pgfsetstrokecolor{currentstroke}%
\pgfsetstrokeopacity{0.000000}%
\pgfsetdash{}{0pt}%
\pgfpathmoveto{\pgfqpoint{0.536502in}{0.394792in}}%
\pgfpathlineto{\pgfqpoint{6.958330in}{0.394792in}}%
\pgfpathlineto{\pgfqpoint{6.958330in}{1.508330in}}%
\pgfpathlineto{\pgfqpoint{0.536502in}{1.508330in}}%
\pgfpathlineto{\pgfqpoint{0.536502in}{0.394792in}}%
\pgfpathclose%
\pgfusepath{fill}%
\end{pgfscope}%
\begin{pgfscope}%
\pgfpathrectangle{\pgfqpoint{0.536502in}{0.394792in}}{\pgfqpoint{6.421828in}{1.113538in}}%
\pgfusepath{clip}%
\pgfsetrectcap%
\pgfsetroundjoin%
\pgfsetlinewidth{0.803000pt}%
\definecolor{currentstroke}{rgb}{1.000000,1.000000,1.000000}%
\pgfsetstrokecolor{currentstroke}%
\pgfsetdash{}{0pt}%
\pgfpathmoveto{\pgfqpoint{0.662421in}{0.394792in}}%
\pgfpathlineto{\pgfqpoint{0.662421in}{1.508330in}}%
\pgfusepath{stroke}%
\end{pgfscope}%
\begin{pgfscope}%
\definecolor{textcolor}{rgb}{0.333333,0.333333,0.333333}%
\pgfsetstrokecolor{textcolor}%
\pgfsetfillcolor{textcolor}%
\pgftext[x=0.662421in,y=0.297570in,,top]{\color{textcolor}\sffamily\fontsize{10.000000}{12.000000}\selectfont 1}%
\end{pgfscope}%
\begin{pgfscope}%
\pgfpathrectangle{\pgfqpoint{0.536502in}{0.394792in}}{\pgfqpoint{6.421828in}{1.113538in}}%
\pgfusepath{clip}%
\pgfsetrectcap%
\pgfsetroundjoin%
\pgfsetlinewidth{0.803000pt}%
\definecolor{currentstroke}{rgb}{1.000000,1.000000,1.000000}%
\pgfsetstrokecolor{currentstroke}%
\pgfsetdash{}{0pt}%
\pgfpathmoveto{\pgfqpoint{3.558539in}{0.394792in}}%
\pgfpathlineto{\pgfqpoint{3.558539in}{1.508330in}}%
\pgfusepath{stroke}%
\end{pgfscope}%
\begin{pgfscope}%
\definecolor{textcolor}{rgb}{0.333333,0.333333,0.333333}%
\pgfsetstrokecolor{textcolor}%
\pgfsetfillcolor{textcolor}%
\pgftext[x=3.558539in,y=0.297570in,,top]{\color{textcolor}\sffamily\fontsize{10.000000}{12.000000}\selectfont 24}%
\end{pgfscope}%
\begin{pgfscope}%
\pgfpathrectangle{\pgfqpoint{0.536502in}{0.394792in}}{\pgfqpoint{6.421828in}{1.113538in}}%
\pgfusepath{clip}%
\pgfsetrectcap%
\pgfsetroundjoin%
\pgfsetlinewidth{0.803000pt}%
\definecolor{currentstroke}{rgb}{1.000000,1.000000,1.000000}%
\pgfsetstrokecolor{currentstroke}%
\pgfsetdash{}{0pt}%
\pgfpathmoveto{\pgfqpoint{3.810375in}{0.394792in}}%
\pgfpathlineto{\pgfqpoint{3.810375in}{1.508330in}}%
\pgfusepath{stroke}%
\end{pgfscope}%
\begin{pgfscope}%
\definecolor{textcolor}{rgb}{0.333333,0.333333,0.333333}%
\pgfsetstrokecolor{textcolor}%
\pgfsetfillcolor{textcolor}%
\pgftext[x=3.810375in,y=0.297570in,,top]{\color{textcolor}\sffamily\fontsize{10.000000}{12.000000}\selectfont 1}%
\end{pgfscope}%
\begin{pgfscope}%
\pgfpathrectangle{\pgfqpoint{0.536502in}{0.394792in}}{\pgfqpoint{6.421828in}{1.113538in}}%
\pgfusepath{clip}%
\pgfsetrectcap%
\pgfsetroundjoin%
\pgfsetlinewidth{0.803000pt}%
\definecolor{currentstroke}{rgb}{1.000000,1.000000,1.000000}%
\pgfsetstrokecolor{currentstroke}%
\pgfsetdash{}{0pt}%
\pgfpathmoveto{\pgfqpoint{6.832412in}{0.394792in}}%
\pgfpathlineto{\pgfqpoint{6.832412in}{1.508330in}}%
\pgfusepath{stroke}%
\end{pgfscope}%
\begin{pgfscope}%
\definecolor{textcolor}{rgb}{0.333333,0.333333,0.333333}%
\pgfsetstrokecolor{textcolor}%
\pgfsetfillcolor{textcolor}%
\pgftext[x=6.832412in,y=0.297570in,,top]{\color{textcolor}\sffamily\fontsize{10.000000}{12.000000}\selectfont 24}%
\end{pgfscope}%
\begin{pgfscope}%
\definecolor{textcolor}{rgb}{0.333333,0.333333,0.333333}%
\pgfsetstrokecolor{textcolor}%
\pgfsetfillcolor{textcolor}%
\pgftext[x=3.651089in,y=0.172085in,,top]{\color{textcolor}\sffamily\fontsize{10.000000}{12.000000}\selectfont Layer}%
\end{pgfscope}%
\begin{pgfscope}%
\pgfpathrectangle{\pgfqpoint{0.536502in}{0.394792in}}{\pgfqpoint{6.421828in}{1.113538in}}%
\pgfusepath{clip}%
\pgfsetrectcap%
\pgfsetroundjoin%
\pgfsetlinewidth{0.803000pt}%
\definecolor{currentstroke}{rgb}{1.000000,1.000000,1.000000}%
\pgfsetstrokecolor{currentstroke}%
\pgfsetdash{}{0pt}%
\pgfpathmoveto{\pgfqpoint{0.536502in}{0.492120in}}%
\pgfpathlineto{\pgfqpoint{6.958330in}{0.492120in}}%
\pgfusepath{stroke}%
\end{pgfscope}%
\begin{pgfscope}%
\definecolor{textcolor}{rgb}{0.333333,0.333333,0.333333}%
\pgfsetstrokecolor{textcolor}%
\pgfsetfillcolor{textcolor}%
\pgftext[x=0.041670in, y=0.439359in, left, base]{\color{textcolor}\sffamily\fontsize{10.000000}{12.000000}\selectfont 0.500}%
\end{pgfscope}%
\begin{pgfscope}%
\pgfpathrectangle{\pgfqpoint{0.536502in}{0.394792in}}{\pgfqpoint{6.421828in}{1.113538in}}%
\pgfusepath{clip}%
\pgfsetrectcap%
\pgfsetroundjoin%
\pgfsetlinewidth{0.803000pt}%
\definecolor{currentstroke}{rgb}{1.000000,1.000000,1.000000}%
\pgfsetstrokecolor{currentstroke}%
\pgfsetdash{}{0pt}%
\pgfpathmoveto{\pgfqpoint{0.536502in}{1.035289in}}%
\pgfpathlineto{\pgfqpoint{6.958330in}{1.035289in}}%
\pgfusepath{stroke}%
\end{pgfscope}%
\begin{pgfscope}%
\definecolor{textcolor}{rgb}{0.333333,0.333333,0.333333}%
\pgfsetstrokecolor{textcolor}%
\pgfsetfillcolor{textcolor}%
\pgftext[x=0.041670in, y=0.982527in, left, base]{\color{textcolor}\sffamily\fontsize{10.000000}{12.000000}\selectfont 0.600}%
\end{pgfscope}%
\begin{pgfscope}%
\pgfpathrectangle{\pgfqpoint{0.536502in}{0.394792in}}{\pgfqpoint{6.421828in}{1.113538in}}%
\pgfusepath{clip}%
\pgfsetrectcap%
\pgfsetroundjoin%
\pgfsetlinewidth{1.003750pt}%
\definecolor{currentstroke}{rgb}{0.886275,0.290196,0.200000}%
\pgfsetstrokecolor{currentstroke}%
\pgfsetdash{}{0pt}%
\pgfpathmoveto{\pgfqpoint{0.662421in}{0.445408in}}%
\pgfpathlineto{\pgfqpoint{0.788339in}{0.445408in}}%
\pgfpathlineto{\pgfqpoint{0.914257in}{0.445408in}}%
\pgfpathlineto{\pgfqpoint{1.040175in}{0.445408in}}%
\pgfpathlineto{\pgfqpoint{1.166093in}{0.445408in}}%
\pgfpathlineto{\pgfqpoint{1.292011in}{0.445408in}}%
\pgfpathlineto{\pgfqpoint{1.417930in}{0.445408in}}%
\pgfpathlineto{\pgfqpoint{1.543848in}{0.445408in}}%
\pgfpathlineto{\pgfqpoint{1.669766in}{0.445408in}}%
\pgfpathlineto{\pgfqpoint{1.795684in}{0.445408in}}%
\pgfpathlineto{\pgfqpoint{1.921602in}{0.445408in}}%
\pgfpathlineto{\pgfqpoint{2.047521in}{0.445408in}}%
\pgfusepath{stroke}%
\end{pgfscope}%
\begin{pgfscope}%
\pgfpathrectangle{\pgfqpoint{0.536502in}{0.394792in}}{\pgfqpoint{6.421828in}{1.113538in}}%
\pgfusepath{clip}%
\pgfsetrectcap%
\pgfsetroundjoin%
\pgfsetlinewidth{1.003750pt}%
\definecolor{currentstroke}{rgb}{0.886275,0.290196,0.200000}%
\pgfsetstrokecolor{currentstroke}%
\pgfsetdash{}{0pt}%
\pgfpathmoveto{\pgfqpoint{3.810375in}{0.445408in}}%
\pgfpathlineto{\pgfqpoint{3.936293in}{0.452104in}}%
\pgfpathlineto{\pgfqpoint{4.062212in}{0.445408in}}%
\pgfpathlineto{\pgfqpoint{4.188130in}{0.445408in}}%
\pgfpathlineto{\pgfqpoint{4.314048in}{0.445408in}}%
\pgfpathlineto{\pgfqpoint{4.439966in}{0.445408in}}%
\pgfpathlineto{\pgfqpoint{4.565884in}{0.452104in}}%
\pgfpathlineto{\pgfqpoint{4.691803in}{0.445408in}}%
\pgfpathlineto{\pgfqpoint{4.817721in}{0.445408in}}%
\pgfpathlineto{\pgfqpoint{4.943639in}{0.445408in}}%
\pgfpathlineto{\pgfqpoint{5.069557in}{0.445408in}}%
\pgfpathlineto{\pgfqpoint{5.195475in}{0.445408in}}%
\pgfusepath{stroke}%
\end{pgfscope}%
\begin{pgfscope}%
\pgfpathrectangle{\pgfqpoint{0.536502in}{0.394792in}}{\pgfqpoint{6.421828in}{1.113538in}}%
\pgfusepath{clip}%
\pgfsetrectcap%
\pgfsetroundjoin%
\pgfsetlinewidth{1.003750pt}%
\definecolor{currentstroke}{rgb}{0.203922,0.541176,0.741176}%
\pgfsetstrokecolor{currentstroke}%
\pgfsetdash{}{0pt}%
\pgfpathmoveto{\pgfqpoint{0.662421in}{0.445408in}}%
\pgfpathlineto{\pgfqpoint{0.788339in}{0.445408in}}%
\pgfpathlineto{\pgfqpoint{0.914257in}{0.445408in}}%
\pgfpathlineto{\pgfqpoint{1.040175in}{0.445408in}}%
\pgfpathlineto{\pgfqpoint{1.166093in}{0.445408in}}%
\pgfpathlineto{\pgfqpoint{1.292011in}{0.445408in}}%
\pgfpathlineto{\pgfqpoint{1.417930in}{0.445408in}}%
\pgfpathlineto{\pgfqpoint{1.543848in}{0.445408in}}%
\pgfpathlineto{\pgfqpoint{1.669766in}{0.445408in}}%
\pgfpathlineto{\pgfqpoint{1.795684in}{0.445408in}}%
\pgfpathlineto{\pgfqpoint{1.921602in}{0.445408in}}%
\pgfpathlineto{\pgfqpoint{2.047521in}{0.445408in}}%
\pgfusepath{stroke}%
\end{pgfscope}%
\begin{pgfscope}%
\pgfpathrectangle{\pgfqpoint{0.536502in}{0.394792in}}{\pgfqpoint{6.421828in}{1.113538in}}%
\pgfusepath{clip}%
\pgfsetrectcap%
\pgfsetroundjoin%
\pgfsetlinewidth{1.003750pt}%
\definecolor{currentstroke}{rgb}{0.203922,0.541176,0.741176}%
\pgfsetstrokecolor{currentstroke}%
\pgfsetdash{}{0pt}%
\pgfpathmoveto{\pgfqpoint{3.810375in}{0.445408in}}%
\pgfpathlineto{\pgfqpoint{3.936293in}{0.445408in}}%
\pgfpathlineto{\pgfqpoint{4.062212in}{0.445408in}}%
\pgfpathlineto{\pgfqpoint{4.188130in}{0.445408in}}%
\pgfpathlineto{\pgfqpoint{4.314048in}{0.445408in}}%
\pgfpathlineto{\pgfqpoint{4.439966in}{0.445408in}}%
\pgfpathlineto{\pgfqpoint{4.565884in}{0.445408in}}%
\pgfpathlineto{\pgfqpoint{4.691803in}{0.445408in}}%
\pgfpathlineto{\pgfqpoint{4.817721in}{0.445408in}}%
\pgfpathlineto{\pgfqpoint{4.943639in}{0.445408in}}%
\pgfpathlineto{\pgfqpoint{5.069557in}{0.445408in}}%
\pgfpathlineto{\pgfqpoint{5.195475in}{0.445408in}}%
\pgfusepath{stroke}%
\end{pgfscope}%
\begin{pgfscope}%
\pgfpathrectangle{\pgfqpoint{0.536502in}{0.394792in}}{\pgfqpoint{6.421828in}{1.113538in}}%
\pgfusepath{clip}%
\pgfsetrectcap%
\pgfsetroundjoin%
\pgfsetlinewidth{1.003750pt}%
\definecolor{currentstroke}{rgb}{0.596078,0.556863,0.835294}%
\pgfsetstrokecolor{currentstroke}%
\pgfsetdash{}{0pt}%
\pgfpathmoveto{\pgfqpoint{0.662421in}{0.445408in}}%
\pgfpathlineto{\pgfqpoint{0.788339in}{0.452104in}}%
\pgfpathlineto{\pgfqpoint{0.914257in}{0.445408in}}%
\pgfpathlineto{\pgfqpoint{1.040175in}{0.445408in}}%
\pgfpathlineto{\pgfqpoint{1.166093in}{0.476462in}}%
\pgfpathlineto{\pgfqpoint{1.292011in}{0.445408in}}%
\pgfpathlineto{\pgfqpoint{1.417930in}{0.445408in}}%
\pgfpathlineto{\pgfqpoint{1.543848in}{0.459064in}}%
\pgfpathlineto{\pgfqpoint{1.669766in}{0.511258in}}%
\pgfpathlineto{\pgfqpoint{1.795684in}{0.483421in}}%
\pgfpathlineto{\pgfqpoint{1.921602in}{0.521697in}}%
\pgfpathlineto{\pgfqpoint{2.047521in}{0.539095in}}%
\pgfpathlineto{\pgfqpoint{2.173439in}{0.500819in}}%
\pgfpathlineto{\pgfqpoint{2.299357in}{0.469503in}}%
\pgfpathlineto{\pgfqpoint{2.425275in}{0.452104in}}%
\pgfpathlineto{\pgfqpoint{2.551193in}{0.445408in}}%
\pgfusepath{stroke}%
\end{pgfscope}%
\begin{pgfscope}%
\pgfpathrectangle{\pgfqpoint{0.536502in}{0.394792in}}{\pgfqpoint{6.421828in}{1.113538in}}%
\pgfusepath{clip}%
\pgfsetrectcap%
\pgfsetroundjoin%
\pgfsetlinewidth{1.003750pt}%
\definecolor{currentstroke}{rgb}{0.596078,0.556863,0.835294}%
\pgfsetstrokecolor{currentstroke}%
\pgfsetdash{}{0pt}%
\pgfpathmoveto{\pgfqpoint{3.810375in}{0.445408in}}%
\pgfpathlineto{\pgfqpoint{3.936293in}{0.448625in}}%
\pgfpathlineto{\pgfqpoint{4.062212in}{0.445408in}}%
\pgfpathlineto{\pgfqpoint{4.188130in}{0.459064in}}%
\pgfpathlineto{\pgfqpoint{4.314048in}{0.490380in}}%
\pgfpathlineto{\pgfqpoint{4.439966in}{0.539095in}}%
\pgfpathlineto{\pgfqpoint{4.565884in}{0.605208in}}%
\pgfpathlineto{\pgfqpoint{4.691803in}{0.640004in}}%
\pgfpathlineto{\pgfqpoint{4.817721in}{0.643484in}}%
\pgfpathlineto{\pgfqpoint{4.943639in}{0.664361in}}%
\pgfpathlineto{\pgfqpoint{5.069557in}{0.674800in}}%
\pgfpathlineto{\pgfqpoint{5.195475in}{0.744393in}}%
\pgfpathlineto{\pgfqpoint{5.321394in}{0.768750in}}%
\pgfpathlineto{\pgfqpoint{5.447312in}{0.855740in}}%
\pgfpathlineto{\pgfqpoint{5.573230in}{1.325489in}}%
\pgfpathlineto{\pgfqpoint{5.699148in}{1.457715in}}%
\pgfusepath{stroke}%
\end{pgfscope}%
\begin{pgfscope}%
\pgfpathrectangle{\pgfqpoint{0.536502in}{0.394792in}}{\pgfqpoint{6.421828in}{1.113538in}}%
\pgfusepath{clip}%
\pgfsetrectcap%
\pgfsetroundjoin%
\pgfsetlinewidth{1.003750pt}%
\definecolor{currentstroke}{rgb}{0.466667,0.466667,0.466667}%
\pgfsetstrokecolor{currentstroke}%
\pgfsetdash{}{0pt}%
\pgfpathmoveto{\pgfqpoint{0.662421in}{0.452104in}}%
\pgfpathlineto{\pgfqpoint{0.788339in}{0.448625in}}%
\pgfpathlineto{\pgfqpoint{0.914257in}{0.448625in}}%
\pgfpathlineto{\pgfqpoint{1.040175in}{0.445408in}}%
\pgfpathlineto{\pgfqpoint{1.166093in}{0.486901in}}%
\pgfpathlineto{\pgfqpoint{1.292011in}{0.472982in}}%
\pgfpathlineto{\pgfqpoint{1.417930in}{0.476462in}}%
\pgfpathlineto{\pgfqpoint{1.543848in}{0.483421in}}%
\pgfpathlineto{\pgfqpoint{1.669766in}{0.445408in}}%
\pgfpathlineto{\pgfqpoint{1.795684in}{0.479941in}}%
\pgfpathlineto{\pgfqpoint{1.921602in}{0.445408in}}%
\pgfpathlineto{\pgfqpoint{2.047521in}{0.459064in}}%
\pgfpathlineto{\pgfqpoint{2.173439in}{0.455584in}}%
\pgfpathlineto{\pgfqpoint{2.299357in}{0.445408in}}%
\pgfpathlineto{\pgfqpoint{2.425275in}{0.472982in}}%
\pgfpathlineto{\pgfqpoint{2.551193in}{0.445408in}}%
\pgfpathlineto{\pgfqpoint{2.677112in}{0.445408in}}%
\pgfpathlineto{\pgfqpoint{2.803030in}{0.445408in}}%
\pgfpathlineto{\pgfqpoint{2.928948in}{0.445408in}}%
\pgfpathlineto{\pgfqpoint{3.054866in}{0.445408in}}%
\pgfpathlineto{\pgfqpoint{3.180784in}{0.445408in}}%
\pgfpathlineto{\pgfqpoint{3.306703in}{0.445408in}}%
\pgfpathlineto{\pgfqpoint{3.432621in}{0.445408in}}%
\pgfpathlineto{\pgfqpoint{3.558539in}{0.445408in}}%
\pgfusepath{stroke}%
\end{pgfscope}%
\begin{pgfscope}%
\pgfpathrectangle{\pgfqpoint{0.536502in}{0.394792in}}{\pgfqpoint{6.421828in}{1.113538in}}%
\pgfusepath{clip}%
\pgfsetrectcap%
\pgfsetroundjoin%
\pgfsetlinewidth{1.003750pt}%
\definecolor{currentstroke}{rgb}{0.466667,0.466667,0.466667}%
\pgfsetstrokecolor{currentstroke}%
\pgfsetdash{}{0pt}%
\pgfpathmoveto{\pgfqpoint{3.810375in}{0.452104in}}%
\pgfpathlineto{\pgfqpoint{3.936293in}{0.448625in}}%
\pgfpathlineto{\pgfqpoint{4.062212in}{0.445408in}}%
\pgfpathlineto{\pgfqpoint{4.188130in}{0.445408in}}%
\pgfpathlineto{\pgfqpoint{4.314048in}{0.445408in}}%
\pgfpathlineto{\pgfqpoint{4.439966in}{0.445408in}}%
\pgfpathlineto{\pgfqpoint{4.565884in}{0.497340in}}%
\pgfpathlineto{\pgfqpoint{4.691803in}{0.542575in}}%
\pgfpathlineto{\pgfqpoint{4.817721in}{0.542575in}}%
\pgfpathlineto{\pgfqpoint{4.943639in}{0.500819in}}%
\pgfpathlineto{\pgfqpoint{5.069557in}{0.511258in}}%
\pgfpathlineto{\pgfqpoint{5.195475in}{0.507778in}}%
\pgfpathlineto{\pgfqpoint{5.321394in}{0.549534in}}%
\pgfpathlineto{\pgfqpoint{5.447312in}{0.640004in}}%
\pgfpathlineto{\pgfqpoint{5.573230in}{0.692198in}}%
\pgfpathlineto{\pgfqpoint{5.699148in}{0.723515in}}%
\pgfpathlineto{\pgfqpoint{5.825066in}{0.786148in}}%
\pgfpathlineto{\pgfqpoint{5.950984in}{0.869659in}}%
\pgfpathlineto{\pgfqpoint{6.076903in}{0.869659in}}%
\pgfpathlineto{\pgfqpoint{6.202821in}{0.890537in}}%
\pgfpathlineto{\pgfqpoint{6.328739in}{0.918374in}}%
\pgfpathlineto{\pgfqpoint{6.454657in}{0.932292in}}%
\pgfpathlineto{\pgfqpoint{6.580575in}{1.109753in}}%
\pgfpathlineto{\pgfqpoint{6.706494in}{1.228060in}}%
\pgfusepath{stroke}%
\end{pgfscope}%
\begin{pgfscope}%
\pgfpathrectangle{\pgfqpoint{0.536502in}{0.394792in}}{\pgfqpoint{6.421828in}{1.113538in}}%
\pgfusepath{clip}%
\pgfsetbuttcap%
\pgfsetroundjoin%
\pgfsetlinewidth{1.003750pt}%
\definecolor{currentstroke}{rgb}{0.000000,0.392157,0.000000}%
\pgfsetstrokecolor{currentstroke}%
\pgfsetdash{{3.700000pt}{1.600000pt}}{0.000000pt}%
\pgfpathmoveto{\pgfqpoint{0.536502in}{0.445408in}}%
\pgfpathlineto{\pgfqpoint{6.958330in}{0.445408in}}%
\pgfusepath{stroke}%
\end{pgfscope}%
\begin{pgfscope}%
\pgfpathrectangle{\pgfqpoint{0.536502in}{0.394792in}}{\pgfqpoint{6.421828in}{1.113538in}}%
\pgfusepath{clip}%
\pgfsetbuttcap%
\pgfsetroundjoin%
\pgfsetlinewidth{1.003750pt}%
\definecolor{currentstroke}{rgb}{0.803922,0.521569,0.247059}%
\pgfsetstrokecolor{currentstroke}%
\pgfsetdash{{3.700000pt}{1.600000pt}}{0.000000pt}%
\pgfpathmoveto{\pgfqpoint{0.536502in}{1.112962in}}%
\pgfpathlineto{\pgfqpoint{6.958330in}{1.112962in}}%
\pgfusepath{stroke}%
\end{pgfscope}%
\begin{pgfscope}%
\pgfpathrectangle{\pgfqpoint{0.536502in}{0.394792in}}{\pgfqpoint{6.421828in}{1.113538in}}%
\pgfusepath{clip}%
\pgfsetrectcap%
\pgfsetroundjoin%
\pgfsetlinewidth{6.022500pt}%
\definecolor{currentstroke}{rgb}{1.000000,1.000000,1.000000}%
\pgfsetstrokecolor{currentstroke}%
\pgfsetdash{}{0pt}%
\pgfpathmoveto{\pgfqpoint{3.684457in}{0.394792in}}%
\pgfpathlineto{\pgfqpoint{3.684457in}{1.508330in}}%
\pgfusepath{stroke}%
\end{pgfscope}%
\begin{pgfscope}%
\pgfsetrectcap%
\pgfsetmiterjoin%
\pgfsetlinewidth{1.003750pt}%
\definecolor{currentstroke}{rgb}{1.000000,1.000000,1.000000}%
\pgfsetstrokecolor{currentstroke}%
\pgfsetdash{}{0pt}%
\pgfpathmoveto{\pgfqpoint{0.536502in}{0.394792in}}%
\pgfpathlineto{\pgfqpoint{0.536502in}{1.508330in}}%
\pgfusepath{stroke}%
\end{pgfscope}%
\begin{pgfscope}%
\pgfsetrectcap%
\pgfsetmiterjoin%
\pgfsetlinewidth{1.003750pt}%
\definecolor{currentstroke}{rgb}{1.000000,1.000000,1.000000}%
\pgfsetstrokecolor{currentstroke}%
\pgfsetdash{}{0pt}%
\pgfpathmoveto{\pgfqpoint{6.958330in}{0.394792in}}%
\pgfpathlineto{\pgfqpoint{6.958330in}{1.508330in}}%
\pgfusepath{stroke}%
\end{pgfscope}%
\begin{pgfscope}%
\pgfsetrectcap%
\pgfsetmiterjoin%
\pgfsetlinewidth{1.003750pt}%
\definecolor{currentstroke}{rgb}{1.000000,1.000000,1.000000}%
\pgfsetstrokecolor{currentstroke}%
\pgfsetdash{}{0pt}%
\pgfpathmoveto{\pgfqpoint{0.536502in}{0.394792in}}%
\pgfpathlineto{\pgfqpoint{6.958330in}{0.394792in}}%
\pgfusepath{stroke}%
\end{pgfscope}%
\begin{pgfscope}%
\pgfsetrectcap%
\pgfsetmiterjoin%
\pgfsetlinewidth{1.003750pt}%
\definecolor{currentstroke}{rgb}{1.000000,1.000000,1.000000}%
\pgfsetstrokecolor{currentstroke}%
\pgfsetdash{}{0pt}%
\pgfpathmoveto{\pgfqpoint{0.536502in}{1.508330in}}%
\pgfpathlineto{\pgfqpoint{6.958330in}{1.508330in}}%
\pgfusepath{stroke}%
\end{pgfscope}%
\end{pgfpicture}%
\makeatother%
\endgroup%

%% file: figures/comparison-legend.pgf
\begingroup%
\makeatletter%
\begin{pgfpicture}%
\pgfpathrectangle{\pgfpointorigin}{\pgfqpoint{7.000000in}{0.500000in}}%
\pgfusepath{use as bounding box, clip}%
\begin{pgfscope}%
\pgfsetbuttcap%
\pgfsetmiterjoin%
\definecolor{currentfill}{rgb}{1.000000,1.000000,1.000000}%
\pgfsetfillcolor{currentfill}%
\pgfsetlinewidth{0.000000pt}%
\definecolor{currentstroke}{rgb}{0.500000,0.500000,0.500000}%
\pgfsetstrokecolor{currentstroke}%
\pgfsetdash{}{0pt}%
\pgfpathmoveto{\pgfqpoint{0.000000in}{0.000000in}}%
\pgfpathlineto{\pgfqpoint{7.000000in}{0.000000in}}%
\pgfpathlineto{\pgfqpoint{7.000000in}{0.500000in}}%
\pgfpathlineto{\pgfqpoint{0.000000in}{0.500000in}}%
\pgfpathlineto{\pgfqpoint{0.000000in}{0.000000in}}%
\pgfpathclose%
\pgfusepath{fill}%
\end{pgfscope}%
\begin{pgfscope}%
\pgfsetbuttcap%
\pgfsetmiterjoin%
\definecolor{currentfill}{rgb}{1.000000,0.960784,0.949020}%
\pgfsetfillcolor{currentfill}%
\pgfsetfillopacity{0.800000}%
\pgfsetlinewidth{0.501875pt}%
\definecolor{currentstroke}{rgb}{0.800000,0.800000,0.800000}%
\pgfsetstrokecolor{currentstroke}%
\pgfsetstrokeopacity{0.800000}%
\pgfsetdash{}{0pt}%
\pgfpathmoveto{\pgfqpoint{2.014638in}{0.070248in}}%
\pgfpathlineto{\pgfqpoint{4.985362in}{0.070248in}}%
\pgfpathquadraticcurveto{\pgfqpoint{5.007585in}{0.070248in}}{\pgfqpoint{5.007585in}{0.092470in}}%
\pgfpathlineto{\pgfqpoint{5.007585in}{0.407530in}}%
\pgfpathquadraticcurveto{\pgfqpoint{5.007585in}{0.429752in}}{\pgfqpoint{4.985362in}{0.429752in}}%
\pgfpathlineto{\pgfqpoint{2.014638in}{0.429752in}}%
\pgfpathquadraticcurveto{\pgfqpoint{1.992415in}{0.429752in}}{\pgfqpoint{1.992415in}{0.407530in}}%
\pgfpathlineto{\pgfqpoint{1.992415in}{0.092470in}}%
\pgfpathquadraticcurveto{\pgfqpoint{1.992415in}{0.070248in}}{\pgfqpoint{2.014638in}{0.070248in}}%
\pgfpathlineto{\pgfqpoint{2.014638in}{0.070248in}}%
\pgfpathclose%
\pgfusepath{stroke,fill}%
\end{pgfscope}%
\begin{pgfscope}%
\pgfsetbuttcap%
\pgfsetroundjoin%
\pgfsetlinewidth{1.003750pt}%
\definecolor{currentstroke}{rgb}{0.000000,0.392157,0.000000}%
\pgfsetstrokecolor{currentstroke}%
\pgfsetdash{{3.700000pt}{1.600000pt}}{0.000000pt}%
\pgfpathmoveto{\pgfqpoint{2.036860in}{0.339778in}}%
\pgfpathlineto{\pgfqpoint{2.147971in}{0.339778in}}%
\pgfpathlineto{\pgfqpoint{2.259082in}{0.339778in}}%
\pgfusepath{stroke}%
\end{pgfscope}%
\begin{pgfscope}%
\definecolor{textcolor}{rgb}{0.000000,0.000000,0.000000}%
\pgfsetstrokecolor{textcolor}%
\pgfsetfillcolor{textcolor}%
\pgftext[x=2.347971in,y=0.300890in,left,base]{\color{textcolor}\sffamily\fontsize{8.000000}{9.600000}\selectfont Baseline}%
\end{pgfscope}%
\begin{pgfscope}%
\pgfsetbuttcap%
\pgfsetroundjoin%
\pgfsetlinewidth{1.003750pt}%
\definecolor{currentstroke}{rgb}{0.803922,0.521569,0.247059}%
\pgfsetstrokecolor{currentstroke}%
\pgfsetdash{{3.700000pt}{1.600000pt}}{0.000000pt}%
\pgfpathmoveto{\pgfqpoint{2.036860in}{0.176693in}}%
\pgfpathlineto{\pgfqpoint{2.147971in}{0.176693in}}%
\pgfpathlineto{\pgfqpoint{2.259082in}{0.176693in}}%
\pgfusepath{stroke}%
\end{pgfscope}%
\begin{pgfscope}%
\definecolor{textcolor}{rgb}{0.000000,0.000000,0.000000}%
\pgfsetstrokecolor{textcolor}%
\pgfsetfillcolor{textcolor}%
\pgftext[x=2.347971in,y=0.137804in,left,base]{\color{textcolor}\sffamily\fontsize{8.000000}{9.600000}\selectfont I3D}%
\end{pgfscope}%
\begin{pgfscope}%
\pgfsetrectcap%
\pgfsetroundjoin%
\pgfsetlinewidth{1.003750pt}%
\definecolor{currentstroke}{rgb}{0.886275,0.290196,0.200000}%
\pgfsetstrokecolor{currentstroke}%
\pgfsetdash{}{0pt}%
\pgfpathmoveto{\pgfqpoint{3.041276in}{0.339778in}}%
\pgfpathlineto{\pgfqpoint{3.152387in}{0.339778in}}%
\pgfpathlineto{\pgfqpoint{3.263498in}{0.339778in}}%
\pgfusepath{stroke}%
\end{pgfscope}%
\begin{pgfscope}%
\definecolor{textcolor}{rgb}{0.000000,0.000000,0.000000}%
\pgfsetstrokecolor{textcolor}%
\pgfsetfillcolor{textcolor}%
\pgftext[x=3.352387in,y=0.300890in,left,base]{\color{textcolor}\sffamily\fontsize{8.000000}{9.600000}\selectfont VideoMAE}%
\end{pgfscope}%
\begin{pgfscope}%
\pgfsetrectcap%
\pgfsetroundjoin%
\pgfsetlinewidth{1.003750pt}%
\definecolor{currentstroke}{rgb}{0.203922,0.541176,0.741176}%
\pgfsetstrokecolor{currentstroke}%
\pgfsetdash{}{0pt}%
\pgfpathmoveto{\pgfqpoint{3.041276in}{0.176693in}}%
\pgfpathlineto{\pgfqpoint{3.152387in}{0.176693in}}%
\pgfpathlineto{\pgfqpoint{3.263498in}{0.176693in}}%
\pgfusepath{stroke}%
\end{pgfscope}%
\begin{pgfscope}%
\definecolor{textcolor}{rgb}{0.000000,0.000000,0.000000}%
\pgfsetstrokecolor{textcolor}%
\pgfsetfillcolor{textcolor}%
\pgftext[x=3.352387in,y=0.137804in,left,base]{\color{textcolor}\sffamily\fontsize{8.000000}{9.600000}\selectfont SVT}%
\end{pgfscope}%
\begin{pgfscope}%
\pgfsetrectcap%
\pgfsetroundjoin%
\pgfsetlinewidth{1.003750pt}%
\definecolor{currentstroke}{rgb}{0.596078,0.556863,0.835294}%
\pgfsetstrokecolor{currentstroke}%
\pgfsetdash{}{0pt}%
\pgfpathmoveto{\pgfqpoint{4.127995in}{0.339778in}}%
\pgfpathlineto{\pgfqpoint{4.239106in}{0.339778in}}%
\pgfpathlineto{\pgfqpoint{4.350217in}{0.339778in}}%
\pgfusepath{stroke}%
\end{pgfscope}%
\begin{pgfscope}%
\definecolor{textcolor}{rgb}{0.000000,0.000000,0.000000}%
\pgfsetstrokecolor{textcolor}%
\pgfsetfillcolor{textcolor}%
\pgftext[x=4.439106in,y=0.300890in,left,base]{\color{textcolor}\sffamily\fontsize{8.000000}{9.600000}\selectfont MaskFeat}%
\end{pgfscope}%
\begin{pgfscope}%
\pgfsetrectcap%
\pgfsetroundjoin%
\pgfsetlinewidth{1.003750pt}%
\definecolor{currentstroke}{rgb}{0.466667,0.466667,0.466667}%
\pgfsetstrokecolor{currentstroke}%
\pgfsetdash{}{0pt}%
\pgfpathmoveto{\pgfqpoint{4.127995in}{0.176693in}}%
\pgfpathlineto{\pgfqpoint{4.239106in}{0.176693in}}%
\pgfpathlineto{\pgfqpoint{4.350217in}{0.176693in}}%
\pgfusepath{stroke}%
\end{pgfscope}%
\begin{pgfscope}%
\definecolor{textcolor}{rgb}{0.000000,0.000000,0.000000}%
\pgfsetstrokecolor{textcolor}%
\pgfsetfillcolor{textcolor}%
\pgftext[x=4.439106in,y=0.137804in,left,base]{\color{textcolor}\sffamily\fontsize{8.000000}{9.600000}\selectfont BEVT}%
\end{pgfscope}%
\end{pgfpicture}%
\makeatother%
\endgroup%

%% file: sections/conclusion.tex
\section{Conclusion}
\label{sec:conclusion}

In this work, we leverage the combined power of self-supervision and transformer-based architectures for the task of isolated sign language recognition and probe them for phonological features. Despite the small amount of data available for ISLR, self-supervised transformers produce state-of-the-art results on gloss-based WLASL2000, by leveraging continuous sign language datasets for pre-training. Probing for sign language phonological features helps us better characterize video models and pre-training tasks. For example, plain ViTs struggle at capturing fine-grained spatial relationships. Self-supervised hierarchical vision transformers behave similarly to self-supervised speech models in terms of layer-wise encoding of linguistic information.

\textbf{Future Work.} Since ISLR is a very limited task in the context of sign language understanding, a natural next step is to extend these models for other sign language tasks, most importantly sign language translation. The biggest challenge in this direction is to extend transformer-based models to more than 16 frames, which is crucial in any continuous sign language task, but requires an immense amount of compute with current attention implementations. As we have seen, probing for ASL phonological features reveals strengths and limitations of various pre-training tasks and architecture choices. Future work on self-supervised sign language models can use these features to better characterize their models, and to further expand our understanding of video models beyond action recognition and detection.

%% file: sections/appendix.tex
\section{Results on Original WLASL2000}
\label{sec:appendix-old-wlasl2000}

Though we emphasize the importance of gloss-based labels for WLASL2000, we include results on the original WLASL2000 labels to show that our results hold in that setting as well. Table~\ref{tab:old-wlasl-perf} compares our models with results in the literature that (1) do not use additional signals during training and (2) use datasets that are publicly available. Additionally, we report results from training MViTv2 on Kinetics with supervision and then fine-tuned for WLASL2000. Our MViTv2 trained with MaskFeat outperforms all video models, second to SAM-SLR~\cite{jiang2021skeleton}, which uses pose.

\begin{table}[]
\centering
\begin{tabular}{llc}
\hline
\textbf{Model} & \textbf{Pre-Training} & \textbf{Top 1 Acc.} \\
\hline
I3D & Sup. Kinetics & 32.48 \\
VideoSwin~\cite{Novopoltsev2023Finetuning} & Sup. Kinetics & 44.58 \\
I3D & Sup. BSL-1K & 46.82 \\
MViTv2 & Sup. Kinetics & 52.43 \\
MViTv2 & Two-Stage MaskFeat & 52.83 \\
SAM-SLR & Pose-based & 58.73 \\
\hline
\end{tabular}	
\caption{Our model, MViTv2 pre-trained using two-stage MaskFeat, compared to supervised methods on the original WLASL2000 dataset.}
\label{tab:old-wlasl-perf}
\end{table}

\section{Linear Probing Graphs: Pre-Training vs. Fine-Tuning}
\label{sec:appendix-comparison-graphs}

Figure~\ref{fig:all-comparison} contains all our figures for linear probing results before and after fine-tuning.

\begin{figure*}
    \caption*{\hspace{0.5in} Before Fine-Tuning \hspace{2in} After Fine-Tuning}
    \begin{subfigure}[b]{\linewidth}
        \centering
        \renewcommand\sffamily{}
        \input{figures/comparison-signtype-nox.pgf}
        \caption{Sign Type}
        \label{fig:all-comparison-signtype}
    \end{subfigure}

    \begin{subfigure}[b]{\linewidth}
        \centering
        \renewcommand\sffamily{}
        \input{figures/comparison-majorlocation-nox.pgf}
        \caption{Major Location}
        \label{fig:all-comparison-majorlocation}
    \end{subfigure}

    \begin{subfigure}[b]{\linewidth}
        \centering
        \renewcommand\sffamily{}
        \input{figures/comparison-minorlocation-nox.pgf}
        \caption{Minor Location}
        \label{fig:all-comparison-minorlocation}
    \end{subfigure}

    \begin{subfigure}[b]{\linewidth}
        \centering
        \renewcommand\sffamily{}
        \input{figures/comparison-secondminorlocation-nox.pgf}
        \caption{Second Minor Location}
        \label{fig:all-comparison-secondminorlocation}
    \end{subfigure}

    \begin{subfigure}[b]{\linewidth}
        \centering
        \renewcommand\sffamily{}
        \input{figures/comparison-contact.pgf}
        \caption{Contact}
        \label{fig:all-comparison-contact}
    \end{subfigure}
\label{fig:all-comparison}
\end{figure*}
\begin{figure*}\ContinuedFloat
    \caption*{\hspace{0.5in} Before Fine-Tuning \hspace{2in} After Fine-Tuning}
    \begin{subfigure}[b]{\linewidth}
        \centering
        \renewcommand\sffamily{}
        \input{figures/comparison-handshape-nox.pgf}
        \caption{Handshape}
        \label{fig:all-comparison-handshape}
    \end{subfigure}

    \begin{subfigure}[b]{\linewidth}
        \centering
        \renewcommand\sffamily{}
        \input{figures/comparison-nondominanthandshape-nox.pgf}
        \caption{Non-Dominant Handshape}
        \label{fig:all-comparison-nondominanthandshape}
    \end{subfigure}

    \begin{subfigure}[b]{\linewidth}
        \centering
        \renewcommand\sffamily{}
        \input{figures/comparison-selectedfinger-nox.pgf}
        \caption{Selected Fingers}
        \label{fig:all-comparison-selectedfinger}
    \end{subfigure}

    \begin{subfigure}[b]{\linewidth}
        \centering
        \renewcommand\sffamily{}
        \input{figures/comparison-flexion-nox.pgf}
        \caption{Flexion}
        \label{fig:all-comparison-flexion}
    \end{subfigure}

    \begin{subfigure}[b]{\linewidth}
        \centering
        \renewcommand\sffamily{}
        \input{figures/comparison-thumbposition.pgf}
        \caption{Thumb Position}
        \label{fig:all-comparison-thumbposition}
    \end{subfigure}
\end{figure*}
\begin{figure*}[t!]\ContinuedFloat
    \caption*{\hspace{0.5in} Before Fine-Tuning \hspace{2in} After Fine-Tuning}

    \begin{subfigure}[b]{\linewidth}
        \centering
        \renewcommand\sffamily{}
        \input{figures/comparison-pathmovement-nox.pgf}
        \caption{Path Movement}
        \label{fig:all-comparison-pathmovement}
    \end{subfigure}

    \begin{subfigure}[b]{\linewidth}
        \centering
        \renewcommand\sffamily{}
        \input{figures/comparison-wristtwist-nox.pgf}
        \caption{Wrist Twist}
        \label{fig:all-comparison-wristtwist}
    \end{subfigure}
\label{fig:all-comparison}
    \begin{subfigure}[b]{\linewidth}
        \centering
        \renewcommand\sffamily{}
        \input{figures/comparison-repeatedmovement.pgf}
        \caption{Repeated Movement}
        \label{fig:all-comparison-repeatedmovement}
    \end{subfigure}
    \centering
    \renewcommand\sffamily{}
    \input{figures/comparison-legend.pgf}
    \caption{Comparison between pre-trained and fine-tuned models on all phonological features. Figures on the left are before fine-tuning and figures on the right are after fine-tuning.}
    \label{fig:all-comparison}
\end{figure*}

%% file: figures/comparison-signtype-nox.pgf
\begingroup%
\makeatletter%
\begin{pgfpicture}%
\pgfpathrectangle{\pgfpointorigin}{\pgfqpoint{7.000000in}{1.250000in}}%
\pgfusepath{use as bounding box, clip}%
\begin{pgfscope}%
\pgfsetbuttcap%
\pgfsetmiterjoin%
\definecolor{currentfill}{rgb}{1.000000,1.000000,1.000000}%
\pgfsetfillcolor{currentfill}%
\pgfsetlinewidth{0.000000pt}%
\definecolor{currentstroke}{rgb}{0.500000,0.500000,0.500000}%
\pgfsetstrokecolor{currentstroke}%
\pgfsetdash{}{0pt}%
\pgfpathmoveto{\pgfqpoint{0.000000in}{0.000000in}}%
\pgfpathlineto{\pgfqpoint{7.000000in}{0.000000in}}%
\pgfpathlineto{\pgfqpoint{7.000000in}{1.250000in}}%
\pgfpathlineto{\pgfqpoint{0.000000in}{1.250000in}}%
\pgfpathlineto{\pgfqpoint{0.000000in}{0.000000in}}%
\pgfpathclose%
\pgfusepath{fill}%
\end{pgfscope}%
\begin{pgfscope}%
\pgfsetbuttcap%
\pgfsetmiterjoin%
\definecolor{currentfill}{rgb}{0.898039,0.898039,0.898039}%
\pgfsetfillcolor{currentfill}%
\pgfsetlinewidth{0.000000pt}%
\definecolor{currentstroke}{rgb}{0.000000,0.000000,0.000000}%
\pgfsetstrokecolor{currentstroke}%
\pgfsetstrokeopacity{0.000000}%
\pgfsetdash{}{0pt}%
\pgfpathmoveto{\pgfqpoint{0.536502in}{0.050071in}}%
\pgfpathlineto{\pgfqpoint{6.958330in}{0.050071in}}%
\pgfpathlineto{\pgfqpoint{6.958330in}{1.208330in}}%
\pgfpathlineto{\pgfqpoint{0.536502in}{1.208330in}}%
\pgfpathlineto{\pgfqpoint{0.536502in}{0.050071in}}%
\pgfpathclose%
\pgfusepath{fill}%
\end{pgfscope}%
\begin{pgfscope}%
\pgfpathrectangle{\pgfqpoint{0.536502in}{0.050071in}}{\pgfqpoint{6.421828in}{1.158259in}}%
\pgfusepath{clip}%
\pgfsetrectcap%
\pgfsetroundjoin%
\pgfsetlinewidth{0.803000pt}%
\definecolor{currentstroke}{rgb}{1.000000,1.000000,1.000000}%
\pgfsetstrokecolor{currentstroke}%
\pgfsetdash{}{0pt}%
\pgfpathmoveto{\pgfqpoint{0.662421in}{0.050071in}}%
\pgfpathlineto{\pgfqpoint{0.662421in}{1.208330in}}%
\pgfusepath{stroke}%
\end{pgfscope}%
\begin{pgfscope}%
\pgfpathrectangle{\pgfqpoint{0.536502in}{0.050071in}}{\pgfqpoint{6.421828in}{1.158259in}}%
\pgfusepath{clip}%
\pgfsetrectcap%
\pgfsetroundjoin%
\pgfsetlinewidth{0.803000pt}%
\definecolor{currentstroke}{rgb}{1.000000,1.000000,1.000000}%
\pgfsetstrokecolor{currentstroke}%
\pgfsetdash{}{0pt}%
\pgfpathmoveto{\pgfqpoint{3.558539in}{0.050071in}}%
\pgfpathlineto{\pgfqpoint{3.558539in}{1.208330in}}%
\pgfusepath{stroke}%
\end{pgfscope}%
\begin{pgfscope}%
\pgfpathrectangle{\pgfqpoint{0.536502in}{0.050071in}}{\pgfqpoint{6.421828in}{1.158259in}}%
\pgfusepath{clip}%
\pgfsetrectcap%
\pgfsetroundjoin%
\pgfsetlinewidth{0.803000pt}%
\definecolor{currentstroke}{rgb}{1.000000,1.000000,1.000000}%
\pgfsetstrokecolor{currentstroke}%
\pgfsetdash{}{0pt}%
\pgfpathmoveto{\pgfqpoint{3.810375in}{0.050071in}}%
\pgfpathlineto{\pgfqpoint{3.810375in}{1.208330in}}%
\pgfusepath{stroke}%
\end{pgfscope}%
\begin{pgfscope}%
\pgfpathrectangle{\pgfqpoint{0.536502in}{0.050071in}}{\pgfqpoint{6.421828in}{1.158259in}}%
\pgfusepath{clip}%
\pgfsetrectcap%
\pgfsetroundjoin%
\pgfsetlinewidth{0.803000pt}%
\definecolor{currentstroke}{rgb}{1.000000,1.000000,1.000000}%
\pgfsetstrokecolor{currentstroke}%
\pgfsetdash{}{0pt}%
\pgfpathmoveto{\pgfqpoint{6.832412in}{0.050071in}}%
\pgfpathlineto{\pgfqpoint{6.832412in}{1.208330in}}%
\pgfusepath{stroke}%
\end{pgfscope}%
\begin{pgfscope}%
\pgfpathrectangle{\pgfqpoint{0.536502in}{0.050071in}}{\pgfqpoint{6.421828in}{1.158259in}}%
\pgfusepath{clip}%
\pgfsetrectcap%
\pgfsetroundjoin%
\pgfsetlinewidth{0.803000pt}%
\definecolor{currentstroke}{rgb}{1.000000,1.000000,1.000000}%
\pgfsetstrokecolor{currentstroke}%
\pgfsetdash{}{0pt}%
\pgfpathmoveto{\pgfqpoint{0.536502in}{0.124072in}}%
\pgfpathlineto{\pgfqpoint{6.958330in}{0.124072in}}%
\pgfusepath{stroke}%
\end{pgfscope}%
\begin{pgfscope}%
\definecolor{textcolor}{rgb}{0.333333,0.333333,0.333333}%
\pgfsetstrokecolor{textcolor}%
\pgfsetfillcolor{textcolor}%
\pgftext[x=0.041670in, y=0.071311in, left, base]{\color{textcolor}\sffamily\fontsize{10.000000}{12.000000}\selectfont 0.400}%
\end{pgfscope}%
\begin{pgfscope}%
\pgfpathrectangle{\pgfqpoint{0.536502in}{0.050071in}}{\pgfqpoint{6.421828in}{1.158259in}}%
\pgfusepath{clip}%
\pgfsetrectcap%
\pgfsetroundjoin%
\pgfsetlinewidth{0.803000pt}%
\definecolor{currentstroke}{rgb}{1.000000,1.000000,1.000000}%
\pgfsetstrokecolor{currentstroke}%
\pgfsetdash{}{0pt}%
\pgfpathmoveto{\pgfqpoint{0.536502in}{0.555453in}}%
\pgfpathlineto{\pgfqpoint{6.958330in}{0.555453in}}%
\pgfusepath{stroke}%
\end{pgfscope}%
\begin{pgfscope}%
\definecolor{textcolor}{rgb}{0.333333,0.333333,0.333333}%
\pgfsetstrokecolor{textcolor}%
\pgfsetfillcolor{textcolor}%
\pgftext[x=0.041670in, y=0.502691in, left, base]{\color{textcolor}\sffamily\fontsize{10.000000}{12.000000}\selectfont 0.600}%
\end{pgfscope}%
\begin{pgfscope}%
\pgfpathrectangle{\pgfqpoint{0.536502in}{0.050071in}}{\pgfqpoint{6.421828in}{1.158259in}}%
\pgfusepath{clip}%
\pgfsetrectcap%
\pgfsetroundjoin%
\pgfsetlinewidth{0.803000pt}%
\definecolor{currentstroke}{rgb}{1.000000,1.000000,1.000000}%
\pgfsetstrokecolor{currentstroke}%
\pgfsetdash{}{0pt}%
\pgfpathmoveto{\pgfqpoint{0.536502in}{0.986833in}}%
\pgfpathlineto{\pgfqpoint{6.958330in}{0.986833in}}%
\pgfusepath{stroke}%
\end{pgfscope}%
\begin{pgfscope}%
\definecolor{textcolor}{rgb}{0.333333,0.333333,0.333333}%
\pgfsetstrokecolor{textcolor}%
\pgfsetfillcolor{textcolor}%
\pgftext[x=0.041670in, y=0.934071in, left, base]{\color{textcolor}\sffamily\fontsize{10.000000}{12.000000}\selectfont 0.800}%
\end{pgfscope}%
\begin{pgfscope}%
\pgfpathrectangle{\pgfqpoint{0.536502in}{0.050071in}}{\pgfqpoint{6.421828in}{1.158259in}}%
\pgfusepath{clip}%
\pgfsetrectcap%
\pgfsetroundjoin%
\pgfsetlinewidth{1.003750pt}%
\definecolor{currentstroke}{rgb}{0.886275,0.290196,0.200000}%
\pgfsetstrokecolor{currentstroke}%
\pgfsetdash{}{0pt}%
\pgfpathmoveto{\pgfqpoint{0.662421in}{0.209188in}}%
\pgfpathlineto{\pgfqpoint{0.788339in}{0.258930in}}%
\pgfpathlineto{\pgfqpoint{0.914257in}{0.292092in}}%
\pgfpathlineto{\pgfqpoint{1.040175in}{0.310055in}}%
\pgfpathlineto{\pgfqpoint{1.166093in}{0.308673in}}%
\pgfpathlineto{\pgfqpoint{1.292011in}{0.392960in}}%
\pgfpathlineto{\pgfqpoint{1.417930in}{0.420594in}}%
\pgfpathlineto{\pgfqpoint{1.543848in}{0.449611in}}%
\pgfpathlineto{\pgfqpoint{1.669766in}{0.470337in}}%
\pgfpathlineto{\pgfqpoint{1.795684in}{0.492445in}}%
\pgfpathlineto{\pgfqpoint{1.921602in}{0.521462in}}%
\pgfpathlineto{\pgfqpoint{2.047521in}{0.540806in}}%
\pgfusepath{stroke}%
\end{pgfscope}%
\begin{pgfscope}%
\pgfpathrectangle{\pgfqpoint{0.536502in}{0.050071in}}{\pgfqpoint{6.421828in}{1.158259in}}%
\pgfusepath{clip}%
\pgfsetrectcap%
\pgfsetroundjoin%
\pgfsetlinewidth{1.003750pt}%
\definecolor{currentstroke}{rgb}{0.886275,0.290196,0.200000}%
\pgfsetstrokecolor{currentstroke}%
\pgfsetdash{}{0pt}%
\pgfpathmoveto{\pgfqpoint{3.810375in}{0.242350in}}%
\pgfpathlineto{\pgfqpoint{3.936293in}{0.256167in}}%
\pgfpathlineto{\pgfqpoint{4.062212in}{0.254785in}}%
\pgfpathlineto{\pgfqpoint{4.188130in}{0.263076in}}%
\pgfpathlineto{\pgfqpoint{4.314048in}{0.352889in}}%
\pgfpathlineto{\pgfqpoint{4.439966in}{0.439939in}}%
\pgfpathlineto{\pgfqpoint{4.565884in}{0.384669in}}%
\pgfpathlineto{\pgfqpoint{4.691803in}{0.502117in}}%
\pgfpathlineto{\pgfqpoint{4.817721in}{0.535279in}}%
\pgfpathlineto{\pgfqpoint{4.943639in}{0.520080in}}%
\pgfpathlineto{\pgfqpoint{5.069557in}{0.478628in}}%
\pgfpathlineto{\pgfqpoint{5.195475in}{0.503499in}}%
\pgfusepath{stroke}%
\end{pgfscope}%
\begin{pgfscope}%
\pgfpathrectangle{\pgfqpoint{0.536502in}{0.050071in}}{\pgfqpoint{6.421828in}{1.158259in}}%
\pgfusepath{clip}%
\pgfsetrectcap%
\pgfsetroundjoin%
\pgfsetlinewidth{1.003750pt}%
\definecolor{currentstroke}{rgb}{0.203922,0.541176,0.741176}%
\pgfsetstrokecolor{currentstroke}%
\pgfsetdash{}{0pt}%
\pgfpathmoveto{\pgfqpoint{0.662421in}{0.341835in}}%
\pgfpathlineto{\pgfqpoint{0.788339in}{0.363943in}}%
\pgfpathlineto{\pgfqpoint{0.914257in}{0.460665in}}%
\pgfpathlineto{\pgfqpoint{1.040175in}{0.457902in}}%
\pgfpathlineto{\pgfqpoint{1.166093in}{0.471719in}}%
\pgfpathlineto{\pgfqpoint{1.292011in}{0.463428in}}%
\pgfpathlineto{\pgfqpoint{1.417930in}{0.506263in}}%
\pgfpathlineto{\pgfqpoint{1.543848in}{0.538043in}}%
\pgfpathlineto{\pgfqpoint{1.669766in}{0.524225in}}%
\pgfpathlineto{\pgfqpoint{1.795684in}{0.564296in}}%
\pgfpathlineto{\pgfqpoint{1.921602in}{0.564296in}}%
\pgfpathlineto{\pgfqpoint{2.047521in}{0.557387in}}%
\pgfusepath{stroke}%
\end{pgfscope}%
\begin{pgfscope}%
\pgfpathrectangle{\pgfqpoint{0.536502in}{0.050071in}}{\pgfqpoint{6.421828in}{1.158259in}}%
\pgfusepath{clip}%
\pgfsetrectcap%
\pgfsetroundjoin%
\pgfsetlinewidth{1.003750pt}%
\definecolor{currentstroke}{rgb}{0.203922,0.541176,0.741176}%
\pgfsetstrokecolor{currentstroke}%
\pgfsetdash{}{0pt}%
\pgfpathmoveto{\pgfqpoint{3.810375in}{0.102719in}}%
\pgfpathlineto{\pgfqpoint{3.936293in}{0.102719in}}%
\pgfpathlineto{\pgfqpoint{4.062212in}{0.102719in}}%
\pgfpathlineto{\pgfqpoint{4.188130in}{0.102719in}}%
\pgfpathlineto{\pgfqpoint{4.314048in}{0.102719in}}%
\pgfpathlineto{\pgfqpoint{4.439966in}{0.102719in}}%
\pgfpathlineto{\pgfqpoint{4.565884in}{0.102719in}}%
\pgfpathlineto{\pgfqpoint{4.691803in}{0.102719in}}%
\pgfpathlineto{\pgfqpoint{4.817721in}{0.102719in}}%
\pgfpathlineto{\pgfqpoint{4.943639in}{0.102719in}}%
\pgfpathlineto{\pgfqpoint{5.069557in}{0.102719in}}%
\pgfpathlineto{\pgfqpoint{5.195475in}{0.102719in}}%
\pgfusepath{stroke}%
\end{pgfscope}%
\begin{pgfscope}%
\pgfpathrectangle{\pgfqpoint{0.536502in}{0.050071in}}{\pgfqpoint{6.421828in}{1.158259in}}%
\pgfusepath{clip}%
\pgfsetrectcap%
\pgfsetroundjoin%
\pgfsetlinewidth{1.003750pt}%
\definecolor{currentstroke}{rgb}{0.596078,0.556863,0.835294}%
\pgfsetstrokecolor{currentstroke}%
\pgfsetdash{}{0pt}%
\pgfpathmoveto{\pgfqpoint{0.662421in}{0.347362in}}%
\pgfpathlineto{\pgfqpoint{0.788339in}{0.531134in}}%
\pgfpathlineto{\pgfqpoint{0.914257in}{0.506263in}}%
\pgfpathlineto{\pgfqpoint{1.040175in}{0.644437in}}%
\pgfpathlineto{\pgfqpoint{1.166093in}{0.754976in}}%
\pgfpathlineto{\pgfqpoint{1.292011in}{0.770175in}}%
\pgfpathlineto{\pgfqpoint{1.417930in}{0.801956in}}%
\pgfpathlineto{\pgfqpoint{1.543848in}{0.842026in}}%
\pgfpathlineto{\pgfqpoint{1.669766in}{0.883478in}}%
\pgfpathlineto{\pgfqpoint{1.795684in}{0.883478in}}%
\pgfpathlineto{\pgfqpoint{1.921602in}{0.873806in}}%
\pgfpathlineto{\pgfqpoint{2.047521in}{0.877951in}}%
\pgfpathlineto{\pgfqpoint{2.173439in}{0.851698in}}%
\pgfpathlineto{\pgfqpoint{2.299357in}{0.813010in}}%
\pgfpathlineto{\pgfqpoint{2.425275in}{0.687271in}}%
\pgfpathlineto{\pgfqpoint{2.551193in}{0.632001in}}%
\pgfusepath{stroke}%
\end{pgfscope}%
\begin{pgfscope}%
\pgfpathrectangle{\pgfqpoint{0.536502in}{0.050071in}}{\pgfqpoint{6.421828in}{1.158259in}}%
\pgfusepath{clip}%
\pgfsetrectcap%
\pgfsetroundjoin%
\pgfsetlinewidth{1.003750pt}%
\definecolor{currentstroke}{rgb}{0.596078,0.556863,0.835294}%
\pgfsetstrokecolor{currentstroke}%
\pgfsetdash{}{0pt}%
\pgfpathmoveto{\pgfqpoint{3.810375in}{0.340453in}}%
\pgfpathlineto{\pgfqpoint{3.936293in}{0.536661in}}%
\pgfpathlineto{\pgfqpoint{4.062212in}{0.551860in}}%
\pgfpathlineto{\pgfqpoint{4.188130in}{0.698325in}}%
\pgfpathlineto{\pgfqpoint{4.314048in}{0.783993in}}%
\pgfpathlineto{\pgfqpoint{4.439966in}{0.829590in}}%
\pgfpathlineto{\pgfqpoint{4.565884in}{0.879333in}}%
\pgfpathlineto{\pgfqpoint{4.691803in}{0.913877in}}%
\pgfpathlineto{\pgfqpoint{4.817721in}{0.953947in}}%
\pgfpathlineto{\pgfqpoint{4.943639in}{0.992636in}}%
\pgfpathlineto{\pgfqpoint{5.069557in}{1.013362in}}%
\pgfpathlineto{\pgfqpoint{5.195475in}{1.042379in}}%
\pgfpathlineto{\pgfqpoint{5.321394in}{1.101794in}}%
\pgfpathlineto{\pgfqpoint{5.447312in}{1.112848in}}%
\pgfpathlineto{\pgfqpoint{5.573230in}{1.155682in}}%
\pgfpathlineto{\pgfqpoint{5.699148in}{0.851698in}}%
\pgfusepath{stroke}%
\end{pgfscope}%
\begin{pgfscope}%
\pgfpathrectangle{\pgfqpoint{0.536502in}{0.050071in}}{\pgfqpoint{6.421828in}{1.158259in}}%
\pgfusepath{clip}%
\pgfsetrectcap%
\pgfsetroundjoin%
\pgfsetlinewidth{1.003750pt}%
\definecolor{currentstroke}{rgb}{0.466667,0.466667,0.466667}%
\pgfsetstrokecolor{currentstroke}%
\pgfsetdash{}{0pt}%
\pgfpathmoveto{\pgfqpoint{0.662421in}{0.488300in}}%
\pgfpathlineto{\pgfqpoint{0.788339in}{0.506263in}}%
\pgfpathlineto{\pgfqpoint{0.914257in}{0.602985in}}%
\pgfpathlineto{\pgfqpoint{1.040175in}{0.652727in}}%
\pgfpathlineto{\pgfqpoint{1.166093in}{0.677599in}}%
\pgfpathlineto{\pgfqpoint{1.292011in}{0.701088in}}%
\pgfpathlineto{\pgfqpoint{1.417930in}{0.766030in}}%
\pgfpathlineto{\pgfqpoint{1.543848in}{0.792283in}}%
\pgfpathlineto{\pgfqpoint{1.669766in}{0.801956in}}%
\pgfpathlineto{\pgfqpoint{1.795684in}{0.810246in}}%
\pgfpathlineto{\pgfqpoint{1.921602in}{0.801956in}}%
\pgfpathlineto{\pgfqpoint{2.047521in}{0.815773in}}%
\pgfpathlineto{\pgfqpoint{2.173439in}{0.821300in}}%
\pgfpathlineto{\pgfqpoint{2.299357in}{0.814391in}}%
\pgfpathlineto{\pgfqpoint{2.425275in}{0.796429in}}%
\pgfpathlineto{\pgfqpoint{2.551193in}{0.788138in}}%
\pgfpathlineto{\pgfqpoint{2.677112in}{0.767412in}}%
\pgfpathlineto{\pgfqpoint{2.803030in}{0.756358in}}%
\pgfpathlineto{\pgfqpoint{2.928948in}{0.742541in}}%
\pgfpathlineto{\pgfqpoint{3.054866in}{0.732868in}}%
\pgfpathlineto{\pgfqpoint{3.180784in}{0.720433in}}%
\pgfpathlineto{\pgfqpoint{3.306703in}{0.723196in}}%
\pgfpathlineto{\pgfqpoint{3.432621in}{0.691416in}}%
\pgfpathlineto{\pgfqpoint{3.558539in}{0.690034in}}%
\pgfusepath{stroke}%
\end{pgfscope}%
\begin{pgfscope}%
\pgfpathrectangle{\pgfqpoint{0.536502in}{0.050071in}}{\pgfqpoint{6.421828in}{1.158259in}}%
\pgfusepath{clip}%
\pgfsetrectcap%
\pgfsetroundjoin%
\pgfsetlinewidth{1.003750pt}%
\definecolor{currentstroke}{rgb}{0.466667,0.466667,0.466667}%
\pgfsetstrokecolor{currentstroke}%
\pgfsetdash{}{0pt}%
\pgfpathmoveto{\pgfqpoint{3.810375in}{0.489682in}}%
\pgfpathlineto{\pgfqpoint{3.936293in}{0.509026in}}%
\pgfpathlineto{\pgfqpoint{4.062212in}{0.619565in}}%
\pgfpathlineto{\pgfqpoint{4.188130in}{0.688653in}}%
\pgfpathlineto{\pgfqpoint{4.314048in}{0.745304in}}%
\pgfpathlineto{\pgfqpoint{4.439966in}{0.783993in}}%
\pgfpathlineto{\pgfqpoint{4.565884in}{0.825445in}}%
\pgfpathlineto{\pgfqpoint{4.691803in}{0.848935in}}%
\pgfpathlineto{\pgfqpoint{4.817721in}{0.862752in}}%
\pgfpathlineto{\pgfqpoint{4.943639in}{0.902823in}}%
\pgfpathlineto{\pgfqpoint{5.069557in}{0.915259in}}%
\pgfpathlineto{\pgfqpoint{5.195475in}{0.977437in}}%
\pgfpathlineto{\pgfqpoint{5.321394in}{0.998163in}}%
\pgfpathlineto{\pgfqpoint{5.447312in}{1.009217in}}%
\pgfpathlineto{\pgfqpoint{5.573230in}{1.011981in}}%
\pgfpathlineto{\pgfqpoint{5.699148in}{1.025798in}}%
\pgfpathlineto{\pgfqpoint{5.825066in}{1.017508in}}%
\pgfpathlineto{\pgfqpoint{5.950984in}{1.058960in}}%
\pgfpathlineto{\pgfqpoint{6.076903in}{1.070014in}}%
\pgfpathlineto{\pgfqpoint{6.202821in}{1.065869in}}%
\pgfpathlineto{\pgfqpoint{6.328739in}{1.063105in}}%
\pgfpathlineto{\pgfqpoint{6.454657in}{1.053433in}}%
\pgfpathlineto{\pgfqpoint{6.580575in}{1.115611in}}%
\pgfpathlineto{\pgfqpoint{6.706494in}{1.116993in}}%
\pgfusepath{stroke}%
\end{pgfscope}%
\begin{pgfscope}%
\pgfpathrectangle{\pgfqpoint{0.536502in}{0.050071in}}{\pgfqpoint{6.421828in}{1.158259in}}%
\pgfusepath{clip}%
\pgfsetbuttcap%
\pgfsetroundjoin%
\pgfsetlinewidth{1.003750pt}%
\definecolor{currentstroke}{rgb}{0.000000,0.392157,0.000000}%
\pgfsetstrokecolor{currentstroke}%
\pgfsetdash{{3.700000pt}{1.600000pt}}{0.000000pt}%
\pgfpathmoveto{\pgfqpoint{0.536502in}{0.102719in}}%
\pgfpathlineto{\pgfqpoint{6.958330in}{0.102719in}}%
\pgfusepath{stroke}%
\end{pgfscope}%
\begin{pgfscope}%
\pgfpathrectangle{\pgfqpoint{0.536502in}{0.050071in}}{\pgfqpoint{6.421828in}{1.158259in}}%
\pgfusepath{clip}%
\pgfsetbuttcap%
\pgfsetroundjoin%
\pgfsetlinewidth{1.003750pt}%
\definecolor{currentstroke}{rgb}{0.803922,0.521569,0.247059}%
\pgfsetstrokecolor{currentstroke}%
\pgfsetdash{{3.700000pt}{1.600000pt}}{0.000000pt}%
\pgfpathmoveto{\pgfqpoint{0.536502in}{1.089286in}}%
\pgfpathlineto{\pgfqpoint{6.958330in}{1.089286in}}%
\pgfusepath{stroke}%
\end{pgfscope}%
\begin{pgfscope}%
\pgfpathrectangle{\pgfqpoint{0.536502in}{0.050071in}}{\pgfqpoint{6.421828in}{1.158259in}}%
\pgfusepath{clip}%
\pgfsetrectcap%
\pgfsetroundjoin%
\pgfsetlinewidth{6.022500pt}%
\definecolor{currentstroke}{rgb}{1.000000,1.000000,1.000000}%
\pgfsetstrokecolor{currentstroke}%
\pgfsetdash{}{0pt}%
\pgfpathmoveto{\pgfqpoint{3.684457in}{0.050071in}}%
\pgfpathlineto{\pgfqpoint{3.684457in}{1.208330in}}%
\pgfusepath{stroke}%
\end{pgfscope}%
\begin{pgfscope}%
\pgfsetrectcap%
\pgfsetmiterjoin%
\pgfsetlinewidth{1.003750pt}%
\definecolor{currentstroke}{rgb}{1.000000,1.000000,1.000000}%
\pgfsetstrokecolor{currentstroke}%
\pgfsetdash{}{0pt}%
\pgfpathmoveto{\pgfqpoint{0.536502in}{0.050071in}}%
\pgfpathlineto{\pgfqpoint{0.536502in}{1.208330in}}%
\pgfusepath{stroke}%
\end{pgfscope}%
\begin{pgfscope}%
\pgfsetrectcap%
\pgfsetmiterjoin%
\pgfsetlinewidth{1.003750pt}%
\definecolor{currentstroke}{rgb}{1.000000,1.000000,1.000000}%
\pgfsetstrokecolor{currentstroke}%
\pgfsetdash{}{0pt}%
\pgfpathmoveto{\pgfqpoint{6.958330in}{0.050071in}}%
\pgfpathlineto{\pgfqpoint{6.958330in}{1.208330in}}%
\pgfusepath{stroke}%
\end{pgfscope}%
\begin{pgfscope}%
\pgfsetrectcap%
\pgfsetmiterjoin%
\pgfsetlinewidth{1.003750pt}%
\definecolor{currentstroke}{rgb}{1.000000,1.000000,1.000000}%
\pgfsetstrokecolor{currentstroke}%
\pgfsetdash{}{0pt}%
\pgfpathmoveto{\pgfqpoint{0.536502in}{0.050071in}}%
\pgfpathlineto{\pgfqpoint{6.958330in}{0.050071in}}%
\pgfusepath{stroke}%
\end{pgfscope}%
\begin{pgfscope}%
\pgfsetrectcap%
\pgfsetmiterjoin%
\pgfsetlinewidth{1.003750pt}%
\definecolor{currentstroke}{rgb}{1.000000,1.000000,1.000000}%
\pgfsetstrokecolor{currentstroke}%
\pgfsetdash{}{0pt}%
\pgfpathmoveto{\pgfqpoint{0.536502in}{1.208330in}}%
\pgfpathlineto{\pgfqpoint{6.958330in}{1.208330in}}%
\pgfusepath{stroke}%
\end{pgfscope}%
\end{pgfpicture}%
\makeatother%
\endgroup%

%% file: figures/comparison-majorlocation-nox.pgf
\begingroup%
\makeatletter%
\begin{pgfpicture}%
\pgfpathrectangle{\pgfpointorigin}{\pgfqpoint{7.000000in}{1.250000in}}%
\pgfusepath{use as bounding box, clip}%
\begin{pgfscope}%
\pgfsetbuttcap%
\pgfsetmiterjoin%
\definecolor{currentfill}{rgb}{1.000000,1.000000,1.000000}%
\pgfsetfillcolor{currentfill}%
\pgfsetlinewidth{0.000000pt}%
\definecolor{currentstroke}{rgb}{0.500000,0.500000,0.500000}%
\pgfsetstrokecolor{currentstroke}%
\pgfsetdash{}{0pt}%
\pgfpathmoveto{\pgfqpoint{0.000000in}{0.000000in}}%
\pgfpathlineto{\pgfqpoint{7.000000in}{0.000000in}}%
\pgfpathlineto{\pgfqpoint{7.000000in}{1.250000in}}%
\pgfpathlineto{\pgfqpoint{0.000000in}{1.250000in}}%
\pgfpathlineto{\pgfqpoint{0.000000in}{0.000000in}}%
\pgfpathclose%
\pgfusepath{fill}%
\end{pgfscope}%
\begin{pgfscope}%
\pgfsetbuttcap%
\pgfsetmiterjoin%
\definecolor{currentfill}{rgb}{0.898039,0.898039,0.898039}%
\pgfsetfillcolor{currentfill}%
\pgfsetlinewidth{0.000000pt}%
\definecolor{currentstroke}{rgb}{0.000000,0.000000,0.000000}%
\pgfsetstrokecolor{currentstroke}%
\pgfsetstrokeopacity{0.000000}%
\pgfsetdash{}{0pt}%
\pgfpathmoveto{\pgfqpoint{0.536502in}{0.041670in}}%
\pgfpathlineto{\pgfqpoint{6.958330in}{0.041670in}}%
\pgfpathlineto{\pgfqpoint{6.958330in}{1.208330in}}%
\pgfpathlineto{\pgfqpoint{0.536502in}{1.208330in}}%
\pgfpathlineto{\pgfqpoint{0.536502in}{0.041670in}}%
\pgfpathclose%
\pgfusepath{fill}%
\end{pgfscope}%
\begin{pgfscope}%
\pgfpathrectangle{\pgfqpoint{0.536502in}{0.041670in}}{\pgfqpoint{6.421828in}{1.166660in}}%
\pgfusepath{clip}%
\pgfsetrectcap%
\pgfsetroundjoin%
\pgfsetlinewidth{0.803000pt}%
\definecolor{currentstroke}{rgb}{1.000000,1.000000,1.000000}%
\pgfsetstrokecolor{currentstroke}%
\pgfsetdash{}{0pt}%
\pgfpathmoveto{\pgfqpoint{0.662421in}{0.041670in}}%
\pgfpathlineto{\pgfqpoint{0.662421in}{1.208330in}}%
\pgfusepath{stroke}%
\end{pgfscope}%
\begin{pgfscope}%
\pgfpathrectangle{\pgfqpoint{0.536502in}{0.041670in}}{\pgfqpoint{6.421828in}{1.166660in}}%
\pgfusepath{clip}%
\pgfsetrectcap%
\pgfsetroundjoin%
\pgfsetlinewidth{0.803000pt}%
\definecolor{currentstroke}{rgb}{1.000000,1.000000,1.000000}%
\pgfsetstrokecolor{currentstroke}%
\pgfsetdash{}{0pt}%
\pgfpathmoveto{\pgfqpoint{3.558539in}{0.041670in}}%
\pgfpathlineto{\pgfqpoint{3.558539in}{1.208330in}}%
\pgfusepath{stroke}%
\end{pgfscope}%
\begin{pgfscope}%
\pgfpathrectangle{\pgfqpoint{0.536502in}{0.041670in}}{\pgfqpoint{6.421828in}{1.166660in}}%
\pgfusepath{clip}%
\pgfsetrectcap%
\pgfsetroundjoin%
\pgfsetlinewidth{0.803000pt}%
\definecolor{currentstroke}{rgb}{1.000000,1.000000,1.000000}%
\pgfsetstrokecolor{currentstroke}%
\pgfsetdash{}{0pt}%
\pgfpathmoveto{\pgfqpoint{3.810375in}{0.041670in}}%
\pgfpathlineto{\pgfqpoint{3.810375in}{1.208330in}}%
\pgfusepath{stroke}%
\end{pgfscope}%
\begin{pgfscope}%
\pgfpathrectangle{\pgfqpoint{0.536502in}{0.041670in}}{\pgfqpoint{6.421828in}{1.166660in}}%
\pgfusepath{clip}%
\pgfsetrectcap%
\pgfsetroundjoin%
\pgfsetlinewidth{0.803000pt}%
\definecolor{currentstroke}{rgb}{1.000000,1.000000,1.000000}%
\pgfsetstrokecolor{currentstroke}%
\pgfsetdash{}{0pt}%
\pgfpathmoveto{\pgfqpoint{6.832412in}{0.041670in}}%
\pgfpathlineto{\pgfqpoint{6.832412in}{1.208330in}}%
\pgfusepath{stroke}%
\end{pgfscope}%
\begin{pgfscope}%
\pgfpathrectangle{\pgfqpoint{0.536502in}{0.041670in}}{\pgfqpoint{6.421828in}{1.166660in}}%
\pgfusepath{clip}%
\pgfsetrectcap%
\pgfsetroundjoin%
\pgfsetlinewidth{0.803000pt}%
\definecolor{currentstroke}{rgb}{1.000000,1.000000,1.000000}%
\pgfsetstrokecolor{currentstroke}%
\pgfsetdash{}{0pt}%
\pgfpathmoveto{\pgfqpoint{0.536502in}{0.209163in}}%
\pgfpathlineto{\pgfqpoint{6.958330in}{0.209163in}}%
\pgfusepath{stroke}%
\end{pgfscope}%
\begin{pgfscope}%
\definecolor{textcolor}{rgb}{0.333333,0.333333,0.333333}%
\pgfsetstrokecolor{textcolor}%
\pgfsetfillcolor{textcolor}%
\pgftext[x=0.041670in, y=0.156402in, left, base]{\color{textcolor}\sffamily\fontsize{10.000000}{12.000000}\selectfont 0.400}%
\end{pgfscope}%
\begin{pgfscope}%
\pgfpathrectangle{\pgfqpoint{0.536502in}{0.041670in}}{\pgfqpoint{6.421828in}{1.166660in}}%
\pgfusepath{clip}%
\pgfsetrectcap%
\pgfsetroundjoin%
\pgfsetlinewidth{0.803000pt}%
\definecolor{currentstroke}{rgb}{1.000000,1.000000,1.000000}%
\pgfsetstrokecolor{currentstroke}%
\pgfsetdash{}{0pt}%
\pgfpathmoveto{\pgfqpoint{0.536502in}{0.591983in}}%
\pgfpathlineto{\pgfqpoint{6.958330in}{0.591983in}}%
\pgfusepath{stroke}%
\end{pgfscope}%
\begin{pgfscope}%
\definecolor{textcolor}{rgb}{0.333333,0.333333,0.333333}%
\pgfsetstrokecolor{textcolor}%
\pgfsetfillcolor{textcolor}%
\pgftext[x=0.041670in, y=0.539222in, left, base]{\color{textcolor}\sffamily\fontsize{10.000000}{12.000000}\selectfont 0.600}%
\end{pgfscope}%
\begin{pgfscope}%
\pgfpathrectangle{\pgfqpoint{0.536502in}{0.041670in}}{\pgfqpoint{6.421828in}{1.166660in}}%
\pgfusepath{clip}%
\pgfsetrectcap%
\pgfsetroundjoin%
\pgfsetlinewidth{0.803000pt}%
\definecolor{currentstroke}{rgb}{1.000000,1.000000,1.000000}%
\pgfsetstrokecolor{currentstroke}%
\pgfsetdash{}{0pt}%
\pgfpathmoveto{\pgfqpoint{0.536502in}{0.974803in}}%
\pgfpathlineto{\pgfqpoint{6.958330in}{0.974803in}}%
\pgfusepath{stroke}%
\end{pgfscope}%
\begin{pgfscope}%
\definecolor{textcolor}{rgb}{0.333333,0.333333,0.333333}%
\pgfsetstrokecolor{textcolor}%
\pgfsetfillcolor{textcolor}%
\pgftext[x=0.041670in, y=0.922042in, left, base]{\color{textcolor}\sffamily\fontsize{10.000000}{12.000000}\selectfont 0.800}%
\end{pgfscope}%
\begin{pgfscope}%
\pgfpathrectangle{\pgfqpoint{0.536502in}{0.041670in}}{\pgfqpoint{6.421828in}{1.166660in}}%
\pgfusepath{clip}%
\pgfsetrectcap%
\pgfsetroundjoin%
\pgfsetlinewidth{1.003750pt}%
\definecolor{currentstroke}{rgb}{0.886275,0.290196,0.200000}%
\pgfsetstrokecolor{currentstroke}%
\pgfsetdash{}{0pt}%
\pgfpathmoveto{\pgfqpoint{0.662421in}{0.148589in}}%
\pgfpathlineto{\pgfqpoint{0.788339in}{0.191506in}}%
\pgfpathlineto{\pgfqpoint{0.914257in}{0.236875in}}%
\pgfpathlineto{\pgfqpoint{1.040175in}{0.222161in}}%
\pgfpathlineto{\pgfqpoint{1.166093in}{0.202542in}}%
\pgfpathlineto{\pgfqpoint{1.292011in}{0.307995in}}%
\pgfpathlineto{\pgfqpoint{1.417930in}{0.381567in}}%
\pgfpathlineto{\pgfqpoint{1.543848in}{0.404865in}}%
\pgfpathlineto{\pgfqpoint{1.669766in}{0.538521in}}%
\pgfpathlineto{\pgfqpoint{1.795684in}{0.540973in}}%
\pgfpathlineto{\pgfqpoint{1.921602in}{0.521354in}}%
\pgfpathlineto{\pgfqpoint{2.047521in}{0.526259in}}%
\pgfusepath{stroke}%
\end{pgfscope}%
\begin{pgfscope}%
\pgfpathrectangle{\pgfqpoint{0.536502in}{0.041670in}}{\pgfqpoint{6.421828in}{1.166660in}}%
\pgfusepath{clip}%
\pgfsetrectcap%
\pgfsetroundjoin%
\pgfsetlinewidth{1.003750pt}%
\definecolor{currentstroke}{rgb}{0.886275,0.290196,0.200000}%
\pgfsetstrokecolor{currentstroke}%
\pgfsetdash{}{0pt}%
\pgfpathmoveto{\pgfqpoint{3.810375in}{0.113029in}}%
\pgfpathlineto{\pgfqpoint{3.936293in}{0.142458in}}%
\pgfpathlineto{\pgfqpoint{4.062212in}{0.143684in}}%
\pgfpathlineto{\pgfqpoint{4.188130in}{0.164529in}}%
\pgfpathlineto{\pgfqpoint{4.314048in}{0.204994in}}%
\pgfpathlineto{\pgfqpoint{4.439966in}{0.225840in}}%
\pgfpathlineto{\pgfqpoint{4.565884in}{0.245459in}}%
\pgfpathlineto{\pgfqpoint{4.691803in}{0.347233in}}%
\pgfpathlineto{\pgfqpoint{4.817721in}{0.384019in}}%
\pgfpathlineto{\pgfqpoint{4.943639in}{0.437972in}}%
\pgfpathlineto{\pgfqpoint{5.069557in}{0.361948in}}%
\pgfpathlineto{\pgfqpoint{5.195475in}{0.325162in}}%
\pgfusepath{stroke}%
\end{pgfscope}%
\begin{pgfscope}%
\pgfpathrectangle{\pgfqpoint{0.536502in}{0.041670in}}{\pgfqpoint{6.421828in}{1.166660in}}%
\pgfusepath{clip}%
\pgfsetrectcap%
\pgfsetroundjoin%
\pgfsetlinewidth{1.003750pt}%
\definecolor{currentstroke}{rgb}{0.203922,0.541176,0.741176}%
\pgfsetstrokecolor{currentstroke}%
\pgfsetdash{}{0pt}%
\pgfpathmoveto{\pgfqpoint{0.662421in}{0.227066in}}%
\pgfpathlineto{\pgfqpoint{0.788339in}{0.298185in}}%
\pgfpathlineto{\pgfqpoint{0.914257in}{0.348460in}}%
\pgfpathlineto{\pgfqpoint{1.040175in}{0.365626in}}%
\pgfpathlineto{\pgfqpoint{1.166093in}{0.413448in}}%
\pgfpathlineto{\pgfqpoint{1.292011in}{0.413448in}}%
\pgfpathlineto{\pgfqpoint{1.417930in}{0.425710in}}%
\pgfpathlineto{\pgfqpoint{1.543848in}{0.417127in}}%
\pgfpathlineto{\pgfqpoint{1.669766in}{0.451461in}}%
\pgfpathlineto{\pgfqpoint{1.795684in}{0.496830in}}%
\pgfpathlineto{\pgfqpoint{1.921602in}{0.482116in}}%
\pgfpathlineto{\pgfqpoint{2.047521in}{0.525033in}}%
\pgfusepath{stroke}%
\end{pgfscope}%
\begin{pgfscope}%
\pgfpathrectangle{\pgfqpoint{0.536502in}{0.041670in}}{\pgfqpoint{6.421828in}{1.166660in}}%
\pgfusepath{clip}%
\pgfsetrectcap%
\pgfsetroundjoin%
\pgfsetlinewidth{1.003750pt}%
\definecolor{currentstroke}{rgb}{0.203922,0.541176,0.741176}%
\pgfsetstrokecolor{currentstroke}%
\pgfsetdash{}{0pt}%
\pgfpathmoveto{\pgfqpoint{3.810375in}{0.094700in}}%
\pgfpathlineto{\pgfqpoint{3.936293in}{0.094700in}}%
\pgfpathlineto{\pgfqpoint{4.062212in}{0.094700in}}%
\pgfpathlineto{\pgfqpoint{4.188130in}{0.094700in}}%
\pgfpathlineto{\pgfqpoint{4.314048in}{0.094700in}}%
\pgfpathlineto{\pgfqpoint{4.439966in}{0.094700in}}%
\pgfpathlineto{\pgfqpoint{4.565884in}{0.094700in}}%
\pgfpathlineto{\pgfqpoint{4.691803in}{0.094700in}}%
\pgfpathlineto{\pgfqpoint{4.817721in}{0.094700in}}%
\pgfpathlineto{\pgfqpoint{4.943639in}{0.094700in}}%
\pgfpathlineto{\pgfqpoint{5.069557in}{0.094700in}}%
\pgfpathlineto{\pgfqpoint{5.195475in}{0.094700in}}%
\pgfusepath{stroke}%
\end{pgfscope}%
\begin{pgfscope}%
\pgfpathrectangle{\pgfqpoint{0.536502in}{0.041670in}}{\pgfqpoint{6.421828in}{1.166660in}}%
\pgfusepath{clip}%
\pgfsetrectcap%
\pgfsetroundjoin%
\pgfsetlinewidth{1.003750pt}%
\definecolor{currentstroke}{rgb}{0.596078,0.556863,0.835294}%
\pgfsetstrokecolor{currentstroke}%
\pgfsetdash{}{0pt}%
\pgfpathmoveto{\pgfqpoint{0.662421in}{0.267530in}}%
\pgfpathlineto{\pgfqpoint{0.788339in}{0.409770in}}%
\pgfpathlineto{\pgfqpoint{0.914257in}{0.412222in}}%
\pgfpathlineto{\pgfqpoint{1.040175in}{0.488247in}}%
\pgfpathlineto{\pgfqpoint{1.166093in}{0.594926in}}%
\pgfpathlineto{\pgfqpoint{1.292011in}{0.629260in}}%
\pgfpathlineto{\pgfqpoint{1.417930in}{0.673403in}}%
\pgfpathlineto{\pgfqpoint{1.543848in}{0.754332in}}%
\pgfpathlineto{\pgfqpoint{1.669766in}{0.784987in}}%
\pgfpathlineto{\pgfqpoint{1.795684in}{0.808285in}}%
\pgfpathlineto{\pgfqpoint{1.921602in}{0.773951in}}%
\pgfpathlineto{\pgfqpoint{2.047521in}{0.794797in}}%
\pgfpathlineto{\pgfqpoint{2.173439in}{0.783761in}}%
\pgfpathlineto{\pgfqpoint{2.299357in}{0.646427in}}%
\pgfpathlineto{\pgfqpoint{2.425275in}{0.578985in}}%
\pgfpathlineto{\pgfqpoint{2.551193in}{0.521354in}}%
\pgfusepath{stroke}%
\end{pgfscope}%
\begin{pgfscope}%
\pgfpathrectangle{\pgfqpoint{0.536502in}{0.041670in}}{\pgfqpoint{6.421828in}{1.166660in}}%
\pgfusepath{clip}%
\pgfsetrectcap%
\pgfsetroundjoin%
\pgfsetlinewidth{1.003750pt}%
\definecolor{currentstroke}{rgb}{0.596078,0.556863,0.835294}%
\pgfsetstrokecolor{currentstroke}%
\pgfsetdash{}{0pt}%
\pgfpathmoveto{\pgfqpoint{3.810375in}{0.231971in}}%
\pgfpathlineto{\pgfqpoint{3.936293in}{0.414674in}}%
\pgfpathlineto{\pgfqpoint{4.062212in}{0.463723in}}%
\pgfpathlineto{\pgfqpoint{4.188130in}{0.533616in}}%
\pgfpathlineto{\pgfqpoint{4.314048in}{0.635391in}}%
\pgfpathlineto{\pgfqpoint{4.439966in}{0.707737in}}%
\pgfpathlineto{\pgfqpoint{4.565884in}{0.777630in}}%
\pgfpathlineto{\pgfqpoint{4.691803in}{0.821773in}}%
\pgfpathlineto{\pgfqpoint{4.817721in}{0.873274in}}%
\pgfpathlineto{\pgfqpoint{4.943639in}{0.892893in}}%
\pgfpathlineto{\pgfqpoint{5.069557in}{0.949298in}}%
\pgfpathlineto{\pgfqpoint{5.195475in}{0.972596in}}%
\pgfpathlineto{\pgfqpoint{5.321394in}{1.025323in}}%
\pgfpathlineto{\pgfqpoint{5.447312in}{1.085407in}}%
\pgfpathlineto{\pgfqpoint{5.573230in}{1.155300in}}%
\pgfpathlineto{\pgfqpoint{5.699148in}{0.772725in}}%
\pgfusepath{stroke}%
\end{pgfscope}%
\begin{pgfscope}%
\pgfpathrectangle{\pgfqpoint{0.536502in}{0.041670in}}{\pgfqpoint{6.421828in}{1.166660in}}%
\pgfusepath{clip}%
\pgfsetrectcap%
\pgfsetroundjoin%
\pgfsetlinewidth{1.003750pt}%
\definecolor{currentstroke}{rgb}{0.466667,0.466667,0.466667}%
\pgfsetstrokecolor{currentstroke}%
\pgfsetdash{}{0pt}%
\pgfpathmoveto{\pgfqpoint{0.662421in}{0.425710in}}%
\pgfpathlineto{\pgfqpoint{0.788339in}{0.442877in}}%
\pgfpathlineto{\pgfqpoint{0.914257in}{0.501735in}}%
\pgfpathlineto{\pgfqpoint{1.040175in}{0.548330in}}%
\pgfpathlineto{\pgfqpoint{1.166093in}{0.578985in}}%
\pgfpathlineto{\pgfqpoint{1.292011in}{0.640296in}}%
\pgfpathlineto{\pgfqpoint{1.417930in}{0.669724in}}%
\pgfpathlineto{\pgfqpoint{1.543848in}{0.685665in}}%
\pgfpathlineto{\pgfqpoint{1.669766in}{0.724903in}}%
\pgfpathlineto{\pgfqpoint{1.795684in}{0.760463in}}%
\pgfpathlineto{\pgfqpoint{1.921602in}{0.775178in}}%
\pgfpathlineto{\pgfqpoint{2.047521in}{0.776404in}}%
\pgfpathlineto{\pgfqpoint{2.173439in}{0.766594in}}%
\pgfpathlineto{\pgfqpoint{2.299357in}{0.762916in}}%
\pgfpathlineto{\pgfqpoint{2.425275in}{0.756785in}}%
\pgfpathlineto{\pgfqpoint{2.551193in}{0.748201in}}%
\pgfpathlineto{\pgfqpoint{2.677112in}{0.729808in}}%
\pgfpathlineto{\pgfqpoint{2.803030in}{0.693022in}}%
\pgfpathlineto{\pgfqpoint{2.928948in}{0.666046in}}%
\pgfpathlineto{\pgfqpoint{3.054866in}{0.663593in}}%
\pgfpathlineto{\pgfqpoint{3.180784in}{0.634165in}}%
\pgfpathlineto{\pgfqpoint{3.306703in}{0.629260in}}%
\pgfpathlineto{\pgfqpoint{3.432621in}{0.604736in}}%
\pgfpathlineto{\pgfqpoint{3.558539in}{0.580212in}}%
\pgfusepath{stroke}%
\end{pgfscope}%
\begin{pgfscope}%
\pgfpathrectangle{\pgfqpoint{0.536502in}{0.041670in}}{\pgfqpoint{6.421828in}{1.166660in}}%
\pgfusepath{clip}%
\pgfsetrectcap%
\pgfsetroundjoin%
\pgfsetlinewidth{1.003750pt}%
\definecolor{currentstroke}{rgb}{0.466667,0.466667,0.466667}%
\pgfsetstrokecolor{currentstroke}%
\pgfsetdash{}{0pt}%
\pgfpathmoveto{\pgfqpoint{3.810375in}{0.420806in}}%
\pgfpathlineto{\pgfqpoint{3.936293in}{0.433068in}}%
\pgfpathlineto{\pgfqpoint{4.062212in}{0.499282in}}%
\pgfpathlineto{\pgfqpoint{4.188130in}{0.576533in}}%
\pgfpathlineto{\pgfqpoint{4.314048in}{0.642748in}}%
\pgfpathlineto{\pgfqpoint{4.439966in}{0.713868in}}%
\pgfpathlineto{\pgfqpoint{4.565884in}{0.745749in}}%
\pgfpathlineto{\pgfqpoint{4.691803in}{0.761689in}}%
\pgfpathlineto{\pgfqpoint{4.817721in}{0.788666in}}%
\pgfpathlineto{\pgfqpoint{4.943639in}{0.832809in}}%
\pgfpathlineto{\pgfqpoint{5.069557in}{0.868369in}}%
\pgfpathlineto{\pgfqpoint{5.195475in}{0.932131in}}%
\pgfpathlineto{\pgfqpoint{5.321394in}{0.972596in}}%
\pgfpathlineto{\pgfqpoint{5.447312in}{0.990989in}}%
\pgfpathlineto{\pgfqpoint{5.573230in}{0.999572in}}%
\pgfpathlineto{\pgfqpoint{5.699148in}{0.994668in}}%
\pgfpathlineto{\pgfqpoint{5.825066in}{1.016739in}}%
\pgfpathlineto{\pgfqpoint{5.950984in}{1.037585in}}%
\pgfpathlineto{\pgfqpoint{6.076903in}{1.055978in}}%
\pgfpathlineto{\pgfqpoint{6.202821in}{1.078049in}}%
\pgfpathlineto{\pgfqpoint{6.328739in}{1.073145in}}%
\pgfpathlineto{\pgfqpoint{6.454657in}{1.048620in}}%
\pgfpathlineto{\pgfqpoint{6.580575in}{1.084180in}}%
\pgfpathlineto{\pgfqpoint{6.706494in}{1.098895in}}%
\pgfusepath{stroke}%
\end{pgfscope}%
\begin{pgfscope}%
\pgfpathrectangle{\pgfqpoint{0.536502in}{0.041670in}}{\pgfqpoint{6.421828in}{1.166660in}}%
\pgfusepath{clip}%
\pgfsetbuttcap%
\pgfsetroundjoin%
\pgfsetlinewidth{1.003750pt}%
\definecolor{currentstroke}{rgb}{0.000000,0.392157,0.000000}%
\pgfsetstrokecolor{currentstroke}%
\pgfsetdash{{3.700000pt}{1.600000pt}}{0.000000pt}%
\pgfpathmoveto{\pgfqpoint{0.536502in}{0.094700in}}%
\pgfpathlineto{\pgfqpoint{6.958330in}{0.094700in}}%
\pgfusepath{stroke}%
\end{pgfscope}%
\begin{pgfscope}%
\pgfpathrectangle{\pgfqpoint{0.536502in}{0.041670in}}{\pgfqpoint{6.421828in}{1.166660in}}%
\pgfusepath{clip}%
\pgfsetbuttcap%
\pgfsetroundjoin%
\pgfsetlinewidth{1.003750pt}%
\definecolor{currentstroke}{rgb}{0.803922,0.521569,0.247059}%
\pgfsetstrokecolor{currentstroke}%
\pgfsetdash{{3.700000pt}{1.600000pt}}{0.000000pt}%
\pgfpathmoveto{\pgfqpoint{0.536502in}{1.053473in}}%
\pgfpathlineto{\pgfqpoint{6.958330in}{1.053473in}}%
\pgfusepath{stroke}%
\end{pgfscope}%
\begin{pgfscope}%
\pgfpathrectangle{\pgfqpoint{0.536502in}{0.041670in}}{\pgfqpoint{6.421828in}{1.166660in}}%
\pgfusepath{clip}%
\pgfsetrectcap%
\pgfsetroundjoin%
\pgfsetlinewidth{6.022500pt}%
\definecolor{currentstroke}{rgb}{1.000000,1.000000,1.000000}%
\pgfsetstrokecolor{currentstroke}%
\pgfsetdash{}{0pt}%
\pgfpathmoveto{\pgfqpoint{3.684457in}{0.041670in}}%
\pgfpathlineto{\pgfqpoint{3.684457in}{1.208330in}}%
\pgfusepath{stroke}%
\end{pgfscope}%
\begin{pgfscope}%
\pgfsetrectcap%
\pgfsetmiterjoin%
\pgfsetlinewidth{1.003750pt}%
\definecolor{currentstroke}{rgb}{1.000000,1.000000,1.000000}%
\pgfsetstrokecolor{currentstroke}%
\pgfsetdash{}{0pt}%
\pgfpathmoveto{\pgfqpoint{0.536502in}{0.041670in}}%
\pgfpathlineto{\pgfqpoint{0.536502in}{1.208330in}}%
\pgfusepath{stroke}%
\end{pgfscope}%
\begin{pgfscope}%
\pgfsetrectcap%
\pgfsetmiterjoin%
\pgfsetlinewidth{1.003750pt}%
\definecolor{currentstroke}{rgb}{1.000000,1.000000,1.000000}%
\pgfsetstrokecolor{currentstroke}%
\pgfsetdash{}{0pt}%
\pgfpathmoveto{\pgfqpoint{6.958330in}{0.041670in}}%
\pgfpathlineto{\pgfqpoint{6.958330in}{1.208330in}}%
\pgfusepath{stroke}%
\end{pgfscope}%
\begin{pgfscope}%
\pgfsetrectcap%
\pgfsetmiterjoin%
\pgfsetlinewidth{1.003750pt}%
\definecolor{currentstroke}{rgb}{1.000000,1.000000,1.000000}%
\pgfsetstrokecolor{currentstroke}%
\pgfsetdash{}{0pt}%
\pgfpathmoveto{\pgfqpoint{0.536502in}{0.041670in}}%
\pgfpathlineto{\pgfqpoint{6.958330in}{0.041670in}}%
\pgfusepath{stroke}%
\end{pgfscope}%
\begin{pgfscope}%
\pgfsetrectcap%
\pgfsetmiterjoin%
\pgfsetlinewidth{1.003750pt}%
\definecolor{currentstroke}{rgb}{1.000000,1.000000,1.000000}%
\pgfsetstrokecolor{currentstroke}%
\pgfsetdash{}{0pt}%
\pgfpathmoveto{\pgfqpoint{0.536502in}{1.208330in}}%
\pgfpathlineto{\pgfqpoint{6.958330in}{1.208330in}}%
\pgfusepath{stroke}%
\end{pgfscope}%
\end{pgfpicture}%
\makeatother%
\endgroup%

%% file: figures/comparison-secondminorlocation-nox.pgf
\begingroup%
\makeatletter%
\begin{pgfpicture}%
\pgfpathrectangle{\pgfpointorigin}{\pgfqpoint{7.000000in}{1.250000in}}%
\pgfusepath{use as bounding box, clip}%
\begin{pgfscope}%
\pgfsetbuttcap%
\pgfsetmiterjoin%
\definecolor{currentfill}{rgb}{1.000000,1.000000,1.000000}%
\pgfsetfillcolor{currentfill}%
\pgfsetlinewidth{0.000000pt}%
\definecolor{currentstroke}{rgb}{0.500000,0.500000,0.500000}%
\pgfsetstrokecolor{currentstroke}%
\pgfsetdash{}{0pt}%
\pgfpathmoveto{\pgfqpoint{0.000000in}{0.000000in}}%
\pgfpathlineto{\pgfqpoint{7.000000in}{0.000000in}}%
\pgfpathlineto{\pgfqpoint{7.000000in}{1.250000in}}%
\pgfpathlineto{\pgfqpoint{0.000000in}{1.250000in}}%
\pgfpathlineto{\pgfqpoint{0.000000in}{0.000000in}}%
\pgfpathclose%
\pgfusepath{fill}%
\end{pgfscope}%
\begin{pgfscope}%
\pgfsetbuttcap%
\pgfsetmiterjoin%
\definecolor{currentfill}{rgb}{0.898039,0.898039,0.898039}%
\pgfsetfillcolor{currentfill}%
\pgfsetlinewidth{0.000000pt}%
\definecolor{currentstroke}{rgb}{0.000000,0.000000,0.000000}%
\pgfsetstrokecolor{currentstroke}%
\pgfsetstrokeopacity{0.000000}%
\pgfsetdash{}{0pt}%
\pgfpathmoveto{\pgfqpoint{0.536502in}{0.041670in}}%
\pgfpathlineto{\pgfqpoint{6.958330in}{0.041670in}}%
\pgfpathlineto{\pgfqpoint{6.958330in}{1.174881in}}%
\pgfpathlineto{\pgfqpoint{0.536502in}{1.174881in}}%
\pgfpathlineto{\pgfqpoint{0.536502in}{0.041670in}}%
\pgfpathclose%
\pgfusepath{fill}%
\end{pgfscope}%
\begin{pgfscope}%
\pgfpathrectangle{\pgfqpoint{0.536502in}{0.041670in}}{\pgfqpoint{6.421828in}{1.133211in}}%
\pgfusepath{clip}%
\pgfsetrectcap%
\pgfsetroundjoin%
\pgfsetlinewidth{0.803000pt}%
\definecolor{currentstroke}{rgb}{1.000000,1.000000,1.000000}%
\pgfsetstrokecolor{currentstroke}%
\pgfsetdash{}{0pt}%
\pgfpathmoveto{\pgfqpoint{0.662421in}{0.041670in}}%
\pgfpathlineto{\pgfqpoint{0.662421in}{1.174881in}}%
\pgfusepath{stroke}%
\end{pgfscope}%
\begin{pgfscope}%
\pgfpathrectangle{\pgfqpoint{0.536502in}{0.041670in}}{\pgfqpoint{6.421828in}{1.133211in}}%
\pgfusepath{clip}%
\pgfsetrectcap%
\pgfsetroundjoin%
\pgfsetlinewidth{0.803000pt}%
\definecolor{currentstroke}{rgb}{1.000000,1.000000,1.000000}%
\pgfsetstrokecolor{currentstroke}%
\pgfsetdash{}{0pt}%
\pgfpathmoveto{\pgfqpoint{3.558539in}{0.041670in}}%
\pgfpathlineto{\pgfqpoint{3.558539in}{1.174881in}}%
\pgfusepath{stroke}%
\end{pgfscope}%
\begin{pgfscope}%
\pgfpathrectangle{\pgfqpoint{0.536502in}{0.041670in}}{\pgfqpoint{6.421828in}{1.133211in}}%
\pgfusepath{clip}%
\pgfsetrectcap%
\pgfsetroundjoin%
\pgfsetlinewidth{0.803000pt}%
\definecolor{currentstroke}{rgb}{1.000000,1.000000,1.000000}%
\pgfsetstrokecolor{currentstroke}%
\pgfsetdash{}{0pt}%
\pgfpathmoveto{\pgfqpoint{3.810375in}{0.041670in}}%
\pgfpathlineto{\pgfqpoint{3.810375in}{1.174881in}}%
\pgfusepath{stroke}%
\end{pgfscope}%
\begin{pgfscope}%
\pgfpathrectangle{\pgfqpoint{0.536502in}{0.041670in}}{\pgfqpoint{6.421828in}{1.133211in}}%
\pgfusepath{clip}%
\pgfsetrectcap%
\pgfsetroundjoin%
\pgfsetlinewidth{0.803000pt}%
\definecolor{currentstroke}{rgb}{1.000000,1.000000,1.000000}%
\pgfsetstrokecolor{currentstroke}%
\pgfsetdash{}{0pt}%
\pgfpathmoveto{\pgfqpoint{6.832412in}{0.041670in}}%
\pgfpathlineto{\pgfqpoint{6.832412in}{1.174881in}}%
\pgfusepath{stroke}%
\end{pgfscope}%
\begin{pgfscope}%
\pgfpathrectangle{\pgfqpoint{0.536502in}{0.041670in}}{\pgfqpoint{6.421828in}{1.133211in}}%
\pgfusepath{clip}%
\pgfsetrectcap%
\pgfsetroundjoin%
\pgfsetlinewidth{0.803000pt}%
\definecolor{currentstroke}{rgb}{1.000000,1.000000,1.000000}%
\pgfsetstrokecolor{currentstroke}%
\pgfsetdash{}{0pt}%
\pgfpathmoveto{\pgfqpoint{0.536502in}{0.283319in}}%
\pgfpathlineto{\pgfqpoint{6.958330in}{0.283319in}}%
\pgfusepath{stroke}%
\end{pgfscope}%
\begin{pgfscope}%
\definecolor{textcolor}{rgb}{0.333333,0.333333,0.333333}%
\pgfsetstrokecolor{textcolor}%
\pgfsetfillcolor{textcolor}%
\pgftext[x=0.041670in, y=0.230558in, left, base]{\color{textcolor}\sffamily\fontsize{10.000000}{12.000000}\selectfont 0.400}%
\end{pgfscope}%
\begin{pgfscope}%
\pgfpathrectangle{\pgfqpoint{0.536502in}{0.041670in}}{\pgfqpoint{6.421828in}{1.133211in}}%
\pgfusepath{clip}%
\pgfsetrectcap%
\pgfsetroundjoin%
\pgfsetlinewidth{0.803000pt}%
\definecolor{currentstroke}{rgb}{1.000000,1.000000,1.000000}%
\pgfsetstrokecolor{currentstroke}%
\pgfsetdash{}{0pt}%
\pgfpathmoveto{\pgfqpoint{0.536502in}{0.719419in}}%
\pgfpathlineto{\pgfqpoint{6.958330in}{0.719419in}}%
\pgfusepath{stroke}%
\end{pgfscope}%
\begin{pgfscope}%
\definecolor{textcolor}{rgb}{0.333333,0.333333,0.333333}%
\pgfsetstrokecolor{textcolor}%
\pgfsetfillcolor{textcolor}%
\pgftext[x=0.041670in, y=0.666658in, left, base]{\color{textcolor}\sffamily\fontsize{10.000000}{12.000000}\selectfont 0.600}%
\end{pgfscope}%
\begin{pgfscope}%
\pgfpathrectangle{\pgfqpoint{0.536502in}{0.041670in}}{\pgfqpoint{6.421828in}{1.133211in}}%
\pgfusepath{clip}%
\pgfsetrectcap%
\pgfsetroundjoin%
\pgfsetlinewidth{0.803000pt}%
\definecolor{currentstroke}{rgb}{1.000000,1.000000,1.000000}%
\pgfsetstrokecolor{currentstroke}%
\pgfsetdash{}{0pt}%
\pgfpathmoveto{\pgfqpoint{0.536502in}{1.155519in}}%
\pgfpathlineto{\pgfqpoint{6.958330in}{1.155519in}}%
\pgfusepath{stroke}%
\end{pgfscope}%
\begin{pgfscope}%
\definecolor{textcolor}{rgb}{0.333333,0.333333,0.333333}%
\pgfsetstrokecolor{textcolor}%
\pgfsetfillcolor{textcolor}%
\pgftext[x=0.041670in, y=1.102758in, left, base]{\color{textcolor}\sffamily\fontsize{10.000000}{12.000000}\selectfont 0.800}%
\end{pgfscope}%
\begin{pgfscope}%
\pgfpathrectangle{\pgfqpoint{0.536502in}{0.041670in}}{\pgfqpoint{6.421828in}{1.133211in}}%
\pgfusepath{clip}%
\pgfsetrectcap%
\pgfsetroundjoin%
\pgfsetlinewidth{1.003750pt}%
\definecolor{currentstroke}{rgb}{0.886275,0.290196,0.200000}%
\pgfsetstrokecolor{currentstroke}%
\pgfsetdash{}{0pt}%
\pgfpathmoveto{\pgfqpoint{0.662421in}{0.093180in}}%
\pgfpathlineto{\pgfqpoint{0.788339in}{0.114191in}}%
\pgfpathlineto{\pgfqpoint{0.914257in}{0.140748in}}%
\pgfpathlineto{\pgfqpoint{1.040175in}{0.171499in}}%
\pgfpathlineto{\pgfqpoint{1.166093in}{0.158919in}}%
\pgfpathlineto{\pgfqpoint{1.292011in}{0.181283in}}%
\pgfpathlineto{\pgfqpoint{1.417930in}{0.205045in}}%
\pgfpathlineto{\pgfqpoint{1.543848in}{0.164510in}}%
\pgfpathlineto{\pgfqpoint{1.669766in}{0.219022in}}%
\pgfpathlineto{\pgfqpoint{1.795684in}{0.262353in}}%
\pgfpathlineto{\pgfqpoint{1.921602in}{0.214829in}}%
\pgfpathlineto{\pgfqpoint{2.047521in}{0.237193in}}%
\pgfusepath{stroke}%
\end{pgfscope}%
\begin{pgfscope}%
\pgfpathrectangle{\pgfqpoint{0.536502in}{0.041670in}}{\pgfqpoint{6.421828in}{1.133211in}}%
\pgfusepath{clip}%
\pgfsetrectcap%
\pgfsetroundjoin%
\pgfsetlinewidth{1.003750pt}%
\definecolor{currentstroke}{rgb}{0.886275,0.290196,0.200000}%
\pgfsetstrokecolor{currentstroke}%
\pgfsetdash{}{0pt}%
\pgfpathmoveto{\pgfqpoint{3.810375in}{0.094622in}}%
\pgfpathlineto{\pgfqpoint{3.936293in}{0.128168in}}%
\pgfpathlineto{\pgfqpoint{4.062212in}{0.195261in}}%
\pgfpathlineto{\pgfqpoint{4.188130in}{0.171499in}}%
\pgfpathlineto{\pgfqpoint{4.314048in}{0.210636in}}%
\pgfpathlineto{\pgfqpoint{4.439966in}{0.219022in}}%
\pgfpathlineto{\pgfqpoint{4.565884in}{0.175692in}}%
\pgfpathlineto{\pgfqpoint{4.691803in}{0.217625in}}%
\pgfpathlineto{\pgfqpoint{4.817721in}{0.241387in}}%
\pgfpathlineto{\pgfqpoint{4.943639in}{0.242784in}}%
\pgfpathlineto{\pgfqpoint{5.069557in}{0.286115in}}%
\pgfpathlineto{\pgfqpoint{5.195475in}{0.245580in}}%
\pgfusepath{stroke}%
\end{pgfscope}%
\begin{pgfscope}%
\pgfpathrectangle{\pgfqpoint{0.536502in}{0.041670in}}{\pgfqpoint{6.421828in}{1.133211in}}%
\pgfusepath{clip}%
\pgfsetrectcap%
\pgfsetroundjoin%
\pgfsetlinewidth{1.003750pt}%
\definecolor{currentstroke}{rgb}{0.203922,0.541176,0.741176}%
\pgfsetstrokecolor{currentstroke}%
\pgfsetdash{}{0pt}%
\pgfpathmoveto{\pgfqpoint{0.662421in}{0.154726in}}%
\pgfpathlineto{\pgfqpoint{0.788339in}{0.156123in}}%
\pgfpathlineto{\pgfqpoint{0.914257in}{0.184078in}}%
\pgfpathlineto{\pgfqpoint{1.040175in}{0.203647in}}%
\pgfpathlineto{\pgfqpoint{1.166093in}{0.206443in}}%
\pgfpathlineto{\pgfqpoint{1.292011in}{0.185476in}}%
\pgfpathlineto{\pgfqpoint{1.417930in}{0.210636in}}%
\pgfpathlineto{\pgfqpoint{1.543848in}{0.203647in}}%
\pgfpathlineto{\pgfqpoint{1.669766in}{0.226011in}}%
\pgfpathlineto{\pgfqpoint{1.795684in}{0.198056in}}%
\pgfpathlineto{\pgfqpoint{1.921602in}{0.196658in}}%
\pgfpathlineto{\pgfqpoint{2.047521in}{0.209238in}}%
\pgfusepath{stroke}%
\end{pgfscope}%
\begin{pgfscope}%
\pgfpathrectangle{\pgfqpoint{0.536502in}{0.041670in}}{\pgfqpoint{6.421828in}{1.133211in}}%
\pgfusepath{clip}%
\pgfsetrectcap%
\pgfsetroundjoin%
\pgfsetlinewidth{1.003750pt}%
\definecolor{currentstroke}{rgb}{0.203922,0.541176,0.741176}%
\pgfsetstrokecolor{currentstroke}%
\pgfsetdash{}{0pt}%
\pgfpathmoveto{\pgfqpoint{3.810375in}{0.093180in}}%
\pgfpathlineto{\pgfqpoint{3.936293in}{0.093180in}}%
\pgfpathlineto{\pgfqpoint{4.062212in}{0.093180in}}%
\pgfpathlineto{\pgfqpoint{4.188130in}{0.093180in}}%
\pgfpathlineto{\pgfqpoint{4.314048in}{0.093180in}}%
\pgfpathlineto{\pgfqpoint{4.439966in}{0.093180in}}%
\pgfpathlineto{\pgfqpoint{4.565884in}{0.093180in}}%
\pgfpathlineto{\pgfqpoint{4.691803in}{0.093180in}}%
\pgfpathlineto{\pgfqpoint{4.817721in}{0.093180in}}%
\pgfpathlineto{\pgfqpoint{4.943639in}{0.093180in}}%
\pgfpathlineto{\pgfqpoint{5.069557in}{0.093180in}}%
\pgfpathlineto{\pgfqpoint{5.195475in}{0.093180in}}%
\pgfusepath{stroke}%
\end{pgfscope}%
\begin{pgfscope}%
\pgfpathrectangle{\pgfqpoint{0.536502in}{0.041670in}}{\pgfqpoint{6.421828in}{1.133211in}}%
\pgfusepath{clip}%
\pgfsetrectcap%
\pgfsetroundjoin%
\pgfsetlinewidth{1.003750pt}%
\definecolor{currentstroke}{rgb}{0.596078,0.556863,0.835294}%
\pgfsetstrokecolor{currentstroke}%
\pgfsetdash{}{0pt}%
\pgfpathmoveto{\pgfqpoint{0.662421in}{0.142146in}}%
\pgfpathlineto{\pgfqpoint{0.788339in}{0.227409in}}%
\pgfpathlineto{\pgfqpoint{0.914257in}{0.263751in}}%
\pgfpathlineto{\pgfqpoint{1.040175in}{0.298695in}}%
\pgfpathlineto{\pgfqpoint{1.166093in}{0.356003in}}%
\pgfpathlineto{\pgfqpoint{1.292011in}{0.393742in}}%
\pgfpathlineto{\pgfqpoint{1.417930in}{0.414708in}}%
\pgfpathlineto{\pgfqpoint{1.543848in}{0.434277in}}%
\pgfpathlineto{\pgfqpoint{1.669766in}{0.460834in}}%
\pgfpathlineto{\pgfqpoint{1.795684in}{0.470619in}}%
\pgfpathlineto{\pgfqpoint{1.921602in}{0.460834in}}%
\pgfpathlineto{\pgfqpoint{2.047521in}{0.442664in}}%
\pgfpathlineto{\pgfqpoint{2.173439in}{0.466425in}}%
\pgfpathlineto{\pgfqpoint{2.299357in}{0.382560in}}%
\pgfpathlineto{\pgfqpoint{2.425275in}{0.302888in}}%
\pgfpathlineto{\pgfqpoint{2.551193in}{0.286115in}}%
\pgfusepath{stroke}%
\end{pgfscope}%
\begin{pgfscope}%
\pgfpathrectangle{\pgfqpoint{0.536502in}{0.041670in}}{\pgfqpoint{6.421828in}{1.133211in}}%
\pgfusepath{clip}%
\pgfsetrectcap%
\pgfsetroundjoin%
\pgfsetlinewidth{1.003750pt}%
\definecolor{currentstroke}{rgb}{0.596078,0.556863,0.835294}%
\pgfsetstrokecolor{currentstroke}%
\pgfsetdash{}{0pt}%
\pgfpathmoveto{\pgfqpoint{3.810375in}{0.157521in}}%
\pgfpathlineto{\pgfqpoint{3.936293in}{0.210636in}}%
\pgfpathlineto{\pgfqpoint{4.062212in}{0.251171in}}%
\pgfpathlineto{\pgfqpoint{4.188130in}{0.353207in}}%
\pgfpathlineto{\pgfqpoint{4.314048in}{0.420299in}}%
\pgfpathlineto{\pgfqpoint{4.439966in}{0.460834in}}%
\pgfpathlineto{\pgfqpoint{4.565884in}{0.519540in}}%
\pgfpathlineto{\pgfqpoint{4.691803in}{0.541904in}}%
\pgfpathlineto{\pgfqpoint{4.817721in}{0.631361in}}%
\pgfpathlineto{\pgfqpoint{4.943639in}{0.670498in}}%
\pgfpathlineto{\pgfqpoint{5.069557in}{0.701249in}}%
\pgfpathlineto{\pgfqpoint{5.195475in}{0.715226in}}%
\pgfpathlineto{\pgfqpoint{5.321394in}{0.818660in}}%
\pgfpathlineto{\pgfqpoint{5.447312in}{0.908117in}}%
\pgfpathlineto{\pgfqpoint{5.573230in}{1.123371in}}%
\pgfpathlineto{\pgfqpoint{5.699148in}{0.975209in}}%
\pgfusepath{stroke}%
\end{pgfscope}%
\begin{pgfscope}%
\pgfpathrectangle{\pgfqpoint{0.536502in}{0.041670in}}{\pgfqpoint{6.421828in}{1.133211in}}%
\pgfusepath{clip}%
\pgfsetrectcap%
\pgfsetroundjoin%
\pgfsetlinewidth{1.003750pt}%
\definecolor{currentstroke}{rgb}{0.466667,0.466667,0.466667}%
\pgfsetstrokecolor{currentstroke}%
\pgfsetdash{}{0pt}%
\pgfpathmoveto{\pgfqpoint{0.662421in}{0.251171in}}%
\pgfpathlineto{\pgfqpoint{0.788339in}{0.258160in}}%
\pgfpathlineto{\pgfqpoint{0.914257in}{0.269342in}}%
\pgfpathlineto{\pgfqpoint{1.040175in}{0.309877in}}%
\pgfpathlineto{\pgfqpoint{1.166093in}{0.312672in}}%
\pgfpathlineto{\pgfqpoint{1.292011in}{0.378367in}}%
\pgfpathlineto{\pgfqpoint{1.417930in}{0.399333in}}%
\pgfpathlineto{\pgfqpoint{1.543848in}{0.413311in}}%
\pgfpathlineto{\pgfqpoint{1.669766in}{0.463630in}}%
\pgfpathlineto{\pgfqpoint{1.795684in}{0.458039in}}%
\pgfpathlineto{\pgfqpoint{1.921602in}{0.451050in}}%
\pgfpathlineto{\pgfqpoint{2.047521in}{0.438470in}}%
\pgfpathlineto{\pgfqpoint{2.173439in}{0.435675in}}%
\pgfpathlineto{\pgfqpoint{2.299357in}{0.441266in}}%
\pgfpathlineto{\pgfqpoint{2.425275in}{0.439868in}}%
\pgfpathlineto{\pgfqpoint{2.551193in}{0.425890in}}%
\pgfpathlineto{\pgfqpoint{2.677112in}{0.409117in}}%
\pgfpathlineto{\pgfqpoint{2.803030in}{0.386753in}}%
\pgfpathlineto{\pgfqpoint{2.928948in}{0.378367in}}%
\pgfpathlineto{\pgfqpoint{3.054866in}{0.376969in}}%
\pgfpathlineto{\pgfqpoint{3.180784in}{0.354605in}}%
\pgfpathlineto{\pgfqpoint{3.306703in}{0.343423in}}%
\pgfpathlineto{\pgfqpoint{3.432621in}{0.337832in}}%
\pgfpathlineto{\pgfqpoint{3.558539in}{0.318263in}}%
\pgfusepath{stroke}%
\end{pgfscope}%
\begin{pgfscope}%
\pgfpathrectangle{\pgfqpoint{0.536502in}{0.041670in}}{\pgfqpoint{6.421828in}{1.133211in}}%
\pgfusepath{clip}%
\pgfsetrectcap%
\pgfsetroundjoin%
\pgfsetlinewidth{1.003750pt}%
\definecolor{currentstroke}{rgb}{0.466667,0.466667,0.466667}%
\pgfsetstrokecolor{currentstroke}%
\pgfsetdash{}{0pt}%
\pgfpathmoveto{\pgfqpoint{3.810375in}{0.251171in}}%
\pgfpathlineto{\pgfqpoint{3.936293in}{0.251171in}}%
\pgfpathlineto{\pgfqpoint{4.062212in}{0.279126in}}%
\pgfpathlineto{\pgfqpoint{4.188130in}{0.314070in}}%
\pgfpathlineto{\pgfqpoint{4.314048in}{0.369980in}}%
\pgfpathlineto{\pgfqpoint{4.439966in}{0.425890in}}%
\pgfpathlineto{\pgfqpoint{4.565884in}{0.439868in}}%
\pgfpathlineto{\pgfqpoint{4.691803in}{0.490187in}}%
\pgfpathlineto{\pgfqpoint{4.817721in}{0.488789in}}%
\pgfpathlineto{\pgfqpoint{4.943639in}{0.525131in}}%
\pgfpathlineto{\pgfqpoint{5.069557in}{0.548893in}}%
\pgfpathlineto{\pgfqpoint{5.195475in}{0.593621in}}%
\pgfpathlineto{\pgfqpoint{5.321394in}{0.655123in}}%
\pgfpathlineto{\pgfqpoint{5.447312in}{0.688669in}}%
\pgfpathlineto{\pgfqpoint{5.573230in}{0.751568in}}%
\pgfpathlineto{\pgfqpoint{5.699148in}{0.771136in}}%
\pgfpathlineto{\pgfqpoint{5.825066in}{0.828444in}}%
\pgfpathlineto{\pgfqpoint{5.950984in}{0.827047in}}%
\pgfpathlineto{\pgfqpoint{6.076903in}{0.881559in}}%
\pgfpathlineto{\pgfqpoint{6.202821in}{0.861991in}}%
\pgfpathlineto{\pgfqpoint{6.328739in}{0.861991in}}%
\pgfpathlineto{\pgfqpoint{6.454657in}{0.863388in}}%
\pgfpathlineto{\pgfqpoint{6.580575in}{0.964027in}}%
\pgfpathlineto{\pgfqpoint{6.706494in}{0.987789in}}%
\pgfusepath{stroke}%
\end{pgfscope}%
\begin{pgfscope}%
\pgfpathrectangle{\pgfqpoint{0.536502in}{0.041670in}}{\pgfqpoint{6.421828in}{1.133211in}}%
\pgfusepath{clip}%
\pgfsetbuttcap%
\pgfsetroundjoin%
\pgfsetlinewidth{1.003750pt}%
\definecolor{currentstroke}{rgb}{0.000000,0.392157,0.000000}%
\pgfsetstrokecolor{currentstroke}%
\pgfsetdash{{3.700000pt}{1.600000pt}}{0.000000pt}%
\pgfpathmoveto{\pgfqpoint{0.536502in}{0.093180in}}%
\pgfpathlineto{\pgfqpoint{6.958330in}{0.093180in}}%
\pgfusepath{stroke}%
\end{pgfscope}%
\begin{pgfscope}%
\pgfpathrectangle{\pgfqpoint{0.536502in}{0.041670in}}{\pgfqpoint{6.421828in}{1.133211in}}%
\pgfusepath{clip}%
\pgfsetbuttcap%
\pgfsetroundjoin%
\pgfsetlinewidth{1.003750pt}%
\definecolor{currentstroke}{rgb}{0.803922,0.521569,0.247059}%
\pgfsetstrokecolor{currentstroke}%
\pgfsetdash{{3.700000pt}{1.600000pt}}{0.000000pt}%
\pgfpathmoveto{\pgfqpoint{0.536502in}{0.919371in}}%
\pgfpathlineto{\pgfqpoint{6.958330in}{0.919371in}}%
\pgfusepath{stroke}%
\end{pgfscope}%
\begin{pgfscope}%
\pgfpathrectangle{\pgfqpoint{0.536502in}{0.041670in}}{\pgfqpoint{6.421828in}{1.133211in}}%
\pgfusepath{clip}%
\pgfsetrectcap%
\pgfsetroundjoin%
\pgfsetlinewidth{6.022500pt}%
\definecolor{currentstroke}{rgb}{1.000000,1.000000,1.000000}%
\pgfsetstrokecolor{currentstroke}%
\pgfsetdash{}{0pt}%
\pgfpathmoveto{\pgfqpoint{3.684457in}{0.041670in}}%
\pgfpathlineto{\pgfqpoint{3.684457in}{1.174881in}}%
\pgfusepath{stroke}%
\end{pgfscope}%
\begin{pgfscope}%
\pgfsetrectcap%
\pgfsetmiterjoin%
\pgfsetlinewidth{1.003750pt}%
\definecolor{currentstroke}{rgb}{1.000000,1.000000,1.000000}%
\pgfsetstrokecolor{currentstroke}%
\pgfsetdash{}{0pt}%
\pgfpathmoveto{\pgfqpoint{0.536502in}{0.041670in}}%
\pgfpathlineto{\pgfqpoint{0.536502in}{1.174881in}}%
\pgfusepath{stroke}%
\end{pgfscope}%
\begin{pgfscope}%
\pgfsetrectcap%
\pgfsetmiterjoin%
\pgfsetlinewidth{1.003750pt}%
\definecolor{currentstroke}{rgb}{1.000000,1.000000,1.000000}%
\pgfsetstrokecolor{currentstroke}%
\pgfsetdash{}{0pt}%
\pgfpathmoveto{\pgfqpoint{6.958330in}{0.041670in}}%
\pgfpathlineto{\pgfqpoint{6.958330in}{1.174881in}}%
\pgfusepath{stroke}%
\end{pgfscope}%
\begin{pgfscope}%
\pgfsetrectcap%
\pgfsetmiterjoin%
\pgfsetlinewidth{1.003750pt}%
\definecolor{currentstroke}{rgb}{1.000000,1.000000,1.000000}%
\pgfsetstrokecolor{currentstroke}%
\pgfsetdash{}{0pt}%
\pgfpathmoveto{\pgfqpoint{0.536502in}{0.041670in}}%
\pgfpathlineto{\pgfqpoint{6.958330in}{0.041670in}}%
\pgfusepath{stroke}%
\end{pgfscope}%
\begin{pgfscope}%
\pgfsetrectcap%
\pgfsetmiterjoin%
\pgfsetlinewidth{1.003750pt}%
\definecolor{currentstroke}{rgb}{1.000000,1.000000,1.000000}%
\pgfsetstrokecolor{currentstroke}%
\pgfsetdash{}{0pt}%
\pgfpathmoveto{\pgfqpoint{0.536502in}{1.174881in}}%
\pgfpathlineto{\pgfqpoint{6.958330in}{1.174881in}}%
\pgfusepath{stroke}%
\end{pgfscope}%
\end{pgfpicture}%
\makeatother%
\endgroup%

%% file: figures/comparison-contact.pgf
\begingroup%
\makeatletter%
\begin{pgfpicture}%
\pgfpathrectangle{\pgfpointorigin}{\pgfqpoint{7.000000in}{1.550000in}}%
\pgfusepath{use as bounding box, clip}%
\begin{pgfscope}%
\pgfsetbuttcap%
\pgfsetmiterjoin%
\definecolor{currentfill}{rgb}{1.000000,1.000000,1.000000}%
\pgfsetfillcolor{currentfill}%
\pgfsetlinewidth{0.000000pt}%
\definecolor{currentstroke}{rgb}{0.500000,0.500000,0.500000}%
\pgfsetstrokecolor{currentstroke}%
\pgfsetdash{}{0pt}%
\pgfpathmoveto{\pgfqpoint{0.000000in}{0.000000in}}%
\pgfpathlineto{\pgfqpoint{7.000000in}{0.000000in}}%
\pgfpathlineto{\pgfqpoint{7.000000in}{1.550000in}}%
\pgfpathlineto{\pgfqpoint{0.000000in}{1.550000in}}%
\pgfpathlineto{\pgfqpoint{0.000000in}{0.000000in}}%
\pgfpathclose%
\pgfusepath{fill}%
\end{pgfscope}%
\begin{pgfscope}%
\pgfsetbuttcap%
\pgfsetmiterjoin%
\definecolor{currentfill}{rgb}{0.898039,0.898039,0.898039}%
\pgfsetfillcolor{currentfill}%
\pgfsetlinewidth{0.000000pt}%
\definecolor{currentstroke}{rgb}{0.000000,0.000000,0.000000}%
\pgfsetstrokecolor{currentstroke}%
\pgfsetstrokeopacity{0.000000}%
\pgfsetdash{}{0pt}%
\pgfpathmoveto{\pgfqpoint{0.536502in}{0.394792in}}%
\pgfpathlineto{\pgfqpoint{6.958330in}{0.394792in}}%
\pgfpathlineto{\pgfqpoint{6.958330in}{1.508330in}}%
\pgfpathlineto{\pgfqpoint{0.536502in}{1.508330in}}%
\pgfpathlineto{\pgfqpoint{0.536502in}{0.394792in}}%
\pgfpathclose%
\pgfusepath{fill}%
\end{pgfscope}%
\begin{pgfscope}%
\pgfpathrectangle{\pgfqpoint{0.536502in}{0.394792in}}{\pgfqpoint{6.421828in}{1.113538in}}%
\pgfusepath{clip}%
\pgfsetrectcap%
\pgfsetroundjoin%
\pgfsetlinewidth{0.803000pt}%
\definecolor{currentstroke}{rgb}{1.000000,1.000000,1.000000}%
\pgfsetstrokecolor{currentstroke}%
\pgfsetdash{}{0pt}%
\pgfpathmoveto{\pgfqpoint{0.662421in}{0.394792in}}%
\pgfpathlineto{\pgfqpoint{0.662421in}{1.508330in}}%
\pgfusepath{stroke}%
\end{pgfscope}%
\begin{pgfscope}%
\definecolor{textcolor}{rgb}{0.333333,0.333333,0.333333}%
\pgfsetstrokecolor{textcolor}%
\pgfsetfillcolor{textcolor}%
\pgftext[x=0.662421in,y=0.297570in,,top]{\color{textcolor}\sffamily\fontsize{10.000000}{12.000000}\selectfont 1}%
\end{pgfscope}%
\begin{pgfscope}%
\pgfpathrectangle{\pgfqpoint{0.536502in}{0.394792in}}{\pgfqpoint{6.421828in}{1.113538in}}%
\pgfusepath{clip}%
\pgfsetrectcap%
\pgfsetroundjoin%
\pgfsetlinewidth{0.803000pt}%
\definecolor{currentstroke}{rgb}{1.000000,1.000000,1.000000}%
\pgfsetstrokecolor{currentstroke}%
\pgfsetdash{}{0pt}%
\pgfpathmoveto{\pgfqpoint{3.558539in}{0.394792in}}%
\pgfpathlineto{\pgfqpoint{3.558539in}{1.508330in}}%
\pgfusepath{stroke}%
\end{pgfscope}%
\begin{pgfscope}%
\definecolor{textcolor}{rgb}{0.333333,0.333333,0.333333}%
\pgfsetstrokecolor{textcolor}%
\pgfsetfillcolor{textcolor}%
\pgftext[x=3.558539in,y=0.297570in,,top]{\color{textcolor}\sffamily\fontsize{10.000000}{12.000000}\selectfont 24}%
\end{pgfscope}%
\begin{pgfscope}%
\pgfpathrectangle{\pgfqpoint{0.536502in}{0.394792in}}{\pgfqpoint{6.421828in}{1.113538in}}%
\pgfusepath{clip}%
\pgfsetrectcap%
\pgfsetroundjoin%
\pgfsetlinewidth{0.803000pt}%
\definecolor{currentstroke}{rgb}{1.000000,1.000000,1.000000}%
\pgfsetstrokecolor{currentstroke}%
\pgfsetdash{}{0pt}%
\pgfpathmoveto{\pgfqpoint{3.810375in}{0.394792in}}%
\pgfpathlineto{\pgfqpoint{3.810375in}{1.508330in}}%
\pgfusepath{stroke}%
\end{pgfscope}%
\begin{pgfscope}%
\definecolor{textcolor}{rgb}{0.333333,0.333333,0.333333}%
\pgfsetstrokecolor{textcolor}%
\pgfsetfillcolor{textcolor}%
\pgftext[x=3.810375in,y=0.297570in,,top]{\color{textcolor}\sffamily\fontsize{10.000000}{12.000000}\selectfont 1}%
\end{pgfscope}%
\begin{pgfscope}%
\pgfpathrectangle{\pgfqpoint{0.536502in}{0.394792in}}{\pgfqpoint{6.421828in}{1.113538in}}%
\pgfusepath{clip}%
\pgfsetrectcap%
\pgfsetroundjoin%
\pgfsetlinewidth{0.803000pt}%
\definecolor{currentstroke}{rgb}{1.000000,1.000000,1.000000}%
\pgfsetstrokecolor{currentstroke}%
\pgfsetdash{}{0pt}%
\pgfpathmoveto{\pgfqpoint{6.832412in}{0.394792in}}%
\pgfpathlineto{\pgfqpoint{6.832412in}{1.508330in}}%
\pgfusepath{stroke}%
\end{pgfscope}%
\begin{pgfscope}%
\definecolor{textcolor}{rgb}{0.333333,0.333333,0.333333}%
\pgfsetstrokecolor{textcolor}%
\pgfsetfillcolor{textcolor}%
\pgftext[x=6.832412in,y=0.297570in,,top]{\color{textcolor}\sffamily\fontsize{10.000000}{12.000000}\selectfont 24}%
\end{pgfscope}%
\begin{pgfscope}%
\definecolor{textcolor}{rgb}{0.333333,0.333333,0.333333}%
\pgfsetstrokecolor{textcolor}%
\pgfsetfillcolor{textcolor}%
\pgftext[x=3.651089in,y=0.172085in,,top]{\color{textcolor}\sffamily\fontsize{10.000000}{12.000000}\selectfont Layer}%
\end{pgfscope}%
\begin{pgfscope}%
\pgfpathrectangle{\pgfqpoint{0.536502in}{0.394792in}}{\pgfqpoint{6.421828in}{1.113538in}}%
\pgfusepath{clip}%
\pgfsetrectcap%
\pgfsetroundjoin%
\pgfsetlinewidth{0.803000pt}%
\definecolor{currentstroke}{rgb}{1.000000,1.000000,1.000000}%
\pgfsetstrokecolor{currentstroke}%
\pgfsetdash{}{0pt}%
\pgfpathmoveto{\pgfqpoint{0.536502in}{0.717781in}}%
\pgfpathlineto{\pgfqpoint{6.958330in}{0.717781in}}%
\pgfusepath{stroke}%
\end{pgfscope}%
\begin{pgfscope}%
\definecolor{textcolor}{rgb}{0.333333,0.333333,0.333333}%
\pgfsetstrokecolor{textcolor}%
\pgfsetfillcolor{textcolor}%
\pgftext[x=0.041670in, y=0.665020in, left, base]{\color{textcolor}\sffamily\fontsize{10.000000}{12.000000}\selectfont 0.700}%
\end{pgfscope}%
\begin{pgfscope}%
\pgfpathrectangle{\pgfqpoint{0.536502in}{0.394792in}}{\pgfqpoint{6.421828in}{1.113538in}}%
\pgfusepath{clip}%
\pgfsetrectcap%
\pgfsetroundjoin%
\pgfsetlinewidth{0.803000pt}%
\definecolor{currentstroke}{rgb}{1.000000,1.000000,1.000000}%
\pgfsetstrokecolor{currentstroke}%
\pgfsetdash{}{0pt}%
\pgfpathmoveto{\pgfqpoint{0.536502in}{1.224051in}}%
\pgfpathlineto{\pgfqpoint{6.958330in}{1.224051in}}%
\pgfusepath{stroke}%
\end{pgfscope}%
\begin{pgfscope}%
\definecolor{textcolor}{rgb}{0.333333,0.333333,0.333333}%
\pgfsetstrokecolor{textcolor}%
\pgfsetfillcolor{textcolor}%
\pgftext[x=0.041670in, y=1.171290in, left, base]{\color{textcolor}\sffamily\fontsize{10.000000}{12.000000}\selectfont 0.800}%
\end{pgfscope}%
\begin{pgfscope}%
\pgfpathrectangle{\pgfqpoint{0.536502in}{0.394792in}}{\pgfqpoint{6.421828in}{1.113538in}}%
\pgfusepath{clip}%
\pgfsetrectcap%
\pgfsetroundjoin%
\pgfsetlinewidth{1.003750pt}%
\definecolor{currentstroke}{rgb}{0.886275,0.290196,0.200000}%
\pgfsetstrokecolor{currentstroke}%
\pgfsetdash{}{0pt}%
\pgfpathmoveto{\pgfqpoint{0.662421in}{0.445408in}}%
\pgfpathlineto{\pgfqpoint{0.788339in}{0.445408in}}%
\pgfpathlineto{\pgfqpoint{0.914257in}{0.445408in}}%
\pgfpathlineto{\pgfqpoint{1.040175in}{0.445408in}}%
\pgfpathlineto{\pgfqpoint{1.166093in}{0.445408in}}%
\pgfpathlineto{\pgfqpoint{1.292011in}{0.471136in}}%
\pgfpathlineto{\pgfqpoint{1.417930in}{0.487363in}}%
\pgfpathlineto{\pgfqpoint{1.543848in}{0.471136in}}%
\pgfpathlineto{\pgfqpoint{1.669766in}{0.523062in}}%
\pgfpathlineto{\pgfqpoint{1.795684in}{0.542534in}}%
\pgfpathlineto{\pgfqpoint{1.921602in}{0.578232in}}%
\pgfpathlineto{\pgfqpoint{2.047521in}{0.549024in}}%
\pgfusepath{stroke}%
\end{pgfscope}%
\begin{pgfscope}%
\pgfpathrectangle{\pgfqpoint{0.536502in}{0.394792in}}{\pgfqpoint{6.421828in}{1.113538in}}%
\pgfusepath{clip}%
\pgfsetrectcap%
\pgfsetroundjoin%
\pgfsetlinewidth{1.003750pt}%
\definecolor{currentstroke}{rgb}{0.886275,0.290196,0.200000}%
\pgfsetstrokecolor{currentstroke}%
\pgfsetdash{}{0pt}%
\pgfpathmoveto{\pgfqpoint{3.810375in}{0.445408in}}%
\pgfpathlineto{\pgfqpoint{3.936293in}{0.445408in}}%
\pgfpathlineto{\pgfqpoint{4.062212in}{0.445408in}}%
\pgfpathlineto{\pgfqpoint{4.188130in}{0.454910in}}%
\pgfpathlineto{\pgfqpoint{4.314048in}{0.445408in}}%
\pgfpathlineto{\pgfqpoint{4.439966in}{0.445408in}}%
\pgfpathlineto{\pgfqpoint{4.565884in}{0.445408in}}%
\pgfpathlineto{\pgfqpoint{4.691803in}{0.477627in}}%
\pgfpathlineto{\pgfqpoint{4.817721in}{0.587968in}}%
\pgfpathlineto{\pgfqpoint{4.943639in}{0.587968in}}%
\pgfpathlineto{\pgfqpoint{5.069557in}{0.643139in}}%
\pgfpathlineto{\pgfqpoint{5.195475in}{0.630157in}}%
\pgfusepath{stroke}%
\end{pgfscope}%
\begin{pgfscope}%
\pgfpathrectangle{\pgfqpoint{0.536502in}{0.394792in}}{\pgfqpoint{6.421828in}{1.113538in}}%
\pgfusepath{clip}%
\pgfsetrectcap%
\pgfsetroundjoin%
\pgfsetlinewidth{1.003750pt}%
\definecolor{currentstroke}{rgb}{0.203922,0.541176,0.741176}%
\pgfsetstrokecolor{currentstroke}%
\pgfsetdash{}{0pt}%
\pgfpathmoveto{\pgfqpoint{0.662421in}{0.445408in}}%
\pgfpathlineto{\pgfqpoint{0.788339in}{0.461401in}}%
\pgfpathlineto{\pgfqpoint{0.914257in}{0.467891in}}%
\pgfpathlineto{\pgfqpoint{1.040175in}{0.542534in}}%
\pgfpathlineto{\pgfqpoint{1.166093in}{0.591213in}}%
\pgfpathlineto{\pgfqpoint{1.292011in}{0.594459in}}%
\pgfpathlineto{\pgfqpoint{1.417930in}{0.545779in}}%
\pgfpathlineto{\pgfqpoint{1.543848in}{0.500344in}}%
\pgfpathlineto{\pgfqpoint{1.669766in}{0.542534in}}%
\pgfpathlineto{\pgfqpoint{1.795684in}{0.626912in}}%
\pgfpathlineto{\pgfqpoint{1.921602in}{0.600949in}}%
\pgfpathlineto{\pgfqpoint{2.047521in}{0.633403in}}%
\pgfusepath{stroke}%
\end{pgfscope}%
\begin{pgfscope}%
\pgfpathrectangle{\pgfqpoint{0.536502in}{0.394792in}}{\pgfqpoint{6.421828in}{1.113538in}}%
\pgfusepath{clip}%
\pgfsetrectcap%
\pgfsetroundjoin%
\pgfsetlinewidth{1.003750pt}%
\definecolor{currentstroke}{rgb}{0.203922,0.541176,0.741176}%
\pgfsetstrokecolor{currentstroke}%
\pgfsetdash{}{0pt}%
\pgfpathmoveto{\pgfqpoint{3.810375in}{0.445408in}}%
\pgfpathlineto{\pgfqpoint{3.936293in}{0.445408in}}%
\pgfpathlineto{\pgfqpoint{4.062212in}{0.445408in}}%
\pgfpathlineto{\pgfqpoint{4.188130in}{0.445408in}}%
\pgfpathlineto{\pgfqpoint{4.314048in}{0.445408in}}%
\pgfpathlineto{\pgfqpoint{4.439966in}{0.445408in}}%
\pgfpathlineto{\pgfqpoint{4.565884in}{0.445408in}}%
\pgfpathlineto{\pgfqpoint{4.691803in}{0.445408in}}%
\pgfpathlineto{\pgfqpoint{4.817721in}{0.445408in}}%
\pgfpathlineto{\pgfqpoint{4.943639in}{0.445408in}}%
\pgfpathlineto{\pgfqpoint{5.069557in}{0.445408in}}%
\pgfpathlineto{\pgfqpoint{5.195475in}{0.445408in}}%
\pgfusepath{stroke}%
\end{pgfscope}%
\begin{pgfscope}%
\pgfpathrectangle{\pgfqpoint{0.536502in}{0.394792in}}{\pgfqpoint{6.421828in}{1.113538in}}%
\pgfusepath{clip}%
\pgfsetrectcap%
\pgfsetroundjoin%
\pgfsetlinewidth{1.003750pt}%
\definecolor{currentstroke}{rgb}{0.596078,0.556863,0.835294}%
\pgfsetstrokecolor{currentstroke}%
\pgfsetdash{}{0pt}%
\pgfpathmoveto{\pgfqpoint{0.662421in}{0.448419in}}%
\pgfpathlineto{\pgfqpoint{0.788339in}{0.506835in}}%
\pgfpathlineto{\pgfqpoint{0.914257in}{0.519816in}}%
\pgfpathlineto{\pgfqpoint{1.040175in}{0.597704in}}%
\pgfpathlineto{\pgfqpoint{1.166093in}{0.724272in}}%
\pgfpathlineto{\pgfqpoint{1.292011in}{0.815141in}}%
\pgfpathlineto{\pgfqpoint{1.417930in}{0.847594in}}%
\pgfpathlineto{\pgfqpoint{1.543848in}{0.831367in}}%
\pgfpathlineto{\pgfqpoint{1.669766in}{0.837858in}}%
\pgfpathlineto{\pgfqpoint{1.795684in}{0.889783in}}%
\pgfpathlineto{\pgfqpoint{1.921602in}{0.811895in}}%
\pgfpathlineto{\pgfqpoint{2.047521in}{0.795669in}}%
\pgfpathlineto{\pgfqpoint{2.173439in}{0.841103in}}%
\pgfpathlineto{\pgfqpoint{2.299357in}{0.675592in}}%
\pgfpathlineto{\pgfqpoint{2.425275in}{0.587968in}}%
\pgfpathlineto{\pgfqpoint{2.551193in}{0.584723in}}%
\pgfusepath{stroke}%
\end{pgfscope}%
\begin{pgfscope}%
\pgfpathrectangle{\pgfqpoint{0.536502in}{0.394792in}}{\pgfqpoint{6.421828in}{1.113538in}}%
\pgfusepath{clip}%
\pgfsetrectcap%
\pgfsetroundjoin%
\pgfsetlinewidth{1.003750pt}%
\definecolor{currentstroke}{rgb}{0.596078,0.556863,0.835294}%
\pgfsetstrokecolor{currentstroke}%
\pgfsetdash{}{0pt}%
\pgfpathmoveto{\pgfqpoint{3.810375in}{0.448419in}}%
\pgfpathlineto{\pgfqpoint{3.936293in}{0.523062in}}%
\pgfpathlineto{\pgfqpoint{4.062212in}{0.500344in}}%
\pgfpathlineto{\pgfqpoint{4.188130in}{0.630157in}}%
\pgfpathlineto{\pgfqpoint{4.314048in}{0.880047in}}%
\pgfpathlineto{\pgfqpoint{4.439966in}{0.889783in}}%
\pgfpathlineto{\pgfqpoint{4.565884in}{0.951444in}}%
\pgfpathlineto{\pgfqpoint{4.691803in}{1.022841in}}%
\pgfpathlineto{\pgfqpoint{4.817721in}{1.035823in}}%
\pgfpathlineto{\pgfqpoint{4.943639in}{1.094238in}}%
\pgfpathlineto{\pgfqpoint{5.069557in}{1.065031in}}%
\pgfpathlineto{\pgfqpoint{5.195475in}{1.103974in}}%
\pgfpathlineto{\pgfqpoint{5.321394in}{1.162390in}}%
\pgfpathlineto{\pgfqpoint{5.447312in}{1.246769in}}%
\pgfpathlineto{\pgfqpoint{5.573230in}{1.415525in}}%
\pgfpathlineto{\pgfqpoint{5.699148in}{1.078012in}}%
\pgfusepath{stroke}%
\end{pgfscope}%
\begin{pgfscope}%
\pgfpathrectangle{\pgfqpoint{0.536502in}{0.394792in}}{\pgfqpoint{6.421828in}{1.113538in}}%
\pgfusepath{clip}%
\pgfsetrectcap%
\pgfsetroundjoin%
\pgfsetlinewidth{1.003750pt}%
\definecolor{currentstroke}{rgb}{0.466667,0.466667,0.466667}%
\pgfsetstrokecolor{currentstroke}%
\pgfsetdash{}{0pt}%
\pgfpathmoveto{\pgfqpoint{0.662421in}{0.506835in}}%
\pgfpathlineto{\pgfqpoint{0.788339in}{0.532798in}}%
\pgfpathlineto{\pgfqpoint{0.914257in}{0.587968in}}%
\pgfpathlineto{\pgfqpoint{1.040175in}{0.600949in}}%
\pgfpathlineto{\pgfqpoint{1.166093in}{0.675592in}}%
\pgfpathlineto{\pgfqpoint{1.292011in}{0.701554in}}%
\pgfpathlineto{\pgfqpoint{1.417930in}{0.792423in}}%
\pgfpathlineto{\pgfqpoint{1.543848in}{0.915746in}}%
\pgfpathlineto{\pgfqpoint{1.669766in}{0.889783in}}%
\pgfpathlineto{\pgfqpoint{1.795684in}{0.928727in}}%
\pgfpathlineto{\pgfqpoint{1.921602in}{0.944954in}}%
\pgfpathlineto{\pgfqpoint{2.047521in}{0.938463in}}%
\pgfpathlineto{\pgfqpoint{2.173439in}{0.906010in}}%
\pgfpathlineto{\pgfqpoint{2.299357in}{0.893028in}}%
\pgfpathlineto{\pgfqpoint{2.425275in}{0.889783in}}%
\pgfpathlineto{\pgfqpoint{2.551193in}{0.886538in}}%
\pgfpathlineto{\pgfqpoint{2.677112in}{0.870311in}}%
\pgfpathlineto{\pgfqpoint{2.803030in}{0.844349in}}%
\pgfpathlineto{\pgfqpoint{2.928948in}{0.789178in}}%
\pgfpathlineto{\pgfqpoint{3.054866in}{0.766461in}}%
\pgfpathlineto{\pgfqpoint{3.180784in}{0.756725in}}%
\pgfpathlineto{\pgfqpoint{3.306703in}{0.734008in}}%
\pgfpathlineto{\pgfqpoint{3.432621in}{0.730762in}}%
\pgfpathlineto{\pgfqpoint{3.558539in}{0.695064in}}%
\pgfusepath{stroke}%
\end{pgfscope}%
\begin{pgfscope}%
\pgfpathrectangle{\pgfqpoint{0.536502in}{0.394792in}}{\pgfqpoint{6.421828in}{1.113538in}}%
\pgfusepath{clip}%
\pgfsetrectcap%
\pgfsetroundjoin%
\pgfsetlinewidth{1.003750pt}%
\definecolor{currentstroke}{rgb}{0.466667,0.466667,0.466667}%
\pgfsetstrokecolor{currentstroke}%
\pgfsetdash{}{0pt}%
\pgfpathmoveto{\pgfqpoint{3.810375in}{0.516571in}}%
\pgfpathlineto{\pgfqpoint{3.936293in}{0.526307in}}%
\pgfpathlineto{\pgfqpoint{4.062212in}{0.597704in}}%
\pgfpathlineto{\pgfqpoint{4.188130in}{0.649629in}}%
\pgfpathlineto{\pgfqpoint{4.314048in}{0.753480in}}%
\pgfpathlineto{\pgfqpoint{4.439966in}{0.805405in}}%
\pgfpathlineto{\pgfqpoint{4.565884in}{0.880047in}}%
\pgfpathlineto{\pgfqpoint{4.691803in}{0.951444in}}%
\pgfpathlineto{\pgfqpoint{4.817721in}{0.974162in}}%
\pgfpathlineto{\pgfqpoint{4.943639in}{0.974162in}}%
\pgfpathlineto{\pgfqpoint{5.069557in}{0.970916in}}%
\pgfpathlineto{\pgfqpoint{5.195475in}{1.048804in}}%
\pgfpathlineto{\pgfqpoint{5.321394in}{1.149409in}}%
\pgfpathlineto{\pgfqpoint{5.447312in}{1.227297in}}%
\pgfpathlineto{\pgfqpoint{5.573230in}{1.250014in}}%
\pgfpathlineto{\pgfqpoint{5.699148in}{1.243523in}}%
\pgfpathlineto{\pgfqpoint{5.825066in}{1.305184in}}%
\pgfpathlineto{\pgfqpoint{5.950984in}{1.337638in}}%
\pgfpathlineto{\pgfqpoint{6.076903in}{1.383072in}}%
\pgfpathlineto{\pgfqpoint{6.202821in}{1.386318in}}%
\pgfpathlineto{\pgfqpoint{6.328739in}{1.383072in}}%
\pgfpathlineto{\pgfqpoint{6.454657in}{1.409035in}}%
\pgfpathlineto{\pgfqpoint{6.580575in}{1.399299in}}%
\pgfpathlineto{\pgfqpoint{6.706494in}{1.457715in}}%
\pgfusepath{stroke}%
\end{pgfscope}%
\begin{pgfscope}%
\pgfpathrectangle{\pgfqpoint{0.536502in}{0.394792in}}{\pgfqpoint{6.421828in}{1.113538in}}%
\pgfusepath{clip}%
\pgfsetbuttcap%
\pgfsetroundjoin%
\pgfsetlinewidth{1.003750pt}%
\definecolor{currentstroke}{rgb}{0.000000,0.392157,0.000000}%
\pgfsetstrokecolor{currentstroke}%
\pgfsetdash{{3.700000pt}{1.600000pt}}{0.000000pt}%
\pgfpathmoveto{\pgfqpoint{0.536502in}{0.445408in}}%
\pgfpathlineto{\pgfqpoint{6.958330in}{0.445408in}}%
\pgfusepath{stroke}%
\end{pgfscope}%
\begin{pgfscope}%
\pgfpathrectangle{\pgfqpoint{0.536502in}{0.394792in}}{\pgfqpoint{6.421828in}{1.113538in}}%
\pgfusepath{clip}%
\pgfsetbuttcap%
\pgfsetroundjoin%
\pgfsetlinewidth{1.003750pt}%
\definecolor{currentstroke}{rgb}{0.803922,0.521569,0.247059}%
\pgfsetstrokecolor{currentstroke}%
\pgfsetdash{{3.700000pt}{1.600000pt}}{0.000000pt}%
\pgfpathmoveto{\pgfqpoint{0.536502in}{1.327837in}}%
\pgfpathlineto{\pgfqpoint{6.958330in}{1.327837in}}%
\pgfusepath{stroke}%
\end{pgfscope}%
\begin{pgfscope}%
\pgfpathrectangle{\pgfqpoint{0.536502in}{0.394792in}}{\pgfqpoint{6.421828in}{1.113538in}}%
\pgfusepath{clip}%
\pgfsetrectcap%
\pgfsetroundjoin%
\pgfsetlinewidth{6.022500pt}%
\definecolor{currentstroke}{rgb}{1.000000,1.000000,1.000000}%
\pgfsetstrokecolor{currentstroke}%
\pgfsetdash{}{0pt}%
\pgfpathmoveto{\pgfqpoint{3.684457in}{0.394792in}}%
\pgfpathlineto{\pgfqpoint{3.684457in}{1.508330in}}%
\pgfusepath{stroke}%
\end{pgfscope}%
\begin{pgfscope}%
\pgfsetrectcap%
\pgfsetmiterjoin%
\pgfsetlinewidth{1.003750pt}%
\definecolor{currentstroke}{rgb}{1.000000,1.000000,1.000000}%
\pgfsetstrokecolor{currentstroke}%
\pgfsetdash{}{0pt}%
\pgfpathmoveto{\pgfqpoint{0.536502in}{0.394792in}}%
\pgfpathlineto{\pgfqpoint{0.536502in}{1.508330in}}%
\pgfusepath{stroke}%
\end{pgfscope}%
\begin{pgfscope}%
\pgfsetrectcap%
\pgfsetmiterjoin%
\pgfsetlinewidth{1.003750pt}%
\definecolor{currentstroke}{rgb}{1.000000,1.000000,1.000000}%
\pgfsetstrokecolor{currentstroke}%
\pgfsetdash{}{0pt}%
\pgfpathmoveto{\pgfqpoint{6.958330in}{0.394792in}}%
\pgfpathlineto{\pgfqpoint{6.958330in}{1.508330in}}%
\pgfusepath{stroke}%
\end{pgfscope}%
\begin{pgfscope}%
\pgfsetrectcap%
\pgfsetmiterjoin%
\pgfsetlinewidth{1.003750pt}%
\definecolor{currentstroke}{rgb}{1.000000,1.000000,1.000000}%
\pgfsetstrokecolor{currentstroke}%
\pgfsetdash{}{0pt}%
\pgfpathmoveto{\pgfqpoint{0.536502in}{0.394792in}}%
\pgfpathlineto{\pgfqpoint{6.958330in}{0.394792in}}%
\pgfusepath{stroke}%
\end{pgfscope}%
\begin{pgfscope}%
\pgfsetrectcap%
\pgfsetmiterjoin%
\pgfsetlinewidth{1.003750pt}%
\definecolor{currentstroke}{rgb}{1.000000,1.000000,1.000000}%
\pgfsetstrokecolor{currentstroke}%
\pgfsetdash{}{0pt}%
\pgfpathmoveto{\pgfqpoint{0.536502in}{1.508330in}}%
\pgfpathlineto{\pgfqpoint{6.958330in}{1.508330in}}%
\pgfusepath{stroke}%
\end{pgfscope}%
\end{pgfpicture}%
\makeatother%
\endgroup%

%% file: figures/comparison-nondominanthandshape-nox.pgf
\begingroup%
\makeatletter%
\begin{pgfpicture}%
\pgfpathrectangle{\pgfpointorigin}{\pgfqpoint{7.000000in}{1.250000in}}%
\pgfusepath{use as bounding box, clip}%
\begin{pgfscope}%
\pgfsetbuttcap%
\pgfsetmiterjoin%
\definecolor{currentfill}{rgb}{1.000000,1.000000,1.000000}%
\pgfsetfillcolor{currentfill}%
\pgfsetlinewidth{0.000000pt}%
\definecolor{currentstroke}{rgb}{0.500000,0.500000,0.500000}%
\pgfsetstrokecolor{currentstroke}%
\pgfsetdash{}{0pt}%
\pgfpathmoveto{\pgfqpoint{0.000000in}{0.000000in}}%
\pgfpathlineto{\pgfqpoint{7.000000in}{0.000000in}}%
\pgfpathlineto{\pgfqpoint{7.000000in}{1.250000in}}%
\pgfpathlineto{\pgfqpoint{0.000000in}{1.250000in}}%
\pgfpathlineto{\pgfqpoint{0.000000in}{0.000000in}}%
\pgfpathclose%
\pgfusepath{fill}%
\end{pgfscope}%
\begin{pgfscope}%
\pgfsetbuttcap%
\pgfsetmiterjoin%
\definecolor{currentfill}{rgb}{0.898039,0.898039,0.898039}%
\pgfsetfillcolor{currentfill}%
\pgfsetlinewidth{0.000000pt}%
\definecolor{currentstroke}{rgb}{0.000000,0.000000,0.000000}%
\pgfsetstrokecolor{currentstroke}%
\pgfsetstrokeopacity{0.000000}%
\pgfsetdash{}{0pt}%
\pgfpathmoveto{\pgfqpoint{0.536502in}{0.041670in}}%
\pgfpathlineto{\pgfqpoint{6.958330in}{0.041670in}}%
\pgfpathlineto{\pgfqpoint{6.958330in}{1.208330in}}%
\pgfpathlineto{\pgfqpoint{0.536502in}{1.208330in}}%
\pgfpathlineto{\pgfqpoint{0.536502in}{0.041670in}}%
\pgfpathclose%
\pgfusepath{fill}%
\end{pgfscope}%
\begin{pgfscope}%
\pgfpathrectangle{\pgfqpoint{0.536502in}{0.041670in}}{\pgfqpoint{6.421828in}{1.166660in}}%
\pgfusepath{clip}%
\pgfsetrectcap%
\pgfsetroundjoin%
\pgfsetlinewidth{0.803000pt}%
\definecolor{currentstroke}{rgb}{1.000000,1.000000,1.000000}%
\pgfsetstrokecolor{currentstroke}%
\pgfsetdash{}{0pt}%
\pgfpathmoveto{\pgfqpoint{0.662421in}{0.041670in}}%
\pgfpathlineto{\pgfqpoint{0.662421in}{1.208330in}}%
\pgfusepath{stroke}%
\end{pgfscope}%
\begin{pgfscope}%
\pgfpathrectangle{\pgfqpoint{0.536502in}{0.041670in}}{\pgfqpoint{6.421828in}{1.166660in}}%
\pgfusepath{clip}%
\pgfsetrectcap%
\pgfsetroundjoin%
\pgfsetlinewidth{0.803000pt}%
\definecolor{currentstroke}{rgb}{1.000000,1.000000,1.000000}%
\pgfsetstrokecolor{currentstroke}%
\pgfsetdash{}{0pt}%
\pgfpathmoveto{\pgfqpoint{3.558539in}{0.041670in}}%
\pgfpathlineto{\pgfqpoint{3.558539in}{1.208330in}}%
\pgfusepath{stroke}%
\end{pgfscope}%
\begin{pgfscope}%
\pgfpathrectangle{\pgfqpoint{0.536502in}{0.041670in}}{\pgfqpoint{6.421828in}{1.166660in}}%
\pgfusepath{clip}%
\pgfsetrectcap%
\pgfsetroundjoin%
\pgfsetlinewidth{0.803000pt}%
\definecolor{currentstroke}{rgb}{1.000000,1.000000,1.000000}%
\pgfsetstrokecolor{currentstroke}%
\pgfsetdash{}{0pt}%
\pgfpathmoveto{\pgfqpoint{3.810375in}{0.041670in}}%
\pgfpathlineto{\pgfqpoint{3.810375in}{1.208330in}}%
\pgfusepath{stroke}%
\end{pgfscope}%
\begin{pgfscope}%
\pgfpathrectangle{\pgfqpoint{0.536502in}{0.041670in}}{\pgfqpoint{6.421828in}{1.166660in}}%
\pgfusepath{clip}%
\pgfsetrectcap%
\pgfsetroundjoin%
\pgfsetlinewidth{0.803000pt}%
\definecolor{currentstroke}{rgb}{1.000000,1.000000,1.000000}%
\pgfsetstrokecolor{currentstroke}%
\pgfsetdash{}{0pt}%
\pgfpathmoveto{\pgfqpoint{6.832412in}{0.041670in}}%
\pgfpathlineto{\pgfqpoint{6.832412in}{1.208330in}}%
\pgfusepath{stroke}%
\end{pgfscope}%
\begin{pgfscope}%
\pgfpathrectangle{\pgfqpoint{0.536502in}{0.041670in}}{\pgfqpoint{6.421828in}{1.166660in}}%
\pgfusepath{clip}%
\pgfsetrectcap%
\pgfsetroundjoin%
\pgfsetlinewidth{0.803000pt}%
\definecolor{currentstroke}{rgb}{1.000000,1.000000,1.000000}%
\pgfsetstrokecolor{currentstroke}%
\pgfsetdash{}{0pt}%
\pgfpathmoveto{\pgfqpoint{0.536502in}{0.154387in}}%
\pgfpathlineto{\pgfqpoint{6.958330in}{0.154387in}}%
\pgfusepath{stroke}%
\end{pgfscope}%
\begin{pgfscope}%
\definecolor{textcolor}{rgb}{0.333333,0.333333,0.333333}%
\pgfsetstrokecolor{textcolor}%
\pgfsetfillcolor{textcolor}%
\pgftext[x=0.041670in, y=0.101626in, left, base]{\color{textcolor}\sffamily\fontsize{10.000000}{12.000000}\selectfont 0.400}%
\end{pgfscope}%
\begin{pgfscope}%
\pgfpathrectangle{\pgfqpoint{0.536502in}{0.041670in}}{\pgfqpoint{6.421828in}{1.166660in}}%
\pgfusepath{clip}%
\pgfsetrectcap%
\pgfsetroundjoin%
\pgfsetlinewidth{0.803000pt}%
\definecolor{currentstroke}{rgb}{1.000000,1.000000,1.000000}%
\pgfsetstrokecolor{currentstroke}%
\pgfsetdash{}{0pt}%
\pgfpathmoveto{\pgfqpoint{0.536502in}{0.631887in}}%
\pgfpathlineto{\pgfqpoint{6.958330in}{0.631887in}}%
\pgfusepath{stroke}%
\end{pgfscope}%
\begin{pgfscope}%
\definecolor{textcolor}{rgb}{0.333333,0.333333,0.333333}%
\pgfsetstrokecolor{textcolor}%
\pgfsetfillcolor{textcolor}%
\pgftext[x=0.041670in, y=0.579125in, left, base]{\color{textcolor}\sffamily\fontsize{10.000000}{12.000000}\selectfont 0.600}%
\end{pgfscope}%
\begin{pgfscope}%
\pgfpathrectangle{\pgfqpoint{0.536502in}{0.041670in}}{\pgfqpoint{6.421828in}{1.166660in}}%
\pgfusepath{clip}%
\pgfsetrectcap%
\pgfsetroundjoin%
\pgfsetlinewidth{0.803000pt}%
\definecolor{currentstroke}{rgb}{1.000000,1.000000,1.000000}%
\pgfsetstrokecolor{currentstroke}%
\pgfsetdash{}{0pt}%
\pgfpathmoveto{\pgfqpoint{0.536502in}{1.109387in}}%
\pgfpathlineto{\pgfqpoint{6.958330in}{1.109387in}}%
\pgfusepath{stroke}%
\end{pgfscope}%
\begin{pgfscope}%
\definecolor{textcolor}{rgb}{0.333333,0.333333,0.333333}%
\pgfsetstrokecolor{textcolor}%
\pgfsetfillcolor{textcolor}%
\pgftext[x=0.041670in, y=1.056625in, left, base]{\color{textcolor}\sffamily\fontsize{10.000000}{12.000000}\selectfont 0.800}%
\end{pgfscope}%
\begin{pgfscope}%
\pgfpathrectangle{\pgfqpoint{0.536502in}{0.041670in}}{\pgfqpoint{6.421828in}{1.166660in}}%
\pgfusepath{clip}%
\pgfsetrectcap%
\pgfsetroundjoin%
\pgfsetlinewidth{1.003750pt}%
\definecolor{currentstroke}{rgb}{0.886275,0.290196,0.200000}%
\pgfsetstrokecolor{currentstroke}%
\pgfsetdash{}{0pt}%
\pgfpathmoveto{\pgfqpoint{0.662421in}{0.094700in}}%
\pgfpathlineto{\pgfqpoint{0.788339in}{0.094700in}}%
\pgfpathlineto{\pgfqpoint{0.914257in}{0.094700in}}%
\pgfpathlineto{\pgfqpoint{1.040175in}{0.094700in}}%
\pgfpathlineto{\pgfqpoint{1.166093in}{0.094700in}}%
\pgfpathlineto{\pgfqpoint{1.292011in}{0.094700in}}%
\pgfpathlineto{\pgfqpoint{1.417930in}{0.094700in}}%
\pgfpathlineto{\pgfqpoint{1.543848in}{0.094700in}}%
\pgfpathlineto{\pgfqpoint{1.669766in}{0.094700in}}%
\pgfpathlineto{\pgfqpoint{1.795684in}{0.094700in}}%
\pgfpathlineto{\pgfqpoint{1.921602in}{0.099291in}}%
\pgfpathlineto{\pgfqpoint{2.047521in}{0.094700in}}%
\pgfusepath{stroke}%
\end{pgfscope}%
\begin{pgfscope}%
\pgfpathrectangle{\pgfqpoint{0.536502in}{0.041670in}}{\pgfqpoint{6.421828in}{1.166660in}}%
\pgfusepath{clip}%
\pgfsetrectcap%
\pgfsetroundjoin%
\pgfsetlinewidth{1.003750pt}%
\definecolor{currentstroke}{rgb}{0.886275,0.290196,0.200000}%
\pgfsetstrokecolor{currentstroke}%
\pgfsetdash{}{0pt}%
\pgfpathmoveto{\pgfqpoint{3.810375in}{0.094700in}}%
\pgfpathlineto{\pgfqpoint{3.936293in}{0.094700in}}%
\pgfpathlineto{\pgfqpoint{4.062212in}{0.094700in}}%
\pgfpathlineto{\pgfqpoint{4.188130in}{0.094700in}}%
\pgfpathlineto{\pgfqpoint{4.314048in}{0.094700in}}%
\pgfpathlineto{\pgfqpoint{4.439966in}{0.094700in}}%
\pgfpathlineto{\pgfqpoint{4.565884in}{0.094700in}}%
\pgfpathlineto{\pgfqpoint{4.691803in}{0.094700in}}%
\pgfpathlineto{\pgfqpoint{4.817721in}{0.096230in}}%
\pgfpathlineto{\pgfqpoint{4.943639in}{0.100822in}}%
\pgfpathlineto{\pgfqpoint{5.069557in}{0.094700in}}%
\pgfpathlineto{\pgfqpoint{5.195475in}{0.094700in}}%
\pgfusepath{stroke}%
\end{pgfscope}%
\begin{pgfscope}%
\pgfpathrectangle{\pgfqpoint{0.536502in}{0.041670in}}{\pgfqpoint{6.421828in}{1.166660in}}%
\pgfusepath{clip}%
\pgfsetrectcap%
\pgfsetroundjoin%
\pgfsetlinewidth{1.003750pt}%
\definecolor{currentstroke}{rgb}{0.203922,0.541176,0.741176}%
\pgfsetstrokecolor{currentstroke}%
\pgfsetdash{}{0pt}%
\pgfpathmoveto{\pgfqpoint{0.662421in}{0.103883in}}%
\pgfpathlineto{\pgfqpoint{0.788339in}{0.100822in}}%
\pgfpathlineto{\pgfqpoint{0.914257in}{0.094700in}}%
\pgfpathlineto{\pgfqpoint{1.040175in}{0.102352in}}%
\pgfpathlineto{\pgfqpoint{1.166093in}{0.105413in}}%
\pgfpathlineto{\pgfqpoint{1.292011in}{0.119187in}}%
\pgfpathlineto{\pgfqpoint{1.417930in}{0.119187in}}%
\pgfpathlineto{\pgfqpoint{1.543848in}{0.148266in}}%
\pgfpathlineto{\pgfqpoint{1.669766in}{0.177344in}}%
\pgfpathlineto{\pgfqpoint{1.795684in}{0.192649in}}%
\pgfpathlineto{\pgfqpoint{1.921602in}{0.211014in}}%
\pgfpathlineto{\pgfqpoint{2.047521in}{0.221727in}}%
\pgfusepath{stroke}%
\end{pgfscope}%
\begin{pgfscope}%
\pgfpathrectangle{\pgfqpoint{0.536502in}{0.041670in}}{\pgfqpoint{6.421828in}{1.166660in}}%
\pgfusepath{clip}%
\pgfsetrectcap%
\pgfsetroundjoin%
\pgfsetlinewidth{1.003750pt}%
\definecolor{currentstroke}{rgb}{0.203922,0.541176,0.741176}%
\pgfsetstrokecolor{currentstroke}%
\pgfsetdash{}{0pt}%
\pgfpathmoveto{\pgfqpoint{3.810375in}{0.094700in}}%
\pgfpathlineto{\pgfqpoint{3.936293in}{0.094700in}}%
\pgfpathlineto{\pgfqpoint{4.062212in}{0.094700in}}%
\pgfpathlineto{\pgfqpoint{4.188130in}{0.094700in}}%
\pgfpathlineto{\pgfqpoint{4.314048in}{0.094700in}}%
\pgfpathlineto{\pgfqpoint{4.439966in}{0.094700in}}%
\pgfpathlineto{\pgfqpoint{4.565884in}{0.094700in}}%
\pgfpathlineto{\pgfqpoint{4.691803in}{0.094700in}}%
\pgfpathlineto{\pgfqpoint{4.817721in}{0.094700in}}%
\pgfpathlineto{\pgfqpoint{4.943639in}{0.094700in}}%
\pgfpathlineto{\pgfqpoint{5.069557in}{0.094700in}}%
\pgfpathlineto{\pgfqpoint{5.195475in}{0.094700in}}%
\pgfusepath{stroke}%
\end{pgfscope}%
\begin{pgfscope}%
\pgfpathrectangle{\pgfqpoint{0.536502in}{0.041670in}}{\pgfqpoint{6.421828in}{1.166660in}}%
\pgfusepath{clip}%
\pgfsetrectcap%
\pgfsetroundjoin%
\pgfsetlinewidth{1.003750pt}%
\definecolor{currentstroke}{rgb}{0.596078,0.556863,0.835294}%
\pgfsetstrokecolor{currentstroke}%
\pgfsetdash{}{0pt}%
\pgfpathmoveto{\pgfqpoint{0.662421in}{0.094700in}}%
\pgfpathlineto{\pgfqpoint{0.788339in}{0.142144in}}%
\pgfpathlineto{\pgfqpoint{0.914257in}{0.152857in}}%
\pgfpathlineto{\pgfqpoint{1.040175in}{0.215605in}}%
\pgfpathlineto{\pgfqpoint{1.166093in}{0.238562in}}%
\pgfpathlineto{\pgfqpoint{1.292011in}{0.269171in}}%
\pgfpathlineto{\pgfqpoint{1.417930in}{0.328858in}}%
\pgfpathlineto{\pgfqpoint{1.543848in}{0.357937in}}%
\pgfpathlineto{\pgfqpoint{1.669766in}{0.383955in}}%
\pgfpathlineto{\pgfqpoint{1.795684in}{0.393137in}}%
\pgfpathlineto{\pgfqpoint{1.921602in}{0.376302in}}%
\pgfpathlineto{\pgfqpoint{2.047521in}{0.373241in}}%
\pgfpathlineto{\pgfqpoint{2.173439in}{0.420685in}}%
\pgfpathlineto{\pgfqpoint{2.299357in}{0.301310in}}%
\pgfpathlineto{\pgfqpoint{2.425275in}{0.217136in}}%
\pgfpathlineto{\pgfqpoint{2.551193in}{0.200301in}}%
\pgfusepath{stroke}%
\end{pgfscope}%
\begin{pgfscope}%
\pgfpathrectangle{\pgfqpoint{0.536502in}{0.041670in}}{\pgfqpoint{6.421828in}{1.166660in}}%
\pgfusepath{clip}%
\pgfsetrectcap%
\pgfsetroundjoin%
\pgfsetlinewidth{1.003750pt}%
\definecolor{currentstroke}{rgb}{0.596078,0.556863,0.835294}%
\pgfsetstrokecolor{currentstroke}%
\pgfsetdash{}{0pt}%
\pgfpathmoveto{\pgfqpoint{3.810375in}{0.094700in}}%
\pgfpathlineto{\pgfqpoint{3.936293in}{0.145205in}}%
\pgfpathlineto{\pgfqpoint{4.062212in}{0.157448in}}%
\pgfpathlineto{\pgfqpoint{4.188130in}{0.230910in}}%
\pgfpathlineto{\pgfqpoint{4.314048in}{0.310493in}}%
\pgfpathlineto{\pgfqpoint{4.439966in}{0.356406in}}%
\pgfpathlineto{\pgfqpoint{4.565884in}{0.409972in}}%
\pgfpathlineto{\pgfqpoint{4.691803in}{0.422216in}}%
\pgfpathlineto{\pgfqpoint{4.817721in}{0.488025in}}%
\pgfpathlineto{\pgfqpoint{4.943639in}{0.512512in}}%
\pgfpathlineto{\pgfqpoint{5.069557in}{0.547712in}}%
\pgfpathlineto{\pgfqpoint{5.195475in}{0.647191in}}%
\pgfpathlineto{\pgfqpoint{5.321394in}{0.761975in}}%
\pgfpathlineto{\pgfqpoint{5.447312in}{0.967055in}}%
\pgfpathlineto{\pgfqpoint{5.573230in}{1.155300in}}%
\pgfpathlineto{\pgfqpoint{5.699148in}{1.066534in}}%
\pgfusepath{stroke}%
\end{pgfscope}%
\begin{pgfscope}%
\pgfpathrectangle{\pgfqpoint{0.536502in}{0.041670in}}{\pgfqpoint{6.421828in}{1.166660in}}%
\pgfusepath{clip}%
\pgfsetrectcap%
\pgfsetroundjoin%
\pgfsetlinewidth{1.003750pt}%
\definecolor{currentstroke}{rgb}{0.466667,0.466667,0.466667}%
\pgfsetstrokecolor{currentstroke}%
\pgfsetdash{}{0pt}%
\pgfpathmoveto{\pgfqpoint{0.662421in}{0.139083in}}%
\pgfpathlineto{\pgfqpoint{0.788339in}{0.152857in}}%
\pgfpathlineto{\pgfqpoint{0.914257in}{0.203362in}}%
\pgfpathlineto{\pgfqpoint{1.040175in}{0.197240in}}%
\pgfpathlineto{\pgfqpoint{1.166093in}{0.227849in}}%
\pgfpathlineto{\pgfqpoint{1.292011in}{0.247745in}}%
\pgfpathlineto{\pgfqpoint{1.417930in}{0.258458in}}%
\pgfpathlineto{\pgfqpoint{1.543848in}{0.295189in}}%
\pgfpathlineto{\pgfqpoint{1.669766in}{0.315084in}}%
\pgfpathlineto{\pgfqpoint{1.795684in}{0.333450in}}%
\pgfpathlineto{\pgfqpoint{1.921602in}{0.347224in}}%
\pgfpathlineto{\pgfqpoint{2.047521in}{0.365589in}}%
\pgfpathlineto{\pgfqpoint{2.173439in}{0.371711in}}%
\pgfpathlineto{\pgfqpoint{2.299357in}{0.371711in}}%
\pgfpathlineto{\pgfqpoint{2.425275in}{0.360998in}}%
\pgfpathlineto{\pgfqpoint{2.551193in}{0.322737in}}%
\pgfpathlineto{\pgfqpoint{2.677112in}{0.299780in}}%
\pgfpathlineto{\pgfqpoint{2.803030in}{0.276823in}}%
\pgfpathlineto{\pgfqpoint{2.928948in}{0.290597in}}%
\pgfpathlineto{\pgfqpoint{3.054866in}{0.273762in}}%
\pgfpathlineto{\pgfqpoint{3.180784in}{0.235501in}}%
\pgfpathlineto{\pgfqpoint{3.306703in}{0.220197in}}%
\pgfpathlineto{\pgfqpoint{3.432621in}{0.229379in}}%
\pgfpathlineto{\pgfqpoint{3.558539in}{0.232440in}}%
\pgfusepath{stroke}%
\end{pgfscope}%
\begin{pgfscope}%
\pgfpathrectangle{\pgfqpoint{0.536502in}{0.041670in}}{\pgfqpoint{6.421828in}{1.166660in}}%
\pgfusepath{clip}%
\pgfsetrectcap%
\pgfsetroundjoin%
\pgfsetlinewidth{1.003750pt}%
\definecolor{currentstroke}{rgb}{0.466667,0.466667,0.466667}%
\pgfsetstrokecolor{currentstroke}%
\pgfsetdash{}{0pt}%
\pgfpathmoveto{\pgfqpoint{3.810375in}{0.140613in}}%
\pgfpathlineto{\pgfqpoint{3.936293in}{0.140613in}}%
\pgfpathlineto{\pgfqpoint{4.062212in}{0.200301in}}%
\pgfpathlineto{\pgfqpoint{4.188130in}{0.220197in}}%
\pgfpathlineto{\pgfqpoint{4.314048in}{0.281415in}}%
\pgfpathlineto{\pgfqpoint{4.439966in}{0.296719in}}%
\pgfpathlineto{\pgfqpoint{4.565884in}{0.347224in}}%
\pgfpathlineto{\pgfqpoint{4.691803in}{0.406911in}}%
\pgfpathlineto{\pgfqpoint{4.817721in}{0.445172in}}%
\pgfpathlineto{\pgfqpoint{4.943639in}{0.481903in}}%
\pgfpathlineto{\pgfqpoint{5.069557in}{0.523225in}}%
\pgfpathlineto{\pgfqpoint{5.195475in}{0.581382in}}%
\pgfpathlineto{\pgfqpoint{5.321394in}{0.625765in}}%
\pgfpathlineto{\pgfqpoint{5.447312in}{0.685453in}}%
\pgfpathlineto{\pgfqpoint{5.573230in}{0.751262in}}%
\pgfpathlineto{\pgfqpoint{5.699148in}{0.801767in}}%
\pgfpathlineto{\pgfqpoint{5.825066in}{0.820132in}}%
\pgfpathlineto{\pgfqpoint{5.950984in}{0.859924in}}%
\pgfpathlineto{\pgfqpoint{6.076903in}{0.853802in}}%
\pgfpathlineto{\pgfqpoint{6.202821in}{0.859924in}}%
\pgfpathlineto{\pgfqpoint{6.328739in}{0.892063in}}%
\pgfpathlineto{\pgfqpoint{6.454657in}{0.882880in}}%
\pgfpathlineto{\pgfqpoint{6.580575in}{0.977768in}}%
\pgfpathlineto{\pgfqpoint{6.706494in}{0.985420in}}%
\pgfusepath{stroke}%
\end{pgfscope}%
\begin{pgfscope}%
\pgfpathrectangle{\pgfqpoint{0.536502in}{0.041670in}}{\pgfqpoint{6.421828in}{1.166660in}}%
\pgfusepath{clip}%
\pgfsetbuttcap%
\pgfsetroundjoin%
\pgfsetlinewidth{1.003750pt}%
\definecolor{currentstroke}{rgb}{0.000000,0.392157,0.000000}%
\pgfsetstrokecolor{currentstroke}%
\pgfsetdash{{3.700000pt}{1.600000pt}}{0.000000pt}%
\pgfpathmoveto{\pgfqpoint{0.536502in}{0.094700in}}%
\pgfpathlineto{\pgfqpoint{6.958330in}{0.094700in}}%
\pgfusepath{stroke}%
\end{pgfscope}%
\begin{pgfscope}%
\pgfpathrectangle{\pgfqpoint{0.536502in}{0.041670in}}{\pgfqpoint{6.421828in}{1.166660in}}%
\pgfusepath{clip}%
\pgfsetbuttcap%
\pgfsetroundjoin%
\pgfsetlinewidth{1.003750pt}%
\definecolor{currentstroke}{rgb}{0.803922,0.521569,0.247059}%
\pgfsetstrokecolor{currentstroke}%
\pgfsetdash{{3.700000pt}{1.600000pt}}{0.000000pt}%
\pgfpathmoveto{\pgfqpoint{0.536502in}{0.933428in}}%
\pgfpathlineto{\pgfqpoint{6.958330in}{0.933428in}}%
\pgfusepath{stroke}%
\end{pgfscope}%
\begin{pgfscope}%
\pgfpathrectangle{\pgfqpoint{0.536502in}{0.041670in}}{\pgfqpoint{6.421828in}{1.166660in}}%
\pgfusepath{clip}%
\pgfsetrectcap%
\pgfsetroundjoin%
\pgfsetlinewidth{6.022500pt}%
\definecolor{currentstroke}{rgb}{1.000000,1.000000,1.000000}%
\pgfsetstrokecolor{currentstroke}%
\pgfsetdash{}{0pt}%
\pgfpathmoveto{\pgfqpoint{3.684457in}{0.041670in}}%
\pgfpathlineto{\pgfqpoint{3.684457in}{1.208330in}}%
\pgfusepath{stroke}%
\end{pgfscope}%
\begin{pgfscope}%
\pgfsetrectcap%
\pgfsetmiterjoin%
\pgfsetlinewidth{1.003750pt}%
\definecolor{currentstroke}{rgb}{1.000000,1.000000,1.000000}%
\pgfsetstrokecolor{currentstroke}%
\pgfsetdash{}{0pt}%
\pgfpathmoveto{\pgfqpoint{0.536502in}{0.041670in}}%
\pgfpathlineto{\pgfqpoint{0.536502in}{1.208330in}}%
\pgfusepath{stroke}%
\end{pgfscope}%
\begin{pgfscope}%
\pgfsetrectcap%
\pgfsetmiterjoin%
\pgfsetlinewidth{1.003750pt}%
\definecolor{currentstroke}{rgb}{1.000000,1.000000,1.000000}%
\pgfsetstrokecolor{currentstroke}%
\pgfsetdash{}{0pt}%
\pgfpathmoveto{\pgfqpoint{6.958330in}{0.041670in}}%
\pgfpathlineto{\pgfqpoint{6.958330in}{1.208330in}}%
\pgfusepath{stroke}%
\end{pgfscope}%
\begin{pgfscope}%
\pgfsetrectcap%
\pgfsetmiterjoin%
\pgfsetlinewidth{1.003750pt}%
\definecolor{currentstroke}{rgb}{1.000000,1.000000,1.000000}%
\pgfsetstrokecolor{currentstroke}%
\pgfsetdash{}{0pt}%
\pgfpathmoveto{\pgfqpoint{0.536502in}{0.041670in}}%
\pgfpathlineto{\pgfqpoint{6.958330in}{0.041670in}}%
\pgfusepath{stroke}%
\end{pgfscope}%
\begin{pgfscope}%
\pgfsetrectcap%
\pgfsetmiterjoin%
\pgfsetlinewidth{1.003750pt}%
\definecolor{currentstroke}{rgb}{1.000000,1.000000,1.000000}%
\pgfsetstrokecolor{currentstroke}%
\pgfsetdash{}{0pt}%
\pgfpathmoveto{\pgfqpoint{0.536502in}{1.208330in}}%
\pgfpathlineto{\pgfqpoint{6.958330in}{1.208330in}}%
\pgfusepath{stroke}%
\end{pgfscope}%
\end{pgfpicture}%
\makeatother%
\endgroup%

%% file: figures/comparison-selectedfinger-nox.pgf
\begingroup%
\makeatletter%
\begin{pgfpicture}%
\pgfpathrectangle{\pgfpointorigin}{\pgfqpoint{7.000000in}{1.250000in}}%
\pgfusepath{use as bounding box, clip}%
\begin{pgfscope}%
\pgfsetbuttcap%
\pgfsetmiterjoin%
\definecolor{currentfill}{rgb}{1.000000,1.000000,1.000000}%
\pgfsetfillcolor{currentfill}%
\pgfsetlinewidth{0.000000pt}%
\definecolor{currentstroke}{rgb}{0.500000,0.500000,0.500000}%
\pgfsetstrokecolor{currentstroke}%
\pgfsetdash{}{0pt}%
\pgfpathmoveto{\pgfqpoint{0.000000in}{0.000000in}}%
\pgfpathlineto{\pgfqpoint{7.000000in}{0.000000in}}%
\pgfpathlineto{\pgfqpoint{7.000000in}{1.250000in}}%
\pgfpathlineto{\pgfqpoint{0.000000in}{1.250000in}}%
\pgfpathlineto{\pgfqpoint{0.000000in}{0.000000in}}%
\pgfpathclose%
\pgfusepath{fill}%
\end{pgfscope}%
\begin{pgfscope}%
\pgfsetbuttcap%
\pgfsetmiterjoin%
\definecolor{currentfill}{rgb}{0.898039,0.898039,0.898039}%
\pgfsetfillcolor{currentfill}%
\pgfsetlinewidth{0.000000pt}%
\definecolor{currentstroke}{rgb}{0.000000,0.000000,0.000000}%
\pgfsetstrokecolor{currentstroke}%
\pgfsetstrokeopacity{0.000000}%
\pgfsetdash{}{0pt}%
\pgfpathmoveto{\pgfqpoint{0.536502in}{0.041670in}}%
\pgfpathlineto{\pgfqpoint{6.958330in}{0.041670in}}%
\pgfpathlineto{\pgfqpoint{6.958330in}{1.208330in}}%
\pgfpathlineto{\pgfqpoint{0.536502in}{1.208330in}}%
\pgfpathlineto{\pgfqpoint{0.536502in}{0.041670in}}%
\pgfpathclose%
\pgfusepath{fill}%
\end{pgfscope}%
\begin{pgfscope}%
\pgfpathrectangle{\pgfqpoint{0.536502in}{0.041670in}}{\pgfqpoint{6.421828in}{1.166660in}}%
\pgfusepath{clip}%
\pgfsetrectcap%
\pgfsetroundjoin%
\pgfsetlinewidth{0.803000pt}%
\definecolor{currentstroke}{rgb}{1.000000,1.000000,1.000000}%
\pgfsetstrokecolor{currentstroke}%
\pgfsetdash{}{0pt}%
\pgfpathmoveto{\pgfqpoint{0.662421in}{0.041670in}}%
\pgfpathlineto{\pgfqpoint{0.662421in}{1.208330in}}%
\pgfusepath{stroke}%
\end{pgfscope}%
\begin{pgfscope}%
\pgfpathrectangle{\pgfqpoint{0.536502in}{0.041670in}}{\pgfqpoint{6.421828in}{1.166660in}}%
\pgfusepath{clip}%
\pgfsetrectcap%
\pgfsetroundjoin%
\pgfsetlinewidth{0.803000pt}%
\definecolor{currentstroke}{rgb}{1.000000,1.000000,1.000000}%
\pgfsetstrokecolor{currentstroke}%
\pgfsetdash{}{0pt}%
\pgfpathmoveto{\pgfqpoint{3.558539in}{0.041670in}}%
\pgfpathlineto{\pgfqpoint{3.558539in}{1.208330in}}%
\pgfusepath{stroke}%
\end{pgfscope}%
\begin{pgfscope}%
\pgfpathrectangle{\pgfqpoint{0.536502in}{0.041670in}}{\pgfqpoint{6.421828in}{1.166660in}}%
\pgfusepath{clip}%
\pgfsetrectcap%
\pgfsetroundjoin%
\pgfsetlinewidth{0.803000pt}%
\definecolor{currentstroke}{rgb}{1.000000,1.000000,1.000000}%
\pgfsetstrokecolor{currentstroke}%
\pgfsetdash{}{0pt}%
\pgfpathmoveto{\pgfqpoint{3.810375in}{0.041670in}}%
\pgfpathlineto{\pgfqpoint{3.810375in}{1.208330in}}%
\pgfusepath{stroke}%
\end{pgfscope}%
\begin{pgfscope}%
\pgfpathrectangle{\pgfqpoint{0.536502in}{0.041670in}}{\pgfqpoint{6.421828in}{1.166660in}}%
\pgfusepath{clip}%
\pgfsetrectcap%
\pgfsetroundjoin%
\pgfsetlinewidth{0.803000pt}%
\definecolor{currentstroke}{rgb}{1.000000,1.000000,1.000000}%
\pgfsetstrokecolor{currentstroke}%
\pgfsetdash{}{0pt}%
\pgfpathmoveto{\pgfqpoint{6.832412in}{0.041670in}}%
\pgfpathlineto{\pgfqpoint{6.832412in}{1.208330in}}%
\pgfusepath{stroke}%
\end{pgfscope}%
\begin{pgfscope}%
\pgfpathrectangle{\pgfqpoint{0.536502in}{0.041670in}}{\pgfqpoint{6.421828in}{1.166660in}}%
\pgfusepath{clip}%
\pgfsetrectcap%
\pgfsetroundjoin%
\pgfsetlinewidth{0.803000pt}%
\definecolor{currentstroke}{rgb}{1.000000,1.000000,1.000000}%
\pgfsetstrokecolor{currentstroke}%
\pgfsetdash{}{0pt}%
\pgfpathmoveto{\pgfqpoint{0.536502in}{0.148243in}}%
\pgfpathlineto{\pgfqpoint{6.958330in}{0.148243in}}%
\pgfusepath{stroke}%
\end{pgfscope}%
\begin{pgfscope}%
\definecolor{textcolor}{rgb}{0.333333,0.333333,0.333333}%
\pgfsetstrokecolor{textcolor}%
\pgfsetfillcolor{textcolor}%
\pgftext[x=0.041670in, y=0.095481in, left, base]{\color{textcolor}\sffamily\fontsize{10.000000}{12.000000}\selectfont 0.500}%
\end{pgfscope}%
\begin{pgfscope}%
\pgfpathrectangle{\pgfqpoint{0.536502in}{0.041670in}}{\pgfqpoint{6.421828in}{1.166660in}}%
\pgfusepath{clip}%
\pgfsetrectcap%
\pgfsetroundjoin%
\pgfsetlinewidth{0.803000pt}%
\definecolor{currentstroke}{rgb}{1.000000,1.000000,1.000000}%
\pgfsetstrokecolor{currentstroke}%
\pgfsetdash{}{0pt}%
\pgfpathmoveto{\pgfqpoint{0.536502in}{0.468857in}}%
\pgfpathlineto{\pgfqpoint{6.958330in}{0.468857in}}%
\pgfusepath{stroke}%
\end{pgfscope}%
\begin{pgfscope}%
\definecolor{textcolor}{rgb}{0.333333,0.333333,0.333333}%
\pgfsetstrokecolor{textcolor}%
\pgfsetfillcolor{textcolor}%
\pgftext[x=0.041670in, y=0.416095in, left, base]{\color{textcolor}\sffamily\fontsize{10.000000}{12.000000}\selectfont 0.600}%
\end{pgfscope}%
\begin{pgfscope}%
\pgfpathrectangle{\pgfqpoint{0.536502in}{0.041670in}}{\pgfqpoint{6.421828in}{1.166660in}}%
\pgfusepath{clip}%
\pgfsetrectcap%
\pgfsetroundjoin%
\pgfsetlinewidth{0.803000pt}%
\definecolor{currentstroke}{rgb}{1.000000,1.000000,1.000000}%
\pgfsetstrokecolor{currentstroke}%
\pgfsetdash{}{0pt}%
\pgfpathmoveto{\pgfqpoint{0.536502in}{0.789471in}}%
\pgfpathlineto{\pgfqpoint{6.958330in}{0.789471in}}%
\pgfusepath{stroke}%
\end{pgfscope}%
\begin{pgfscope}%
\definecolor{textcolor}{rgb}{0.333333,0.333333,0.333333}%
\pgfsetstrokecolor{textcolor}%
\pgfsetfillcolor{textcolor}%
\pgftext[x=0.041670in, y=0.736709in, left, base]{\color{textcolor}\sffamily\fontsize{10.000000}{12.000000}\selectfont 0.700}%
\end{pgfscope}%
\begin{pgfscope}%
\pgfpathrectangle{\pgfqpoint{0.536502in}{0.041670in}}{\pgfqpoint{6.421828in}{1.166660in}}%
\pgfusepath{clip}%
\pgfsetrectcap%
\pgfsetroundjoin%
\pgfsetlinewidth{0.803000pt}%
\definecolor{currentstroke}{rgb}{1.000000,1.000000,1.000000}%
\pgfsetstrokecolor{currentstroke}%
\pgfsetdash{}{0pt}%
\pgfpathmoveto{\pgfqpoint{0.536502in}{1.110085in}}%
\pgfpathlineto{\pgfqpoint{6.958330in}{1.110085in}}%
\pgfusepath{stroke}%
\end{pgfscope}%
\begin{pgfscope}%
\definecolor{textcolor}{rgb}{0.333333,0.333333,0.333333}%
\pgfsetstrokecolor{textcolor}%
\pgfsetfillcolor{textcolor}%
\pgftext[x=0.041670in, y=1.057324in, left, base]{\color{textcolor}\sffamily\fontsize{10.000000}{12.000000}\selectfont 0.800}%
\end{pgfscope}%
\begin{pgfscope}%
\pgfpathrectangle{\pgfqpoint{0.536502in}{0.041670in}}{\pgfqpoint{6.421828in}{1.166660in}}%
\pgfusepath{clip}%
\pgfsetrectcap%
\pgfsetroundjoin%
\pgfsetlinewidth{1.003750pt}%
\definecolor{currentstroke}{rgb}{0.886275,0.290196,0.200000}%
\pgfsetstrokecolor{currentstroke}%
\pgfsetdash{}{0pt}%
\pgfpathmoveto{\pgfqpoint{0.662421in}{0.094700in}}%
\pgfpathlineto{\pgfqpoint{0.788339in}{0.094700in}}%
\pgfpathlineto{\pgfqpoint{0.914257in}{0.094700in}}%
\pgfpathlineto{\pgfqpoint{1.040175in}{0.094700in}}%
\pgfpathlineto{\pgfqpoint{1.166093in}{0.094700in}}%
\pgfpathlineto{\pgfqpoint{1.292011in}{0.094700in}}%
\pgfpathlineto{\pgfqpoint{1.417930in}{0.094700in}}%
\pgfpathlineto{\pgfqpoint{1.543848in}{0.103028in}}%
\pgfpathlineto{\pgfqpoint{1.669766in}{0.094700in}}%
\pgfpathlineto{\pgfqpoint{1.795684in}{0.094700in}}%
\pgfpathlineto{\pgfqpoint{1.921602in}{0.094700in}}%
\pgfpathlineto{\pgfqpoint{2.047521in}{0.098917in}}%
\pgfusepath{stroke}%
\end{pgfscope}%
\begin{pgfscope}%
\pgfpathrectangle{\pgfqpoint{0.536502in}{0.041670in}}{\pgfqpoint{6.421828in}{1.166660in}}%
\pgfusepath{clip}%
\pgfsetrectcap%
\pgfsetroundjoin%
\pgfsetlinewidth{1.003750pt}%
\definecolor{currentstroke}{rgb}{0.886275,0.290196,0.200000}%
\pgfsetstrokecolor{currentstroke}%
\pgfsetdash{}{0pt}%
\pgfpathmoveto{\pgfqpoint{3.810375in}{0.094700in}}%
\pgfpathlineto{\pgfqpoint{3.936293in}{0.100973in}}%
\pgfpathlineto{\pgfqpoint{4.062212in}{0.094700in}}%
\pgfpathlineto{\pgfqpoint{4.188130in}{0.094700in}}%
\pgfpathlineto{\pgfqpoint{4.314048in}{0.094700in}}%
\pgfpathlineto{\pgfqpoint{4.439966in}{0.103028in}}%
\pgfpathlineto{\pgfqpoint{4.565884in}{0.096862in}}%
\pgfpathlineto{\pgfqpoint{4.691803in}{0.094807in}}%
\pgfpathlineto{\pgfqpoint{4.817721in}{0.094700in}}%
\pgfpathlineto{\pgfqpoint{4.943639in}{0.098917in}}%
\pgfpathlineto{\pgfqpoint{5.069557in}{0.103028in}}%
\pgfpathlineto{\pgfqpoint{5.195475in}{0.094807in}}%
\pgfusepath{stroke}%
\end{pgfscope}%
\begin{pgfscope}%
\pgfpathrectangle{\pgfqpoint{0.536502in}{0.041670in}}{\pgfqpoint{6.421828in}{1.166660in}}%
\pgfusepath{clip}%
\pgfsetrectcap%
\pgfsetroundjoin%
\pgfsetlinewidth{1.003750pt}%
\definecolor{currentstroke}{rgb}{0.203922,0.541176,0.741176}%
\pgfsetstrokecolor{currentstroke}%
\pgfsetdash{}{0pt}%
\pgfpathmoveto{\pgfqpoint{0.662421in}{0.094700in}}%
\pgfpathlineto{\pgfqpoint{0.788339in}{0.094700in}}%
\pgfpathlineto{\pgfqpoint{0.914257in}{0.094700in}}%
\pgfpathlineto{\pgfqpoint{1.040175in}{0.094700in}}%
\pgfpathlineto{\pgfqpoint{1.166093in}{0.094700in}}%
\pgfpathlineto{\pgfqpoint{1.292011in}{0.094700in}}%
\pgfpathlineto{\pgfqpoint{1.417930in}{0.094700in}}%
\pgfpathlineto{\pgfqpoint{1.543848in}{0.094700in}}%
\pgfpathlineto{\pgfqpoint{1.669766in}{0.094700in}}%
\pgfpathlineto{\pgfqpoint{1.795684in}{0.113304in}}%
\pgfpathlineto{\pgfqpoint{1.921602in}{0.131801in}}%
\pgfpathlineto{\pgfqpoint{2.047521in}{0.119470in}}%
\pgfusepath{stroke}%
\end{pgfscope}%
\begin{pgfscope}%
\pgfpathrectangle{\pgfqpoint{0.536502in}{0.041670in}}{\pgfqpoint{6.421828in}{1.166660in}}%
\pgfusepath{clip}%
\pgfsetrectcap%
\pgfsetroundjoin%
\pgfsetlinewidth{1.003750pt}%
\definecolor{currentstroke}{rgb}{0.203922,0.541176,0.741176}%
\pgfsetstrokecolor{currentstroke}%
\pgfsetdash{}{0pt}%
\pgfpathmoveto{\pgfqpoint{3.810375in}{0.094700in}}%
\pgfpathlineto{\pgfqpoint{3.936293in}{0.094700in}}%
\pgfpathlineto{\pgfqpoint{4.062212in}{0.094700in}}%
\pgfpathlineto{\pgfqpoint{4.188130in}{0.094700in}}%
\pgfpathlineto{\pgfqpoint{4.314048in}{0.094700in}}%
\pgfpathlineto{\pgfqpoint{4.439966in}{0.094700in}}%
\pgfpathlineto{\pgfqpoint{4.565884in}{0.094700in}}%
\pgfpathlineto{\pgfqpoint{4.691803in}{0.094700in}}%
\pgfpathlineto{\pgfqpoint{4.817721in}{0.094700in}}%
\pgfpathlineto{\pgfqpoint{4.943639in}{0.094807in}}%
\pgfpathlineto{\pgfqpoint{5.069557in}{0.094700in}}%
\pgfpathlineto{\pgfqpoint{5.195475in}{0.094700in}}%
\pgfusepath{stroke}%
\end{pgfscope}%
\begin{pgfscope}%
\pgfpathrectangle{\pgfqpoint{0.536502in}{0.041670in}}{\pgfqpoint{6.421828in}{1.166660in}}%
\pgfusepath{clip}%
\pgfsetrectcap%
\pgfsetroundjoin%
\pgfsetlinewidth{1.003750pt}%
\definecolor{currentstroke}{rgb}{0.596078,0.556863,0.835294}%
\pgfsetstrokecolor{currentstroke}%
\pgfsetdash{}{0pt}%
\pgfpathmoveto{\pgfqpoint{0.662421in}{0.094807in}}%
\pgfpathlineto{\pgfqpoint{0.788339in}{0.105083in}}%
\pgfpathlineto{\pgfqpoint{0.914257in}{0.096862in}}%
\pgfpathlineto{\pgfqpoint{1.040175in}{0.094700in}}%
\pgfpathlineto{\pgfqpoint{1.166093in}{0.119470in}}%
\pgfpathlineto{\pgfqpoint{1.292011in}{0.094700in}}%
\pgfpathlineto{\pgfqpoint{1.417930in}{0.094700in}}%
\pgfpathlineto{\pgfqpoint{1.543848in}{0.117414in}}%
\pgfpathlineto{\pgfqpoint{1.669766in}{0.119470in}}%
\pgfpathlineto{\pgfqpoint{1.795684in}{0.121525in}}%
\pgfpathlineto{\pgfqpoint{1.921602in}{0.103028in}}%
\pgfpathlineto{\pgfqpoint{2.047521in}{0.107138in}}%
\pgfpathlineto{\pgfqpoint{2.173439in}{0.103028in}}%
\pgfpathlineto{\pgfqpoint{2.299357in}{0.115359in}}%
\pgfpathlineto{\pgfqpoint{2.425275in}{0.094700in}}%
\pgfpathlineto{\pgfqpoint{2.551193in}{0.094700in}}%
\pgfusepath{stroke}%
\end{pgfscope}%
\begin{pgfscope}%
\pgfpathrectangle{\pgfqpoint{0.536502in}{0.041670in}}{\pgfqpoint{6.421828in}{1.166660in}}%
\pgfusepath{clip}%
\pgfsetrectcap%
\pgfsetroundjoin%
\pgfsetlinewidth{1.003750pt}%
\definecolor{currentstroke}{rgb}{0.596078,0.556863,0.835294}%
\pgfsetstrokecolor{currentstroke}%
\pgfsetdash{}{0pt}%
\pgfpathmoveto{\pgfqpoint{3.810375in}{0.094807in}}%
\pgfpathlineto{\pgfqpoint{3.936293in}{0.103028in}}%
\pgfpathlineto{\pgfqpoint{4.062212in}{0.096862in}}%
\pgfpathlineto{\pgfqpoint{4.188130in}{0.094807in}}%
\pgfpathlineto{\pgfqpoint{4.314048in}{0.094700in}}%
\pgfpathlineto{\pgfqpoint{4.439966in}{0.103028in}}%
\pgfpathlineto{\pgfqpoint{4.565884in}{0.140022in}}%
\pgfpathlineto{\pgfqpoint{4.691803in}{0.146187in}}%
\pgfpathlineto{\pgfqpoint{4.817721in}{0.148243in}}%
\pgfpathlineto{\pgfqpoint{4.943639in}{0.224286in}}%
\pgfpathlineto{\pgfqpoint{5.069557in}{0.306494in}}%
\pgfpathlineto{\pgfqpoint{5.195475in}{0.507906in}}%
\pgfpathlineto{\pgfqpoint{5.321394in}{0.694931in}}%
\pgfpathlineto{\pgfqpoint{5.447312in}{0.836741in}}%
\pgfpathlineto{\pgfqpoint{5.573230in}{1.155300in}}%
\pgfpathlineto{\pgfqpoint{5.699148in}{0.939502in}}%
\pgfusepath{stroke}%
\end{pgfscope}%
\begin{pgfscope}%
\pgfpathrectangle{\pgfqpoint{0.536502in}{0.041670in}}{\pgfqpoint{6.421828in}{1.166660in}}%
\pgfusepath{clip}%
\pgfsetrectcap%
\pgfsetroundjoin%
\pgfsetlinewidth{1.003750pt}%
\definecolor{currentstroke}{rgb}{0.466667,0.466667,0.466667}%
\pgfsetstrokecolor{currentstroke}%
\pgfsetdash{}{0pt}%
\pgfpathmoveto{\pgfqpoint{0.662421in}{0.094807in}}%
\pgfpathlineto{\pgfqpoint{0.788339in}{0.094700in}}%
\pgfpathlineto{\pgfqpoint{0.914257in}{0.096862in}}%
\pgfpathlineto{\pgfqpoint{1.040175in}{0.109193in}}%
\pgfpathlineto{\pgfqpoint{1.166093in}{0.094807in}}%
\pgfpathlineto{\pgfqpoint{1.292011in}{0.115359in}}%
\pgfpathlineto{\pgfqpoint{1.417930in}{0.119470in}}%
\pgfpathlineto{\pgfqpoint{1.543848in}{0.094700in}}%
\pgfpathlineto{\pgfqpoint{1.669766in}{0.094700in}}%
\pgfpathlineto{\pgfqpoint{1.795684in}{0.094700in}}%
\pgfpathlineto{\pgfqpoint{1.921602in}{0.094700in}}%
\pgfpathlineto{\pgfqpoint{2.047521in}{0.094700in}}%
\pgfpathlineto{\pgfqpoint{2.173439in}{0.094700in}}%
\pgfpathlineto{\pgfqpoint{2.299357in}{0.094700in}}%
\pgfpathlineto{\pgfqpoint{2.425275in}{0.094700in}}%
\pgfpathlineto{\pgfqpoint{2.551193in}{0.094700in}}%
\pgfpathlineto{\pgfqpoint{2.677112in}{0.094700in}}%
\pgfpathlineto{\pgfqpoint{2.803030in}{0.094700in}}%
\pgfpathlineto{\pgfqpoint{2.928948in}{0.094700in}}%
\pgfpathlineto{\pgfqpoint{3.054866in}{0.094700in}}%
\pgfpathlineto{\pgfqpoint{3.180784in}{0.094700in}}%
\pgfpathlineto{\pgfqpoint{3.306703in}{0.094700in}}%
\pgfpathlineto{\pgfqpoint{3.432621in}{0.094700in}}%
\pgfpathlineto{\pgfqpoint{3.558539in}{0.094700in}}%
\pgfusepath{stroke}%
\end{pgfscope}%
\begin{pgfscope}%
\pgfpathrectangle{\pgfqpoint{0.536502in}{0.041670in}}{\pgfqpoint{6.421828in}{1.166660in}}%
\pgfusepath{clip}%
\pgfsetrectcap%
\pgfsetroundjoin%
\pgfsetlinewidth{1.003750pt}%
\definecolor{currentstroke}{rgb}{0.466667,0.466667,0.466667}%
\pgfsetstrokecolor{currentstroke}%
\pgfsetdash{}{0pt}%
\pgfpathmoveto{\pgfqpoint{3.810375in}{0.098917in}}%
\pgfpathlineto{\pgfqpoint{3.936293in}{0.096862in}}%
\pgfpathlineto{\pgfqpoint{4.062212in}{0.105083in}}%
\pgfpathlineto{\pgfqpoint{4.188130in}{0.117414in}}%
\pgfpathlineto{\pgfqpoint{4.314048in}{0.094700in}}%
\pgfpathlineto{\pgfqpoint{4.439966in}{0.094700in}}%
\pgfpathlineto{\pgfqpoint{4.565884in}{0.094700in}}%
\pgfpathlineto{\pgfqpoint{4.691803in}{0.111249in}}%
\pgfpathlineto{\pgfqpoint{4.817721in}{0.199623in}}%
\pgfpathlineto{\pgfqpoint{4.943639in}{0.236617in}}%
\pgfpathlineto{\pgfqpoint{5.069557in}{0.283887in}}%
\pgfpathlineto{\pgfqpoint{5.195475in}{0.265390in}}%
\pgfpathlineto{\pgfqpoint{5.321394in}{0.322936in}}%
\pgfpathlineto{\pgfqpoint{5.447312in}{0.396924in}}%
\pgfpathlineto{\pgfqpoint{5.573230in}{0.495575in}}%
\pgfpathlineto{\pgfqpoint{5.699148in}{0.555176in}}%
\pgfpathlineto{\pgfqpoint{5.825066in}{0.592170in}}%
\pgfpathlineto{\pgfqpoint{5.950984in}{0.631219in}}%
\pgfpathlineto{\pgfqpoint{6.076903in}{0.670268in}}%
\pgfpathlineto{\pgfqpoint{6.202821in}{0.659992in}}%
\pgfpathlineto{\pgfqpoint{6.328739in}{0.692876in}}%
\pgfpathlineto{\pgfqpoint{6.454657in}{0.723704in}}%
\pgfpathlineto{\pgfqpoint{6.580575in}{0.781250in}}%
\pgfpathlineto{\pgfqpoint{6.706494in}{0.824410in}}%
\pgfusepath{stroke}%
\end{pgfscope}%
\begin{pgfscope}%
\pgfpathrectangle{\pgfqpoint{0.536502in}{0.041670in}}{\pgfqpoint{6.421828in}{1.166660in}}%
\pgfusepath{clip}%
\pgfsetbuttcap%
\pgfsetroundjoin%
\pgfsetlinewidth{1.003750pt}%
\definecolor{currentstroke}{rgb}{0.000000,0.392157,0.000000}%
\pgfsetstrokecolor{currentstroke}%
\pgfsetdash{{3.700000pt}{1.600000pt}}{0.000000pt}%
\pgfpathmoveto{\pgfqpoint{0.536502in}{0.094700in}}%
\pgfpathlineto{\pgfqpoint{6.958330in}{0.094700in}}%
\pgfusepath{stroke}%
\end{pgfscope}%
\begin{pgfscope}%
\pgfpathrectangle{\pgfqpoint{0.536502in}{0.041670in}}{\pgfqpoint{6.421828in}{1.166660in}}%
\pgfusepath{clip}%
\pgfsetbuttcap%
\pgfsetroundjoin%
\pgfsetlinewidth{1.003750pt}%
\definecolor{currentstroke}{rgb}{0.803922,0.521569,0.247059}%
\pgfsetstrokecolor{currentstroke}%
\pgfsetdash{{3.700000pt}{1.600000pt}}{0.000000pt}%
\pgfpathmoveto{\pgfqpoint{0.536502in}{0.881808in}}%
\pgfpathlineto{\pgfqpoint{6.958330in}{0.881808in}}%
\pgfusepath{stroke}%
\end{pgfscope}%
\begin{pgfscope}%
\pgfpathrectangle{\pgfqpoint{0.536502in}{0.041670in}}{\pgfqpoint{6.421828in}{1.166660in}}%
\pgfusepath{clip}%
\pgfsetrectcap%
\pgfsetroundjoin%
\pgfsetlinewidth{6.022500pt}%
\definecolor{currentstroke}{rgb}{1.000000,1.000000,1.000000}%
\pgfsetstrokecolor{currentstroke}%
\pgfsetdash{}{0pt}%
\pgfpathmoveto{\pgfqpoint{3.684457in}{0.041670in}}%
\pgfpathlineto{\pgfqpoint{3.684457in}{1.208330in}}%
\pgfusepath{stroke}%
\end{pgfscope}%
\begin{pgfscope}%
\pgfsetrectcap%
\pgfsetmiterjoin%
\pgfsetlinewidth{1.003750pt}%
\definecolor{currentstroke}{rgb}{1.000000,1.000000,1.000000}%
\pgfsetstrokecolor{currentstroke}%
\pgfsetdash{}{0pt}%
\pgfpathmoveto{\pgfqpoint{0.536502in}{0.041670in}}%
\pgfpathlineto{\pgfqpoint{0.536502in}{1.208330in}}%
\pgfusepath{stroke}%
\end{pgfscope}%
\begin{pgfscope}%
\pgfsetrectcap%
\pgfsetmiterjoin%
\pgfsetlinewidth{1.003750pt}%
\definecolor{currentstroke}{rgb}{1.000000,1.000000,1.000000}%
\pgfsetstrokecolor{currentstroke}%
\pgfsetdash{}{0pt}%
\pgfpathmoveto{\pgfqpoint{6.958330in}{0.041670in}}%
\pgfpathlineto{\pgfqpoint{6.958330in}{1.208330in}}%
\pgfusepath{stroke}%
\end{pgfscope}%
\begin{pgfscope}%
\pgfsetrectcap%
\pgfsetmiterjoin%
\pgfsetlinewidth{1.003750pt}%
\definecolor{currentstroke}{rgb}{1.000000,1.000000,1.000000}%
\pgfsetstrokecolor{currentstroke}%
\pgfsetdash{}{0pt}%
\pgfpathmoveto{\pgfqpoint{0.536502in}{0.041670in}}%
\pgfpathlineto{\pgfqpoint{6.958330in}{0.041670in}}%
\pgfusepath{stroke}%
\end{pgfscope}%
\begin{pgfscope}%
\pgfsetrectcap%
\pgfsetmiterjoin%
\pgfsetlinewidth{1.003750pt}%
\definecolor{currentstroke}{rgb}{1.000000,1.000000,1.000000}%
\pgfsetstrokecolor{currentstroke}%
\pgfsetdash{}{0pt}%
\pgfpathmoveto{\pgfqpoint{0.536502in}{1.208330in}}%
\pgfpathlineto{\pgfqpoint{6.958330in}{1.208330in}}%
\pgfusepath{stroke}%
\end{pgfscope}%
\end{pgfpicture}%
\makeatother%
\endgroup%

%% file: figures/comparison-flexion-nox.pgf
\begingroup%
\makeatletter%
\begin{pgfpicture}%
\pgfpathrectangle{\pgfpointorigin}{\pgfqpoint{7.000000in}{1.250000in}}%
\pgfusepath{use as bounding box, clip}%
\begin{pgfscope}%
\pgfsetbuttcap%
\pgfsetmiterjoin%
\definecolor{currentfill}{rgb}{1.000000,1.000000,1.000000}%
\pgfsetfillcolor{currentfill}%
\pgfsetlinewidth{0.000000pt}%
\definecolor{currentstroke}{rgb}{0.500000,0.500000,0.500000}%
\pgfsetstrokecolor{currentstroke}%
\pgfsetdash{}{0pt}%
\pgfpathmoveto{\pgfqpoint{0.000000in}{0.000000in}}%
\pgfpathlineto{\pgfqpoint{7.000000in}{0.000000in}}%
\pgfpathlineto{\pgfqpoint{7.000000in}{1.250000in}}%
\pgfpathlineto{\pgfqpoint{0.000000in}{1.250000in}}%
\pgfpathlineto{\pgfqpoint{0.000000in}{0.000000in}}%
\pgfpathclose%
\pgfusepath{fill}%
\end{pgfscope}%
\begin{pgfscope}%
\pgfsetbuttcap%
\pgfsetmiterjoin%
\definecolor{currentfill}{rgb}{0.898039,0.898039,0.898039}%
\pgfsetfillcolor{currentfill}%
\pgfsetlinewidth{0.000000pt}%
\definecolor{currentstroke}{rgb}{0.000000,0.000000,0.000000}%
\pgfsetstrokecolor{currentstroke}%
\pgfsetstrokeopacity{0.000000}%
\pgfsetdash{}{0pt}%
\pgfpathmoveto{\pgfqpoint{0.536502in}{0.041670in}}%
\pgfpathlineto{\pgfqpoint{6.958330in}{0.041670in}}%
\pgfpathlineto{\pgfqpoint{6.958330in}{1.162945in}}%
\pgfpathlineto{\pgfqpoint{0.536502in}{1.162945in}}%
\pgfpathlineto{\pgfqpoint{0.536502in}{0.041670in}}%
\pgfpathclose%
\pgfusepath{fill}%
\end{pgfscope}%
\begin{pgfscope}%
\pgfpathrectangle{\pgfqpoint{0.536502in}{0.041670in}}{\pgfqpoint{6.421828in}{1.121275in}}%
\pgfusepath{clip}%
\pgfsetrectcap%
\pgfsetroundjoin%
\pgfsetlinewidth{0.803000pt}%
\definecolor{currentstroke}{rgb}{1.000000,1.000000,1.000000}%
\pgfsetstrokecolor{currentstroke}%
\pgfsetdash{}{0pt}%
\pgfpathmoveto{\pgfqpoint{0.662421in}{0.041670in}}%
\pgfpathlineto{\pgfqpoint{0.662421in}{1.162945in}}%
\pgfusepath{stroke}%
\end{pgfscope}%
\begin{pgfscope}%
\pgfpathrectangle{\pgfqpoint{0.536502in}{0.041670in}}{\pgfqpoint{6.421828in}{1.121275in}}%
\pgfusepath{clip}%
\pgfsetrectcap%
\pgfsetroundjoin%
\pgfsetlinewidth{0.803000pt}%
\definecolor{currentstroke}{rgb}{1.000000,1.000000,1.000000}%
\pgfsetstrokecolor{currentstroke}%
\pgfsetdash{}{0pt}%
\pgfpathmoveto{\pgfqpoint{3.558539in}{0.041670in}}%
\pgfpathlineto{\pgfqpoint{3.558539in}{1.162945in}}%
\pgfusepath{stroke}%
\end{pgfscope}%
\begin{pgfscope}%
\pgfpathrectangle{\pgfqpoint{0.536502in}{0.041670in}}{\pgfqpoint{6.421828in}{1.121275in}}%
\pgfusepath{clip}%
\pgfsetrectcap%
\pgfsetroundjoin%
\pgfsetlinewidth{0.803000pt}%
\definecolor{currentstroke}{rgb}{1.000000,1.000000,1.000000}%
\pgfsetstrokecolor{currentstroke}%
\pgfsetdash{}{0pt}%
\pgfpathmoveto{\pgfqpoint{3.810375in}{0.041670in}}%
\pgfpathlineto{\pgfqpoint{3.810375in}{1.162945in}}%
\pgfusepath{stroke}%
\end{pgfscope}%
\begin{pgfscope}%
\pgfpathrectangle{\pgfqpoint{0.536502in}{0.041670in}}{\pgfqpoint{6.421828in}{1.121275in}}%
\pgfusepath{clip}%
\pgfsetrectcap%
\pgfsetroundjoin%
\pgfsetlinewidth{0.803000pt}%
\definecolor{currentstroke}{rgb}{1.000000,1.000000,1.000000}%
\pgfsetstrokecolor{currentstroke}%
\pgfsetdash{}{0pt}%
\pgfpathmoveto{\pgfqpoint{6.832412in}{0.041670in}}%
\pgfpathlineto{\pgfqpoint{6.832412in}{1.162945in}}%
\pgfusepath{stroke}%
\end{pgfscope}%
\begin{pgfscope}%
\pgfpathrectangle{\pgfqpoint{0.536502in}{0.041670in}}{\pgfqpoint{6.421828in}{1.121275in}}%
\pgfusepath{clip}%
\pgfsetrectcap%
\pgfsetroundjoin%
\pgfsetlinewidth{0.803000pt}%
\definecolor{currentstroke}{rgb}{1.000000,1.000000,1.000000}%
\pgfsetstrokecolor{currentstroke}%
\pgfsetdash{}{0pt}%
\pgfpathmoveto{\pgfqpoint{0.536502in}{0.400107in}}%
\pgfpathlineto{\pgfqpoint{6.958330in}{0.400107in}}%
\pgfusepath{stroke}%
\end{pgfscope}%
\begin{pgfscope}%
\definecolor{textcolor}{rgb}{0.333333,0.333333,0.333333}%
\pgfsetstrokecolor{textcolor}%
\pgfsetfillcolor{textcolor}%
\pgftext[x=0.041670in, y=0.347346in, left, base]{\color{textcolor}\sffamily\fontsize{10.000000}{12.000000}\selectfont 0.500}%
\end{pgfscope}%
\begin{pgfscope}%
\pgfpathrectangle{\pgfqpoint{0.536502in}{0.041670in}}{\pgfqpoint{6.421828in}{1.121275in}}%
\pgfusepath{clip}%
\pgfsetrectcap%
\pgfsetroundjoin%
\pgfsetlinewidth{0.803000pt}%
\definecolor{currentstroke}{rgb}{1.000000,1.000000,1.000000}%
\pgfsetstrokecolor{currentstroke}%
\pgfsetdash{}{0pt}%
\pgfpathmoveto{\pgfqpoint{0.536502in}{0.777834in}}%
\pgfpathlineto{\pgfqpoint{6.958330in}{0.777834in}}%
\pgfusepath{stroke}%
\end{pgfscope}%
\begin{pgfscope}%
\definecolor{textcolor}{rgb}{0.333333,0.333333,0.333333}%
\pgfsetstrokecolor{textcolor}%
\pgfsetfillcolor{textcolor}%
\pgftext[x=0.041670in, y=0.725073in, left, base]{\color{textcolor}\sffamily\fontsize{10.000000}{12.000000}\selectfont 0.600}%
\end{pgfscope}%
\begin{pgfscope}%
\pgfpathrectangle{\pgfqpoint{0.536502in}{0.041670in}}{\pgfqpoint{6.421828in}{1.121275in}}%
\pgfusepath{clip}%
\pgfsetrectcap%
\pgfsetroundjoin%
\pgfsetlinewidth{0.803000pt}%
\definecolor{currentstroke}{rgb}{1.000000,1.000000,1.000000}%
\pgfsetstrokecolor{currentstroke}%
\pgfsetdash{}{0pt}%
\pgfpathmoveto{\pgfqpoint{0.536502in}{1.155562in}}%
\pgfpathlineto{\pgfqpoint{6.958330in}{1.155562in}}%
\pgfusepath{stroke}%
\end{pgfscope}%
\begin{pgfscope}%
\definecolor{textcolor}{rgb}{0.333333,0.333333,0.333333}%
\pgfsetstrokecolor{textcolor}%
\pgfsetfillcolor{textcolor}%
\pgftext[x=0.041670in, y=1.102800in, left, base]{\color{textcolor}\sffamily\fontsize{10.000000}{12.000000}\selectfont 0.700}%
\end{pgfscope}%
\begin{pgfscope}%
\pgfpathrectangle{\pgfqpoint{0.536502in}{0.041670in}}{\pgfqpoint{6.421828in}{1.121275in}}%
\pgfusepath{clip}%
\pgfsetrectcap%
\pgfsetroundjoin%
\pgfsetlinewidth{1.003750pt}%
\definecolor{currentstroke}{rgb}{0.886275,0.290196,0.200000}%
\pgfsetstrokecolor{currentstroke}%
\pgfsetdash{}{0pt}%
\pgfpathmoveto{\pgfqpoint{0.662421in}{0.092637in}}%
\pgfpathlineto{\pgfqpoint{0.788339in}{0.092637in}}%
\pgfpathlineto{\pgfqpoint{0.914257in}{0.092637in}}%
\pgfpathlineto{\pgfqpoint{1.040175in}{0.092637in}}%
\pgfpathlineto{\pgfqpoint{1.166093in}{0.092637in}}%
\pgfpathlineto{\pgfqpoint{1.292011in}{0.092637in}}%
\pgfpathlineto{\pgfqpoint{1.417930in}{0.092637in}}%
\pgfpathlineto{\pgfqpoint{1.543848in}{0.092637in}}%
\pgfpathlineto{\pgfqpoint{1.669766in}{0.095020in}}%
\pgfpathlineto{\pgfqpoint{1.795684in}{0.095020in}}%
\pgfpathlineto{\pgfqpoint{1.921602in}{0.095020in}}%
\pgfpathlineto{\pgfqpoint{2.047521in}{0.097441in}}%
\pgfusepath{stroke}%
\end{pgfscope}%
\begin{pgfscope}%
\pgfpathrectangle{\pgfqpoint{0.536502in}{0.041670in}}{\pgfqpoint{6.421828in}{1.121275in}}%
\pgfusepath{clip}%
\pgfsetrectcap%
\pgfsetroundjoin%
\pgfsetlinewidth{1.003750pt}%
\definecolor{currentstroke}{rgb}{0.886275,0.290196,0.200000}%
\pgfsetstrokecolor{currentstroke}%
\pgfsetdash{}{0pt}%
\pgfpathmoveto{\pgfqpoint{3.810375in}{0.092637in}}%
\pgfpathlineto{\pgfqpoint{3.936293in}{0.092637in}}%
\pgfpathlineto{\pgfqpoint{4.062212in}{0.092637in}}%
\pgfpathlineto{\pgfqpoint{4.188130in}{0.092637in}}%
\pgfpathlineto{\pgfqpoint{4.314048in}{0.092637in}}%
\pgfpathlineto{\pgfqpoint{4.439966in}{0.092637in}}%
\pgfpathlineto{\pgfqpoint{4.565884in}{0.092637in}}%
\pgfpathlineto{\pgfqpoint{4.691803in}{0.092637in}}%
\pgfpathlineto{\pgfqpoint{4.817721in}{0.092637in}}%
\pgfpathlineto{\pgfqpoint{4.943639in}{0.092637in}}%
\pgfpathlineto{\pgfqpoint{5.069557in}{0.092637in}}%
\pgfpathlineto{\pgfqpoint{5.195475in}{0.092637in}}%
\pgfusepath{stroke}%
\end{pgfscope}%
\begin{pgfscope}%
\pgfpathrectangle{\pgfqpoint{0.536502in}{0.041670in}}{\pgfqpoint{6.421828in}{1.121275in}}%
\pgfusepath{clip}%
\pgfsetrectcap%
\pgfsetroundjoin%
\pgfsetlinewidth{1.003750pt}%
\definecolor{currentstroke}{rgb}{0.203922,0.541176,0.741176}%
\pgfsetstrokecolor{currentstroke}%
\pgfsetdash{}{0pt}%
\pgfpathmoveto{\pgfqpoint{0.662421in}{0.092637in}}%
\pgfpathlineto{\pgfqpoint{0.788339in}{0.092637in}}%
\pgfpathlineto{\pgfqpoint{0.914257in}{0.092637in}}%
\pgfpathlineto{\pgfqpoint{1.040175in}{0.092637in}}%
\pgfpathlineto{\pgfqpoint{1.166093in}{0.092637in}}%
\pgfpathlineto{\pgfqpoint{1.292011in}{0.092637in}}%
\pgfpathlineto{\pgfqpoint{1.417930in}{0.092637in}}%
\pgfpathlineto{\pgfqpoint{1.543848in}{0.092637in}}%
\pgfpathlineto{\pgfqpoint{1.669766in}{0.092637in}}%
\pgfpathlineto{\pgfqpoint{1.795684in}{0.092637in}}%
\pgfpathlineto{\pgfqpoint{1.921602in}{0.092637in}}%
\pgfpathlineto{\pgfqpoint{2.047521in}{0.092637in}}%
\pgfusepath{stroke}%
\end{pgfscope}%
\begin{pgfscope}%
\pgfpathrectangle{\pgfqpoint{0.536502in}{0.041670in}}{\pgfqpoint{6.421828in}{1.121275in}}%
\pgfusepath{clip}%
\pgfsetrectcap%
\pgfsetroundjoin%
\pgfsetlinewidth{1.003750pt}%
\definecolor{currentstroke}{rgb}{0.203922,0.541176,0.741176}%
\pgfsetstrokecolor{currentstroke}%
\pgfsetdash{}{0pt}%
\pgfpathmoveto{\pgfqpoint{3.810375in}{0.092637in}}%
\pgfpathlineto{\pgfqpoint{3.936293in}{0.092637in}}%
\pgfpathlineto{\pgfqpoint{4.062212in}{0.092637in}}%
\pgfpathlineto{\pgfqpoint{4.188130in}{0.092637in}}%
\pgfpathlineto{\pgfqpoint{4.314048in}{0.092637in}}%
\pgfpathlineto{\pgfqpoint{4.439966in}{0.092637in}}%
\pgfpathlineto{\pgfqpoint{4.565884in}{0.092637in}}%
\pgfpathlineto{\pgfqpoint{4.691803in}{0.092637in}}%
\pgfpathlineto{\pgfqpoint{4.817721in}{0.092637in}}%
\pgfpathlineto{\pgfqpoint{4.943639in}{0.092637in}}%
\pgfpathlineto{\pgfqpoint{5.069557in}{0.092637in}}%
\pgfpathlineto{\pgfqpoint{5.195475in}{0.092637in}}%
\pgfusepath{stroke}%
\end{pgfscope}%
\begin{pgfscope}%
\pgfpathrectangle{\pgfqpoint{0.536502in}{0.041670in}}{\pgfqpoint{6.421828in}{1.121275in}}%
\pgfusepath{clip}%
\pgfsetrectcap%
\pgfsetroundjoin%
\pgfsetlinewidth{1.003750pt}%
\definecolor{currentstroke}{rgb}{0.596078,0.556863,0.835294}%
\pgfsetstrokecolor{currentstroke}%
\pgfsetdash{}{0pt}%
\pgfpathmoveto{\pgfqpoint{0.662421in}{0.092637in}}%
\pgfpathlineto{\pgfqpoint{0.788339in}{0.092637in}}%
\pgfpathlineto{\pgfqpoint{0.914257in}{0.092637in}}%
\pgfpathlineto{\pgfqpoint{1.040175in}{0.092637in}}%
\pgfpathlineto{\pgfqpoint{1.166093in}{0.095020in}}%
\pgfpathlineto{\pgfqpoint{1.292011in}{0.095020in}}%
\pgfpathlineto{\pgfqpoint{1.417930in}{0.104705in}}%
\pgfpathlineto{\pgfqpoint{1.543848in}{0.092637in}}%
\pgfpathlineto{\pgfqpoint{1.669766in}{0.092637in}}%
\pgfpathlineto{\pgfqpoint{1.795684in}{0.097441in}}%
\pgfpathlineto{\pgfqpoint{1.921602in}{0.095020in}}%
\pgfpathlineto{\pgfqpoint{2.047521in}{0.092637in}}%
\pgfpathlineto{\pgfqpoint{2.173439in}{0.092637in}}%
\pgfpathlineto{\pgfqpoint{2.299357in}{0.092637in}}%
\pgfpathlineto{\pgfqpoint{2.425275in}{0.092637in}}%
\pgfpathlineto{\pgfqpoint{2.551193in}{0.092637in}}%
\pgfusepath{stroke}%
\end{pgfscope}%
\begin{pgfscope}%
\pgfpathrectangle{\pgfqpoint{0.536502in}{0.041670in}}{\pgfqpoint{6.421828in}{1.121275in}}%
\pgfusepath{clip}%
\pgfsetrectcap%
\pgfsetroundjoin%
\pgfsetlinewidth{1.003750pt}%
\definecolor{currentstroke}{rgb}{0.596078,0.556863,0.835294}%
\pgfsetstrokecolor{currentstroke}%
\pgfsetdash{}{0pt}%
\pgfpathmoveto{\pgfqpoint{3.810375in}{0.092637in}}%
\pgfpathlineto{\pgfqpoint{3.936293in}{0.092637in}}%
\pgfpathlineto{\pgfqpoint{4.062212in}{0.092637in}}%
\pgfpathlineto{\pgfqpoint{4.188130in}{0.092637in}}%
\pgfpathlineto{\pgfqpoint{4.314048in}{0.099862in}}%
\pgfpathlineto{\pgfqpoint{4.439966in}{0.104705in}}%
\pgfpathlineto{\pgfqpoint{4.565884in}{0.092637in}}%
\pgfpathlineto{\pgfqpoint{4.691803in}{0.092637in}}%
\pgfpathlineto{\pgfqpoint{4.817721in}{0.107126in}}%
\pgfpathlineto{\pgfqpoint{4.943639in}{0.124076in}}%
\pgfpathlineto{\pgfqpoint{5.069557in}{0.189451in}}%
\pgfpathlineto{\pgfqpoint{5.195475in}{0.276619in}}%
\pgfpathlineto{\pgfqpoint{5.321394in}{0.484854in}}%
\pgfpathlineto{\pgfqpoint{5.447312in}{0.683403in}}%
\pgfpathlineto{\pgfqpoint{5.573230in}{1.109556in}}%
\pgfpathlineto{\pgfqpoint{5.699148in}{1.111978in}}%
\pgfusepath{stroke}%
\end{pgfscope}%
\begin{pgfscope}%
\pgfpathrectangle{\pgfqpoint{0.536502in}{0.041670in}}{\pgfqpoint{6.421828in}{1.121275in}}%
\pgfusepath{clip}%
\pgfsetrectcap%
\pgfsetroundjoin%
\pgfsetlinewidth{1.003750pt}%
\definecolor{currentstroke}{rgb}{0.466667,0.466667,0.466667}%
\pgfsetstrokecolor{currentstroke}%
\pgfsetdash{}{0pt}%
\pgfpathmoveto{\pgfqpoint{0.662421in}{0.092637in}}%
\pgfpathlineto{\pgfqpoint{0.788339in}{0.092637in}}%
\pgfpathlineto{\pgfqpoint{0.914257in}{0.092637in}}%
\pgfpathlineto{\pgfqpoint{1.040175in}{0.092637in}}%
\pgfpathlineto{\pgfqpoint{1.166093in}{0.092637in}}%
\pgfpathlineto{\pgfqpoint{1.292011in}{0.092637in}}%
\pgfpathlineto{\pgfqpoint{1.417930in}{0.092637in}}%
\pgfpathlineto{\pgfqpoint{1.543848in}{0.092637in}}%
\pgfpathlineto{\pgfqpoint{1.669766in}{0.092637in}}%
\pgfpathlineto{\pgfqpoint{1.795684in}{0.092637in}}%
\pgfpathlineto{\pgfqpoint{1.921602in}{0.092637in}}%
\pgfpathlineto{\pgfqpoint{2.047521in}{0.092637in}}%
\pgfpathlineto{\pgfqpoint{2.173439in}{0.092637in}}%
\pgfpathlineto{\pgfqpoint{2.299357in}{0.092637in}}%
\pgfpathlineto{\pgfqpoint{2.425275in}{0.092637in}}%
\pgfpathlineto{\pgfqpoint{2.551193in}{0.092637in}}%
\pgfpathlineto{\pgfqpoint{2.677112in}{0.092637in}}%
\pgfpathlineto{\pgfqpoint{2.803030in}{0.092637in}}%
\pgfpathlineto{\pgfqpoint{2.928948in}{0.092637in}}%
\pgfpathlineto{\pgfqpoint{3.054866in}{0.092637in}}%
\pgfpathlineto{\pgfqpoint{3.180784in}{0.092637in}}%
\pgfpathlineto{\pgfqpoint{3.306703in}{0.092637in}}%
\pgfpathlineto{\pgfqpoint{3.432621in}{0.092637in}}%
\pgfpathlineto{\pgfqpoint{3.558539in}{0.092637in}}%
\pgfusepath{stroke}%
\end{pgfscope}%
\begin{pgfscope}%
\pgfpathrectangle{\pgfqpoint{0.536502in}{0.041670in}}{\pgfqpoint{6.421828in}{1.121275in}}%
\pgfusepath{clip}%
\pgfsetrectcap%
\pgfsetroundjoin%
\pgfsetlinewidth{1.003750pt}%
\definecolor{currentstroke}{rgb}{0.466667,0.466667,0.466667}%
\pgfsetstrokecolor{currentstroke}%
\pgfsetdash{}{0pt}%
\pgfpathmoveto{\pgfqpoint{3.810375in}{0.092637in}}%
\pgfpathlineto{\pgfqpoint{3.936293in}{0.092637in}}%
\pgfpathlineto{\pgfqpoint{4.062212in}{0.092637in}}%
\pgfpathlineto{\pgfqpoint{4.188130in}{0.092637in}}%
\pgfpathlineto{\pgfqpoint{4.314048in}{0.092637in}}%
\pgfpathlineto{\pgfqpoint{4.439966in}{0.092637in}}%
\pgfpathlineto{\pgfqpoint{4.565884in}{0.092637in}}%
\pgfpathlineto{\pgfqpoint{4.691803in}{0.099862in}}%
\pgfpathlineto{\pgfqpoint{4.817721in}{0.099862in}}%
\pgfpathlineto{\pgfqpoint{4.943639in}{0.104705in}}%
\pgfpathlineto{\pgfqpoint{5.069557in}{0.143446in}}%
\pgfpathlineto{\pgfqpoint{5.195475in}{0.196715in}}%
\pgfpathlineto{\pgfqpoint{5.321394in}{0.199137in}}%
\pgfpathlineto{\pgfqpoint{5.447312in}{0.300833in}}%
\pgfpathlineto{\pgfqpoint{5.573230in}{0.286305in}}%
\pgfpathlineto{\pgfqpoint{5.699148in}{0.356523in}}%
\pgfpathlineto{\pgfqpoint{5.825066in}{0.470326in}}%
\pgfpathlineto{\pgfqpoint{5.950984in}{0.499382in}}%
\pgfpathlineto{\pgfqpoint{6.076903in}{0.540544in}}%
\pgfpathlineto{\pgfqpoint{6.202821in}{0.618027in}}%
\pgfpathlineto{\pgfqpoint{6.328739in}{0.644661in}}%
\pgfpathlineto{\pgfqpoint{6.454657in}{0.596235in}}%
\pgfpathlineto{\pgfqpoint{6.580575in}{0.707616in}}%
\pgfpathlineto{\pgfqpoint{6.706494in}{0.799626in}}%
\pgfusepath{stroke}%
\end{pgfscope}%
\begin{pgfscope}%
\pgfpathrectangle{\pgfqpoint{0.536502in}{0.041670in}}{\pgfqpoint{6.421828in}{1.121275in}}%
\pgfusepath{clip}%
\pgfsetbuttcap%
\pgfsetroundjoin%
\pgfsetlinewidth{1.003750pt}%
\definecolor{currentstroke}{rgb}{0.000000,0.392157,0.000000}%
\pgfsetstrokecolor{currentstroke}%
\pgfsetdash{{3.700000pt}{1.600000pt}}{0.000000pt}%
\pgfpathmoveto{\pgfqpoint{0.536502in}{0.092637in}}%
\pgfpathlineto{\pgfqpoint{6.958330in}{0.092637in}}%
\pgfusepath{stroke}%
\end{pgfscope}%
\begin{pgfscope}%
\pgfpathrectangle{\pgfqpoint{0.536502in}{0.041670in}}{\pgfqpoint{6.421828in}{1.121275in}}%
\pgfusepath{clip}%
\pgfsetbuttcap%
\pgfsetroundjoin%
\pgfsetlinewidth{1.003750pt}%
\definecolor{currentstroke}{rgb}{0.803922,0.521569,0.247059}%
\pgfsetstrokecolor{currentstroke}%
\pgfsetdash{{3.700000pt}{1.600000pt}}{0.000000pt}%
\pgfpathmoveto{\pgfqpoint{0.536502in}{0.915705in}}%
\pgfpathlineto{\pgfqpoint{6.958330in}{0.915705in}}%
\pgfusepath{stroke}%
\end{pgfscope}%
\begin{pgfscope}%
\pgfpathrectangle{\pgfqpoint{0.536502in}{0.041670in}}{\pgfqpoint{6.421828in}{1.121275in}}%
\pgfusepath{clip}%
\pgfsetrectcap%
\pgfsetroundjoin%
\pgfsetlinewidth{6.022500pt}%
\definecolor{currentstroke}{rgb}{1.000000,1.000000,1.000000}%
\pgfsetstrokecolor{currentstroke}%
\pgfsetdash{}{0pt}%
\pgfpathmoveto{\pgfqpoint{3.684457in}{0.041670in}}%
\pgfpathlineto{\pgfqpoint{3.684457in}{1.162945in}}%
\pgfusepath{stroke}%
\end{pgfscope}%
\begin{pgfscope}%
\pgfsetrectcap%
\pgfsetmiterjoin%
\pgfsetlinewidth{1.003750pt}%
\definecolor{currentstroke}{rgb}{1.000000,1.000000,1.000000}%
\pgfsetstrokecolor{currentstroke}%
\pgfsetdash{}{0pt}%
\pgfpathmoveto{\pgfqpoint{0.536502in}{0.041670in}}%
\pgfpathlineto{\pgfqpoint{0.536502in}{1.162945in}}%
\pgfusepath{stroke}%
\end{pgfscope}%
\begin{pgfscope}%
\pgfsetrectcap%
\pgfsetmiterjoin%
\pgfsetlinewidth{1.003750pt}%
\definecolor{currentstroke}{rgb}{1.000000,1.000000,1.000000}%
\pgfsetstrokecolor{currentstroke}%
\pgfsetdash{}{0pt}%
\pgfpathmoveto{\pgfqpoint{6.958330in}{0.041670in}}%
\pgfpathlineto{\pgfqpoint{6.958330in}{1.162945in}}%
\pgfusepath{stroke}%
\end{pgfscope}%
\begin{pgfscope}%
\pgfsetrectcap%
\pgfsetmiterjoin%
\pgfsetlinewidth{1.003750pt}%
\definecolor{currentstroke}{rgb}{1.000000,1.000000,1.000000}%
\pgfsetstrokecolor{currentstroke}%
\pgfsetdash{}{0pt}%
\pgfpathmoveto{\pgfqpoint{0.536502in}{0.041670in}}%
\pgfpathlineto{\pgfqpoint{6.958330in}{0.041670in}}%
\pgfusepath{stroke}%
\end{pgfscope}%
\begin{pgfscope}%
\pgfsetrectcap%
\pgfsetmiterjoin%
\pgfsetlinewidth{1.003750pt}%
\definecolor{currentstroke}{rgb}{1.000000,1.000000,1.000000}%
\pgfsetstrokecolor{currentstroke}%
\pgfsetdash{}{0pt}%
\pgfpathmoveto{\pgfqpoint{0.536502in}{1.162945in}}%
\pgfpathlineto{\pgfqpoint{6.958330in}{1.162945in}}%
\pgfusepath{stroke}%
\end{pgfscope}%
\end{pgfpicture}%
\makeatother%
\endgroup%

%% file: figures/comparison-thumbposition.pgf
\begingroup%
\makeatletter%
\begin{pgfpicture}%
\pgfpathrectangle{\pgfpointorigin}{\pgfqpoint{7.000000in}{1.550000in}}%
\pgfusepath{use as bounding box, clip}%
\begin{pgfscope}%
\pgfsetbuttcap%
\pgfsetmiterjoin%
\definecolor{currentfill}{rgb}{1.000000,1.000000,1.000000}%
\pgfsetfillcolor{currentfill}%
\pgfsetlinewidth{0.000000pt}%
\definecolor{currentstroke}{rgb}{0.500000,0.500000,0.500000}%
\pgfsetstrokecolor{currentstroke}%
\pgfsetdash{}{0pt}%
\pgfpathmoveto{\pgfqpoint{0.000000in}{0.000000in}}%
\pgfpathlineto{\pgfqpoint{7.000000in}{0.000000in}}%
\pgfpathlineto{\pgfqpoint{7.000000in}{1.550000in}}%
\pgfpathlineto{\pgfqpoint{0.000000in}{1.550000in}}%
\pgfpathlineto{\pgfqpoint{0.000000in}{0.000000in}}%
\pgfpathclose%
\pgfusepath{fill}%
\end{pgfscope}%
\begin{pgfscope}%
\pgfsetbuttcap%
\pgfsetmiterjoin%
\definecolor{currentfill}{rgb}{0.898039,0.898039,0.898039}%
\pgfsetfillcolor{currentfill}%
\pgfsetlinewidth{0.000000pt}%
\definecolor{currentstroke}{rgb}{0.000000,0.000000,0.000000}%
\pgfsetstrokecolor{currentstroke}%
\pgfsetstrokeopacity{0.000000}%
\pgfsetdash{}{0pt}%
\pgfpathmoveto{\pgfqpoint{0.536502in}{0.394792in}}%
\pgfpathlineto{\pgfqpoint{6.958330in}{0.394792in}}%
\pgfpathlineto{\pgfqpoint{6.958330in}{1.508330in}}%
\pgfpathlineto{\pgfqpoint{0.536502in}{1.508330in}}%
\pgfpathlineto{\pgfqpoint{0.536502in}{0.394792in}}%
\pgfpathclose%
\pgfusepath{fill}%
\end{pgfscope}%
\begin{pgfscope}%
\pgfpathrectangle{\pgfqpoint{0.536502in}{0.394792in}}{\pgfqpoint{6.421828in}{1.113538in}}%
\pgfusepath{clip}%
\pgfsetrectcap%
\pgfsetroundjoin%
\pgfsetlinewidth{0.803000pt}%
\definecolor{currentstroke}{rgb}{1.000000,1.000000,1.000000}%
\pgfsetstrokecolor{currentstroke}%
\pgfsetdash{}{0pt}%
\pgfpathmoveto{\pgfqpoint{0.662421in}{0.394792in}}%
\pgfpathlineto{\pgfqpoint{0.662421in}{1.508330in}}%
\pgfusepath{stroke}%
\end{pgfscope}%
\begin{pgfscope}%
\definecolor{textcolor}{rgb}{0.333333,0.333333,0.333333}%
\pgfsetstrokecolor{textcolor}%
\pgfsetfillcolor{textcolor}%
\pgftext[x=0.662421in,y=0.297570in,,top]{\color{textcolor}\sffamily\fontsize{10.000000}{12.000000}\selectfont 1}%
\end{pgfscope}%
\begin{pgfscope}%
\pgfpathrectangle{\pgfqpoint{0.536502in}{0.394792in}}{\pgfqpoint{6.421828in}{1.113538in}}%
\pgfusepath{clip}%
\pgfsetrectcap%
\pgfsetroundjoin%
\pgfsetlinewidth{0.803000pt}%
\definecolor{currentstroke}{rgb}{1.000000,1.000000,1.000000}%
\pgfsetstrokecolor{currentstroke}%
\pgfsetdash{}{0pt}%
\pgfpathmoveto{\pgfqpoint{3.558539in}{0.394792in}}%
\pgfpathlineto{\pgfqpoint{3.558539in}{1.508330in}}%
\pgfusepath{stroke}%
\end{pgfscope}%
\begin{pgfscope}%
\definecolor{textcolor}{rgb}{0.333333,0.333333,0.333333}%
\pgfsetstrokecolor{textcolor}%
\pgfsetfillcolor{textcolor}%
\pgftext[x=3.558539in,y=0.297570in,,top]{\color{textcolor}\sffamily\fontsize{10.000000}{12.000000}\selectfont 24}%
\end{pgfscope}%
\begin{pgfscope}%
\pgfpathrectangle{\pgfqpoint{0.536502in}{0.394792in}}{\pgfqpoint{6.421828in}{1.113538in}}%
\pgfusepath{clip}%
\pgfsetrectcap%
\pgfsetroundjoin%
\pgfsetlinewidth{0.803000pt}%
\definecolor{currentstroke}{rgb}{1.000000,1.000000,1.000000}%
\pgfsetstrokecolor{currentstroke}%
\pgfsetdash{}{0pt}%
\pgfpathmoveto{\pgfqpoint{3.810375in}{0.394792in}}%
\pgfpathlineto{\pgfqpoint{3.810375in}{1.508330in}}%
\pgfusepath{stroke}%
\end{pgfscope}%
\begin{pgfscope}%
\definecolor{textcolor}{rgb}{0.333333,0.333333,0.333333}%
\pgfsetstrokecolor{textcolor}%
\pgfsetfillcolor{textcolor}%
\pgftext[x=3.810375in,y=0.297570in,,top]{\color{textcolor}\sffamily\fontsize{10.000000}{12.000000}\selectfont 1}%
\end{pgfscope}%
\begin{pgfscope}%
\pgfpathrectangle{\pgfqpoint{0.536502in}{0.394792in}}{\pgfqpoint{6.421828in}{1.113538in}}%
\pgfusepath{clip}%
\pgfsetrectcap%
\pgfsetroundjoin%
\pgfsetlinewidth{0.803000pt}%
\definecolor{currentstroke}{rgb}{1.000000,1.000000,1.000000}%
\pgfsetstrokecolor{currentstroke}%
\pgfsetdash{}{0pt}%
\pgfpathmoveto{\pgfqpoint{6.832412in}{0.394792in}}%
\pgfpathlineto{\pgfqpoint{6.832412in}{1.508330in}}%
\pgfusepath{stroke}%
\end{pgfscope}%
\begin{pgfscope}%
\definecolor{textcolor}{rgb}{0.333333,0.333333,0.333333}%
\pgfsetstrokecolor{textcolor}%
\pgfsetfillcolor{textcolor}%
\pgftext[x=6.832412in,y=0.297570in,,top]{\color{textcolor}\sffamily\fontsize{10.000000}{12.000000}\selectfont 24}%
\end{pgfscope}%
\begin{pgfscope}%
\definecolor{textcolor}{rgb}{0.333333,0.333333,0.333333}%
\pgfsetstrokecolor{textcolor}%
\pgfsetfillcolor{textcolor}%
\pgftext[x=3.651089in,y=0.172085in,,top]{\color{textcolor}\sffamily\fontsize{10.000000}{12.000000}\selectfont Layer}%
\end{pgfscope}%
\begin{pgfscope}%
\pgfpathrectangle{\pgfqpoint{0.536502in}{0.394792in}}{\pgfqpoint{6.421828in}{1.113538in}}%
\pgfusepath{clip}%
\pgfsetrectcap%
\pgfsetroundjoin%
\pgfsetlinewidth{0.803000pt}%
\definecolor{currentstroke}{rgb}{1.000000,1.000000,1.000000}%
\pgfsetstrokecolor{currentstroke}%
\pgfsetdash{}{0pt}%
\pgfpathmoveto{\pgfqpoint{0.536502in}{0.396960in}}%
\pgfpathlineto{\pgfqpoint{6.958330in}{0.396960in}}%
\pgfusepath{stroke}%
\end{pgfscope}%
\begin{pgfscope}%
\definecolor{textcolor}{rgb}{0.333333,0.333333,0.333333}%
\pgfsetstrokecolor{textcolor}%
\pgfsetfillcolor{textcolor}%
\pgftext[x=0.041670in, y=0.344198in, left, base]{\color{textcolor}\sffamily\fontsize{10.000000}{12.000000}\selectfont 0.650}%
\end{pgfscope}%
\begin{pgfscope}%
\pgfpathrectangle{\pgfqpoint{0.536502in}{0.394792in}}{\pgfqpoint{6.421828in}{1.113538in}}%
\pgfusepath{clip}%
\pgfsetrectcap%
\pgfsetroundjoin%
\pgfsetlinewidth{0.803000pt}%
\definecolor{currentstroke}{rgb}{1.000000,1.000000,1.000000}%
\pgfsetstrokecolor{currentstroke}%
\pgfsetdash{}{0pt}%
\pgfpathmoveto{\pgfqpoint{0.536502in}{0.711556in}}%
\pgfpathlineto{\pgfqpoint{6.958330in}{0.711556in}}%
\pgfusepath{stroke}%
\end{pgfscope}%
\begin{pgfscope}%
\definecolor{textcolor}{rgb}{0.333333,0.333333,0.333333}%
\pgfsetstrokecolor{textcolor}%
\pgfsetfillcolor{textcolor}%
\pgftext[x=0.041670in, y=0.658795in, left, base]{\color{textcolor}\sffamily\fontsize{10.000000}{12.000000}\selectfont 0.700}%
\end{pgfscope}%
\begin{pgfscope}%
\pgfpathrectangle{\pgfqpoint{0.536502in}{0.394792in}}{\pgfqpoint{6.421828in}{1.113538in}}%
\pgfusepath{clip}%
\pgfsetrectcap%
\pgfsetroundjoin%
\pgfsetlinewidth{0.803000pt}%
\definecolor{currentstroke}{rgb}{1.000000,1.000000,1.000000}%
\pgfsetstrokecolor{currentstroke}%
\pgfsetdash{}{0pt}%
\pgfpathmoveto{\pgfqpoint{0.536502in}{1.026153in}}%
\pgfpathlineto{\pgfqpoint{6.958330in}{1.026153in}}%
\pgfusepath{stroke}%
\end{pgfscope}%
\begin{pgfscope}%
\definecolor{textcolor}{rgb}{0.333333,0.333333,0.333333}%
\pgfsetstrokecolor{textcolor}%
\pgfsetfillcolor{textcolor}%
\pgftext[x=0.041670in, y=0.973391in, left, base]{\color{textcolor}\sffamily\fontsize{10.000000}{12.000000}\selectfont 0.750}%
\end{pgfscope}%
\begin{pgfscope}%
\pgfpathrectangle{\pgfqpoint{0.536502in}{0.394792in}}{\pgfqpoint{6.421828in}{1.113538in}}%
\pgfusepath{clip}%
\pgfsetrectcap%
\pgfsetroundjoin%
\pgfsetlinewidth{0.803000pt}%
\definecolor{currentstroke}{rgb}{1.000000,1.000000,1.000000}%
\pgfsetstrokecolor{currentstroke}%
\pgfsetdash{}{0pt}%
\pgfpathmoveto{\pgfqpoint{0.536502in}{1.340749in}}%
\pgfpathlineto{\pgfqpoint{6.958330in}{1.340749in}}%
\pgfusepath{stroke}%
\end{pgfscope}%
\begin{pgfscope}%
\definecolor{textcolor}{rgb}{0.333333,0.333333,0.333333}%
\pgfsetstrokecolor{textcolor}%
\pgfsetfillcolor{textcolor}%
\pgftext[x=0.041670in, y=1.287988in, left, base]{\color{textcolor}\sffamily\fontsize{10.000000}{12.000000}\selectfont 0.800}%
\end{pgfscope}%
\begin{pgfscope}%
\pgfpathrectangle{\pgfqpoint{0.536502in}{0.394792in}}{\pgfqpoint{6.421828in}{1.113538in}}%
\pgfusepath{clip}%
\pgfsetrectcap%
\pgfsetroundjoin%
\pgfsetlinewidth{1.003750pt}%
\definecolor{currentstroke}{rgb}{0.886275,0.290196,0.200000}%
\pgfsetstrokecolor{currentstroke}%
\pgfsetdash{}{0pt}%
\pgfpathmoveto{\pgfqpoint{0.662421in}{0.445408in}}%
\pgfpathlineto{\pgfqpoint{0.788339in}{0.449392in}}%
\pgfpathlineto{\pgfqpoint{0.914257in}{0.445408in}}%
\pgfpathlineto{\pgfqpoint{1.040175in}{0.445408in}}%
\pgfpathlineto{\pgfqpoint{1.166093in}{0.445408in}}%
\pgfpathlineto{\pgfqpoint{1.292011in}{0.445408in}}%
\pgfpathlineto{\pgfqpoint{1.417930in}{0.445408in}}%
\pgfpathlineto{\pgfqpoint{1.543848in}{0.445408in}}%
\pgfpathlineto{\pgfqpoint{1.669766in}{0.445408in}}%
\pgfpathlineto{\pgfqpoint{1.795684in}{0.445408in}}%
\pgfpathlineto{\pgfqpoint{1.921602in}{0.445408in}}%
\pgfpathlineto{\pgfqpoint{2.047521in}{0.445408in}}%
\pgfusepath{stroke}%
\end{pgfscope}%
\begin{pgfscope}%
\pgfpathrectangle{\pgfqpoint{0.536502in}{0.394792in}}{\pgfqpoint{6.421828in}{1.113538in}}%
\pgfusepath{clip}%
\pgfsetrectcap%
\pgfsetroundjoin%
\pgfsetlinewidth{1.003750pt}%
\definecolor{currentstroke}{rgb}{0.886275,0.290196,0.200000}%
\pgfsetstrokecolor{currentstroke}%
\pgfsetdash{}{0pt}%
\pgfpathmoveto{\pgfqpoint{3.810375in}{0.445408in}}%
\pgfpathlineto{\pgfqpoint{3.936293in}{0.445408in}}%
\pgfpathlineto{\pgfqpoint{4.062212in}{0.445408in}}%
\pgfpathlineto{\pgfqpoint{4.188130in}{0.457459in}}%
\pgfpathlineto{\pgfqpoint{4.314048in}{0.445408in}}%
\pgfpathlineto{\pgfqpoint{4.439966in}{0.453426in}}%
\pgfpathlineto{\pgfqpoint{4.565884in}{0.445408in}}%
\pgfpathlineto{\pgfqpoint{4.691803in}{0.445408in}}%
\pgfpathlineto{\pgfqpoint{4.817721in}{0.445408in}}%
\pgfpathlineto{\pgfqpoint{4.943639in}{0.445408in}}%
\pgfpathlineto{\pgfqpoint{5.069557in}{0.449392in}}%
\pgfpathlineto{\pgfqpoint{5.195475in}{0.445408in}}%
\pgfusepath{stroke}%
\end{pgfscope}%
\begin{pgfscope}%
\pgfpathrectangle{\pgfqpoint{0.536502in}{0.394792in}}{\pgfqpoint{6.421828in}{1.113538in}}%
\pgfusepath{clip}%
\pgfsetrectcap%
\pgfsetroundjoin%
\pgfsetlinewidth{1.003750pt}%
\definecolor{currentstroke}{rgb}{0.203922,0.541176,0.741176}%
\pgfsetstrokecolor{currentstroke}%
\pgfsetdash{}{0pt}%
\pgfpathmoveto{\pgfqpoint{0.662421in}{0.469559in}}%
\pgfpathlineto{\pgfqpoint{0.788339in}{0.445408in}}%
\pgfpathlineto{\pgfqpoint{0.914257in}{0.445408in}}%
\pgfpathlineto{\pgfqpoint{1.040175in}{0.445408in}}%
\pgfpathlineto{\pgfqpoint{1.166093in}{0.445408in}}%
\pgfpathlineto{\pgfqpoint{1.292011in}{0.445408in}}%
\pgfpathlineto{\pgfqpoint{1.417930in}{0.445408in}}%
\pgfpathlineto{\pgfqpoint{1.543848in}{0.445408in}}%
\pgfpathlineto{\pgfqpoint{1.669766in}{0.445408in}}%
\pgfpathlineto{\pgfqpoint{1.795684in}{0.501825in}}%
\pgfpathlineto{\pgfqpoint{1.921602in}{0.449392in}}%
\pgfpathlineto{\pgfqpoint{2.047521in}{0.445408in}}%
\pgfusepath{stroke}%
\end{pgfscope}%
\begin{pgfscope}%
\pgfpathrectangle{\pgfqpoint{0.536502in}{0.394792in}}{\pgfqpoint{6.421828in}{1.113538in}}%
\pgfusepath{clip}%
\pgfsetrectcap%
\pgfsetroundjoin%
\pgfsetlinewidth{1.003750pt}%
\definecolor{currentstroke}{rgb}{0.203922,0.541176,0.741176}%
\pgfsetstrokecolor{currentstroke}%
\pgfsetdash{}{0pt}%
\pgfpathmoveto{\pgfqpoint{3.810375in}{0.445408in}}%
\pgfpathlineto{\pgfqpoint{3.936293in}{0.445408in}}%
\pgfpathlineto{\pgfqpoint{4.062212in}{0.445408in}}%
\pgfpathlineto{\pgfqpoint{4.188130in}{0.445408in}}%
\pgfpathlineto{\pgfqpoint{4.314048in}{0.445408in}}%
\pgfpathlineto{\pgfqpoint{4.439966in}{0.445408in}}%
\pgfpathlineto{\pgfqpoint{4.565884in}{0.445408in}}%
\pgfpathlineto{\pgfqpoint{4.691803in}{0.445408in}}%
\pgfpathlineto{\pgfqpoint{4.817721in}{0.445408in}}%
\pgfpathlineto{\pgfqpoint{4.943639in}{0.445408in}}%
\pgfpathlineto{\pgfqpoint{5.069557in}{0.445408in}}%
\pgfpathlineto{\pgfqpoint{5.195475in}{0.445408in}}%
\pgfusepath{stroke}%
\end{pgfscope}%
\begin{pgfscope}%
\pgfpathrectangle{\pgfqpoint{0.536502in}{0.394792in}}{\pgfqpoint{6.421828in}{1.113538in}}%
\pgfusepath{clip}%
\pgfsetrectcap%
\pgfsetroundjoin%
\pgfsetlinewidth{1.003750pt}%
\definecolor{currentstroke}{rgb}{0.596078,0.556863,0.835294}%
\pgfsetstrokecolor{currentstroke}%
\pgfsetdash{}{0pt}%
\pgfpathmoveto{\pgfqpoint{0.662421in}{0.445408in}}%
\pgfpathlineto{\pgfqpoint{0.788339in}{0.453426in}}%
\pgfpathlineto{\pgfqpoint{0.914257in}{0.445408in}}%
\pgfpathlineto{\pgfqpoint{1.040175in}{0.445408in}}%
\pgfpathlineto{\pgfqpoint{1.166093in}{0.469559in}}%
\pgfpathlineto{\pgfqpoint{1.292011in}{0.493759in}}%
\pgfpathlineto{\pgfqpoint{1.417930in}{0.449392in}}%
\pgfpathlineto{\pgfqpoint{1.543848in}{0.445408in}}%
\pgfpathlineto{\pgfqpoint{1.669766in}{0.457459in}}%
\pgfpathlineto{\pgfqpoint{1.795684in}{0.513925in}}%
\pgfpathlineto{\pgfqpoint{1.921602in}{0.509892in}}%
\pgfpathlineto{\pgfqpoint{2.047521in}{0.493759in}}%
\pgfpathlineto{\pgfqpoint{2.173439in}{0.489725in}}%
\pgfpathlineto{\pgfqpoint{2.299357in}{0.453426in}}%
\pgfpathlineto{\pgfqpoint{2.425275in}{0.445408in}}%
\pgfpathlineto{\pgfqpoint{2.551193in}{0.445408in}}%
\pgfusepath{stroke}%
\end{pgfscope}%
\begin{pgfscope}%
\pgfpathrectangle{\pgfqpoint{0.536502in}{0.394792in}}{\pgfqpoint{6.421828in}{1.113538in}}%
\pgfusepath{clip}%
\pgfsetrectcap%
\pgfsetroundjoin%
\pgfsetlinewidth{1.003750pt}%
\definecolor{currentstroke}{rgb}{0.596078,0.556863,0.835294}%
\pgfsetstrokecolor{currentstroke}%
\pgfsetdash{}{0pt}%
\pgfpathmoveto{\pgfqpoint{3.810375in}{0.445408in}}%
\pgfpathlineto{\pgfqpoint{3.936293in}{0.445408in}}%
\pgfpathlineto{\pgfqpoint{4.062212in}{0.445408in}}%
\pgfpathlineto{\pgfqpoint{4.188130in}{0.453426in}}%
\pgfpathlineto{\pgfqpoint{4.314048in}{0.477625in}}%
\pgfpathlineto{\pgfqpoint{4.439966in}{0.562325in}}%
\pgfpathlineto{\pgfqpoint{4.565884in}{0.489725in}}%
\pgfpathlineto{\pgfqpoint{4.691803in}{0.509892in}}%
\pgfpathlineto{\pgfqpoint{4.817721in}{0.489725in}}%
\pgfpathlineto{\pgfqpoint{4.943639in}{0.695423in}}%
\pgfpathlineto{\pgfqpoint{5.069557in}{0.723656in}}%
\pgfpathlineto{\pgfqpoint{5.195475in}{0.868854in}}%
\pgfpathlineto{\pgfqpoint{5.321394in}{1.151185in}}%
\pgfpathlineto{\pgfqpoint{5.447312in}{1.304450in}}%
\pgfpathlineto{\pgfqpoint{5.573230in}{1.457715in}}%
\pgfpathlineto{\pgfqpoint{5.699148in}{1.223784in}}%
\pgfusepath{stroke}%
\end{pgfscope}%
\begin{pgfscope}%
\pgfpathrectangle{\pgfqpoint{0.536502in}{0.394792in}}{\pgfqpoint{6.421828in}{1.113538in}}%
\pgfusepath{clip}%
\pgfsetrectcap%
\pgfsetroundjoin%
\pgfsetlinewidth{1.003750pt}%
\definecolor{currentstroke}{rgb}{0.466667,0.466667,0.466667}%
\pgfsetstrokecolor{currentstroke}%
\pgfsetdash{}{0pt}%
\pgfpathmoveto{\pgfqpoint{0.662421in}{0.453426in}}%
\pgfpathlineto{\pgfqpoint{0.788339in}{0.445408in}}%
\pgfpathlineto{\pgfqpoint{0.914257in}{0.445408in}}%
\pgfpathlineto{\pgfqpoint{1.040175in}{0.445408in}}%
\pgfpathlineto{\pgfqpoint{1.166093in}{0.445408in}}%
\pgfpathlineto{\pgfqpoint{1.292011in}{0.445408in}}%
\pgfpathlineto{\pgfqpoint{1.417930in}{0.469559in}}%
\pgfpathlineto{\pgfqpoint{1.543848in}{0.526025in}}%
\pgfpathlineto{\pgfqpoint{1.669766in}{0.546191in}}%
\pgfpathlineto{\pgfqpoint{1.795684in}{0.477625in}}%
\pgfpathlineto{\pgfqpoint{1.921602in}{0.489725in}}%
\pgfpathlineto{\pgfqpoint{2.047521in}{0.546191in}}%
\pgfpathlineto{\pgfqpoint{2.173439in}{0.534092in}}%
\pgfpathlineto{\pgfqpoint{2.299357in}{0.501825in}}%
\pgfpathlineto{\pgfqpoint{2.425275in}{0.513925in}}%
\pgfpathlineto{\pgfqpoint{2.551193in}{0.445408in}}%
\pgfpathlineto{\pgfqpoint{2.677112in}{0.465526in}}%
\pgfpathlineto{\pgfqpoint{2.803030in}{0.445408in}}%
\pgfpathlineto{\pgfqpoint{2.928948in}{0.445408in}}%
\pgfpathlineto{\pgfqpoint{3.054866in}{0.445408in}}%
\pgfpathlineto{\pgfqpoint{3.180784in}{0.501825in}}%
\pgfpathlineto{\pgfqpoint{3.306703in}{0.465526in}}%
\pgfpathlineto{\pgfqpoint{3.432621in}{0.445408in}}%
\pgfpathlineto{\pgfqpoint{3.558539in}{0.445408in}}%
\pgfusepath{stroke}%
\end{pgfscope}%
\begin{pgfscope}%
\pgfpathrectangle{\pgfqpoint{0.536502in}{0.394792in}}{\pgfqpoint{6.421828in}{1.113538in}}%
\pgfusepath{clip}%
\pgfsetrectcap%
\pgfsetroundjoin%
\pgfsetlinewidth{1.003750pt}%
\definecolor{currentstroke}{rgb}{0.466667,0.466667,0.466667}%
\pgfsetstrokecolor{currentstroke}%
\pgfsetdash{}{0pt}%
\pgfpathmoveto{\pgfqpoint{3.810375in}{0.453426in}}%
\pgfpathlineto{\pgfqpoint{3.936293in}{0.445408in}}%
\pgfpathlineto{\pgfqpoint{4.062212in}{0.453426in}}%
\pgfpathlineto{\pgfqpoint{4.188130in}{0.469559in}}%
\pgfpathlineto{\pgfqpoint{4.314048in}{0.473592in}}%
\pgfpathlineto{\pgfqpoint{4.439966in}{0.461492in}}%
\pgfpathlineto{\pgfqpoint{4.565884in}{0.614757in}}%
\pgfpathlineto{\pgfqpoint{4.691803in}{0.642990in}}%
\pgfpathlineto{\pgfqpoint{4.817721in}{0.602657in}}%
\pgfpathlineto{\pgfqpoint{4.943639in}{0.671223in}}%
\pgfpathlineto{\pgfqpoint{5.069557in}{0.687356in}}%
\pgfpathlineto{\pgfqpoint{5.195475in}{0.711556in}}%
\pgfpathlineto{\pgfqpoint{5.321394in}{0.808355in}}%
\pgfpathlineto{\pgfqpoint{5.447312in}{0.852721in}}%
\pgfpathlineto{\pgfqpoint{5.573230in}{0.884988in}}%
\pgfpathlineto{\pgfqpoint{5.699148in}{0.929354in}}%
\pgfpathlineto{\pgfqpoint{5.825066in}{0.957587in}}%
\pgfpathlineto{\pgfqpoint{5.950984in}{0.993886in}}%
\pgfpathlineto{\pgfqpoint{6.076903in}{1.026153in}}%
\pgfpathlineto{\pgfqpoint{6.202821in}{0.997920in}}%
\pgfpathlineto{\pgfqpoint{6.328739in}{1.034219in}}%
\pgfpathlineto{\pgfqpoint{6.454657in}{1.070519in}}%
\pgfpathlineto{\pgfqpoint{6.580575in}{1.191518in}}%
\pgfpathlineto{\pgfqpoint{6.706494in}{1.296383in}}%
\pgfusepath{stroke}%
\end{pgfscope}%
\begin{pgfscope}%
\pgfpathrectangle{\pgfqpoint{0.536502in}{0.394792in}}{\pgfqpoint{6.421828in}{1.113538in}}%
\pgfusepath{clip}%
\pgfsetbuttcap%
\pgfsetroundjoin%
\pgfsetlinewidth{1.003750pt}%
\definecolor{currentstroke}{rgb}{0.000000,0.392157,0.000000}%
\pgfsetstrokecolor{currentstroke}%
\pgfsetdash{{3.700000pt}{1.600000pt}}{0.000000pt}%
\pgfpathmoveto{\pgfqpoint{0.536502in}{0.445408in}}%
\pgfpathlineto{\pgfqpoint{6.958330in}{0.445408in}}%
\pgfusepath{stroke}%
\end{pgfscope}%
\begin{pgfscope}%
\pgfpathrectangle{\pgfqpoint{0.536502in}{0.394792in}}{\pgfqpoint{6.421828in}{1.113538in}}%
\pgfusepath{clip}%
\pgfsetbuttcap%
\pgfsetroundjoin%
\pgfsetlinewidth{1.003750pt}%
\definecolor{currentstroke}{rgb}{0.803922,0.521569,0.247059}%
\pgfsetstrokecolor{currentstroke}%
\pgfsetdash{{3.700000pt}{1.600000pt}}{0.000000pt}%
\pgfpathmoveto{\pgfqpoint{0.536502in}{1.255808in}}%
\pgfpathlineto{\pgfqpoint{6.958330in}{1.255808in}}%
\pgfusepath{stroke}%
\end{pgfscope}%
\begin{pgfscope}%
\pgfpathrectangle{\pgfqpoint{0.536502in}{0.394792in}}{\pgfqpoint{6.421828in}{1.113538in}}%
\pgfusepath{clip}%
\pgfsetrectcap%
\pgfsetroundjoin%
\pgfsetlinewidth{6.022500pt}%
\definecolor{currentstroke}{rgb}{1.000000,1.000000,1.000000}%
\pgfsetstrokecolor{currentstroke}%
\pgfsetdash{}{0pt}%
\pgfpathmoveto{\pgfqpoint{3.684457in}{0.394792in}}%
\pgfpathlineto{\pgfqpoint{3.684457in}{1.508330in}}%
\pgfusepath{stroke}%
\end{pgfscope}%
\begin{pgfscope}%
\pgfsetrectcap%
\pgfsetmiterjoin%
\pgfsetlinewidth{1.003750pt}%
\definecolor{currentstroke}{rgb}{1.000000,1.000000,1.000000}%
\pgfsetstrokecolor{currentstroke}%
\pgfsetdash{}{0pt}%
\pgfpathmoveto{\pgfqpoint{0.536502in}{0.394792in}}%
\pgfpathlineto{\pgfqpoint{0.536502in}{1.508330in}}%
\pgfusepath{stroke}%
\end{pgfscope}%
\begin{pgfscope}%
\pgfsetrectcap%
\pgfsetmiterjoin%
\pgfsetlinewidth{1.003750pt}%
\definecolor{currentstroke}{rgb}{1.000000,1.000000,1.000000}%
\pgfsetstrokecolor{currentstroke}%
\pgfsetdash{}{0pt}%
\pgfpathmoveto{\pgfqpoint{6.958330in}{0.394792in}}%
\pgfpathlineto{\pgfqpoint{6.958330in}{1.508330in}}%
\pgfusepath{stroke}%
\end{pgfscope}%
\begin{pgfscope}%
\pgfsetrectcap%
\pgfsetmiterjoin%
\pgfsetlinewidth{1.003750pt}%
\definecolor{currentstroke}{rgb}{1.000000,1.000000,1.000000}%
\pgfsetstrokecolor{currentstroke}%
\pgfsetdash{}{0pt}%
\pgfpathmoveto{\pgfqpoint{0.536502in}{0.394792in}}%
\pgfpathlineto{\pgfqpoint{6.958330in}{0.394792in}}%
\pgfusepath{stroke}%
\end{pgfscope}%
\begin{pgfscope}%
\pgfsetrectcap%
\pgfsetmiterjoin%
\pgfsetlinewidth{1.003750pt}%
\definecolor{currentstroke}{rgb}{1.000000,1.000000,1.000000}%
\pgfsetstrokecolor{currentstroke}%
\pgfsetdash{}{0pt}%
\pgfpathmoveto{\pgfqpoint{0.536502in}{1.508330in}}%
\pgfpathlineto{\pgfqpoint{6.958330in}{1.508330in}}%
\pgfusepath{stroke}%
\end{pgfscope}%
\end{pgfpicture}%
\makeatother%
\endgroup%

%% file: figures/comparison-pathmovement-nox.pgf
\begingroup%
\makeatletter%
\begin{pgfpicture}%
\pgfpathrectangle{\pgfpointorigin}{\pgfqpoint{7.000000in}{1.250000in}}%
\pgfusepath{use as bounding box, clip}%
\begin{pgfscope}%
\pgfsetbuttcap%
\pgfsetmiterjoin%
\definecolor{currentfill}{rgb}{1.000000,1.000000,1.000000}%
\pgfsetfillcolor{currentfill}%
\pgfsetlinewidth{0.000000pt}%
\definecolor{currentstroke}{rgb}{0.500000,0.500000,0.500000}%
\pgfsetstrokecolor{currentstroke}%
\pgfsetdash{}{0pt}%
\pgfpathmoveto{\pgfqpoint{0.000000in}{0.000000in}}%
\pgfpathlineto{\pgfqpoint{7.000000in}{0.000000in}}%
\pgfpathlineto{\pgfqpoint{7.000000in}{1.250000in}}%
\pgfpathlineto{\pgfqpoint{0.000000in}{1.250000in}}%
\pgfpathlineto{\pgfqpoint{0.000000in}{0.000000in}}%
\pgfpathclose%
\pgfusepath{fill}%
\end{pgfscope}%
\begin{pgfscope}%
\pgfsetbuttcap%
\pgfsetmiterjoin%
\definecolor{currentfill}{rgb}{0.898039,0.898039,0.898039}%
\pgfsetfillcolor{currentfill}%
\pgfsetlinewidth{0.000000pt}%
\definecolor{currentstroke}{rgb}{0.000000,0.000000,0.000000}%
\pgfsetstrokecolor{currentstroke}%
\pgfsetstrokeopacity{0.000000}%
\pgfsetdash{}{0pt}%
\pgfpathmoveto{\pgfqpoint{0.536502in}{0.041670in}}%
\pgfpathlineto{\pgfqpoint{6.958330in}{0.041670in}}%
\pgfpathlineto{\pgfqpoint{6.958330in}{1.208330in}}%
\pgfpathlineto{\pgfqpoint{0.536502in}{1.208330in}}%
\pgfpathlineto{\pgfqpoint{0.536502in}{0.041670in}}%
\pgfpathclose%
\pgfusepath{fill}%
\end{pgfscope}%
\begin{pgfscope}%
\pgfpathrectangle{\pgfqpoint{0.536502in}{0.041670in}}{\pgfqpoint{6.421828in}{1.166660in}}%
\pgfusepath{clip}%
\pgfsetrectcap%
\pgfsetroundjoin%
\pgfsetlinewidth{0.803000pt}%
\definecolor{currentstroke}{rgb}{1.000000,1.000000,1.000000}%
\pgfsetstrokecolor{currentstroke}%
\pgfsetdash{}{0pt}%
\pgfpathmoveto{\pgfqpoint{0.662421in}{0.041670in}}%
\pgfpathlineto{\pgfqpoint{0.662421in}{1.208330in}}%
\pgfusepath{stroke}%
\end{pgfscope}%
\begin{pgfscope}%
\pgfpathrectangle{\pgfqpoint{0.536502in}{0.041670in}}{\pgfqpoint{6.421828in}{1.166660in}}%
\pgfusepath{clip}%
\pgfsetrectcap%
\pgfsetroundjoin%
\pgfsetlinewidth{0.803000pt}%
\definecolor{currentstroke}{rgb}{1.000000,1.000000,1.000000}%
\pgfsetstrokecolor{currentstroke}%
\pgfsetdash{}{0pt}%
\pgfpathmoveto{\pgfqpoint{3.558539in}{0.041670in}}%
\pgfpathlineto{\pgfqpoint{3.558539in}{1.208330in}}%
\pgfusepath{stroke}%
\end{pgfscope}%
\begin{pgfscope}%
\pgfpathrectangle{\pgfqpoint{0.536502in}{0.041670in}}{\pgfqpoint{6.421828in}{1.166660in}}%
\pgfusepath{clip}%
\pgfsetrectcap%
\pgfsetroundjoin%
\pgfsetlinewidth{0.803000pt}%
\definecolor{currentstroke}{rgb}{1.000000,1.000000,1.000000}%
\pgfsetstrokecolor{currentstroke}%
\pgfsetdash{}{0pt}%
\pgfpathmoveto{\pgfqpoint{3.810375in}{0.041670in}}%
\pgfpathlineto{\pgfqpoint{3.810375in}{1.208330in}}%
\pgfusepath{stroke}%
\end{pgfscope}%
\begin{pgfscope}%
\pgfpathrectangle{\pgfqpoint{0.536502in}{0.041670in}}{\pgfqpoint{6.421828in}{1.166660in}}%
\pgfusepath{clip}%
\pgfsetrectcap%
\pgfsetroundjoin%
\pgfsetlinewidth{0.803000pt}%
\definecolor{currentstroke}{rgb}{1.000000,1.000000,1.000000}%
\pgfsetstrokecolor{currentstroke}%
\pgfsetdash{}{0pt}%
\pgfpathmoveto{\pgfqpoint{6.832412in}{0.041670in}}%
\pgfpathlineto{\pgfqpoint{6.832412in}{1.208330in}}%
\pgfusepath{stroke}%
\end{pgfscope}%
\begin{pgfscope}%
\pgfpathrectangle{\pgfqpoint{0.536502in}{0.041670in}}{\pgfqpoint{6.421828in}{1.166660in}}%
\pgfusepath{clip}%
\pgfsetrectcap%
\pgfsetroundjoin%
\pgfsetlinewidth{0.803000pt}%
\definecolor{currentstroke}{rgb}{1.000000,1.000000,1.000000}%
\pgfsetstrokecolor{currentstroke}%
\pgfsetdash{}{0pt}%
\pgfpathmoveto{\pgfqpoint{0.536502in}{0.143641in}}%
\pgfpathlineto{\pgfqpoint{6.958330in}{0.143641in}}%
\pgfusepath{stroke}%
\end{pgfscope}%
\begin{pgfscope}%
\definecolor{textcolor}{rgb}{0.333333,0.333333,0.333333}%
\pgfsetstrokecolor{textcolor}%
\pgfsetfillcolor{textcolor}%
\pgftext[x=0.041670in, y=0.090879in, left, base]{\color{textcolor}\sffamily\fontsize{10.000000}{12.000000}\selectfont 0.500}%
\end{pgfscope}%
\begin{pgfscope}%
\pgfpathrectangle{\pgfqpoint{0.536502in}{0.041670in}}{\pgfqpoint{6.421828in}{1.166660in}}%
\pgfusepath{clip}%
\pgfsetrectcap%
\pgfsetroundjoin%
\pgfsetlinewidth{0.803000pt}%
\definecolor{currentstroke}{rgb}{1.000000,1.000000,1.000000}%
\pgfsetstrokecolor{currentstroke}%
\pgfsetdash{}{0pt}%
\pgfpathmoveto{\pgfqpoint{0.536502in}{0.712722in}}%
\pgfpathlineto{\pgfqpoint{6.958330in}{0.712722in}}%
\pgfusepath{stroke}%
\end{pgfscope}%
\begin{pgfscope}%
\definecolor{textcolor}{rgb}{0.333333,0.333333,0.333333}%
\pgfsetstrokecolor{textcolor}%
\pgfsetfillcolor{textcolor}%
\pgftext[x=0.041670in, y=0.659960in, left, base]{\color{textcolor}\sffamily\fontsize{10.000000}{12.000000}\selectfont 0.600}%
\end{pgfscope}%
\begin{pgfscope}%
\pgfpathrectangle{\pgfqpoint{0.536502in}{0.041670in}}{\pgfqpoint{6.421828in}{1.166660in}}%
\pgfusepath{clip}%
\pgfsetrectcap%
\pgfsetroundjoin%
\pgfsetlinewidth{1.003750pt}%
\definecolor{currentstroke}{rgb}{0.886275,0.290196,0.200000}%
\pgfsetstrokecolor{currentstroke}%
\pgfsetdash{}{0pt}%
\pgfpathmoveto{\pgfqpoint{0.662421in}{0.094700in}}%
\pgfpathlineto{\pgfqpoint{0.788339in}{0.094700in}}%
\pgfpathlineto{\pgfqpoint{0.914257in}{0.094700in}}%
\pgfpathlineto{\pgfqpoint{1.040175in}{0.094700in}}%
\pgfpathlineto{\pgfqpoint{1.166093in}{0.094700in}}%
\pgfpathlineto{\pgfqpoint{1.292011in}{0.094700in}}%
\pgfpathlineto{\pgfqpoint{1.417930in}{0.094700in}}%
\pgfpathlineto{\pgfqpoint{1.543848in}{0.094700in}}%
\pgfpathlineto{\pgfqpoint{1.669766in}{0.094700in}}%
\pgfpathlineto{\pgfqpoint{1.795684in}{0.094700in}}%
\pgfpathlineto{\pgfqpoint{1.921602in}{0.094700in}}%
\pgfpathlineto{\pgfqpoint{2.047521in}{0.094700in}}%
\pgfusepath{stroke}%
\end{pgfscope}%
\begin{pgfscope}%
\pgfpathrectangle{\pgfqpoint{0.536502in}{0.041670in}}{\pgfqpoint{6.421828in}{1.166660in}}%
\pgfusepath{clip}%
\pgfsetrectcap%
\pgfsetroundjoin%
\pgfsetlinewidth{1.003750pt}%
\definecolor{currentstroke}{rgb}{0.886275,0.290196,0.200000}%
\pgfsetstrokecolor{currentstroke}%
\pgfsetdash{}{0pt}%
\pgfpathmoveto{\pgfqpoint{3.810375in}{0.094700in}}%
\pgfpathlineto{\pgfqpoint{3.936293in}{0.101716in}}%
\pgfpathlineto{\pgfqpoint{4.062212in}{0.094700in}}%
\pgfpathlineto{\pgfqpoint{4.188130in}{0.094700in}}%
\pgfpathlineto{\pgfqpoint{4.314048in}{0.094700in}}%
\pgfpathlineto{\pgfqpoint{4.439966in}{0.094700in}}%
\pgfpathlineto{\pgfqpoint{4.565884in}{0.101716in}}%
\pgfpathlineto{\pgfqpoint{4.691803in}{0.094700in}}%
\pgfpathlineto{\pgfqpoint{4.817721in}{0.094700in}}%
\pgfpathlineto{\pgfqpoint{4.943639in}{0.094700in}}%
\pgfpathlineto{\pgfqpoint{5.069557in}{0.094700in}}%
\pgfpathlineto{\pgfqpoint{5.195475in}{0.094700in}}%
\pgfusepath{stroke}%
\end{pgfscope}%
\begin{pgfscope}%
\pgfpathrectangle{\pgfqpoint{0.536502in}{0.041670in}}{\pgfqpoint{6.421828in}{1.166660in}}%
\pgfusepath{clip}%
\pgfsetrectcap%
\pgfsetroundjoin%
\pgfsetlinewidth{1.003750pt}%
\definecolor{currentstroke}{rgb}{0.203922,0.541176,0.741176}%
\pgfsetstrokecolor{currentstroke}%
\pgfsetdash{}{0pt}%
\pgfpathmoveto{\pgfqpoint{0.662421in}{0.094700in}}%
\pgfpathlineto{\pgfqpoint{0.788339in}{0.094700in}}%
\pgfpathlineto{\pgfqpoint{0.914257in}{0.094700in}}%
\pgfpathlineto{\pgfqpoint{1.040175in}{0.094700in}}%
\pgfpathlineto{\pgfqpoint{1.166093in}{0.094700in}}%
\pgfpathlineto{\pgfqpoint{1.292011in}{0.094700in}}%
\pgfpathlineto{\pgfqpoint{1.417930in}{0.094700in}}%
\pgfpathlineto{\pgfqpoint{1.543848in}{0.094700in}}%
\pgfpathlineto{\pgfqpoint{1.669766in}{0.094700in}}%
\pgfpathlineto{\pgfqpoint{1.795684in}{0.094700in}}%
\pgfpathlineto{\pgfqpoint{1.921602in}{0.094700in}}%
\pgfpathlineto{\pgfqpoint{2.047521in}{0.094700in}}%
\pgfusepath{stroke}%
\end{pgfscope}%
\begin{pgfscope}%
\pgfpathrectangle{\pgfqpoint{0.536502in}{0.041670in}}{\pgfqpoint{6.421828in}{1.166660in}}%
\pgfusepath{clip}%
\pgfsetrectcap%
\pgfsetroundjoin%
\pgfsetlinewidth{1.003750pt}%
\definecolor{currentstroke}{rgb}{0.203922,0.541176,0.741176}%
\pgfsetstrokecolor{currentstroke}%
\pgfsetdash{}{0pt}%
\pgfpathmoveto{\pgfqpoint{3.810375in}{0.094700in}}%
\pgfpathlineto{\pgfqpoint{3.936293in}{0.094700in}}%
\pgfpathlineto{\pgfqpoint{4.062212in}{0.094700in}}%
\pgfpathlineto{\pgfqpoint{4.188130in}{0.094700in}}%
\pgfpathlineto{\pgfqpoint{4.314048in}{0.094700in}}%
\pgfpathlineto{\pgfqpoint{4.439966in}{0.094700in}}%
\pgfpathlineto{\pgfqpoint{4.565884in}{0.094700in}}%
\pgfpathlineto{\pgfqpoint{4.691803in}{0.094700in}}%
\pgfpathlineto{\pgfqpoint{4.817721in}{0.094700in}}%
\pgfpathlineto{\pgfqpoint{4.943639in}{0.094700in}}%
\pgfpathlineto{\pgfqpoint{5.069557in}{0.094700in}}%
\pgfpathlineto{\pgfqpoint{5.195475in}{0.094700in}}%
\pgfusepath{stroke}%
\end{pgfscope}%
\begin{pgfscope}%
\pgfpathrectangle{\pgfqpoint{0.536502in}{0.041670in}}{\pgfqpoint{6.421828in}{1.166660in}}%
\pgfusepath{clip}%
\pgfsetrectcap%
\pgfsetroundjoin%
\pgfsetlinewidth{1.003750pt}%
\definecolor{currentstroke}{rgb}{0.596078,0.556863,0.835294}%
\pgfsetstrokecolor{currentstroke}%
\pgfsetdash{}{0pt}%
\pgfpathmoveto{\pgfqpoint{0.662421in}{0.094700in}}%
\pgfpathlineto{\pgfqpoint{0.788339in}{0.101716in}}%
\pgfpathlineto{\pgfqpoint{0.914257in}{0.094700in}}%
\pgfpathlineto{\pgfqpoint{1.040175in}{0.094700in}}%
\pgfpathlineto{\pgfqpoint{1.166093in}{0.127236in}}%
\pgfpathlineto{\pgfqpoint{1.292011in}{0.094700in}}%
\pgfpathlineto{\pgfqpoint{1.417930in}{0.094700in}}%
\pgfpathlineto{\pgfqpoint{1.543848in}{0.109008in}}%
\pgfpathlineto{\pgfqpoint{1.669766in}{0.163692in}}%
\pgfpathlineto{\pgfqpoint{1.795684in}{0.134527in}}%
\pgfpathlineto{\pgfqpoint{1.921602in}{0.174629in}}%
\pgfpathlineto{\pgfqpoint{2.047521in}{0.192857in}}%
\pgfpathlineto{\pgfqpoint{2.173439in}{0.152755in}}%
\pgfpathlineto{\pgfqpoint{2.299357in}{0.119944in}}%
\pgfpathlineto{\pgfqpoint{2.425275in}{0.101716in}}%
\pgfpathlineto{\pgfqpoint{2.551193in}{0.094700in}}%
\pgfusepath{stroke}%
\end{pgfscope}%
\begin{pgfscope}%
\pgfpathrectangle{\pgfqpoint{0.536502in}{0.041670in}}{\pgfqpoint{6.421828in}{1.166660in}}%
\pgfusepath{clip}%
\pgfsetrectcap%
\pgfsetroundjoin%
\pgfsetlinewidth{1.003750pt}%
\definecolor{currentstroke}{rgb}{0.596078,0.556863,0.835294}%
\pgfsetstrokecolor{currentstroke}%
\pgfsetdash{}{0pt}%
\pgfpathmoveto{\pgfqpoint{3.810375in}{0.094700in}}%
\pgfpathlineto{\pgfqpoint{3.936293in}{0.098071in}}%
\pgfpathlineto{\pgfqpoint{4.062212in}{0.094700in}}%
\pgfpathlineto{\pgfqpoint{4.188130in}{0.109008in}}%
\pgfpathlineto{\pgfqpoint{4.314048in}{0.141818in}}%
\pgfpathlineto{\pgfqpoint{4.439966in}{0.192857in}}%
\pgfpathlineto{\pgfqpoint{4.565884in}{0.262124in}}%
\pgfpathlineto{\pgfqpoint{4.691803in}{0.298580in}}%
\pgfpathlineto{\pgfqpoint{4.817721in}{0.302225in}}%
\pgfpathlineto{\pgfqpoint{4.943639in}{0.324099in}}%
\pgfpathlineto{\pgfqpoint{5.069557in}{0.335036in}}%
\pgfpathlineto{\pgfqpoint{5.195475in}{0.407948in}}%
\pgfpathlineto{\pgfqpoint{5.321394in}{0.433468in}}%
\pgfpathlineto{\pgfqpoint{5.447312in}{0.524608in}}%
\pgfpathlineto{\pgfqpoint{5.573230in}{1.016767in}}%
\pgfpathlineto{\pgfqpoint{5.699148in}{1.155300in}}%
\pgfusepath{stroke}%
\end{pgfscope}%
\begin{pgfscope}%
\pgfpathrectangle{\pgfqpoint{0.536502in}{0.041670in}}{\pgfqpoint{6.421828in}{1.166660in}}%
\pgfusepath{clip}%
\pgfsetrectcap%
\pgfsetroundjoin%
\pgfsetlinewidth{1.003750pt}%
\definecolor{currentstroke}{rgb}{0.466667,0.466667,0.466667}%
\pgfsetstrokecolor{currentstroke}%
\pgfsetdash{}{0pt}%
\pgfpathmoveto{\pgfqpoint{0.662421in}{0.101716in}}%
\pgfpathlineto{\pgfqpoint{0.788339in}{0.098071in}}%
\pgfpathlineto{\pgfqpoint{0.914257in}{0.098071in}}%
\pgfpathlineto{\pgfqpoint{1.040175in}{0.094700in}}%
\pgfpathlineto{\pgfqpoint{1.166093in}{0.138173in}}%
\pgfpathlineto{\pgfqpoint{1.292011in}{0.123590in}}%
\pgfpathlineto{\pgfqpoint{1.417930in}{0.127236in}}%
\pgfpathlineto{\pgfqpoint{1.543848in}{0.134527in}}%
\pgfpathlineto{\pgfqpoint{1.669766in}{0.094700in}}%
\pgfpathlineto{\pgfqpoint{1.795684in}{0.130881in}}%
\pgfpathlineto{\pgfqpoint{1.921602in}{0.094700in}}%
\pgfpathlineto{\pgfqpoint{2.047521in}{0.109008in}}%
\pgfpathlineto{\pgfqpoint{2.173439in}{0.105362in}}%
\pgfpathlineto{\pgfqpoint{2.299357in}{0.094700in}}%
\pgfpathlineto{\pgfqpoint{2.425275in}{0.123590in}}%
\pgfpathlineto{\pgfqpoint{2.551193in}{0.094700in}}%
\pgfpathlineto{\pgfqpoint{2.677112in}{0.094700in}}%
\pgfpathlineto{\pgfqpoint{2.803030in}{0.094700in}}%
\pgfpathlineto{\pgfqpoint{2.928948in}{0.094700in}}%
\pgfpathlineto{\pgfqpoint{3.054866in}{0.094700in}}%
\pgfpathlineto{\pgfqpoint{3.180784in}{0.094700in}}%
\pgfpathlineto{\pgfqpoint{3.306703in}{0.094700in}}%
\pgfpathlineto{\pgfqpoint{3.432621in}{0.094700in}}%
\pgfpathlineto{\pgfqpoint{3.558539in}{0.094700in}}%
\pgfusepath{stroke}%
\end{pgfscope}%
\begin{pgfscope}%
\pgfpathrectangle{\pgfqpoint{0.536502in}{0.041670in}}{\pgfqpoint{6.421828in}{1.166660in}}%
\pgfusepath{clip}%
\pgfsetrectcap%
\pgfsetroundjoin%
\pgfsetlinewidth{1.003750pt}%
\definecolor{currentstroke}{rgb}{0.466667,0.466667,0.466667}%
\pgfsetstrokecolor{currentstroke}%
\pgfsetdash{}{0pt}%
\pgfpathmoveto{\pgfqpoint{3.810375in}{0.101716in}}%
\pgfpathlineto{\pgfqpoint{3.936293in}{0.098071in}}%
\pgfpathlineto{\pgfqpoint{4.062212in}{0.094700in}}%
\pgfpathlineto{\pgfqpoint{4.188130in}{0.094700in}}%
\pgfpathlineto{\pgfqpoint{4.314048in}{0.094700in}}%
\pgfpathlineto{\pgfqpoint{4.439966in}{0.094700in}}%
\pgfpathlineto{\pgfqpoint{4.565884in}{0.149109in}}%
\pgfpathlineto{\pgfqpoint{4.691803in}{0.196502in}}%
\pgfpathlineto{\pgfqpoint{4.817721in}{0.196502in}}%
\pgfpathlineto{\pgfqpoint{4.943639in}{0.152755in}}%
\pgfpathlineto{\pgfqpoint{5.069557in}{0.163692in}}%
\pgfpathlineto{\pgfqpoint{5.195475in}{0.160046in}}%
\pgfpathlineto{\pgfqpoint{5.321394in}{0.203794in}}%
\pgfpathlineto{\pgfqpoint{5.447312in}{0.298580in}}%
\pgfpathlineto{\pgfqpoint{5.573230in}{0.353264in}}%
\pgfpathlineto{\pgfqpoint{5.699148in}{0.386075in}}%
\pgfpathlineto{\pgfqpoint{5.825066in}{0.451696in}}%
\pgfpathlineto{\pgfqpoint{5.950984in}{0.539191in}}%
\pgfpathlineto{\pgfqpoint{6.076903in}{0.539191in}}%
\pgfpathlineto{\pgfqpoint{6.202821in}{0.561064in}}%
\pgfpathlineto{\pgfqpoint{6.328739in}{0.590229in}}%
\pgfpathlineto{\pgfqpoint{6.454657in}{0.604812in}}%
\pgfpathlineto{\pgfqpoint{6.580575in}{0.790738in}}%
\pgfpathlineto{\pgfqpoint{6.706494in}{0.914689in}}%
\pgfusepath{stroke}%
\end{pgfscope}%
\begin{pgfscope}%
\pgfpathrectangle{\pgfqpoint{0.536502in}{0.041670in}}{\pgfqpoint{6.421828in}{1.166660in}}%
\pgfusepath{clip}%
\pgfsetbuttcap%
\pgfsetroundjoin%
\pgfsetlinewidth{1.003750pt}%
\definecolor{currentstroke}{rgb}{0.000000,0.392157,0.000000}%
\pgfsetstrokecolor{currentstroke}%
\pgfsetdash{{3.700000pt}{1.600000pt}}{0.000000pt}%
\pgfpathmoveto{\pgfqpoint{0.536502in}{0.094700in}}%
\pgfpathlineto{\pgfqpoint{6.958330in}{0.094700in}}%
\pgfusepath{stroke}%
\end{pgfscope}%
\begin{pgfscope}%
\pgfpathrectangle{\pgfqpoint{0.536502in}{0.041670in}}{\pgfqpoint{6.421828in}{1.166660in}}%
\pgfusepath{clip}%
\pgfsetbuttcap%
\pgfsetroundjoin%
\pgfsetlinewidth{1.003750pt}%
\definecolor{currentstroke}{rgb}{0.803922,0.521569,0.247059}%
\pgfsetstrokecolor{currentstroke}%
\pgfsetdash{{3.700000pt}{1.600000pt}}{0.000000pt}%
\pgfpathmoveto{\pgfqpoint{0.536502in}{0.794101in}}%
\pgfpathlineto{\pgfqpoint{6.958330in}{0.794101in}}%
\pgfusepath{stroke}%
\end{pgfscope}%
\begin{pgfscope}%
\pgfpathrectangle{\pgfqpoint{0.536502in}{0.041670in}}{\pgfqpoint{6.421828in}{1.166660in}}%
\pgfusepath{clip}%
\pgfsetrectcap%
\pgfsetroundjoin%
\pgfsetlinewidth{6.022500pt}%
\definecolor{currentstroke}{rgb}{1.000000,1.000000,1.000000}%
\pgfsetstrokecolor{currentstroke}%
\pgfsetdash{}{0pt}%
\pgfpathmoveto{\pgfqpoint{3.684457in}{0.041670in}}%
\pgfpathlineto{\pgfqpoint{3.684457in}{1.208330in}}%
\pgfusepath{stroke}%
\end{pgfscope}%
\begin{pgfscope}%
\pgfsetrectcap%
\pgfsetmiterjoin%
\pgfsetlinewidth{1.003750pt}%
\definecolor{currentstroke}{rgb}{1.000000,1.000000,1.000000}%
\pgfsetstrokecolor{currentstroke}%
\pgfsetdash{}{0pt}%
\pgfpathmoveto{\pgfqpoint{0.536502in}{0.041670in}}%
\pgfpathlineto{\pgfqpoint{0.536502in}{1.208330in}}%
\pgfusepath{stroke}%
\end{pgfscope}%
\begin{pgfscope}%
\pgfsetrectcap%
\pgfsetmiterjoin%
\pgfsetlinewidth{1.003750pt}%
\definecolor{currentstroke}{rgb}{1.000000,1.000000,1.000000}%
\pgfsetstrokecolor{currentstroke}%
\pgfsetdash{}{0pt}%
\pgfpathmoveto{\pgfqpoint{6.958330in}{0.041670in}}%
\pgfpathlineto{\pgfqpoint{6.958330in}{1.208330in}}%
\pgfusepath{stroke}%
\end{pgfscope}%
\begin{pgfscope}%
\pgfsetrectcap%
\pgfsetmiterjoin%
\pgfsetlinewidth{1.003750pt}%
\definecolor{currentstroke}{rgb}{1.000000,1.000000,1.000000}%
\pgfsetstrokecolor{currentstroke}%
\pgfsetdash{}{0pt}%
\pgfpathmoveto{\pgfqpoint{0.536502in}{0.041670in}}%
\pgfpathlineto{\pgfqpoint{6.958330in}{0.041670in}}%
\pgfusepath{stroke}%
\end{pgfscope}%
\begin{pgfscope}%
\pgfsetrectcap%
\pgfsetmiterjoin%
\pgfsetlinewidth{1.003750pt}%
\definecolor{currentstroke}{rgb}{1.000000,1.000000,1.000000}%
\pgfsetstrokecolor{currentstroke}%
\pgfsetdash{}{0pt}%
\pgfpathmoveto{\pgfqpoint{0.536502in}{1.208330in}}%
\pgfpathlineto{\pgfqpoint{6.958330in}{1.208330in}}%
\pgfusepath{stroke}%
\end{pgfscope}%
\end{pgfpicture}%
\makeatother%
\endgroup%

%% file: figures/comparison-wristtwist-nox.pgf
\begingroup%
\makeatletter%
\begin{pgfpicture}%
\pgfpathrectangle{\pgfpointorigin}{\pgfqpoint{7.000000in}{1.250000in}}%
\pgfusepath{use as bounding box, clip}%
\begin{pgfscope}%
\pgfsetbuttcap%
\pgfsetmiterjoin%
\definecolor{currentfill}{rgb}{1.000000,1.000000,1.000000}%
\pgfsetfillcolor{currentfill}%
\pgfsetlinewidth{0.000000pt}%
\definecolor{currentstroke}{rgb}{0.500000,0.500000,0.500000}%
\pgfsetstrokecolor{currentstroke}%
\pgfsetdash{}{0pt}%
\pgfpathmoveto{\pgfqpoint{0.000000in}{0.000000in}}%
\pgfpathlineto{\pgfqpoint{7.000000in}{0.000000in}}%
\pgfpathlineto{\pgfqpoint{7.000000in}{1.250000in}}%
\pgfpathlineto{\pgfqpoint{0.000000in}{1.250000in}}%
\pgfpathlineto{\pgfqpoint{0.000000in}{0.000000in}}%
\pgfpathclose%
\pgfusepath{fill}%
\end{pgfscope}%
\begin{pgfscope}%
\pgfsetbuttcap%
\pgfsetmiterjoin%
\definecolor{currentfill}{rgb}{0.898039,0.898039,0.898039}%
\pgfsetfillcolor{currentfill}%
\pgfsetlinewidth{0.000000pt}%
\definecolor{currentstroke}{rgb}{0.000000,0.000000,0.000000}%
\pgfsetstrokecolor{currentstroke}%
\pgfsetstrokeopacity{0.000000}%
\pgfsetdash{}{0pt}%
\pgfpathmoveto{\pgfqpoint{0.536502in}{0.041670in}}%
\pgfpathlineto{\pgfqpoint{6.958330in}{0.041670in}}%
\pgfpathlineto{\pgfqpoint{6.958330in}{1.208330in}}%
\pgfpathlineto{\pgfqpoint{0.536502in}{1.208330in}}%
\pgfpathlineto{\pgfqpoint{0.536502in}{0.041670in}}%
\pgfpathclose%
\pgfusepath{fill}%
\end{pgfscope}%
\begin{pgfscope}%
\pgfpathrectangle{\pgfqpoint{0.536502in}{0.041670in}}{\pgfqpoint{6.421828in}{1.166660in}}%
\pgfusepath{clip}%
\pgfsetrectcap%
\pgfsetroundjoin%
\pgfsetlinewidth{0.803000pt}%
\definecolor{currentstroke}{rgb}{1.000000,1.000000,1.000000}%
\pgfsetstrokecolor{currentstroke}%
\pgfsetdash{}{0pt}%
\pgfpathmoveto{\pgfqpoint{0.662421in}{0.041670in}}%
\pgfpathlineto{\pgfqpoint{0.662421in}{1.208330in}}%
\pgfusepath{stroke}%
\end{pgfscope}%
\begin{pgfscope}%
\pgfpathrectangle{\pgfqpoint{0.536502in}{0.041670in}}{\pgfqpoint{6.421828in}{1.166660in}}%
\pgfusepath{clip}%
\pgfsetrectcap%
\pgfsetroundjoin%
\pgfsetlinewidth{0.803000pt}%
\definecolor{currentstroke}{rgb}{1.000000,1.000000,1.000000}%
\pgfsetstrokecolor{currentstroke}%
\pgfsetdash{}{0pt}%
\pgfpathmoveto{\pgfqpoint{3.558539in}{0.041670in}}%
\pgfpathlineto{\pgfqpoint{3.558539in}{1.208330in}}%
\pgfusepath{stroke}%
\end{pgfscope}%
\begin{pgfscope}%
\pgfpathrectangle{\pgfqpoint{0.536502in}{0.041670in}}{\pgfqpoint{6.421828in}{1.166660in}}%
\pgfusepath{clip}%
\pgfsetrectcap%
\pgfsetroundjoin%
\pgfsetlinewidth{0.803000pt}%
\definecolor{currentstroke}{rgb}{1.000000,1.000000,1.000000}%
\pgfsetstrokecolor{currentstroke}%
\pgfsetdash{}{0pt}%
\pgfpathmoveto{\pgfqpoint{3.810375in}{0.041670in}}%
\pgfpathlineto{\pgfqpoint{3.810375in}{1.208330in}}%
\pgfusepath{stroke}%
\end{pgfscope}%
\begin{pgfscope}%
\pgfpathrectangle{\pgfqpoint{0.536502in}{0.041670in}}{\pgfqpoint{6.421828in}{1.166660in}}%
\pgfusepath{clip}%
\pgfsetrectcap%
\pgfsetroundjoin%
\pgfsetlinewidth{0.803000pt}%
\definecolor{currentstroke}{rgb}{1.000000,1.000000,1.000000}%
\pgfsetstrokecolor{currentstroke}%
\pgfsetdash{}{0pt}%
\pgfpathmoveto{\pgfqpoint{6.832412in}{0.041670in}}%
\pgfpathlineto{\pgfqpoint{6.832412in}{1.208330in}}%
\pgfusepath{stroke}%
\end{pgfscope}%
\begin{pgfscope}%
\pgfpathrectangle{\pgfqpoint{0.536502in}{0.041670in}}{\pgfqpoint{6.421828in}{1.166660in}}%
\pgfusepath{clip}%
\pgfsetrectcap%
\pgfsetroundjoin%
\pgfsetlinewidth{0.803000pt}%
\definecolor{currentstroke}{rgb}{1.000000,1.000000,1.000000}%
\pgfsetstrokecolor{currentstroke}%
\pgfsetdash{}{0pt}%
\pgfpathmoveto{\pgfqpoint{0.536502in}{0.362116in}}%
\pgfpathlineto{\pgfqpoint{6.958330in}{0.362116in}}%
\pgfusepath{stroke}%
\end{pgfscope}%
\begin{pgfscope}%
\definecolor{textcolor}{rgb}{0.333333,0.333333,0.333333}%
\pgfsetstrokecolor{textcolor}%
\pgfsetfillcolor{textcolor}%
\pgftext[x=0.041670in, y=0.309355in, left, base]{\color{textcolor}\sffamily\fontsize{10.000000}{12.000000}\selectfont 0.840}%
\end{pgfscope}%
\begin{pgfscope}%
\pgfpathrectangle{\pgfqpoint{0.536502in}{0.041670in}}{\pgfqpoint{6.421828in}{1.166660in}}%
\pgfusepath{clip}%
\pgfsetrectcap%
\pgfsetroundjoin%
\pgfsetlinewidth{0.803000pt}%
\definecolor{currentstroke}{rgb}{1.000000,1.000000,1.000000}%
\pgfsetstrokecolor{currentstroke}%
\pgfsetdash{}{0pt}%
\pgfpathmoveto{\pgfqpoint{0.536502in}{0.815364in}}%
\pgfpathlineto{\pgfqpoint{6.958330in}{0.815364in}}%
\pgfusepath{stroke}%
\end{pgfscope}%
\begin{pgfscope}%
\definecolor{textcolor}{rgb}{0.333333,0.333333,0.333333}%
\pgfsetstrokecolor{textcolor}%
\pgfsetfillcolor{textcolor}%
\pgftext[x=0.041670in, y=0.762603in, left, base]{\color{textcolor}\sffamily\fontsize{10.000000}{12.000000}\selectfont 0.860}%
\end{pgfscope}%
\begin{pgfscope}%
\pgfpathrectangle{\pgfqpoint{0.536502in}{0.041670in}}{\pgfqpoint{6.421828in}{1.166660in}}%
\pgfusepath{clip}%
\pgfsetrectcap%
\pgfsetroundjoin%
\pgfsetlinewidth{1.003750pt}%
\definecolor{currentstroke}{rgb}{0.886275,0.290196,0.200000}%
\pgfsetstrokecolor{currentstroke}%
\pgfsetdash{}{0pt}%
\pgfpathmoveto{\pgfqpoint{0.662421in}{0.094816in}}%
\pgfpathlineto{\pgfqpoint{0.788339in}{0.094816in}}%
\pgfpathlineto{\pgfqpoint{0.914257in}{0.094816in}}%
\pgfpathlineto{\pgfqpoint{1.040175in}{0.094816in}}%
\pgfpathlineto{\pgfqpoint{1.166093in}{0.094816in}}%
\pgfpathlineto{\pgfqpoint{1.292011in}{0.094816in}}%
\pgfpathlineto{\pgfqpoint{1.417930in}{0.094816in}}%
\pgfpathlineto{\pgfqpoint{1.543848in}{0.094816in}}%
\pgfpathlineto{\pgfqpoint{1.669766in}{0.094816in}}%
\pgfpathlineto{\pgfqpoint{1.795684in}{0.094816in}}%
\pgfpathlineto{\pgfqpoint{1.921602in}{0.094816in}}%
\pgfpathlineto{\pgfqpoint{2.047521in}{0.094816in}}%
\pgfusepath{stroke}%
\end{pgfscope}%
\begin{pgfscope}%
\pgfpathrectangle{\pgfqpoint{0.536502in}{0.041670in}}{\pgfqpoint{6.421828in}{1.166660in}}%
\pgfusepath{clip}%
\pgfsetrectcap%
\pgfsetroundjoin%
\pgfsetlinewidth{1.003750pt}%
\definecolor{currentstroke}{rgb}{0.886275,0.290196,0.200000}%
\pgfsetstrokecolor{currentstroke}%
\pgfsetdash{}{0pt}%
\pgfpathmoveto{\pgfqpoint{3.810375in}{0.094816in}}%
\pgfpathlineto{\pgfqpoint{3.936293in}{0.094816in}}%
\pgfpathlineto{\pgfqpoint{4.062212in}{0.094816in}}%
\pgfpathlineto{\pgfqpoint{4.188130in}{0.094816in}}%
\pgfpathlineto{\pgfqpoint{4.314048in}{0.094816in}}%
\pgfpathlineto{\pgfqpoint{4.439966in}{0.094816in}}%
\pgfpathlineto{\pgfqpoint{4.565884in}{0.094816in}}%
\pgfpathlineto{\pgfqpoint{4.691803in}{0.094816in}}%
\pgfpathlineto{\pgfqpoint{4.817721in}{0.094816in}}%
\pgfpathlineto{\pgfqpoint{4.943639in}{0.094816in}}%
\pgfpathlineto{\pgfqpoint{5.069557in}{0.094816in}}%
\pgfpathlineto{\pgfqpoint{5.195475in}{0.094816in}}%
\pgfusepath{stroke}%
\end{pgfscope}%
\begin{pgfscope}%
\pgfpathrectangle{\pgfqpoint{0.536502in}{0.041670in}}{\pgfqpoint{6.421828in}{1.166660in}}%
\pgfusepath{clip}%
\pgfsetrectcap%
\pgfsetroundjoin%
\pgfsetlinewidth{1.003750pt}%
\definecolor{currentstroke}{rgb}{0.203922,0.541176,0.741176}%
\pgfsetstrokecolor{currentstroke}%
\pgfsetdash{}{0pt}%
\pgfpathmoveto{\pgfqpoint{0.662421in}{0.094700in}}%
\pgfpathlineto{\pgfqpoint{0.788339in}{0.094700in}}%
\pgfpathlineto{\pgfqpoint{0.914257in}{0.094700in}}%
\pgfpathlineto{\pgfqpoint{1.040175in}{0.094816in}}%
\pgfpathlineto{\pgfqpoint{1.166093in}{0.094816in}}%
\pgfpathlineto{\pgfqpoint{1.292011in}{0.094816in}}%
\pgfpathlineto{\pgfqpoint{1.417930in}{0.109343in}}%
\pgfpathlineto{\pgfqpoint{1.543848in}{0.123871in}}%
\pgfpathlineto{\pgfqpoint{1.669766in}{0.138398in}}%
\pgfpathlineto{\pgfqpoint{1.795684in}{0.123871in}}%
\pgfpathlineto{\pgfqpoint{1.921602in}{0.094816in}}%
\pgfpathlineto{\pgfqpoint{2.047521in}{0.094816in}}%
\pgfusepath{stroke}%
\end{pgfscope}%
\begin{pgfscope}%
\pgfpathrectangle{\pgfqpoint{0.536502in}{0.041670in}}{\pgfqpoint{6.421828in}{1.166660in}}%
\pgfusepath{clip}%
\pgfsetrectcap%
\pgfsetroundjoin%
\pgfsetlinewidth{1.003750pt}%
\definecolor{currentstroke}{rgb}{0.203922,0.541176,0.741176}%
\pgfsetstrokecolor{currentstroke}%
\pgfsetdash{}{0pt}%
\pgfpathmoveto{\pgfqpoint{3.810375in}{0.094816in}}%
\pgfpathlineto{\pgfqpoint{3.936293in}{0.094816in}}%
\pgfpathlineto{\pgfqpoint{4.062212in}{0.094816in}}%
\pgfpathlineto{\pgfqpoint{4.188130in}{0.094816in}}%
\pgfpathlineto{\pgfqpoint{4.314048in}{0.094816in}}%
\pgfpathlineto{\pgfqpoint{4.439966in}{0.094816in}}%
\pgfpathlineto{\pgfqpoint{4.565884in}{0.094700in}}%
\pgfpathlineto{\pgfqpoint{4.691803in}{0.094700in}}%
\pgfpathlineto{\pgfqpoint{4.817721in}{0.094816in}}%
\pgfpathlineto{\pgfqpoint{4.943639in}{0.094700in}}%
\pgfpathlineto{\pgfqpoint{5.069557in}{0.094816in}}%
\pgfpathlineto{\pgfqpoint{5.195475in}{0.094816in}}%
\pgfusepath{stroke}%
\end{pgfscope}%
\begin{pgfscope}%
\pgfpathrectangle{\pgfqpoint{0.536502in}{0.041670in}}{\pgfqpoint{6.421828in}{1.166660in}}%
\pgfusepath{clip}%
\pgfsetrectcap%
\pgfsetroundjoin%
\pgfsetlinewidth{1.003750pt}%
\definecolor{currentstroke}{rgb}{0.596078,0.556863,0.835294}%
\pgfsetstrokecolor{currentstroke}%
\pgfsetdash{}{0pt}%
\pgfpathmoveto{\pgfqpoint{0.662421in}{0.094816in}}%
\pgfpathlineto{\pgfqpoint{0.788339in}{0.094816in}}%
\pgfpathlineto{\pgfqpoint{0.914257in}{0.094816in}}%
\pgfpathlineto{\pgfqpoint{1.040175in}{0.094700in}}%
\pgfpathlineto{\pgfqpoint{1.166093in}{0.094700in}}%
\pgfpathlineto{\pgfqpoint{1.292011in}{0.094816in}}%
\pgfpathlineto{\pgfqpoint{1.417930in}{0.094700in}}%
\pgfpathlineto{\pgfqpoint{1.543848in}{0.094700in}}%
\pgfpathlineto{\pgfqpoint{1.669766in}{0.094700in}}%
\pgfpathlineto{\pgfqpoint{1.795684in}{0.109343in}}%
\pgfpathlineto{\pgfqpoint{1.921602in}{0.094700in}}%
\pgfpathlineto{\pgfqpoint{2.047521in}{0.094700in}}%
\pgfpathlineto{\pgfqpoint{2.173439in}{0.094700in}}%
\pgfpathlineto{\pgfqpoint{2.299357in}{0.109343in}}%
\pgfpathlineto{\pgfqpoint{2.425275in}{0.094816in}}%
\pgfpathlineto{\pgfqpoint{2.551193in}{0.109343in}}%
\pgfusepath{stroke}%
\end{pgfscope}%
\begin{pgfscope}%
\pgfpathrectangle{\pgfqpoint{0.536502in}{0.041670in}}{\pgfqpoint{6.421828in}{1.166660in}}%
\pgfusepath{clip}%
\pgfsetrectcap%
\pgfsetroundjoin%
\pgfsetlinewidth{1.003750pt}%
\definecolor{currentstroke}{rgb}{0.596078,0.556863,0.835294}%
\pgfsetstrokecolor{currentstroke}%
\pgfsetdash{}{0pt}%
\pgfpathmoveto{\pgfqpoint{3.810375in}{0.094816in}}%
\pgfpathlineto{\pgfqpoint{3.936293in}{0.094816in}}%
\pgfpathlineto{\pgfqpoint{4.062212in}{0.094816in}}%
\pgfpathlineto{\pgfqpoint{4.188130in}{0.094816in}}%
\pgfpathlineto{\pgfqpoint{4.314048in}{0.094700in}}%
\pgfpathlineto{\pgfqpoint{4.439966in}{0.109343in}}%
\pgfpathlineto{\pgfqpoint{4.565884in}{0.094816in}}%
\pgfpathlineto{\pgfqpoint{4.691803in}{0.094700in}}%
\pgfpathlineto{\pgfqpoint{4.817721in}{0.123871in}}%
\pgfpathlineto{\pgfqpoint{4.943639in}{0.094700in}}%
\pgfpathlineto{\pgfqpoint{5.069557in}{0.094700in}}%
\pgfpathlineto{\pgfqpoint{5.195475in}{0.094700in}}%
\pgfpathlineto{\pgfqpoint{5.321394in}{0.152925in}}%
\pgfpathlineto{\pgfqpoint{5.447312in}{0.167452in}}%
\pgfpathlineto{\pgfqpoint{5.573230in}{1.053610in}}%
\pgfpathlineto{\pgfqpoint{5.699148in}{1.010028in}}%
\pgfusepath{stroke}%
\end{pgfscope}%
\begin{pgfscope}%
\pgfpathrectangle{\pgfqpoint{0.536502in}{0.041670in}}{\pgfqpoint{6.421828in}{1.166660in}}%
\pgfusepath{clip}%
\pgfsetrectcap%
\pgfsetroundjoin%
\pgfsetlinewidth{1.003750pt}%
\definecolor{currentstroke}{rgb}{0.466667,0.466667,0.466667}%
\pgfsetstrokecolor{currentstroke}%
\pgfsetdash{}{0pt}%
\pgfpathmoveto{\pgfqpoint{0.662421in}{0.094816in}}%
\pgfpathlineto{\pgfqpoint{0.788339in}{0.094816in}}%
\pgfpathlineto{\pgfqpoint{0.914257in}{0.094816in}}%
\pgfpathlineto{\pgfqpoint{1.040175in}{0.094816in}}%
\pgfpathlineto{\pgfqpoint{1.166093in}{0.094816in}}%
\pgfpathlineto{\pgfqpoint{1.292011in}{0.094700in}}%
\pgfpathlineto{\pgfqpoint{1.417930in}{0.094816in}}%
\pgfpathlineto{\pgfqpoint{1.543848in}{0.094700in}}%
\pgfpathlineto{\pgfqpoint{1.669766in}{0.094700in}}%
\pgfpathlineto{\pgfqpoint{1.795684in}{0.094700in}}%
\pgfpathlineto{\pgfqpoint{1.921602in}{0.094700in}}%
\pgfpathlineto{\pgfqpoint{2.047521in}{0.094700in}}%
\pgfpathlineto{\pgfqpoint{2.173439in}{0.094700in}}%
\pgfpathlineto{\pgfqpoint{2.299357in}{0.094700in}}%
\pgfpathlineto{\pgfqpoint{2.425275in}{0.094700in}}%
\pgfpathlineto{\pgfqpoint{2.551193in}{0.094700in}}%
\pgfpathlineto{\pgfqpoint{2.677112in}{0.094816in}}%
\pgfpathlineto{\pgfqpoint{2.803030in}{0.109343in}}%
\pgfpathlineto{\pgfqpoint{2.928948in}{0.094816in}}%
\pgfpathlineto{\pgfqpoint{3.054866in}{0.094816in}}%
\pgfpathlineto{\pgfqpoint{3.180784in}{0.094816in}}%
\pgfpathlineto{\pgfqpoint{3.306703in}{0.094700in}}%
\pgfpathlineto{\pgfqpoint{3.432621in}{0.094816in}}%
\pgfpathlineto{\pgfqpoint{3.558539in}{0.094816in}}%
\pgfusepath{stroke}%
\end{pgfscope}%
\begin{pgfscope}%
\pgfpathrectangle{\pgfqpoint{0.536502in}{0.041670in}}{\pgfqpoint{6.421828in}{1.166660in}}%
\pgfusepath{clip}%
\pgfsetrectcap%
\pgfsetroundjoin%
\pgfsetlinewidth{1.003750pt}%
\definecolor{currentstroke}{rgb}{0.466667,0.466667,0.466667}%
\pgfsetstrokecolor{currentstroke}%
\pgfsetdash{}{0pt}%
\pgfpathmoveto{\pgfqpoint{3.810375in}{0.094816in}}%
\pgfpathlineto{\pgfqpoint{3.936293in}{0.094816in}}%
\pgfpathlineto{\pgfqpoint{4.062212in}{0.094816in}}%
\pgfpathlineto{\pgfqpoint{4.188130in}{0.094816in}}%
\pgfpathlineto{\pgfqpoint{4.314048in}{0.094700in}}%
\pgfpathlineto{\pgfqpoint{4.439966in}{0.094700in}}%
\pgfpathlineto{\pgfqpoint{4.565884in}{0.094700in}}%
\pgfpathlineto{\pgfqpoint{4.691803in}{0.094700in}}%
\pgfpathlineto{\pgfqpoint{4.817721in}{0.094700in}}%
\pgfpathlineto{\pgfqpoint{4.943639in}{0.094700in}}%
\pgfpathlineto{\pgfqpoint{5.069557in}{0.094700in}}%
\pgfpathlineto{\pgfqpoint{5.195475in}{0.094700in}}%
\pgfpathlineto{\pgfqpoint{5.321394in}{0.094700in}}%
\pgfpathlineto{\pgfqpoint{5.447312in}{0.094700in}}%
\pgfpathlineto{\pgfqpoint{5.573230in}{0.138398in}}%
\pgfpathlineto{\pgfqpoint{5.699148in}{0.283669in}}%
\pgfpathlineto{\pgfqpoint{5.825066in}{0.457996in}}%
\pgfpathlineto{\pgfqpoint{5.950984in}{0.399887in}}%
\pgfpathlineto{\pgfqpoint{6.076903in}{0.734012in}}%
\pgfpathlineto{\pgfqpoint{6.202821in}{0.821175in}}%
\pgfpathlineto{\pgfqpoint{6.328739in}{0.574213in}}%
\pgfpathlineto{\pgfqpoint{6.454657in}{0.661376in}}%
\pgfpathlineto{\pgfqpoint{6.580575in}{1.155300in}}%
\pgfpathlineto{\pgfqpoint{6.706494in}{1.140773in}}%
\pgfusepath{stroke}%
\end{pgfscope}%
\begin{pgfscope}%
\pgfpathrectangle{\pgfqpoint{0.536502in}{0.041670in}}{\pgfqpoint{6.421828in}{1.166660in}}%
\pgfusepath{clip}%
\pgfsetbuttcap%
\pgfsetroundjoin%
\pgfsetlinewidth{1.003750pt}%
\definecolor{currentstroke}{rgb}{0.000000,0.392157,0.000000}%
\pgfsetstrokecolor{currentstroke}%
\pgfsetdash{{3.700000pt}{1.600000pt}}{0.000000pt}%
\pgfpathmoveto{\pgfqpoint{0.536502in}{0.094700in}}%
\pgfpathlineto{\pgfqpoint{6.958330in}{0.094700in}}%
\pgfusepath{stroke}%
\end{pgfscope}%
\begin{pgfscope}%
\pgfpathrectangle{\pgfqpoint{0.536502in}{0.041670in}}{\pgfqpoint{6.421828in}{1.166660in}}%
\pgfusepath{clip}%
\pgfsetbuttcap%
\pgfsetroundjoin%
\pgfsetlinewidth{1.003750pt}%
\definecolor{currentstroke}{rgb}{0.803922,0.521569,0.247059}%
\pgfsetstrokecolor{currentstroke}%
\pgfsetdash{{3.700000pt}{1.600000pt}}{0.000000pt}%
\pgfpathmoveto{\pgfqpoint{0.536502in}{0.661260in}}%
\pgfpathlineto{\pgfqpoint{6.958330in}{0.661260in}}%
\pgfusepath{stroke}%
\end{pgfscope}%
\begin{pgfscope}%
\pgfpathrectangle{\pgfqpoint{0.536502in}{0.041670in}}{\pgfqpoint{6.421828in}{1.166660in}}%
\pgfusepath{clip}%
\pgfsetrectcap%
\pgfsetroundjoin%
\pgfsetlinewidth{6.022500pt}%
\definecolor{currentstroke}{rgb}{1.000000,1.000000,1.000000}%
\pgfsetstrokecolor{currentstroke}%
\pgfsetdash{}{0pt}%
\pgfpathmoveto{\pgfqpoint{3.684457in}{0.041670in}}%
\pgfpathlineto{\pgfqpoint{3.684457in}{1.208330in}}%
\pgfusepath{stroke}%
\end{pgfscope}%
\begin{pgfscope}%
\pgfsetrectcap%
\pgfsetmiterjoin%
\pgfsetlinewidth{1.003750pt}%
\definecolor{currentstroke}{rgb}{1.000000,1.000000,1.000000}%
\pgfsetstrokecolor{currentstroke}%
\pgfsetdash{}{0pt}%
\pgfpathmoveto{\pgfqpoint{0.536502in}{0.041670in}}%
\pgfpathlineto{\pgfqpoint{0.536502in}{1.208330in}}%
\pgfusepath{stroke}%
\end{pgfscope}%
\begin{pgfscope}%
\pgfsetrectcap%
\pgfsetmiterjoin%
\pgfsetlinewidth{1.003750pt}%
\definecolor{currentstroke}{rgb}{1.000000,1.000000,1.000000}%
\pgfsetstrokecolor{currentstroke}%
\pgfsetdash{}{0pt}%
\pgfpathmoveto{\pgfqpoint{6.958330in}{0.041670in}}%
\pgfpathlineto{\pgfqpoint{6.958330in}{1.208330in}}%
\pgfusepath{stroke}%
\end{pgfscope}%
\begin{pgfscope}%
\pgfsetrectcap%
\pgfsetmiterjoin%
\pgfsetlinewidth{1.003750pt}%
\definecolor{currentstroke}{rgb}{1.000000,1.000000,1.000000}%
\pgfsetstrokecolor{currentstroke}%
\pgfsetdash{}{0pt}%
\pgfpathmoveto{\pgfqpoint{0.536502in}{0.041670in}}%
\pgfpathlineto{\pgfqpoint{6.958330in}{0.041670in}}%
\pgfusepath{stroke}%
\end{pgfscope}%
\begin{pgfscope}%
\pgfsetrectcap%
\pgfsetmiterjoin%
\pgfsetlinewidth{1.003750pt}%
\definecolor{currentstroke}{rgb}{1.000000,1.000000,1.000000}%
\pgfsetstrokecolor{currentstroke}%
\pgfsetdash{}{0pt}%
\pgfpathmoveto{\pgfqpoint{0.536502in}{1.208330in}}%
\pgfpathlineto{\pgfqpoint{6.958330in}{1.208330in}}%
\pgfusepath{stroke}%
\end{pgfscope}%
\end{pgfpicture}%
\makeatother%
\endgroup%

%% file: figures/comparison-repeatedmovement.pgf
\begingroup%
\makeatletter%
\begin{pgfpicture}%
\pgfpathrectangle{\pgfpointorigin}{\pgfqpoint{7.000000in}{1.550000in}}%
\pgfusepath{use as bounding box, clip}%
\begin{pgfscope}%
\pgfsetbuttcap%
\pgfsetmiterjoin%
\definecolor{currentfill}{rgb}{1.000000,1.000000,1.000000}%
\pgfsetfillcolor{currentfill}%
\pgfsetlinewidth{0.000000pt}%
\definecolor{currentstroke}{rgb}{0.500000,0.500000,0.500000}%
\pgfsetstrokecolor{currentstroke}%
\pgfsetdash{}{0pt}%
\pgfpathmoveto{\pgfqpoint{0.000000in}{0.000000in}}%
\pgfpathlineto{\pgfqpoint{7.000000in}{0.000000in}}%
\pgfpathlineto{\pgfqpoint{7.000000in}{1.550000in}}%
\pgfpathlineto{\pgfqpoint{0.000000in}{1.550000in}}%
\pgfpathlineto{\pgfqpoint{0.000000in}{0.000000in}}%
\pgfpathclose%
\pgfusepath{fill}%
\end{pgfscope}%
\begin{pgfscope}%
\pgfsetbuttcap%
\pgfsetmiterjoin%
\definecolor{currentfill}{rgb}{0.898039,0.898039,0.898039}%
\pgfsetfillcolor{currentfill}%
\pgfsetlinewidth{0.000000pt}%
\definecolor{currentstroke}{rgb}{0.000000,0.000000,0.000000}%
\pgfsetstrokecolor{currentstroke}%
\pgfsetstrokeopacity{0.000000}%
\pgfsetdash{}{0pt}%
\pgfpathmoveto{\pgfqpoint{0.536502in}{0.394792in}}%
\pgfpathlineto{\pgfqpoint{6.958330in}{0.394792in}}%
\pgfpathlineto{\pgfqpoint{6.958330in}{1.508330in}}%
\pgfpathlineto{\pgfqpoint{0.536502in}{1.508330in}}%
\pgfpathlineto{\pgfqpoint{0.536502in}{0.394792in}}%
\pgfpathclose%
\pgfusepath{fill}%
\end{pgfscope}%
\begin{pgfscope}%
\pgfpathrectangle{\pgfqpoint{0.536502in}{0.394792in}}{\pgfqpoint{6.421828in}{1.113538in}}%
\pgfusepath{clip}%
\pgfsetrectcap%
\pgfsetroundjoin%
\pgfsetlinewidth{0.803000pt}%
\definecolor{currentstroke}{rgb}{1.000000,1.000000,1.000000}%
\pgfsetstrokecolor{currentstroke}%
\pgfsetdash{}{0pt}%
\pgfpathmoveto{\pgfqpoint{0.662421in}{0.394792in}}%
\pgfpathlineto{\pgfqpoint{0.662421in}{1.508330in}}%
\pgfusepath{stroke}%
\end{pgfscope}%
\begin{pgfscope}%
\definecolor{textcolor}{rgb}{0.333333,0.333333,0.333333}%
\pgfsetstrokecolor{textcolor}%
\pgfsetfillcolor{textcolor}%
\pgftext[x=0.662421in,y=0.297570in,,top]{\color{textcolor}\sffamily\fontsize{10.000000}{12.000000}\selectfont 1}%
\end{pgfscope}%
\begin{pgfscope}%
\pgfpathrectangle{\pgfqpoint{0.536502in}{0.394792in}}{\pgfqpoint{6.421828in}{1.113538in}}%
\pgfusepath{clip}%
\pgfsetrectcap%
\pgfsetroundjoin%
\pgfsetlinewidth{0.803000pt}%
\definecolor{currentstroke}{rgb}{1.000000,1.000000,1.000000}%
\pgfsetstrokecolor{currentstroke}%
\pgfsetdash{}{0pt}%
\pgfpathmoveto{\pgfqpoint{3.558539in}{0.394792in}}%
\pgfpathlineto{\pgfqpoint{3.558539in}{1.508330in}}%
\pgfusepath{stroke}%
\end{pgfscope}%
\begin{pgfscope}%
\definecolor{textcolor}{rgb}{0.333333,0.333333,0.333333}%
\pgfsetstrokecolor{textcolor}%
\pgfsetfillcolor{textcolor}%
\pgftext[x=3.558539in,y=0.297570in,,top]{\color{textcolor}\sffamily\fontsize{10.000000}{12.000000}\selectfont 24}%
\end{pgfscope}%
\begin{pgfscope}%
\pgfpathrectangle{\pgfqpoint{0.536502in}{0.394792in}}{\pgfqpoint{6.421828in}{1.113538in}}%
\pgfusepath{clip}%
\pgfsetrectcap%
\pgfsetroundjoin%
\pgfsetlinewidth{0.803000pt}%
\definecolor{currentstroke}{rgb}{1.000000,1.000000,1.000000}%
\pgfsetstrokecolor{currentstroke}%
\pgfsetdash{}{0pt}%
\pgfpathmoveto{\pgfqpoint{3.810375in}{0.394792in}}%
\pgfpathlineto{\pgfqpoint{3.810375in}{1.508330in}}%
\pgfusepath{stroke}%
\end{pgfscope}%
\begin{pgfscope}%
\definecolor{textcolor}{rgb}{0.333333,0.333333,0.333333}%
\pgfsetstrokecolor{textcolor}%
\pgfsetfillcolor{textcolor}%
\pgftext[x=3.810375in,y=0.297570in,,top]{\color{textcolor}\sffamily\fontsize{10.000000}{12.000000}\selectfont 1}%
\end{pgfscope}%
\begin{pgfscope}%
\pgfpathrectangle{\pgfqpoint{0.536502in}{0.394792in}}{\pgfqpoint{6.421828in}{1.113538in}}%
\pgfusepath{clip}%
\pgfsetrectcap%
\pgfsetroundjoin%
\pgfsetlinewidth{0.803000pt}%
\definecolor{currentstroke}{rgb}{1.000000,1.000000,1.000000}%
\pgfsetstrokecolor{currentstroke}%
\pgfsetdash{}{0pt}%
\pgfpathmoveto{\pgfqpoint{6.832412in}{0.394792in}}%
\pgfpathlineto{\pgfqpoint{6.832412in}{1.508330in}}%
\pgfusepath{stroke}%
\end{pgfscope}%
\begin{pgfscope}%
\definecolor{textcolor}{rgb}{0.333333,0.333333,0.333333}%
\pgfsetstrokecolor{textcolor}%
\pgfsetfillcolor{textcolor}%
\pgftext[x=6.832412in,y=0.297570in,,top]{\color{textcolor}\sffamily\fontsize{10.000000}{12.000000}\selectfont 24}%
\end{pgfscope}%
\begin{pgfscope}%
\definecolor{textcolor}{rgb}{0.333333,0.333333,0.333333}%
\pgfsetstrokecolor{textcolor}%
\pgfsetfillcolor{textcolor}%
\pgftext[x=3.651089in,y=0.172085in,,top]{\color{textcolor}\sffamily\fontsize{10.000000}{12.000000}\selectfont Layer}%
\end{pgfscope}%
\begin{pgfscope}%
\pgfpathrectangle{\pgfqpoint{0.536502in}{0.394792in}}{\pgfqpoint{6.421828in}{1.113538in}}%
\pgfusepath{clip}%
\pgfsetrectcap%
\pgfsetroundjoin%
\pgfsetlinewidth{0.803000pt}%
\definecolor{currentstroke}{rgb}{1.000000,1.000000,1.000000}%
\pgfsetstrokecolor{currentstroke}%
\pgfsetdash{}{0pt}%
\pgfpathmoveto{\pgfqpoint{0.536502in}{0.711917in}}%
\pgfpathlineto{\pgfqpoint{6.958330in}{0.711917in}}%
\pgfusepath{stroke}%
\end{pgfscope}%
\begin{pgfscope}%
\definecolor{textcolor}{rgb}{0.333333,0.333333,0.333333}%
\pgfsetstrokecolor{textcolor}%
\pgfsetfillcolor{textcolor}%
\pgftext[x=0.041670in, y=0.659156in, left, base]{\color{textcolor}\sffamily\fontsize{10.000000}{12.000000}\selectfont 0.600}%
\end{pgfscope}%
\begin{pgfscope}%
\pgfpathrectangle{\pgfqpoint{0.536502in}{0.394792in}}{\pgfqpoint{6.421828in}{1.113538in}}%
\pgfusepath{clip}%
\pgfsetrectcap%
\pgfsetroundjoin%
\pgfsetlinewidth{0.803000pt}%
\definecolor{currentstroke}{rgb}{1.000000,1.000000,1.000000}%
\pgfsetstrokecolor{currentstroke}%
\pgfsetdash{}{0pt}%
\pgfpathmoveto{\pgfqpoint{0.536502in}{1.050128in}}%
\pgfpathlineto{\pgfqpoint{6.958330in}{1.050128in}}%
\pgfusepath{stroke}%
\end{pgfscope}%
\begin{pgfscope}%
\definecolor{textcolor}{rgb}{0.333333,0.333333,0.333333}%
\pgfsetstrokecolor{textcolor}%
\pgfsetfillcolor{textcolor}%
\pgftext[x=0.041670in, y=0.997366in, left, base]{\color{textcolor}\sffamily\fontsize{10.000000}{12.000000}\selectfont 0.700}%
\end{pgfscope}%
\begin{pgfscope}%
\pgfpathrectangle{\pgfqpoint{0.536502in}{0.394792in}}{\pgfqpoint{6.421828in}{1.113538in}}%
\pgfusepath{clip}%
\pgfsetrectcap%
\pgfsetroundjoin%
\pgfsetlinewidth{0.803000pt}%
\definecolor{currentstroke}{rgb}{1.000000,1.000000,1.000000}%
\pgfsetstrokecolor{currentstroke}%
\pgfsetdash{}{0pt}%
\pgfpathmoveto{\pgfqpoint{0.536502in}{1.388338in}}%
\pgfpathlineto{\pgfqpoint{6.958330in}{1.388338in}}%
\pgfusepath{stroke}%
\end{pgfscope}%
\begin{pgfscope}%
\definecolor{textcolor}{rgb}{0.333333,0.333333,0.333333}%
\pgfsetstrokecolor{textcolor}%
\pgfsetfillcolor{textcolor}%
\pgftext[x=0.041670in, y=1.335577in, left, base]{\color{textcolor}\sffamily\fontsize{10.000000}{12.000000}\selectfont 0.800}%
\end{pgfscope}%
\begin{pgfscope}%
\pgfpathrectangle{\pgfqpoint{0.536502in}{0.394792in}}{\pgfqpoint{6.421828in}{1.113538in}}%
\pgfusepath{clip}%
\pgfsetrectcap%
\pgfsetroundjoin%
\pgfsetlinewidth{1.003750pt}%
\definecolor{currentstroke}{rgb}{0.886275,0.290196,0.200000}%
\pgfsetstrokecolor{currentstroke}%
\pgfsetdash{}{0pt}%
\pgfpathmoveto{\pgfqpoint{0.662421in}{0.445408in}}%
\pgfpathlineto{\pgfqpoint{0.788339in}{0.445408in}}%
\pgfpathlineto{\pgfqpoint{0.914257in}{0.445408in}}%
\pgfpathlineto{\pgfqpoint{1.040175in}{0.529804in}}%
\pgfpathlineto{\pgfqpoint{1.166093in}{0.577500in}}%
\pgfpathlineto{\pgfqpoint{1.292011in}{0.614357in}}%
\pgfpathlineto{\pgfqpoint{1.417930in}{0.627365in}}%
\pgfpathlineto{\pgfqpoint{1.543848in}{0.581836in}}%
\pgfpathlineto{\pgfqpoint{1.669766in}{0.601349in}}%
\pgfpathlineto{\pgfqpoint{1.795684in}{0.592677in}}%
\pgfpathlineto{\pgfqpoint{1.921602in}{0.579668in}}%
\pgfpathlineto{\pgfqpoint{2.047521in}{0.523300in}}%
\pgfusepath{stroke}%
\end{pgfscope}%
\begin{pgfscope}%
\pgfpathrectangle{\pgfqpoint{0.536502in}{0.394792in}}{\pgfqpoint{6.421828in}{1.113538in}}%
\pgfusepath{clip}%
\pgfsetrectcap%
\pgfsetroundjoin%
\pgfsetlinewidth{1.003750pt}%
\definecolor{currentstroke}{rgb}{0.886275,0.290196,0.200000}%
\pgfsetstrokecolor{currentstroke}%
\pgfsetdash{}{0pt}%
\pgfpathmoveto{\pgfqpoint{3.810375in}{0.445408in}}%
\pgfpathlineto{\pgfqpoint{3.936293in}{0.458260in}}%
\pgfpathlineto{\pgfqpoint{4.062212in}{0.492948in}}%
\pgfpathlineto{\pgfqpoint{4.188130in}{0.542812in}}%
\pgfpathlineto{\pgfqpoint{4.314048in}{0.614357in}}%
\pgfpathlineto{\pgfqpoint{4.439966in}{0.642541in}}%
\pgfpathlineto{\pgfqpoint{4.565884in}{0.727093in}}%
\pgfpathlineto{\pgfqpoint{4.691803in}{0.692405in}}%
\pgfpathlineto{\pgfqpoint{4.817721in}{0.675061in}}%
\pgfpathlineto{\pgfqpoint{4.943639in}{0.698909in}}%
\pgfpathlineto{\pgfqpoint{5.069557in}{0.642541in}}%
\pgfpathlineto{\pgfqpoint{5.195475in}{0.610021in}}%
\pgfusepath{stroke}%
\end{pgfscope}%
\begin{pgfscope}%
\pgfpathrectangle{\pgfqpoint{0.536502in}{0.394792in}}{\pgfqpoint{6.421828in}{1.113538in}}%
\pgfusepath{clip}%
\pgfsetrectcap%
\pgfsetroundjoin%
\pgfsetlinewidth{1.003750pt}%
\definecolor{currentstroke}{rgb}{0.203922,0.541176,0.741176}%
\pgfsetstrokecolor{currentstroke}%
\pgfsetdash{}{0pt}%
\pgfpathmoveto{\pgfqpoint{0.662421in}{0.531972in}}%
\pgfpathlineto{\pgfqpoint{0.788339in}{0.573164in}}%
\pgfpathlineto{\pgfqpoint{0.914257in}{0.516796in}}%
\pgfpathlineto{\pgfqpoint{1.040175in}{0.553652in}}%
\pgfpathlineto{\pgfqpoint{1.166093in}{0.581836in}}%
\pgfpathlineto{\pgfqpoint{1.292011in}{0.623029in}}%
\pgfpathlineto{\pgfqpoint{1.417930in}{0.603517in}}%
\pgfpathlineto{\pgfqpoint{1.543848in}{0.607853in}}%
\pgfpathlineto{\pgfqpoint{1.669766in}{0.627365in}}%
\pgfpathlineto{\pgfqpoint{1.795684in}{0.597013in}}%
\pgfpathlineto{\pgfqpoint{1.921602in}{0.633869in}}%
\pgfpathlineto{\pgfqpoint{2.047521in}{0.642541in}}%
\pgfusepath{stroke}%
\end{pgfscope}%
\begin{pgfscope}%
\pgfpathrectangle{\pgfqpoint{0.536502in}{0.394792in}}{\pgfqpoint{6.421828in}{1.113538in}}%
\pgfusepath{clip}%
\pgfsetrectcap%
\pgfsetroundjoin%
\pgfsetlinewidth{1.003750pt}%
\definecolor{currentstroke}{rgb}{0.203922,0.541176,0.741176}%
\pgfsetstrokecolor{currentstroke}%
\pgfsetdash{}{0pt}%
\pgfpathmoveto{\pgfqpoint{3.810375in}{0.445408in}}%
\pgfpathlineto{\pgfqpoint{3.936293in}{0.445408in}}%
\pgfpathlineto{\pgfqpoint{4.062212in}{0.445408in}}%
\pgfpathlineto{\pgfqpoint{4.188130in}{0.445408in}}%
\pgfpathlineto{\pgfqpoint{4.314048in}{0.445408in}}%
\pgfpathlineto{\pgfqpoint{4.439966in}{0.445408in}}%
\pgfpathlineto{\pgfqpoint{4.565884in}{0.445408in}}%
\pgfpathlineto{\pgfqpoint{4.691803in}{0.445408in}}%
\pgfpathlineto{\pgfqpoint{4.817721in}{0.445408in}}%
\pgfpathlineto{\pgfqpoint{4.943639in}{0.445408in}}%
\pgfpathlineto{\pgfqpoint{5.069557in}{0.445408in}}%
\pgfpathlineto{\pgfqpoint{5.195475in}{0.445408in}}%
\pgfusepath{stroke}%
\end{pgfscope}%
\begin{pgfscope}%
\pgfpathrectangle{\pgfqpoint{0.536502in}{0.394792in}}{\pgfqpoint{6.421828in}{1.113538in}}%
\pgfusepath{clip}%
\pgfsetrectcap%
\pgfsetroundjoin%
\pgfsetlinewidth{1.003750pt}%
\definecolor{currentstroke}{rgb}{0.596078,0.556863,0.835294}%
\pgfsetstrokecolor{currentstroke}%
\pgfsetdash{}{0pt}%
\pgfpathmoveto{\pgfqpoint{0.662421in}{0.512460in}}%
\pgfpathlineto{\pgfqpoint{0.788339in}{0.612189in}}%
\pgfpathlineto{\pgfqpoint{0.914257in}{0.636037in}}%
\pgfpathlineto{\pgfqpoint{1.040175in}{0.683733in}}%
\pgfpathlineto{\pgfqpoint{1.166093in}{0.855006in}}%
\pgfpathlineto{\pgfqpoint{1.292011in}{0.891863in}}%
\pgfpathlineto{\pgfqpoint{1.417930in}{0.911375in}}%
\pgfpathlineto{\pgfqpoint{1.543848in}{0.961239in}}%
\pgfpathlineto{\pgfqpoint{1.669766in}{0.891863in}}%
\pgfpathlineto{\pgfqpoint{1.795684in}{0.900535in}}%
\pgfpathlineto{\pgfqpoint{1.921602in}{0.922215in}}%
\pgfpathlineto{\pgfqpoint{2.047521in}{0.852838in}}%
\pgfpathlineto{\pgfqpoint{2.173439in}{0.820318in}}%
\pgfpathlineto{\pgfqpoint{2.299357in}{0.776958in}}%
\pgfpathlineto{\pgfqpoint{2.425275in}{0.724925in}}%
\pgfpathlineto{\pgfqpoint{2.551193in}{0.716253in}}%
\pgfusepath{stroke}%
\end{pgfscope}%
\begin{pgfscope}%
\pgfpathrectangle{\pgfqpoint{0.536502in}{0.394792in}}{\pgfqpoint{6.421828in}{1.113538in}}%
\pgfusepath{clip}%
\pgfsetrectcap%
\pgfsetroundjoin%
\pgfsetlinewidth{1.003750pt}%
\definecolor{currentstroke}{rgb}{0.596078,0.556863,0.835294}%
\pgfsetstrokecolor{currentstroke}%
\pgfsetdash{}{0pt}%
\pgfpathmoveto{\pgfqpoint{3.810375in}{0.538476in}}%
\pgfpathlineto{\pgfqpoint{3.936293in}{0.644709in}}%
\pgfpathlineto{\pgfqpoint{4.062212in}{0.688069in}}%
\pgfpathlineto{\pgfqpoint{4.188130in}{0.865846in}}%
\pgfpathlineto{\pgfqpoint{4.314048in}{0.933055in}}%
\pgfpathlineto{\pgfqpoint{4.439966in}{1.126008in}}%
\pgfpathlineto{\pgfqpoint{4.565884in}{1.206225in}}%
\pgfpathlineto{\pgfqpoint{4.691803in}{1.240913in}}%
\pgfpathlineto{\pgfqpoint{4.817721in}{1.251753in}}%
\pgfpathlineto{\pgfqpoint{4.943639in}{1.240913in}}%
\pgfpathlineto{\pgfqpoint{5.069557in}{1.225737in}}%
\pgfpathlineto{\pgfqpoint{5.195475in}{1.217065in}}%
\pgfpathlineto{\pgfqpoint{5.321394in}{1.336306in}}%
\pgfpathlineto{\pgfqpoint{5.447312in}{1.429530in}}%
\pgfpathlineto{\pgfqpoint{5.573230in}{1.457715in}}%
\pgfpathlineto{\pgfqpoint{5.699148in}{1.104328in}}%
\pgfusepath{stroke}%
\end{pgfscope}%
\begin{pgfscope}%
\pgfpathrectangle{\pgfqpoint{0.536502in}{0.394792in}}{\pgfqpoint{6.421828in}{1.113538in}}%
\pgfusepath{clip}%
\pgfsetrectcap%
\pgfsetroundjoin%
\pgfsetlinewidth{1.003750pt}%
\definecolor{currentstroke}{rgb}{0.466667,0.466667,0.466667}%
\pgfsetstrokecolor{currentstroke}%
\pgfsetdash{}{0pt}%
\pgfpathmoveto{\pgfqpoint{0.662421in}{0.590509in}}%
\pgfpathlineto{\pgfqpoint{0.788339in}{0.653381in}}%
\pgfpathlineto{\pgfqpoint{0.914257in}{0.874519in}}%
\pgfpathlineto{\pgfqpoint{1.040175in}{0.954735in}}%
\pgfpathlineto{\pgfqpoint{1.166093in}{0.855006in}}%
\pgfpathlineto{\pgfqpoint{1.292011in}{0.911375in}}%
\pgfpathlineto{\pgfqpoint{1.417930in}{0.909207in}}%
\pgfpathlineto{\pgfqpoint{1.543848in}{0.920047in}}%
\pgfpathlineto{\pgfqpoint{1.669766in}{0.928719in}}%
\pgfpathlineto{\pgfqpoint{1.795684in}{0.946063in}}%
\pgfpathlineto{\pgfqpoint{1.921602in}{0.922215in}}%
\pgfpathlineto{\pgfqpoint{2.047521in}{0.909207in}}%
\pgfpathlineto{\pgfqpoint{2.173439in}{0.902703in}}%
\pgfpathlineto{\pgfqpoint{2.299357in}{0.920047in}}%
\pgfpathlineto{\pgfqpoint{2.425275in}{0.852838in}}%
\pgfpathlineto{\pgfqpoint{2.551193in}{0.878855in}}%
\pgfpathlineto{\pgfqpoint{2.677112in}{0.824654in}}%
\pgfpathlineto{\pgfqpoint{2.803030in}{0.772622in}}%
\pgfpathlineto{\pgfqpoint{2.928948in}{0.807310in}}%
\pgfpathlineto{\pgfqpoint{3.054866in}{0.768286in}}%
\pgfpathlineto{\pgfqpoint{3.180784in}{0.731430in}}%
\pgfpathlineto{\pgfqpoint{3.306703in}{0.657717in}}%
\pgfpathlineto{\pgfqpoint{3.432621in}{0.683733in}}%
\pgfpathlineto{\pgfqpoint{3.558539in}{0.627365in}}%
\pgfusepath{stroke}%
\end{pgfscope}%
\begin{pgfscope}%
\pgfpathrectangle{\pgfqpoint{0.536502in}{0.394792in}}{\pgfqpoint{6.421828in}{1.113538in}}%
\pgfusepath{clip}%
\pgfsetrectcap%
\pgfsetroundjoin%
\pgfsetlinewidth{1.003750pt}%
\definecolor{currentstroke}{rgb}{0.466667,0.466667,0.466667}%
\pgfsetstrokecolor{currentstroke}%
\pgfsetdash{}{0pt}%
\pgfpathmoveto{\pgfqpoint{3.810375in}{0.575332in}}%
\pgfpathlineto{\pgfqpoint{3.936293in}{0.666389in}}%
\pgfpathlineto{\pgfqpoint{4.062212in}{0.922215in}}%
\pgfpathlineto{\pgfqpoint{4.188130in}{0.924383in}}%
\pgfpathlineto{\pgfqpoint{4.314048in}{0.935223in}}%
\pgfpathlineto{\pgfqpoint{4.439966in}{0.920047in}}%
\pgfpathlineto{\pgfqpoint{4.565884in}{0.987255in}}%
\pgfpathlineto{\pgfqpoint{4.691803in}{1.030616in}}%
\pgfpathlineto{\pgfqpoint{4.817721in}{1.021944in}}%
\pgfpathlineto{\pgfqpoint{4.943639in}{1.041456in}}%
\pgfpathlineto{\pgfqpoint{5.069557in}{1.071808in}}%
\pgfpathlineto{\pgfqpoint{5.195475in}{1.095656in}}%
\pgfpathlineto{\pgfqpoint{5.321394in}{1.145520in}}%
\pgfpathlineto{\pgfqpoint{5.447312in}{1.219233in}}%
\pgfpathlineto{\pgfqpoint{5.573230in}{1.173705in}}%
\pgfpathlineto{\pgfqpoint{5.699148in}{1.191049in}}%
\pgfpathlineto{\pgfqpoint{5.825066in}{1.212729in}}%
\pgfpathlineto{\pgfqpoint{5.950984in}{1.256089in}}%
\pgfpathlineto{\pgfqpoint{6.076903in}{1.245249in}}%
\pgfpathlineto{\pgfqpoint{6.202821in}{1.238745in}}%
\pgfpathlineto{\pgfqpoint{6.328739in}{1.199721in}}%
\pgfpathlineto{\pgfqpoint{6.454657in}{1.221401in}}%
\pgfpathlineto{\pgfqpoint{6.580575in}{1.275601in}}%
\pgfpathlineto{\pgfqpoint{6.706494in}{1.325466in}}%
\pgfusepath{stroke}%
\end{pgfscope}%
\begin{pgfscope}%
\pgfpathrectangle{\pgfqpoint{0.536502in}{0.394792in}}{\pgfqpoint{6.421828in}{1.113538in}}%
\pgfusepath{clip}%
\pgfsetbuttcap%
\pgfsetroundjoin%
\pgfsetlinewidth{1.003750pt}%
\definecolor{currentstroke}{rgb}{0.000000,0.392157,0.000000}%
\pgfsetstrokecolor{currentstroke}%
\pgfsetdash{{3.700000pt}{1.600000pt}}{0.000000pt}%
\pgfpathmoveto{\pgfqpoint{0.536502in}{0.445408in}}%
\pgfpathlineto{\pgfqpoint{6.958330in}{0.445408in}}%
\pgfusepath{stroke}%
\end{pgfscope}%
\begin{pgfscope}%
\pgfpathrectangle{\pgfqpoint{0.536502in}{0.394792in}}{\pgfqpoint{6.421828in}{1.113538in}}%
\pgfusepath{clip}%
\pgfsetbuttcap%
\pgfsetroundjoin%
\pgfsetlinewidth{1.003750pt}%
\definecolor{currentstroke}{rgb}{0.803922,0.521569,0.247059}%
\pgfsetstrokecolor{currentstroke}%
\pgfsetdash{{3.700000pt}{1.600000pt}}{0.000000pt}%
\pgfpathmoveto{\pgfqpoint{0.536502in}{1.379545in}}%
\pgfpathlineto{\pgfqpoint{6.958330in}{1.379545in}}%
\pgfusepath{stroke}%
\end{pgfscope}%
\begin{pgfscope}%
\pgfpathrectangle{\pgfqpoint{0.536502in}{0.394792in}}{\pgfqpoint{6.421828in}{1.113538in}}%
\pgfusepath{clip}%
\pgfsetrectcap%
\pgfsetroundjoin%
\pgfsetlinewidth{6.022500pt}%
\definecolor{currentstroke}{rgb}{1.000000,1.000000,1.000000}%
\pgfsetstrokecolor{currentstroke}%
\pgfsetdash{}{0pt}%
\pgfpathmoveto{\pgfqpoint{3.684457in}{0.394792in}}%
\pgfpathlineto{\pgfqpoint{3.684457in}{1.508330in}}%
\pgfusepath{stroke}%
\end{pgfscope}%
\begin{pgfscope}%
\pgfsetrectcap%
\pgfsetmiterjoin%
\pgfsetlinewidth{1.003750pt}%
\definecolor{currentstroke}{rgb}{1.000000,1.000000,1.000000}%
\pgfsetstrokecolor{currentstroke}%
\pgfsetdash{}{0pt}%
\pgfpathmoveto{\pgfqpoint{0.536502in}{0.394792in}}%
\pgfpathlineto{\pgfqpoint{0.536502in}{1.508330in}}%
\pgfusepath{stroke}%
\end{pgfscope}%
\begin{pgfscope}%
\pgfsetrectcap%
\pgfsetmiterjoin%
\pgfsetlinewidth{1.003750pt}%
\definecolor{currentstroke}{rgb}{1.000000,1.000000,1.000000}%
\pgfsetstrokecolor{currentstroke}%
\pgfsetdash{}{0pt}%
\pgfpathmoveto{\pgfqpoint{6.958330in}{0.394792in}}%
\pgfpathlineto{\pgfqpoint{6.958330in}{1.508330in}}%
\pgfusepath{stroke}%
\end{pgfscope}%
\begin{pgfscope}%
\pgfsetrectcap%
\pgfsetmiterjoin%
\pgfsetlinewidth{1.003750pt}%
\definecolor{currentstroke}{rgb}{1.000000,1.000000,1.000000}%
\pgfsetstrokecolor{currentstroke}%
\pgfsetdash{}{0pt}%
\pgfpathmoveto{\pgfqpoint{0.536502in}{0.394792in}}%
\pgfpathlineto{\pgfqpoint{6.958330in}{0.394792in}}%
\pgfusepath{stroke}%
\end{pgfscope}%
\begin{pgfscope}%
\pgfsetrectcap%
\pgfsetmiterjoin%
\pgfsetlinewidth{1.003750pt}%
\definecolor{currentstroke}{rgb}{1.000000,1.000000,1.000000}%
\pgfsetstrokecolor{currentstroke}%
\pgfsetdash{}{0pt}%
\pgfpathmoveto{\pgfqpoint{0.536502in}{1.508330in}}%
\pgfpathlineto{\pgfqpoint{6.958330in}{1.508330in}}%
\pgfusepath{stroke}%
\end{pgfscope}%
\end{pgfpicture}%
\makeatother%
\endgroup%